\documentclass[11pt]{acmsiggraph}

\usepackage[utf8]{inputenc}	

\usepackage{anyfontsize}

\usepackage[english]{babel}	

\usepackage{microtype}		

\usepackage{enumerate}		

\usepackage{times}			

\usepackage{amsmath,amssymb,amsthm}

\usepackage{graphicx} 

\usepackage{overpic}

\usepackage{pgf}

\usepackage{pgfplots}
\newlength\figureheight 
\newlength\figurewidth 

\usepackage{tikz,tkz-euclide,tkz-graph}
\usetikzlibrary{calc}
\usetkzobj{all}

\usepackage{xypic}

\usepackage{tabularx}

\usepackage{caption}
\usepackage{booktabs}		

\usepackage{multirow}

\usepackage[english]{varioref}

\usepackage{hyperref} 
\hypersetup{colorlinks,citecolor=black,filecolor=black,linkcolor=black,urlcolor=black}

\usepackage{algorithmicx,algorithm,algpseudocode}

\newcommand{\bb}[1]{\boldsymbol{\mathrm{#1}}}

\title{Shape-from-intrinsic operator}
\author{Davide Boscaini
\and
Davide Eynard
\and
Michael M. Bronstein}
\pdfauthor{Davide Boscaini}
\affiliation{Institute of Computational Science, Faculty of Informatics, University of Lugano}

\begin{document}

\teaser{

\begin{minipage}{\textwidth}
\hspace{6cm}
\setlength\figureheight{2cm} 
\setlength\figurewidth{9cm} 
%
%
\begin{tikzpicture}

\begin{axis}[%
width=\figurewidth,
height=\figureheight,
unbounded coords=jump,
scale only axis,
xtick={50,100,150,200,250,300},
xmin=0,
xmax=300,
every outer y axis line/.append style={red},
every y tick label/.append style={font=\color{red}\fontsize{6}{8}\selectfont},
ymode=log,
ymin=1,
ymax=110,
yminorticks=true,
ytick style={red},
ylabel={MfO energy},
ylabel near ticks,
ylabel style={red},
axis x line*=bottom,
axis y line*=left,
ylabel style={font=\fontsize{8}{10}\selectfont},
xlabel style={font=\fontsize{8}{10}\selectfont},
every x tick label/.append style={font=\fontsize{6}{8}\selectfont},
]
\addplot [
color=red,
solid,
line width=2.0pt,
forget plot
]
table[row sep=crcr]{
2 104.002999895734\\
3 89.892601661537\\
4 82.7044416376202\\
5 77.6111961371986\\
6 73.3085831381206\\
7 69.5842582832049\\
8 66.3153828882026\\
9 63.419514547691\\
10 60.8331899965227\\
NaN NaN\\
21 77.1828626565453\\
22 69.855482385884\\
23 65.0485123790808\\
24 60.8991182772538\\
25 57.2517366407823\\
26 54.0232316441198\\
27 51.155358833863\\
28 48.6021607089288\\
29 46.3248867279605\\
30 44.2893915435722\\
NaN NaN\\
41 47.7501313656224\\
42 45.1419392609285\\
43 42.8915620151656\\
44 40.8878832562863\\
45 39.0881941996773\\
46 37.4633699130223\\
47 35.9902182277689\\
48 34.6490989474684\\
49 33.4229452799265\\
50 32.2968035902284\\
NaN NaN\\
61 33.8236140006436\\
62 32.3780551895627\\
63 31.1295838465891\\
64 29.9992081238084\\
65 28.9633018633378\\
66 28.0078486557636\\
67 27.1224226301695\\
68 26.2985216260174\\
69 25.5289409322959\\
70 24.8074795684374\\
NaN NaN\\
81 25.6271646582136\\
82 24.6931702723822\\
83 23.8882248256784\\
84 23.1559299480976\\
85 22.4779565252588\\
86 21.8447080899105\\
87 21.2498882920808\\
88 20.6888105003313\\
89 20.1577288218149\\
90 19.6535268869112\\
NaN NaN\\
101 20.1447335703203\\
102 19.4963500509977\\
103 18.9309248365186\\
104 18.4142148575726\\
105 17.9327515696976\\
106 17.4794990264281\\
107 17.0500926246248\\
108 16.6414892094615\\
109 16.2513892229768\\
110 15.8779574319531\\
NaN NaN\\
121 16.1799770327202\\
122 15.7126981136635\\
123 15.2965817437094\\
124 14.912852414004\\
125 14.5528388381897\\
126 14.2117519788604\\
127 13.8865966734706\\
128 13.5753162682766\\
129 13.276393920027\\
130 12.9886485285634\\
NaN NaN\\
141 13.1723743776805\\
142 12.8259560584961\\
143 12.5105441511364\\
144 12.2163621435584\\
145 11.938286864393\\
146 11.6732834042418\\
147 11.4193817363328\\
148 11.1752070335861\\
149 10.9397419760934\\
150 10.7121973008091\\
NaN NaN\\
161 10.8215943544982\\
162 10.5582945958739\\
163 10.3139483645249\\
164 10.0835535696677\\
165 9.8642011187897\\
166 9.65404429885655\\
167 9.45182760813271\\
168 9.25664812967787\\
169 9.06782526954205\\
170 8.88482540003096\\
NaN NaN\\
181 8.9487097787324\\
182 8.74416201515924\\
183 8.55153429168501\\
184 8.36824119571505\\
185 8.19264266984029\\
186 8.02362981765138\\
187 7.86041281328437\\
188 7.70240383409332\\
189 7.54914862536111\\
190 7.40028461947209\\
NaN NaN\\
201 7.43742032116364\\
202 7.2755579886992\\
203 7.12149662912122\\
204 6.97385803804179\\
205 6.83170309939246\\
206 6.69436087049825\\
207 6.56133242516147\\
208 6.43223368583404\\
209 6.30675991109333\\
210 6.18466282584741\\
NaN NaN\\
221 6.20695553269925\\
222 6.07692686430599\\
223 5.95222997478056\\
224 5.83209734990158\\
225 5.71597457080895\\
226 5.60344650657052\\
227 5.49419245631267\\
228 5.38795774477529\\
229 5.28453511988102\\
230 5.18375221882297\\
NaN NaN\\
241 5.19852117611429\\
242 5.0927894109419\\
243 4.9908570316659\\
244 4.89227461701826\\
245 4.79670175262928\\
246 4.70387330908277\\
247 4.61357761389346\\
248 4.52564187842314\\
249 4.4399222108454\\
250 4.35629661610102\\
NaN NaN\\
261 4.36787987269582\\
262 4.2810566335308\\
263 4.19704365336962\\
264 4.11556364161025\\
265 4.03639830411795\\
266 3.95937196119781\\
267 3.88434038318596\\
268 3.81118299903452\\
269 3.73979734614565\\
270 3.670095044881\\
NaN NaN\\
281 3.68094231000783\\
282 3.60906859631246\\
283 3.53934103265185\\
284 3.47157945780056\\
285 3.4056375211429\\
286 3.3413942095208\\
287 3.27874782283618\\
288 3.21761161055195\\
289 3.15791055918795\\
290 3.0995789925206\\
};
\end{axis}

\begin{axis}[%
width=\figurewidth,
height=\figureheight,
unbounded coords=jump,
scale only axis,
xtick={50,100,150,200,250,300},
xticklabels={$50$,$100$,$150$,$200$,$250$,$300$},
xmin=0,
xmax=300,
xlabel={{\fontsize{8}{10}\selectfont inner iterations}},
every outer y axis line/.append style={green!50!black},
every y tick label/.append style={font=\color{green!50!black}\fontsize{6}{8}\selectfont},
ymode=log,
ymin=4e-05,
ymax=10,
yminorticks=true,
ytickten = {-4,-3,-2,-1,0},
ytick style={green!50!black},
ylabel={MDS stress},
ylabel near ticks,
ylabel style={font=\color{green!50!black}\fontsize{8}{10}\selectfont},
axis x line*=bottom,
axis y line*=right,
every x tick label/.append style={font=\fontsize{6}{8}\selectfont}
]
\addplot [
color=green!50!black,
solid,
line width=2.0pt,
forget plot
]
table[row sep=crcr]{
11 1.14325155951389\\
12 0.785232900445892\\
13 0.561035135987013\\
14 0.362481137790535\\
15 0.243144544515889\\
16 0.17541637385248\\
17 0.132085396935081\\
18 0.10330180303276\\
19 0.0840923184039628\\
20 0.0711155164540947\\
NaN NaN\\
31 0.0155122702117095\\
32 0.0122790910580929\\
33 0.0105300583860422\\
34 0.00929093284558514\\
35 0.00834431580432646\\
36 0.00759220307429321\\
37 0.00697800584765347\\
38 0.00646560812220487\\
39 0.0060306733987524\\
40 0.00565617490120074\\
NaN NaN\\
51 0.00663448338407673\\
52 0.00528906527282349\\
53 0.00460618239456427\\
54 0.0041202080987017\\
55 0.00374303225459145\\
56 0.00343812623321606\\
57 0.00318500582231364\\
58 0.00297067397829953\\
59 0.00278633358052505\\
60 0.00262576857228695\\
NaN NaN\\
71 0.00347448922699246\\
72 0.0027807061233627\\
73 0.00245592210191692\\
74 0.00222559669544813\\
75 0.00204457013635893\\
76 0.00189588482821332\\
77 0.00177046250325336\\
78 0.00166264261528911\\
79 0.00156860692801091\\
80 0.00148564711821896\\
NaN NaN\\
91 0.00212800062102016\\
92 0.00169960038738968\\
93 0.00151412679284627\\
94 0.00138433776673024\\
95 0.00128186426547809\\
96 0.00119692223604123\\
97 0.00112452752769081\\
98 0.00106164405471178\\
99 0.00100624546409238\\
100 0.00095690104686345\\
NaN NaN\\
111 0.00144268550723485\\
112 0.00113937556857143\\
113 0.00101535947012024\\
114 0.000930666091040301\\
115 0.00086440515853682\\
116 0.000809635072520744\\
117 0.000762939009352256\\
118 0.000722289142670305\\
119 0.000686360286011245\\
120 0.000654233553004203\\
NaN NaN\\
131 0.00104225888829776\\
132 0.000806559748451461\\
133 0.000713480505874072\\
134 0.000652006299608208\\
135 0.000604983431594363\\
136 0.000566681305621022\\
137 0.000534327312143558\\
138 0.000506318436736083\\
139 0.000481636652412722\\
140 0.000459595708797628\\
NaN NaN\\
151 0.000785295337207052\\
152 0.000591909856124747\\
153 0.000516937892804304\\
154 0.000469179901069502\\
155 0.000433731242975845\\
156 0.000405493887934473\\
157 0.000382018433587474\\
158 0.00036191946438894\\
159 0.000344341066083618\\
160 0.000328722741159905\\
NaN NaN\\
171 0.000610368568647754\\
172 0.0004471766964611\\
173 0.000384413999153447\\
174 0.000345739537541996\\
175 0.000317924109805632\\
176 0.000296323065476861\\
177 0.000278709587385111\\
178 0.000263844310386334\\
179 0.000250978145310486\\
180 0.000239632222395607\\
NaN NaN\\
191 0.000486790722000258\\
192 0.000347152686130714\\
193 0.000293495661600457\\
194 0.000261326139083137\\
195 0.00023885054042264\\
196 0.000221829363921223\\
197 0.00020822927732601\\
198 0.000196930882044852\\
199 0.000187268210383526\\
200 0.000178823146967764\\
NaN NaN\\
211 0.000397083018272402\\
212 0.000276773575464013\\
213 0.000230337624090389\\
214 0.000203068193433584\\
215 0.000184474984999384\\
216 0.000170708687209995\\
217 0.000159919390860279\\
218 0.000151095535466631\\
219 0.000143641688741194\\
220 0.000137188720228121\\
NaN NaN\\
231 0.000330345568572394\\
232 0.000226334961376142\\
233 0.000185843851924529\\
234 0.000162401593129397\\
235 0.000146718127288046\\
236 0.000135322325047826\\
237 0.000126540533450288\\
238 0.000119460798291953\\
239 0.000113549839118551\\
240 0.000108479770321116\\
NaN NaN\\
251 0.000279468972698345\\
252 0.000189408445889408\\
253 0.000153934064512093\\
254 0.000133570523810681\\
255 0.00012013071776298\\
256 0.000110505426005325\\
257 0.00010318897461947\\
258 9.73617796756519e-05\\
259 9.25462111513957e-05\\
260 8.84499989709638e-05\\
NaN NaN\\
271 0.000239730801746094\\
272 0.000161709732038584\\
273 0.00013054307254328\\
274 0.000112719956443202\\
275 0.000101059398388093\\
276 9.27932281529765e-05\\
277 8.65739276993099e-05\\
278 8.16673256413382e-05\\
279 7.76459902945685e-05\\
280 7.42489186170685e-05\\
NaN NaN\\
291 0.000207965882224832\\
292 0.000140382277401218\\
293 0.000112960175195342\\
294 9.72789947792552e-05\\
295 8.70677582773211e-05\\
296 7.98752220228234e-05\\
297 7.45010762835963e-05\\
298 7.02897666610294e-05\\
299 6.68593030283977e-05\\
300 6.39764502001517e-05\\
};
\end{axis}
\end{tikzpicture}%

\end{minipage}

\vspace{-3.5cm}
%
%
\begin{minipage}{\textwidth}
	\begin{overpic}
	[trim=12cm 2cm 17cm 2cm,clip,width=0.16\columnwidth]{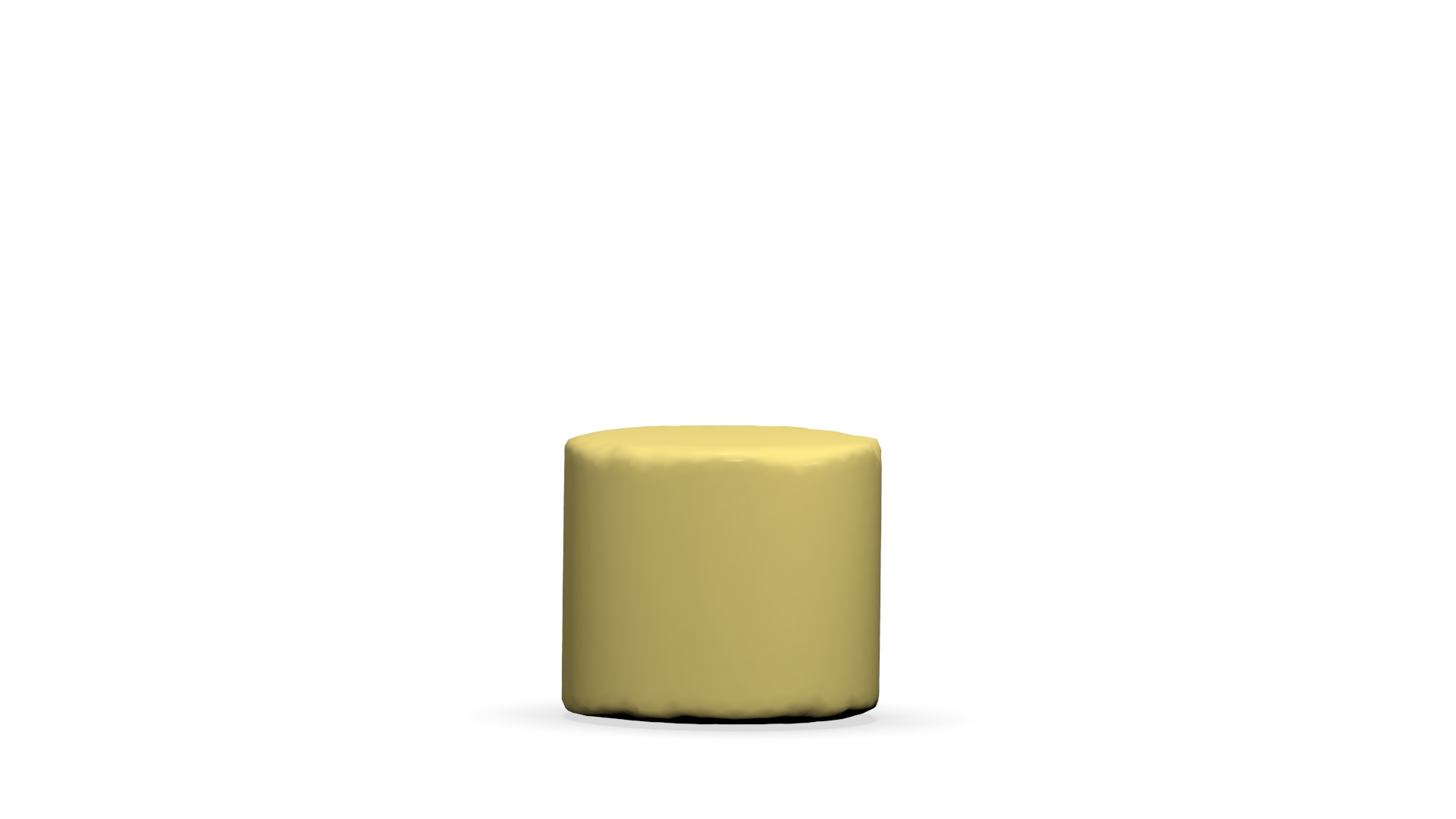}
	\put(15,48){{\fontsize{4}{6}\selectfont $\diagram\toul\enddiagram$}}
	\put(10,68){{\fontsize{8}{10}\selectfont $D_{A,B}$}}
	\end{overpic}
	\begin{overpic}
	[trim=14cm 2cm 15cm 2cm,clip,width=0.16\columnwidth]{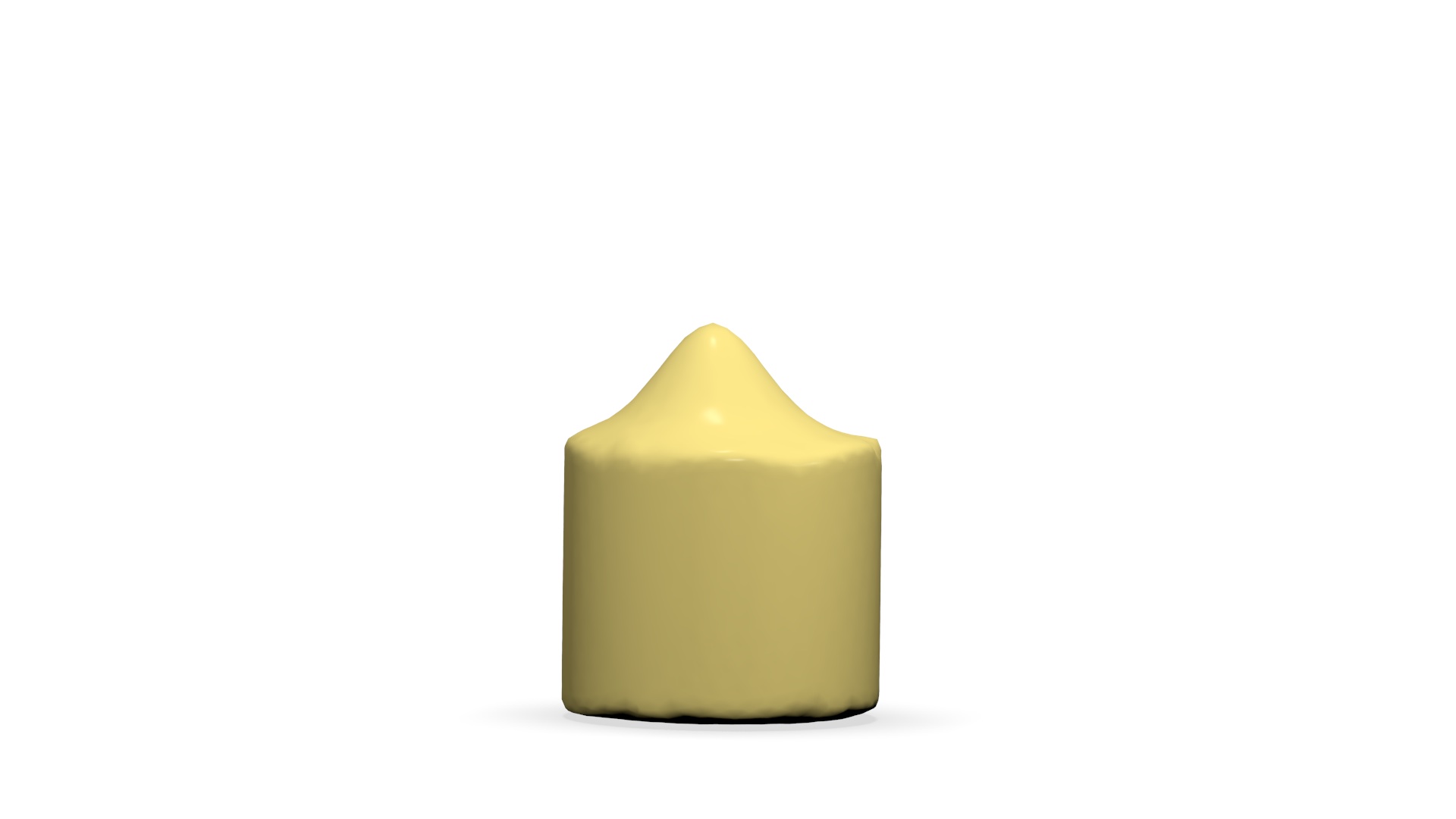}
	\put(-12,20){{\fontsize{12}{13}\selectfont $\longrightarrow$}}
	\put(-6.5,27){{\fontsize{6}{8}\selectfont $F$}}
	\end{overpic}
\end{minipage} \\
%
%
\vspace{1cm}
\begin{minipage}{\textwidth}
	\begin{overpic}
	[trim=10cm 2cm 16cm 11cm,clip,width=0.16\columnwidth]{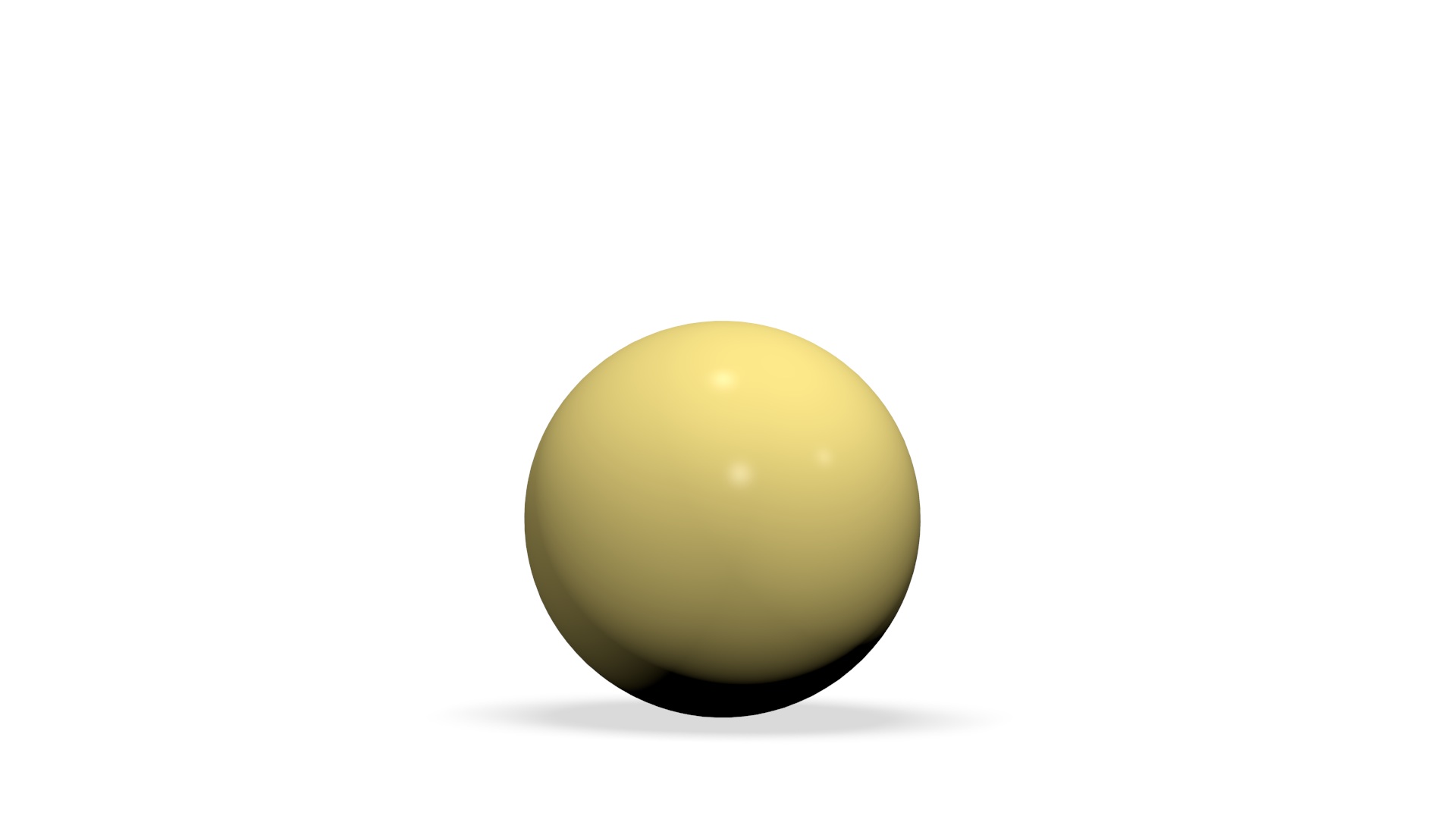}
	\put(50,92.5){{\fontsize{6}{8}\selectfont $A$}}
	\put(50,-8){{\fontsize{6}{8}\selectfont $C$}}
	\put(52,65){{\fontsize{13}{15}\selectfont $\downarrow$}}
	\put(59,69){{\fontsize{6}{8}\selectfont $G$}}
	\put(15,53){{\fontsize{4}{6}\selectfont $\diagram\toul\enddiagram$}}
	\put(10,74){{\fontsize{8}{10}\selectfont $D_{C,X}$}}
	\end{overpic}
	\begin{overpic}
	[trim=12cm 2cm 14cm 11cm,clip,width=0.16\columnwidth]{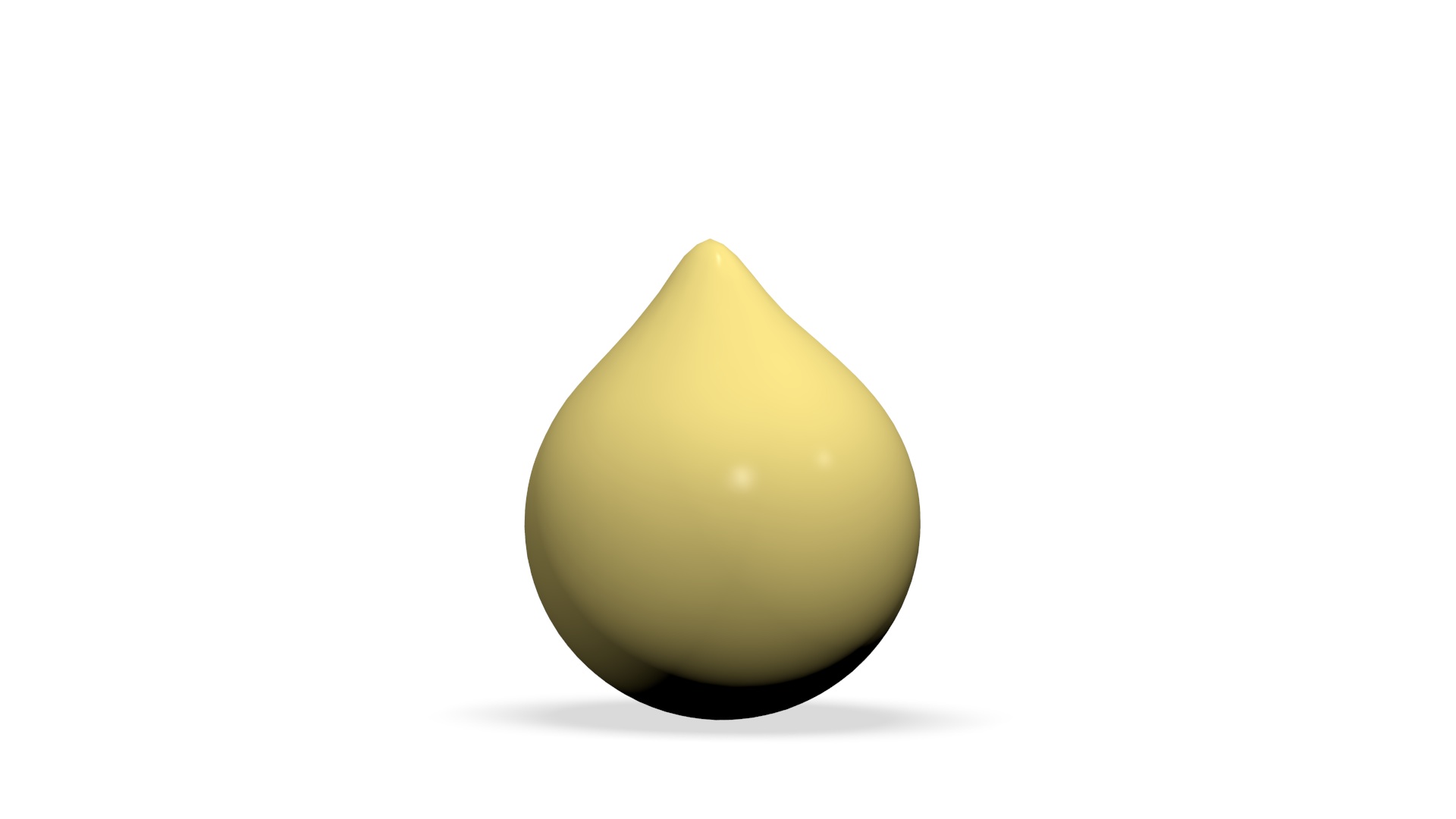}
	\put(46,92.5){{\fontsize{6}{8}\selectfont $B$}}
	\put(44,-8){{\fontsize{6}{8}\selectfont $X=\;?$}}
	\put(15,-11){\color{gray}\line(1,0){72.5}} 
	\put(15,-11){\color{gray}\line(0,1){78}} 
	\put(15,67){\color{gray}\line(1,0){72.5}} 
	\put(87.5,-11){\color{gray}\line(0,1){78}} 
	\put(-12,21){{\fontsize{12}{13}\selectfont $\longrightarrow$}}
	\put(-3,28){{\fontsize{6}{8}\selectfont $I$}}
	\end{overpic}
	\begin{overpic}
	[trim=3cm 2cm 23cm 11cm,clip,width=0.16\columnwidth]{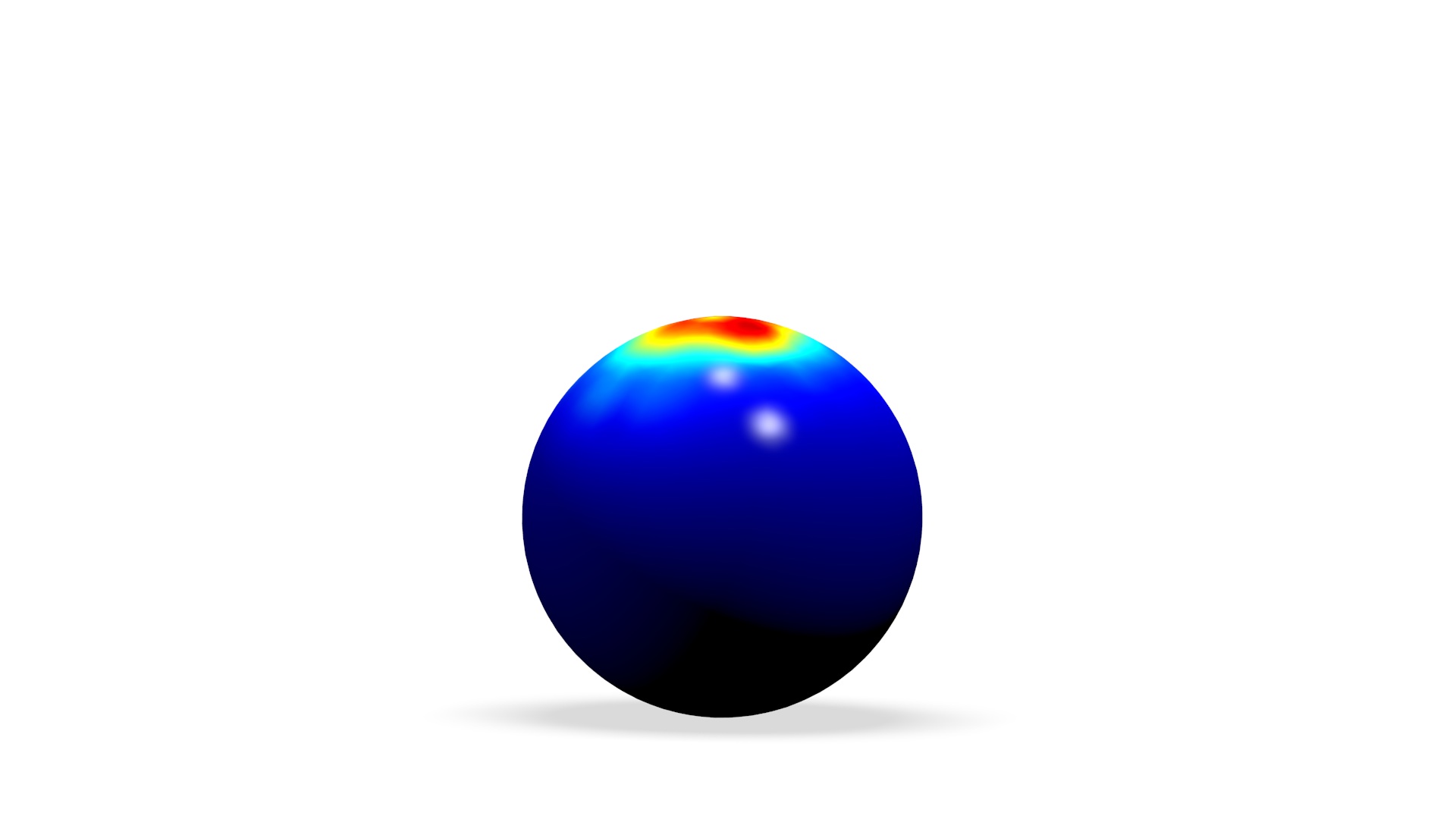}
	\put(70,-8){{\fontsize{6}{8}\selectfont $1$}}
	%
	\multiput(64.5,96.5)(0,10){7}{\color{gray}\line(0,1){5}} 
	\put(52,95){{\fontsize{5}{7}\selectfont $\underbrace{}$}}
	\put(45,80){\color{red}{\fontsize{8}{10}\selectfont MfO}}
	\put(46,70){\color{red}{\fontsize{8}{10}\selectfont iters}}
	%
	\multiput(75,96.5)(0,10){5}{\color{gray}\line(0,1){5}} 
	\put(65,95){{\fontsize{5}{7}\selectfont $\underbrace{}$}}
	\put(67,80){\color{green!50!black}{\fontsize{8}{10}\selectfont MDS}}
	\put(68,70){\color{green!50!black}{\fontsize{8}{10}\selectfont iters}}
	\put(53,167){{\fontsize{9.5}{12}\selectfont $\overbrace{}$}}
	\put(47,177){{\fontsize{8}{10}\selectfont outer iter}}
	\end{overpic}
	\begin{overpic}
	[trim=7cm 2cm 19cm 11cm,clip,width=0.16\columnwidth]{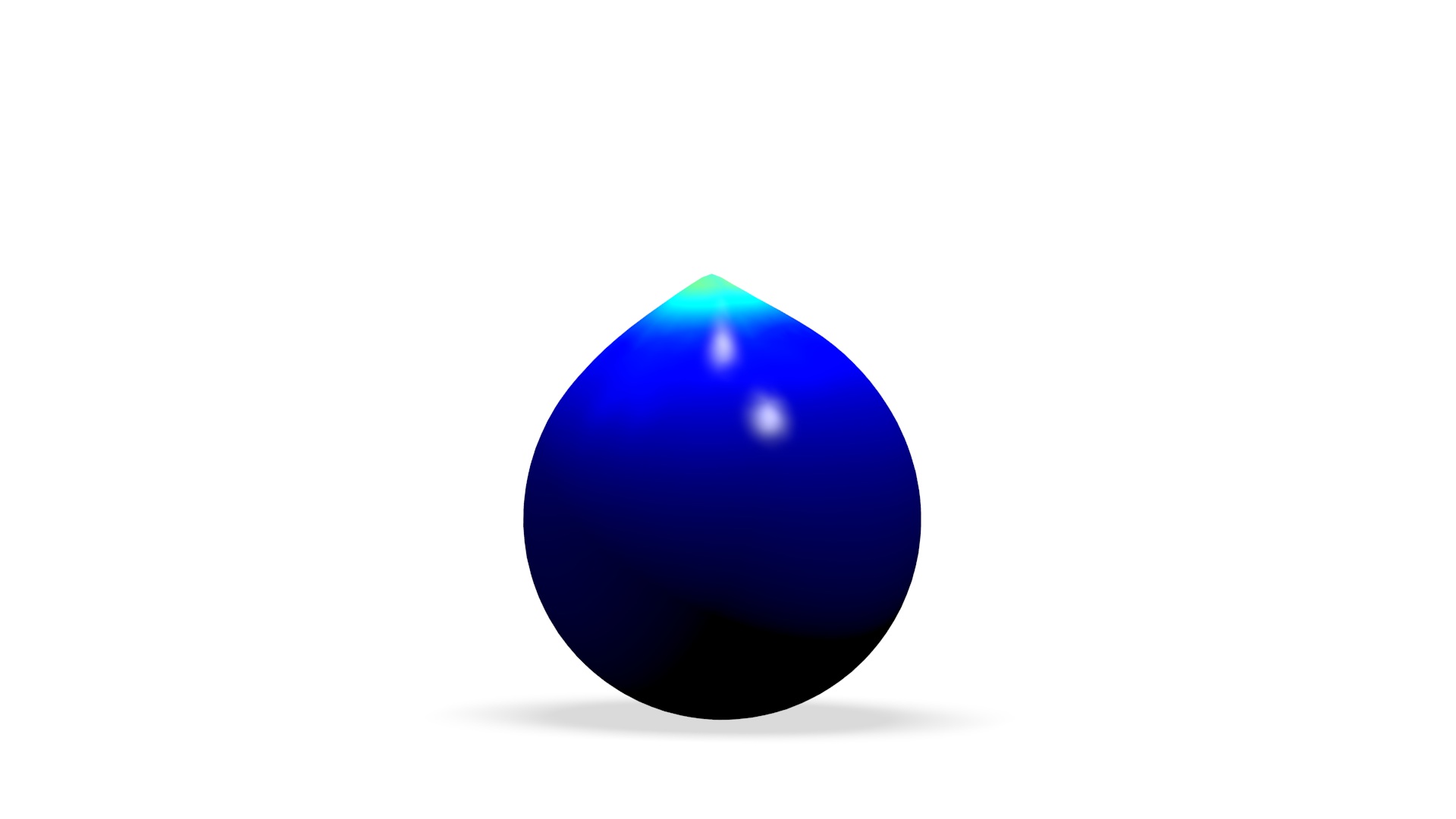}
	\put(60,-8){{\fontsize{6}{8}\selectfont $5$}}
	\put(84,-20){{\fontsize{8}{10}\selectfont outer iterations}}
	\end{overpic}
	\begin{overpic}
	[trim=12cm 2cm 14cm 11cm,clip,width=0.16\columnwidth]{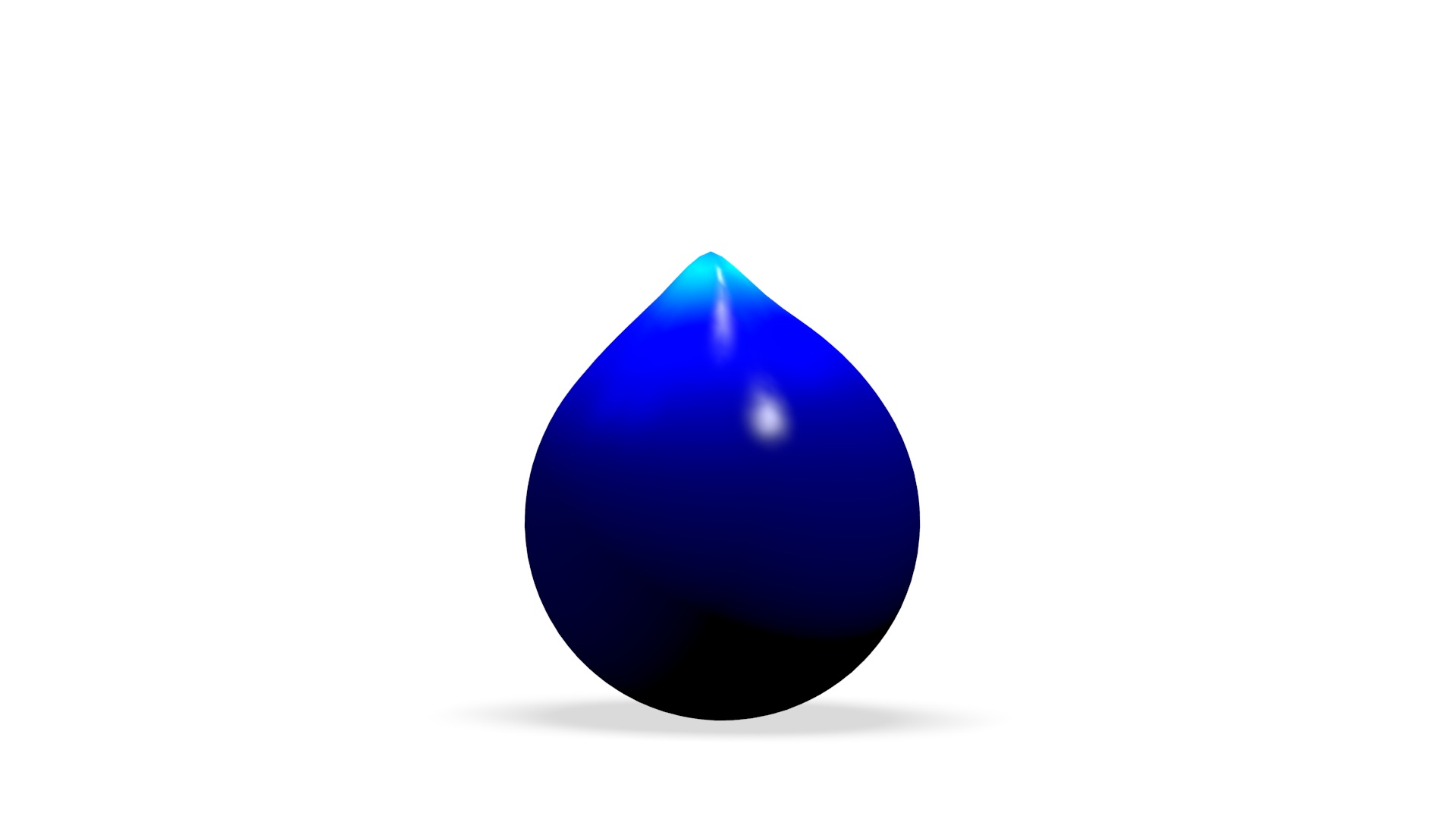}
	\put(46,-8){{\fontsize{6}{8}\selectfont $10$}}
	\end{overpic}
	\begin{overpic}
	[trim=18cm 2cm 8cm 11cm,clip,width=0.16\columnwidth]{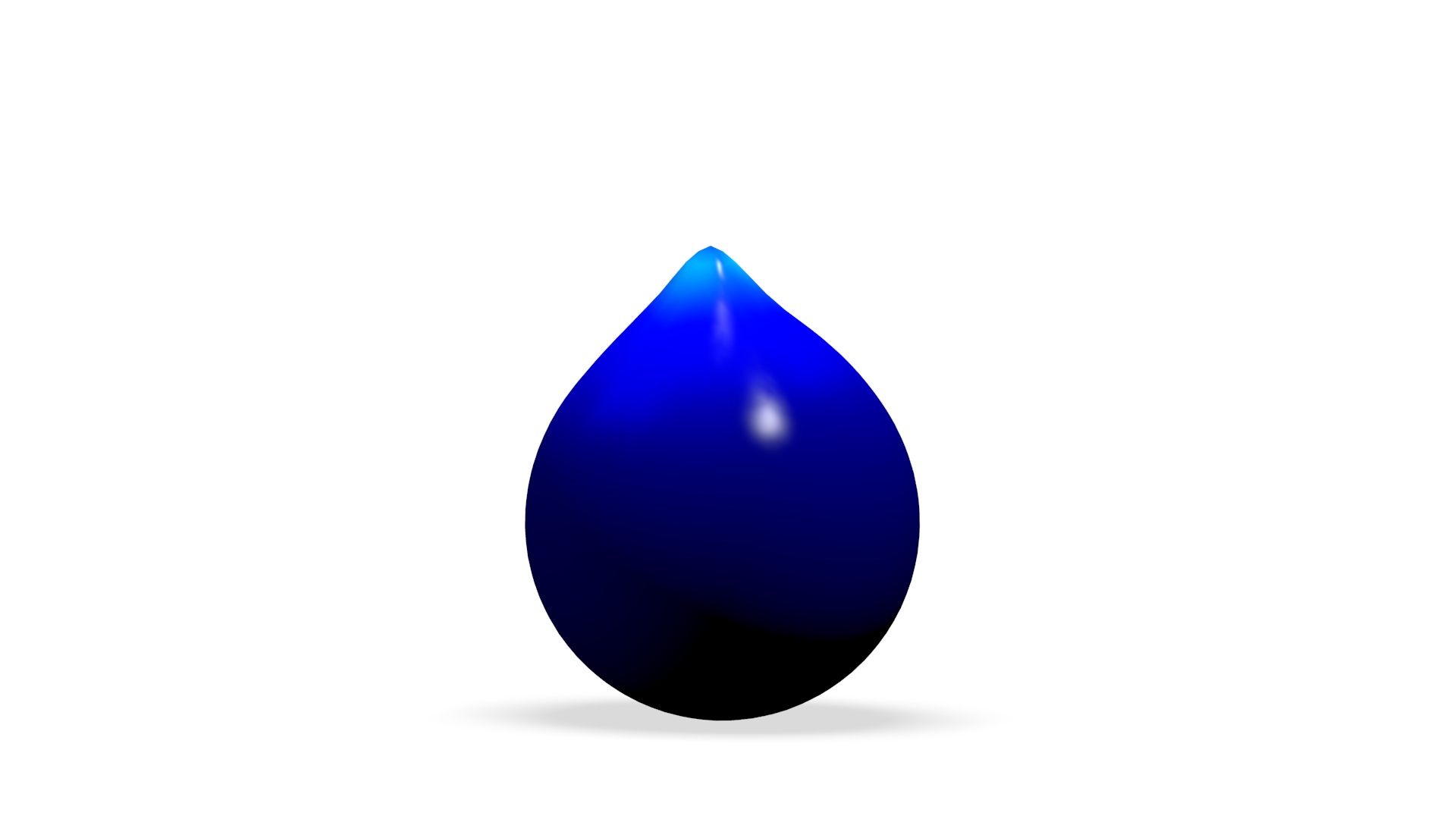}
	\put(35,-8){{\fontsize{6}{8}\selectfont $15$}}
	\put(10,15){\begin{overpic}[width=0.125\textwidth,angle=180]{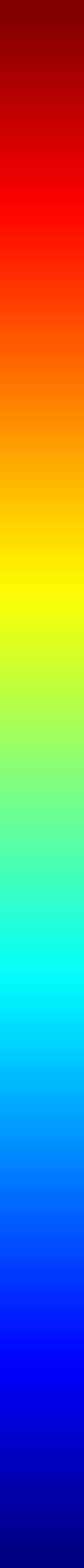}\end{overpic}}
	\put(90,5){\fontsize{6}{8}\selectfont $0$}
	\put(90,55){\fontsize{6}{8}\selectfont $10$}
	\end{overpic}
\end{minipage}

\vspace{0.5cm}

\caption{Left: shape analogy synthesis problem. Given two shapes $A$, $B$ related by a functional map $F$, the difference between them is described by means of the shape difference operator $D_{A,B}$, which captures how the respective inner products change under the functional map. Given another shape $C$ related to $A$ by a functional map $G$, the goal is to find a new shape $X$ that would be analogous to $B$, in the sense that $D_{C,X} = D_{A,B}$. 
Right: recovery of the unknown analogous shape $X$ from the difference operator using the solution of the shape-from-operator problem proposed in this paper. Our approach starts with the embedding of the shape $C$ and alternates inner iterations of minimization of the energy describing the misfit of the two operators w.r.t. to the discrete metric (metric-from-operator or MfO, Algorithm~\ref{algo:mfo}), and minimization of the stress of embedding the metric into $\mathbb{R}^3$ by a few iterations of a multidimensional scaling (MDS, Algorithm~\ref{algo:smacof}).   
Repeating for several outer iterations produces a monotonously decreasing energy (right, top), which results in shape deformation into the desired result (right, bottom). Color shows the vertex-wise energy $\epsilon_i$; hotter colors correspond to larger values (log scale). 
See text for details. 
}
\label{fig:teaser}
}

\maketitle

\begin{abstract}

Shape-from-X is an important class of problems in the fields of geometry processing, computer graphics, and vision, attempting to recover the structure of a shape from some observations. 
In this paper, we formulate the problem of shape-from-operator (SfO), recovering an embedding of a mesh from intrinsic differential operators defined on the mesh. 
Particularly interesting instances of our SfO problem include synthesis of shape analogies, shape-from-Laplacian reconstruction, and shape exaggeration. 
Numerically, we approach the SfO problem by splitting it into two optimization sub-problems that are applied in an alternating scheme: metric-from-operator (reconstruction of the discrete metric from the intrinsic operator) and embedding-from-metric (finding a shape embedding that would realize a given metric, a setting of the multidimensional scaling problem). 
%
%

\end{abstract}

\section{Introduction}
Shape reconstruction problems, colloquially known as `Shape-from-X', have been a topic of intensive research in computer vision, graphics, and geometry processing for several decades. 
Classical examples of `X' include motion \cite{Poelman1997,Kanatani1985,Snavely2006}, shading \cite{IkeuchiHorn1981,Valgaerts2012,Yu2013}, photometric stereo \cite{Woodham1989}, as well as more exotic examples such as texture \cite{ikeuchi1984,Rosenholtz1997,Forsyth01}, contour \cite{Witkin1980,brady1984extremum} and sketches \cite{Karpenko2006}.

A recent line of works by Maks Ovsjanikov and co-authors have brought operator-based approaches to geometric processing and analysis problems such as correspondence \cite{ovsjanikov2012functional}, signal processing on manifolds \cite{azencot2013operator}, and quantifying differences between shapes \cite{Rustamov2013}. 
In the latter paper, shape differences are modeled by an intrinsic linear operator, which allows to tell in a convenient way not only how different two shapes are, but also where and in which way they are different.
A particularly appealing use of shapes difference operators is to describe {\em shape analogies}, i.e. to tell how much the difference between shapes $A$ and $B$ is similar to the difference between $C$ and $D$, even if $A$ and $C$ themselves are very different (for example, a sphere and a cylinder in Figure~\ref{fig:teaser}). 
\footnote{Roughly speaking, shape difference operator defines the notion of `$B-A$' for shapes.  The shapes are analogous if `$B-A = D-C$'. 
The problem of shape analogy synthesis is how to define `$D = C + (B-A)$'. 
 }
However, while the authors show convincingly in their work how to use difference operators to {\em describe} analogies between given shapes, the challenging question how to {\em generate} such analogies (i.e., given $A, B$ and $C$, synthesize $D$) remains unanswered.

\textbf{Main contributions.}
In this paper, we study the problem of shape reconstruction from intrinsic differential operators, such as Laplacians or the aforementioned shape difference operators. 
By `shape reconstruction' we intend finding an embedding of the shape in the 3D space inducing a Riemannian metric, that, in turn, induces intrinsic operators with desired properties. 
Particularly interesting instances of our shape-from-operator problem (SfO) include synthesis of shape analogies, shape-from-Laplacian reconstruction, and shape exaggeration.

Numerically, we approach the SfO problem by splitting it into two optimization sub-problems: metric-from-operator (reconstruction of the Riemannian metric from the intrinsic operator, which, in the case of shapes discretized as triangular meshes, is represented by edge lengths) and embedding-from-metric (finding a shape embedding that would realize a given metric). 
These sub-problems are applied in an alternating way, producing the desired shape (see example in Figure~\ref{fig:teaser}). 

%
%

The rest of the paper is organized as follows. In Section~\ref{sec:related}, we overview some of the related works. 
Section~\ref{sec:backgr} introduces the notation and mathematical setting of our problem. 
In Section~\ref{sec:prob} we formulate the shape-from-operator problem, and consider two of its particular settings: shape-from-Laplacian and shape-from-difference operator. We also discuss a  numerical optimization scheme for solving this problem. 
Section~\ref{sec:res} provides experimental validation of the proposed approach. We show examples of shape reconstruction from Laplacian, shape analogy synthesis, and shape caricaturization. 
Limitations and failure cases are discussed in Section~\ref{sec:disc}, which concludes the paper.

\section{Related work}
\label{sec:related}

As already noted, shape-from-X problems have been of interest in various communities for a long time, and our problem can be regarded as another animal in this zoo. 
Recently, a few works appeared questioning what structures can be recovered from the Laplacian. 
A well-known fact in differential geometry is that the Laplace-Beltrami operator is fully determined by the Riemannian metric, and, conversely, the metric is determined by the Laplace-Beltrami operator (or a heat kernel constructed from it) \cite{rosenberg1997laplacian}. 
In the discrete setting, the length of edges of a triangular mesh plays the role of the metric, and fully determines intrinsic discrete Laplacians, e.g. cotangent weights \cite{Pinkall1993,meyer2003:ddg}. Zeng et al. \cite{Zeng2012} showed that the converse also holds for discrete metrics, and formulated the problem of discrete metric reconstruction from the Laplacian. It was shown later by \cite{deGoes2014} that this problem boils down to minimizing the conformal energy.

At the other end, we have problems generally referred to as multidimensional scaling (MDS) \cite{kruskal1964multidimensional,borg2005modern,aflalo2013spectral}, consisting of finding a configuration of points in the Euclidean space that realize, as isometrically as possible, some given distance structure. In our terminology, MDS problems can be regarded as problems of shape-from-metric reconstruction.  
In a sense, our SfO problem is a marriage between these two problems.

Several applications we discuss in relation to our problem have been considered from other perspectives. 
Methods for shape deformation and pose transfer have been proposed by \cite{Sumner2004,sorkine2004laplacian,Rong2008},
Analysis and transfer of shape style have been presented by \cite{Welnicka2011,Ma2014,alhashim_sig14}. 
Finally, shape exaggeration and caricaturization have been studied in several recent works (see e.g., \cite{Lewiner2011,Clarke2011}). 

\section{Background}
\label{sec:backgr}

\subsection{Basic definitions} 

Throughout the paper, we denote by $\bb{A} = (a_{ij})$, $\bb{a} = (a_i)$, and $a$ matrices, vectors, and scalars, respectively. $\|\bb{A}\|_\mathrm{F} = \sqrt{ \sum_{ij} \lvert a \rvert^2_{ij}}$ denotes the Frobenius norm of a matrix.

We model a $3$D shape as a simply-connected smooth compact two-dimensional surface $X$ without boundary. We denote by $T_x X$ the tangent space at point $x$ and define the Riemannian metric as the inner product $\langle \cdot, \cdot \rangle_{T_x X} \colon T_x X \times T_x X \to \mathbb{R}$ on the tangent space. 
We denote by $L^2(X)$ the space of square-integrable functions and by $H^1(X)$ the Sobolev space of weakly differentiable functions on $X$, respectively, and define the standard inner products 
\begin{align}
\label{eq:inprod1}
\langle f, g \rangle_{L^2(X)} &= \int_X f(x) g(x) da(x); \\
\langle f, g \rangle_{H^1(X)} &= \int_X \langle \nabla f(x) , \nabla g(x)\rangle_{T_x X} da(x) 
\label{eq:inprod}
\end{align}
on these spaces (here $da$ denotes the area element induced by the Riemannian metric).  
The \emph{Laplace-Beltrami operator} $\Delta f$ is defined through the \emph{Stokes formula}, 
\begin{equation}
\label{eq:stokes}
\langle \Delta f, g \rangle_{L^2(X)} = \langle f, g \rangle_{H^1(X)}, 
\end{equation}
and is \emph{intrinsic}, i.e., expressible entirely in terms of the Riemannian metric.

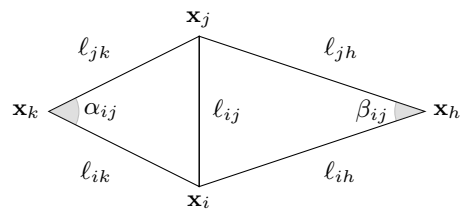
\begin{figure}[t!]
\centering
\begin{tikzpicture}[x=1cm,y=1cm]
\coordinate[label=below:$\bb{x}_i$] (xi) at (0,0) {};
\coordinate[label=above:$\bb{x}_j$] (xj) at (0,2) {};
\coordinate[label=left:$\bb{x}_k$] (xk) at (-2,1) {};
\coordinate[label=right:$\bb{x}_h$] (xh) at (3,1) {};
\draw (xi)--(xj)--(xk)--cycle;
\tkzLabelSegment[right=2pt](xi,xj){$\ell_{ij}$}
\tkzLabelSegment[above left=2pt](xj,xk){$\ell_{jk}$}
\tkzLabelSegment[below left=2pt](xk,xi){$\ell_{ik}$}
\draw (xi)--(xj)--(xh)--cycle;
\tkzLabelSegment[above right=2pt](xj,xh){$\ell_{jh}$}
\tkzLabelSegment[below right=2pt](xh,xi){$\ell_{ih}$}
\begin{scope}
\path[clip] (xk)--(xi)--(xj);
\fill[gray, opacity=0.25, draw=black] (xk) circle (4mm);
\node at ($(xk)+(0:7mm)$) {$\alpha_{ij}$};
\end{scope}
\begin{scope}
\path[clip] (xh)--(xj)--(xi);
\fill[gray, opacity=0.25, draw=black] (xh) circle (4mm);
\node at ($(xh)+(0:-7mm)$) {$\beta_{ij}$};
\end{scope}
\end{tikzpicture}
\caption{Definitions used in the paper: edge $ij$ has length $\ell_{ij}$. Angles $\alpha_{ij}$ and $\beta_{ij}$ are opposite to edge $ij$. Triangle  $ikj$ has area $A_{ijk}$. 
}
\label{fig:cot_weights}
\end{figure}

\subsection{Discrete metrics and Laplacians}
 
In the discrete setting, the surface $X$ is approximated by a manifold triangular mesh $(V,E,F)$ with vertices $V = \{1, \hdots, n\}$, in which each edge $ij \in E$ is shared by exactly two triangular faces ($ikj$ and $ihj \in F$; see Figure~\ref{fig:cot_weights} for this and the following definitions). 
A real function $f \colon X \to \mathbb{R}$ on the surface is sampled on the vertices of the mesh and can be identified with an $n$-dimensional vector $\bb{f} = (f_1, \hdots, f_n)^\top$. 
A discrete Riemannian metric is defined by assigning each edge $ij$ a \emph{length} $\ell_{ij} > 0$, satisfying the strong triangle inequality\footnote{
We require a strong version of the triangle inequality to avoid flat triangles. 
},  
\begin{equation}
\label{eq:triangineq}
\begin{aligned}
\ell_{ij} + \ell_{jk} - \ell_{ki} &> 0, \\
\ell_{jk} + \ell_{ki} - \ell_{ij} &> 0, \\ 
\ell_{ki} + \ell_{ij} - \ell_{jk} &> 0, 
\end{aligned}
\end{equation}
for all $ijk \in F$. 
We denote by $\bb{\ell} = (\ell_{ij \in E})$ the vector of edge lengths of size $\lvert E \rvert$, representing the discrete metric.

The standard inner product on the space of functions on the mesh is discretized as $\langle \bb{f}, \bb{g} \rangle_{L^2(X)} = \bb{f}^\top \bb{A} \bb{g}$, where 
\begin{equation}
\label{eq:lap_area}
\begin{aligned}
\bb{A} &= \mathrm{diag}(a_1, \hdots, a_n) \\
a_i &= \frac{1}{3} \sum_{jk: ijk \in F} A_{ijk}
\end{aligned}
\end{equation}
is the local area element equal to one third of the sum of the areas of triangles sharing the vertex $i$, and $A_{ijk}$ denotes the area of triangle $ijk$. 
Using Heron's formula for triangle area, we can express 
\begin{equation}
\label{eq:heron}
\begin{aligned}
A_{ijk} &= \sqrt{s(s-\ell_{ik})(s-\ell_{kj})(s-\ell_{ij})},\\
s &= (\ell_{ik} + \ell_{kj} +\ell_{ij})/2,
\end{aligned}
\end{equation}
entirely in terms of the discrete metric.

The discrete version of the Laplace-Beltrami operator is given as an $n\times n$ matrix $\bb{L} = \bb{A}^{-1}\bb{W}$, where $\bb{W}$ is a matrix of edge-wise weights (also referred to as {\em stiffness matrix}), satisfying 
$w_{ij} = 0$ if $ij \notin E$ and $w_{ii} = -\sum_{j\neq i}w_{ij}$. Such a matrix has a constant eigenvector with the corresponding null eigenvalue. 
In particular, we are interested in Laplacian operators that are intrinsic, i.e., expressible entirely in terms of the edge lengths $\bb{\ell}$, and consider weights given by 
\begin{equation}
\label{eq:cotan_}
w_{ij}(\bb{\ell}) = \frac{ -\ell_{ij}^2 + \ell_{jk}^2 + \ell_{ki}^2 }{8 A_{ijk}} + \frac{ -\ell_{ij}^2 + \ell_{jh}^2 + \ell_{hi}^2 }{8 A_{ijh}}
\end{equation}
for $ij \in E$.

An \emph{embedding} is the geometric realization of the mesh $(V,E,F)$ in $\mathbb{R}^3$ specified by providing the three-dimensional coordinates $\bb{x}_i$ for each vertex $i \in V$ (we will hereinafter represent the embedding by an $n\times 3$ matrix $\bb{X}$). Such an embedding induces a metric 
\begin{equation}
\bb{\ell}(\bb{X}) = (\| \bb{x}_i - \bb{x}_j \| : ij \in E).
\end{equation}
With this metric, it is easy to verify that formula~(\ref{eq:cotan_}) becomes the standard cotangent weight \cite{Pinkall1993,meyer2003:ddg} 
\begin{equation}
\label{eq:cotan}
w_{ij} = 
\begin{cases}
(\cot \alpha _{ij} + \cot \beta _{ij})/2 & i\neq j;  \\		
-\sum_{k\neq i} w_{ik} & i = j, 
\end{cases}
\end{equation}
since $\cot \alpha_{ij} = (-\ell_{ij}^2 + \ell_{jk}^2 + \ell_{ki}^2 )/(4 A_{ijk})$ \cite{Jacobson2012}. 
Thus, an embedding $\bb{X}$ defines a discrete metric $\bb{\ell}(\bb{X})$, and consequently, a discrete Laplacian $\bb{W}(\bb{\ell}(\bb{X}))$. 

In the following, with slight abuse of notation, we will use $X$ to also refer to the triangular mesh approximating the underlying smooth surface, depending on the context. Also, we will use $X$
and $\bb{X}$ interchangeably when referring to a 3D shape.

\subsection{Metric-from-Laplacian}
 
Several recent works considered the reconstruction of shape intrinsic geometry from a Laplacian operator. 
Zeng et al. \cite{Zeng2012} showed that the cotangent Laplacian and the discrete Riemannian metric (unique up to a scaling) represented by edge lengths 
are mutually defined by each other, and that the set of all discrete metrics that can be defined on a triangular mesh is convex.  
The authors showed that it is possible to find a discrete metric $\bb{\ell}$ that realizes a given `reference' Laplacian defined through edge weights $\bb{\bar{W}} = (\bar{w}_{ij})$, by minimizing the convex energy given implicitly by 
\begin{equation}\label{eq:energy_variational}
\mathcal{E}_{\mathrm{imp}}(\bb{\ell}) = \int_{\bb{\ell}_0}^{\bb{\ell}} \sum_{ij} (\bar{w}_{ij} - w_{ij}(\bb{\ell})) \, d\ell_{ij}
\end{equation}
where $w_{ij}(\bb{\ell})$ are the metric-dependent weights defined according to~(\ref{eq:cotan_}).

De Goes et al. \cite{deGoes2014} derived a closed-form expression of  (\ref{eq:energy_variational}), which turns out to be the classical \emph{conformal energy} 
\begin{equation}\label{eq:conformal_energy}
\mathcal{E}_{\mathrm{conf}}(\bb\ell) = \frac{1}{2}\sum_{ij}(w_{ij}(\bb{\ell})-\bar{w}_{ij})\ell^2_{ij} .
\end{equation}

\subsection{Shape difference operators } 
Let us now consider two shapes $X$ and $Y$ related by a point-wise bijective map $t \colon Y \to X$. 
Ovsjanikov et al. \cite{ovsjanikov2012functional} showed that $t$ induces a linear \emph{functional map} $F \colon L^2(X) \to L^2(Y)$, by which a function $f$ on $X$ is translated into a function $F f = f \circ t$ on $Y$. 
Note that in general $F$ is not necessarily a point-wise map, in the sense that a delta-function on $X$ can be mapped to a `blob' on $Y$.

Rustamov et al. \cite{Rustamov2013} showed that the difference between shapes $X$ and $Y$ can be represented in the form of a linear operator on $L^2(X)$ that describes how the respective inner products change under the functional map. 
Let $\langle \cdot, \cdot \rangle_{X}$ and $\langle \cdot, \cdot \rangle_{Y}$ denote some inner products on $L^2(X)$ and $L^2(Y)$, respectively. 
Then, the \emph{shape difference operator} is a unique linear operator $D_{X,Y} \colon  L^2(X) \to L^2(X)$ satisfying 
\begin{equation}
\label{eq:shapediff}
\langle f, D_{X,Y}g \rangle_X = \langle F f ,F g \rangle_Y,   
\end{equation}
for all $f, g\in L^2(X)$. 
The shape difference operator depends on the choice of the inner products $\langle \cdot, \cdot \rangle_{X}$ and  $\langle \cdot, \cdot \rangle_{Y}$. 
Rustamov et al. \cite{Rustamov2013} considered the two inner products~(\ref{eq:inprod1}) and~(\ref{eq:inprod}). The former gives rise to the \emph{area-based} shape difference denoted by $V_{X,Y}$, while the latter results in the \emph{conformal} shape difference $R_{X,Y}$ (we refer the reader to \cite{Rustamov2013} for derivations and technical details).

In the discrete setting, shapes $X$ and $Y$ are represented as triangular meshes with $n$ and $m$ vertices, respectively. The functional correspondence is represented by an $m \times n$ matrix $\bb{F}$, and the inner products are discretized as $\langle \bb{f}, \bb{g} \rangle_{X} = \bb{f}^\top \bb{H}_X \bb{g}$ and $\langle \bb{p}, \bb{q} \rangle_{Y} = \bb{p}^\top \bb{H}_Y \bb{q}$ (here $\bb{H}_X$ and $\bb{H}_Y$ are $n\times n$ and $m\times m$ positive-definite matrices, respectively). 
For the two aforementioned choices of inner products, the area-based difference operator is given by an $n\times n$ matrix 
\begin{equation}
\bb{V}_{X,Y} = \bb{A}_X^{-1}\bb{F}^\top \bb{A}_Y\bb{F}, 
\end{equation}
where $\bb{A}$ is defined as in (\ref{eq:lap_area}). 
The conformal shape difference operator is 
\begin{equation}
\bb{R}_{X,Y} = \bb{W}_X^{\dagger}\bb{F}^\top \bb{W}_Y\bb{F}, 
\end{equation}
where $\bb{W}$ is the matrix of cotangent weights~(\ref{eq:cotan}) and $^\dagger$ denotes the Moore-Penrose pseudoinverse.\footnote{
Note that while the matrix $\bb{A}$ is invertible, $\bb{W}$ is rank-deficient (it has one zero eigenvalue) and thus is only pseudo-invertible. 
}  
We note that both operators $\bb{V}_{X,Y}$ and $\bb{R}_{X,Y}$ are intrinsic, since matrices $\bb{A}$ and $\bb{W}$ are expressed only in terms of edge lengths.

\section{Shape-from-Operator}
\label{sec:prob}

Rustamov et al. \cite{Rustamov2013} employed the shape difference operators framework to study shape analogies. Let $A$ and $B$ be shapes related by a functional map $\bb{F}$, giving rise to the shape difference operator $\bb{D}_{A,B}$ (area-based, conformal, or both), and let $C$ be another shape related to $A$ by a functional map $\bb{G}$. Then, one would like to know what would shape $X$ be such that the difference $\bb{D}_{C,X}$ is equal to $\bb{D}_{A,B}$ under the functional map $\bb{G}$? 
In other words, one wants to find an {\em analogy} of the difference between $A$ and $B$ (see Figure~\ref{fig:teaser}). 
To find such analogies, Rustamov et al. \cite{Rustamov2013} considered a finite collection of shapes $X_1, \hdots, X_K$ and picked up the shape minimizing the energy 
\begin{multline}
\label{eq:analogy}
X^* = \mathop{\mathrm{argmin}}_{X \in \{X_1, \hdots, X_K\} }
\| \bb{V}_{C,X}\bb{G} - \bb{G}\bb{V}_{A,B} \|_{\mathrm{F}}^2 \\
+ 
\| \bb{R}_{C,X}\bb{G} - \bb{G}\bb{R}_{A,B} \|_{\mathrm{F}}^2. 
\end{multline}
%
%
The important question how to {\em generate} $X$ from the given difference operator (rather than browsing through a collection of shapes) was left open.

\subsection{Problem formulation}

This question, together with the works of \cite{Zeng2012,deGoes2014}, is the main inspiration for our present work. 
More broadly, we consider the following problem we call {\em shape-from-operator} (SfO): find an embedding $\bb{X}$ of the shape, such that the discrete metric $\bb{\ell}(\bb{X})$ it induces would make an intrinsic operator $\bb{Q}(\bb{\ell}(\bb{X}))$ satisfy some property or a set of properties (for example, one may wish to make $\bb{Q}(\bb{\ell}(\bb{X}))$ as similar as possible to some given reference operator $\bar{\bb{Q}}$). 
%
%
In this paper, we consider a class of SfO problems of the form 
\begin{multline}
\label{eq:sfo}
\bb{X}^* = \mathop{\mathrm{argmin}}_{\bb{X} \in \mathbb{R}^{n\times 3} }
\lambda\| \bb{H}_1 \bb{A}(\bb{\ell}(\bb{X})) \bb{K}_1 - \bb{J}_1 \|_{\mathrm{F}}^2 \\
+ (1- \lambda)\| \bb{H}_2\bb{W}(\bb{\ell}(\bb{X})) \bb{K}_2 - \bb{J}_2 \|_{\mathrm{F}}^2, 
\end{multline}
where $0 \leq \lambda \leq 1$ and $\bb{H}_i$, $\bb{K}_i$, and $\bb{J}_i$ are some given matrices of dimensions $m\times n$, $n\times l$, and $m \times l$, respectively.  
We denote the energy minimized in~(\ref{eq:sfo}) by $\mathcal{E}(\bb{X})$.

\textbf{Shape-from-Laplacian} is a particular setting of the SfO problem, wherein one is given 
a pair of meshes $A, B$ related by the functional map $\bb{F}$. The embedding of $B$ is not given, but instead, we are given its 
cotangent weights matrix $\bb{W}_B$. 
The goal is to find an embedding $\bb{X}$ (which can be regarded as a deformation of shape $A$) that induces a Laplacian $\bb{W}(\bb{\ell}(\bb{X}))$ as close as possible to $\bb{W}_B$ under the functional map $\bb{F}$, by minimizing 
\begin{equation}
\mathcal{E}_{\mathrm{lap}}(\bb{X}) =
\|  \bb{W}(\bb{\ell}(\bb{X}))\bb{F} - \bb{W}_B \bb{F}\|_{\mathrm{F}}^2 
\label{eq:sflap}
\end{equation}
It is easy to see that problem~(\ref{eq:sflap}) is a particular case of~(\ref{eq:sfo}) when using $\lambda = 0$, $\bb{H}_2 = \bb{I}$, $\bb{K}_2 = \bb{F}$, and $\bb{J}_2 = \bb{W}_{B}\bb{F}$.  
%
Note that our shape-from-Laplacian problem is different from the metric-from-Laplacian problems considered by \cite{Zeng2012} and \cite{deGoes2014} in the sense that we are additionally looking for an embedding that realizes the discrete metric.  
Secondly, unlike \cite{Zeng2012,deGoes2014}, we allow for arbitrary (not necessarily bijective) correspondence between $A$ and $B$.

\textbf{Shape-from-difference operator} is the problem of synthesizing the shape analogy~(\ref{eq:analogy}), where we use the embedding of the given shape $C$ as an initialization, and try to deform it to obtain the desired shape $X$. Importantly, this means that the two meshes $C$ and $X$ are compatible and the functional correspondence between them is identity. We thus have 
simpler expressions for $\bb{V}_{C,X}(\bb{X}) = \bb{A}_{C}^{-1}\bb{A}(\bb{\ell}(\bb{X}))$ and 
$\bb{R}_{C,X} = \bb{W}_{C}^{\dagger}\bb{W}(\bb{\ell}(\bb{X}))$ leading to the energy 
\begin{multline}
\label{eq:analogy_}
\mathcal{E}_{\mathrm{dif}}(\bb{X}) =  \lambda\| \bb{A}_{C}^{-1}\bb{A}(\bb{\ell}(\bb{X})) \bb{G} - \bb{G}\bb{V}_{A,B} \|_{\mathrm{F}}^2 \\
+ (1- \lambda)\| \bb{W}_{C}^\dagger\bb{W}(\bb{\ell}(\bb{X})) \bb{G} - \bb{G}\bb{R}_{A,B} \|_{\mathrm{F}}^2 
\end{multline}
that is also a particular case of~(\ref{eq:sfo}) with $\bb{H}_1 = \bb{A}_C^{-1}$,  $\bb{H}_2 = \bb{W}_C^{\dagger}$, $\bb{K}_1 = \bb{K}_2 = \bb{G}$, $\bb{J}_1 = \bb{G}\bb{V}_{A,B}$, and 
$\bb{J}_2 = \bb{G}\bb{R}_{A,B}$.

\subsection{Numerical optimization }
\label{sec:num}

In the SfO problem~(\ref{eq:sfo}), we have a two-level dependence ($\bb{W}$ or $\bb{A}$ depending on $\bb{\ell}$, which in turn depends on the embedding $\bb{X}$), making the optimization directly w.r.t. the embedding coordinates $\bb{X}$ extremely hard. 
Instead, we split the problem into two stages: first, optimize $\mathcal{E}$ w.r.t. to the discrete metric $\bb{\ell}$, and then recover the embedding $\bb{X}$ from the metric $\bb{\ell}$. 

\textbf{Embedding-from-metric} is a special setting of the {\em multidimensional scaling} (MDS) problem \cite{kruskal1964multidimensional,borg2005modern}: given a metric $\bb{\ell}$, find its Euclidean realization by minimizing the {\em stress} 
\begin{equation}
\begin{aligned}
\label{eq:mds}
\bb{X}^* &= \mathop{\mathrm{argmin}}_{\bb{X}\in \mathbb{R}^{n\times 3}} \sum_{ij \in E} (\| \bb{x}_{i} - \bb{x}_j \| - \ell_{ij})^2 \\
&= \mathop{\mathrm{argmin}}_{\mathbb{X}\in \bb{R}^{n\times 3}} \sum_{i>j} v_{ij} (\| \bb{x}_{i} - \bb{x}_j \| - \ell_{ij})^2,
\end{aligned}
\end{equation}
where $v_{ij} = 1$ if $ij \in E$ and zero otherwise. 
A classical approach for solving~(\ref{eq:mds}) is the iterative SMACOF Algorithm~\ref{algo:smacof} \cite{de1977applications} based on the fixed-point iteration of the form 
\begin{equation}
\bb{X} \leftarrow \bb{Z}^\dagger \bb{B}(\bb{X})\bb{X}, 
\label{eq:smacof}
\end{equation}
where  
\begin{equation}
\bb{Z} =
\begin{cases}
-v_{ij} & i\neq j \\
\sum_{j\neq i} v_{ij} & i=j
\end{cases}
\end{equation}
\begin{equation}
\bb{B}(\bb{X}) =
\begin{cases}
-\frac{v_{ij}\ell_{ij}}{\|\bb{x}_i - \bb{x}_j\|} & i\neq j \,\,\, \text{and} \,\,\, \bb{x}_i \neq \bb{x}_j \\
0 & i\neq j \,\,\, \text{and} \,\,\, \bb{x}_i = \bb{x}_j \\
-\sum_{j\neq i} b_{ij} & i=j
\end{cases}
\end{equation}
are $n\times n$ matrices (the matrix $\bb{Z}^\dagger$ depends only on the mesh connectivity and is pre-computed). 
It can be shown \cite{bronstein2006multigrid} that iteration~(\ref{eq:smacof}) is equivalent to steepest descent with constant step size, and is guaranteed to produce a non-increasing sequence of stress values \cite{borg2005modern}. 
The complexity of a SMACOF iteration is $\mathcal{O}(n^2)$.

\textbf{Metric-from-operator (MfO)}  is the problem 
\begin{equation}
\label{eq:metric_rec}
\bb{\ell} = \mathop{\mathrm{argmin}}_{\bb{\ell}\in \mathbb{R}^{|E|}} \mathcal{E}(\bb{\ell}) \,\,\, \text{s.t.} \,\,\, (\ref{eq:triangineq}),  
\end{equation}
where we have to restrict the search to all the valid discrete metrics\footnote{If the triangle inequality is violated, the energy $\mathcal{E}$ is not well-defined, as Heron's formula would produce imaginary triangle area values.}  satisfying the triangle inequality~(\ref{eq:triangineq}). 
Metric-from-Laplacian restoration of \cite{Zeng2012} and \cite{deGoes2014} can be seen as particular settings of this problem. 
Optimization~(\ref{eq:metric_rec}) can be carried out using standard gradient descent-type algorithm. 
The gradient of the energy $\mathcal{E}(\bb{\ell})$ is given in the Appendix; the overall complexity of the computation of the energy and its gradient is $\mathcal{O}(n^3)$. 



The main complication of solving~(\ref{eq:metric_rec}) is guaranteeing that the metric $\bb{\ell}$ remains valid throughout all the optimization iterations. 
At the same time, we note that all metrics $\bb{\ell}(\bb{X})$ arising from Euclidean embeddings $\bb{X}$ satisfy the triangle inequality by definition. This brings us to adopting the following optimization scheme:

\textbf{Alternating optimization. } We perform optimization w.r.t. to the metric~(\ref{eq:metric_rec}) and the embedding~(\ref{eq:mds}) alternatingly.  
Initializing $\bb{X}$ with the embedding coordinates $\bb{C}$ of the shape $C$, we compute the metric $\ell(\bb{X})$. We make $N_\mathrm{MfO}$ steps of safeguarded optimization Algorithm~\ref{algo:mfo} to improve the metric. Then, we compute the embedding of the improved metric performing $N_\mathrm{MDS}$ steps of the SMACOF Agorithm~\ref{algo:smacof}, starting with the current embedding. These steps are repeated for $N$ outer iterations,  as outlined in Algorithm~\ref{algo:alt}.

A convergence example of such a scheme in the shape-from-difference operator problem is shown in Figure~\ref{fig:teaser} (green curve shows the values of the stress at each internal SMACOF iteration; red curve shows the values of the energy $\mathcal{E}_\mathrm{dif}$ on each internal iteration of Algorithm~\ref{algo:mfo}).

We should note that our optimization problem~(\ref{eq:analogy_}) is non-convex and thus the described optimization method does not guarantee global convergence. However, we typically have a good initialization ($\bb{X}$ is initialized by the embedding $\bb{C}$), which shows in practice good convergence properties.

\begin{algorithm}[!htb]
\caption{Safeguarded optimization for metric-from-operator recovery (internal iterations of the alternating minimization Algorithm~\ref{algo:alt}).  }\label{algo:mfo}
\begin{algorithmic}[0]
\State \textbf{Inputs:} initial valid metric $\bb{\ell}_0$, initial step size $\mu_0$
\State \textbf{Output:} improved valid metric $\bb{\ell}$
\vspace{0.1cm}
\end{algorithmic}
\begin{algorithmic}[1]
\State Initialize $\bb{\ell} \leftarrow \bb{\ell}_0$
	\State \textbf{for} $k = 1, \hdots, N_\mathrm{MfO}$ \textbf{do}
		\State \hspace{0.45cm} Set step size $\mu \leftarrow \mu_0$
	%
		\State \hspace{0.45cm} \textbf{while} $\bb{\ell}$ invalid \textbf{or} $\mathcal{E}( \bb{\ell} ) < \mathcal{E}( \bb{\ell} - \mu \nabla \mathcal{E}(\bb{\ell}) )$ \textbf{do}
		\State \hspace{1.35cm} $\mu \leftarrow \mu/2$
		\State \hspace{0.45cm} \textbf{end while}
		\State \hspace{0.45cm}$\bb{\ell} \leftarrow \bb{\ell} - \mu \nabla \mathcal{E}(\bb{\ell})$

	\State  \textbf{end for}
	\vspace{0.1cm}	

\end{algorithmic}
\end{algorithm}

\begin{algorithm}[!htb]
\caption{SMACOF algorithm for embedding-from-metric recovery (internal iterations of the alternating minimization Algorithm~\ref{algo:alt}).  }\label{algo:smacof}
\begin{algorithmic}[0]
\State \textbf{Inputs:} metric $\bb{\ell}$, initial embedding $\bb{X}_0$
\State \textbf{Output:} embedding $\bb{X}$
\vspace{0.1cm}
\end{algorithmic}
\begin{algorithmic}[1]
\State Initialize $\bb{X} \leftarrow \bb{X}_0$
	\State \textbf{for} $k = 1, \hdots, N_\mathrm{MDS}$ \textbf{do}
		\State \hspace{0.45cm} 
		$\bb{X} \leftarrow \bb{Z}^\dagger \bb{B}(\bb{X})\bb{X}$, 
	
	\State \textbf{end for}
	\vspace{0.1cm}	

\end{algorithmic}
\end{algorithm}

\begin{algorithm}[!htb]
\caption{Alternating minimization scheme for shape-from-operator synthesis. }\label{algo:alt}
\begin{algorithmic}[0]
\State \textbf{Inputs:} shape difference operators $\bb{V}_{A,B}$, $\bb{R}_{A,B}$, functional maps $\bb{F}$, $\bb{G}$, embedding $\bb{C}$ of shape C. 
\State \textbf{Output:} embedding $\bb{X}$ 
\vspace{0.1cm}
\end{algorithmic}
\begin{algorithmic}[1]
\State Initialize the embedding $\bb{X} \leftarrow \bb{C}$
\State \textbf{for} $i = 1, \hdots, N$ \textbf{do}
		\State \hspace{0.45cm} Compute metric  $\bb{\ell}(\bb{X})$ from the embedding $\bb{X}$
		\State \hspace{0.45cm} Improve metric $\bb{\ell}$ using Algorithm~\ref{algo:mfo} 
		\State \hspace{0.45cm} Compute embedding $\bb{X}$ from metric $\bb{\ell}$ using Algorithm~\ref{algo:smacof} 
	\vspace{0.1cm}

\State \textbf{end for}			
\end{algorithmic}
\end{algorithm}

\section{Results and applications}\label{sec:res}

\begin{figure*}[t!]
\centering

\begingroup%
  \makeatletter%
  \providecommand\color[2][]{%
    \errmessage{(Inkscape) Color is used for the text in Inkscape, but the package 'color.sty' is not loaded}%
    \renewcommand\color[2][]{}%
  }%
  \providecommand\transparent[1]{%
    \errmessage{(Inkscape) Transparency is used (non-zero) for the text in Inkscape, but the package 'transparent.sty' is not loaded}%
    \renewcommand\transparent[1]{}%
  }%
  \providecommand\rotatebox[2]{#2}%
  \ifx\svgwidth\undefined%
    \setlength{\unitlength}{482.48027344bp}%
    \ifx\svgscale\undefined%
      \relax%
    \else%
      \setlength{\unitlength}{\unitlength * \real{\svgscale}}%
    \fi%
  \else%
    \setlength{\unitlength}{\svgwidth}%
  \fi%
  \global\let\svgwidth\undefined%
  \global\let\svgscale\undefined%
  \makeatother%
  \begin{picture}(1,0.67792128)%
    \put(0,0){\includegraphics[width=\unitlength]{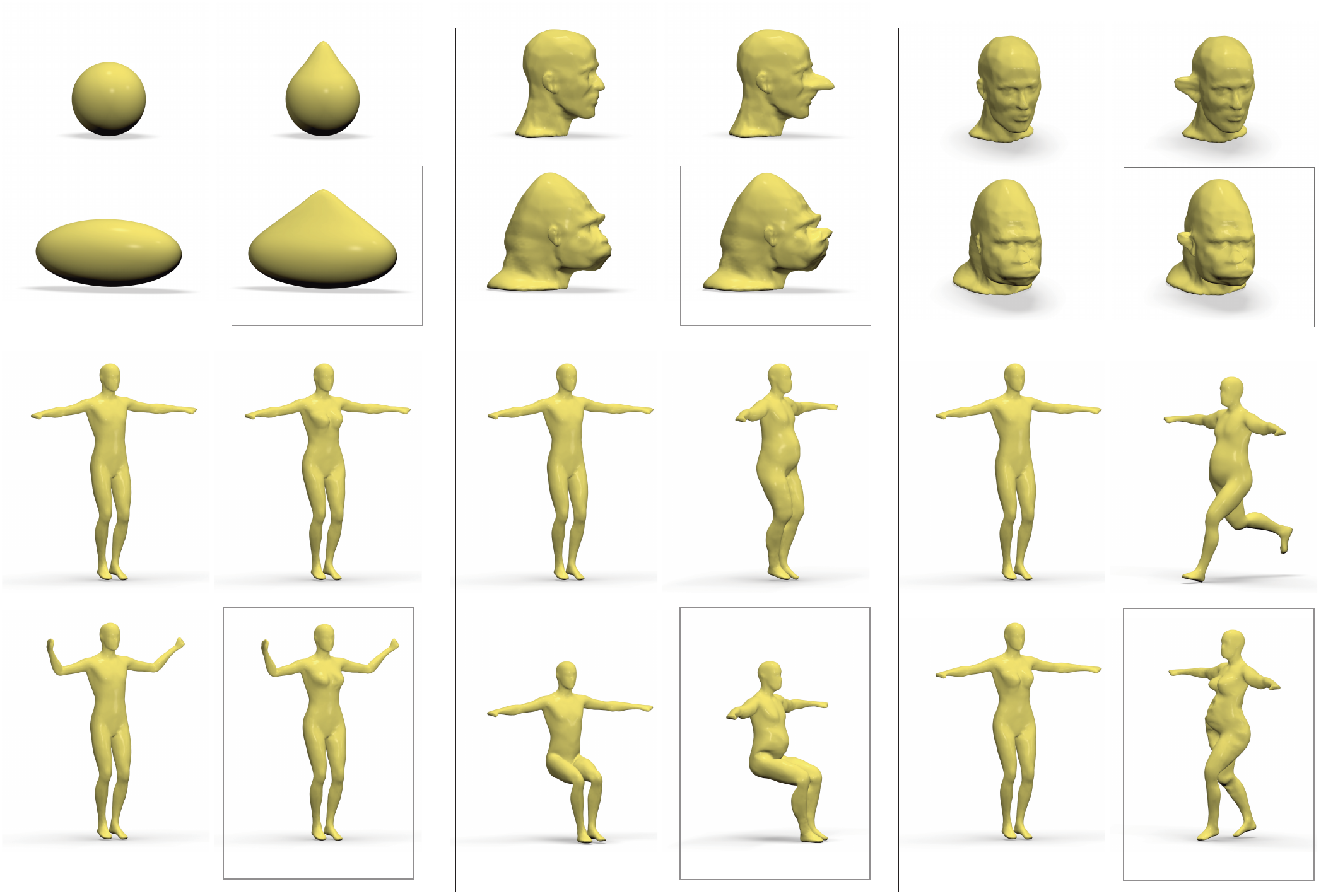}}%
    \put(0.07822537,0.55921089){\color[rgb]{0,0,0}\makebox(0,0)[lb]{\smash{\fontsize{6}{8}\selectfont$A$}}}%
    \put(0.23938428,0.4418093){\color[rgb]{0,0,0}\makebox(0,0)[lb]{\smash{\fontsize{6}{8}\selectfont$X$}}}%
    \put(0.07709786,0.44210116){\color[rgb]{0,0,0}\makebox(0,0)[lb]{\smash{\fontsize{6}{8}\selectfont$C$}}}%
    \put(0.23992814,0.55921089){\color[rgb]{0,0,0}\makebox(0,0)[lb]{\smash{\fontsize{6}{8}\selectfont$B$}}}%
    \put(0.41849549,0.55921089){\color[rgb]{0,0,0}\makebox(0,0)[lb]{\smash{\fontsize{6}{8}\selectfont$A$}}}%
    \put(0.57633823,0.4418093){\color[rgb]{0,0,0}\makebox(0,0)[lb]{\smash{\fontsize{6}{8}\selectfont$X$}}}%
    \put(0.417368,0.44210116){\color[rgb]{0,0,0}\makebox(0,0)[lb]{\smash{\fontsize{6}{8}\selectfont$C$}}}%
    \put(0.57688209,0.55921089){\color[rgb]{0,0,0}\makebox(0,0)[lb]{\smash{\fontsize{6}{8}\selectfont$B$}}}%
    \put(0.75811607,0.55921089){\color[rgb]{0,0,0}\makebox(0,0)[lb]{\smash{\fontsize{6}{8}\selectfont$A$}}}%
    \put(0.91927495,0.4418093){\color[rgb]{0,0,0}\makebox(0,0)[lb]{\smash{\fontsize{6}{8}\selectfont$X$}}}%
    \put(0.75698858,0.44210116){\color[rgb]{0,0,0}\makebox(0,0)[lb]{\smash{\fontsize{6}{8}\selectfont$C$}}}%
    \put(0.91981882,0.55921089){\color[rgb]{0,0,0}\makebox(0,0)[lb]{\smash{\fontsize{6}{8}\selectfont$B$}}}%
    \put(0.07822537,0.22300089){\color[rgb]{0,0,0}\makebox(0,0)[lb]{\smash{\fontsize{6}{8}\selectfont$A$}}}%
    \put(0.23938428,0.02222929){\color[rgb]{0,0,0}\makebox(0,0)[lb]{\smash{\fontsize{6}{8}\selectfont$X$}}}%
    \put(0.07709786,0.02252111){\color[rgb]{0,0,0}\makebox(0,0)[lb]{\smash{\fontsize{6}{8}\selectfont$C$}}}%
    \put(0.23992814,0.22300089){\color[rgb]{0,0,0}\makebox(0,0)[lb]{\smash{\fontsize{6}{8}\selectfont$B$}}}%
    \put(0.41849549,0.22300089){\color[rgb]{0,0,0}\makebox(0,0)[lb]{\smash{\fontsize{6}{8}\selectfont$A$}}}%
    \put(0.57633823,0.02222929){\color[rgb]{0,0,0}\makebox(0,0)[lb]{\smash{\fontsize{6}{8}\selectfont$X$}}}%
    \put(0.417368,0.02252111){\color[rgb]{0,0,0}\makebox(0,0)[lb]{\smash{\fontsize{6}{8}\selectfont$C$}}}%
    \put(0.57688209,0.22300089){\color[rgb]{0,0,0}\makebox(0,0)[lb]{\smash{\fontsize{6}{8}\selectfont$B$}}}%
    \put(0.75811607,0.22300089){\color[rgb]{0,0,0}\makebox(0,0)[lb]{\smash{\fontsize{6}{8}\selectfont$A$}}}%
    \put(0.91927495,0.02222929){\color[rgb]{0,0,0}\makebox(0,0)[lb]{\smash{\fontsize{6}{8}\selectfont$X$}}}%
    \put(0.75698858,0.02252111){\color[rgb]{0,0,0}\makebox(0,0)[lb]{\smash{\fontsize{6}{8}\selectfont$C$}}}%
    \put(0.91981882,0.22300089){\color[rgb]{0,0,0}\makebox(0,0)[lb]{\smash{\fontsize{6}{8}\selectfont$B$}}}%
  \end{picture}%
\endgroup%

\caption{Synthesis of shape analogies (given $A$, $B$, and $C$, find $X$) by solving the shape-from-difference operator problem using the optimization method described in this paper.}
\label{fig:analogies1}
\end{figure*}

In this section, we show the applications of our approach for shape synthesis from intrinsic differential operators, considering the shape-from-Laplacian and shape-from-difference operator problems described in Section~\ref{sec:prob}. 
As we noted, both problems are part of the same framework, so we use the general problem of shape-from-difference operator in various settings.

We used shapes from TOSCA \cite{tosca} and AIM@SHAPE \cite{aimatshape} datasets, as well as from Gabriel Peyr{\'e}'s graph toolbox \cite{Peyre_toolbox}. 
All the shapes were downsampled and isotropically remeshed to $1$K--$3.5$K vertices. Shape deformations were created using Blender. 
The optimization scheme of Section~\ref{sec:num} was implemented in MATLAB and executed on a MacPro  
machine with 2.6GHz CPU and 16GB RAM. 
Typical timing for a mesh with $3.5$K vertices was 3.43 sec for a SMACOF iteration and 2.48 sec for an MfO iteration. In most of our experiments, we used the values $\lambda = 0.5$, $N_\mathrm{MfO}=1$--$10$, and $N_\mathrm{MDS}=10$. 

%

\begin{figure*}[t!]
\centering
%
\vspace{1cm}
\begin{minipage}{\textwidth}
	\begin{overpic}
	[trim=7cm 0cm 15cm 0cm,clip,width=0.12\columnwidth]{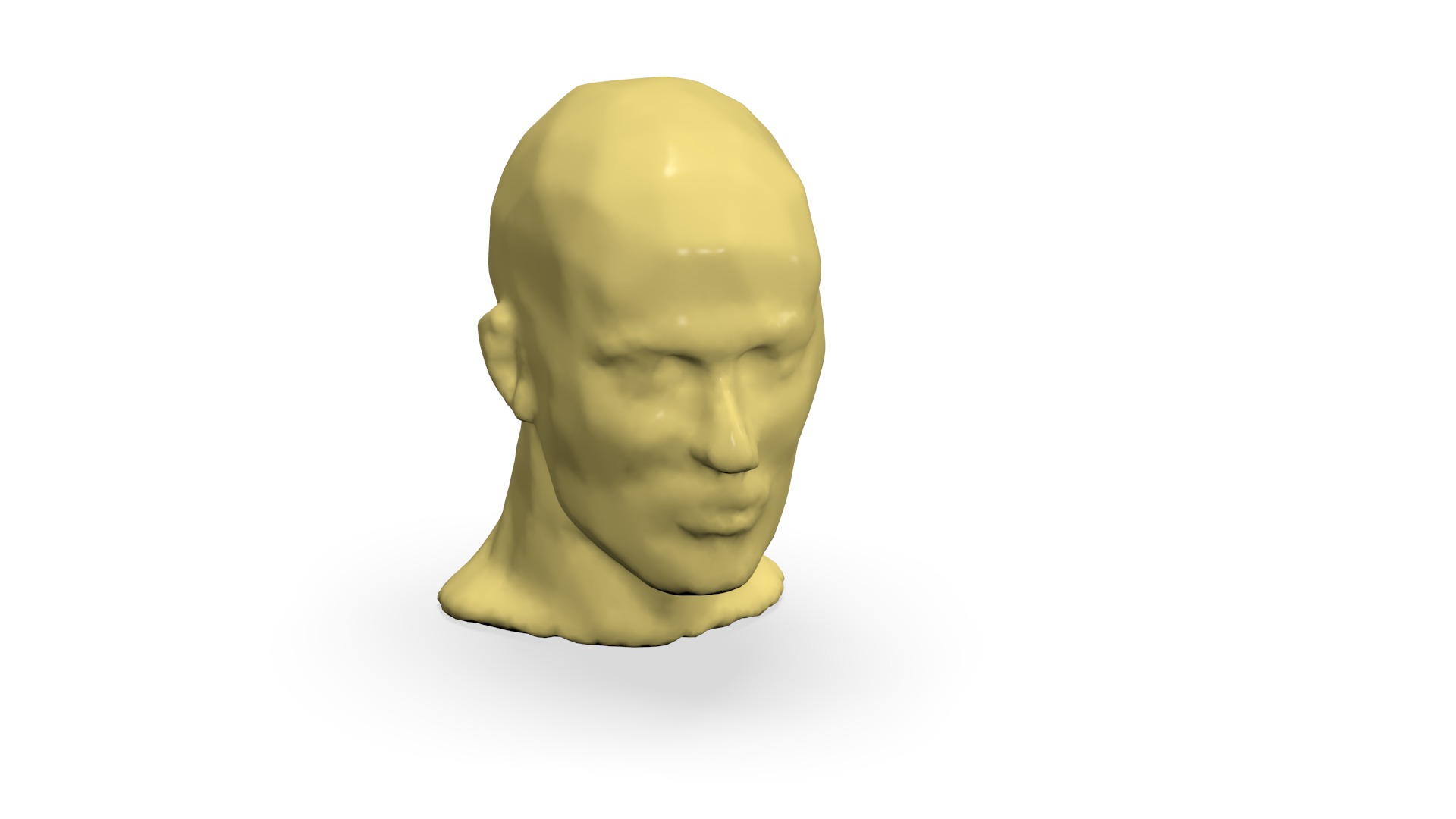}
	\end{overpic}
	\begin{overpic}
	[trim=7cm 0cm 15cm 0cm,clip,width=0.12\columnwidth]{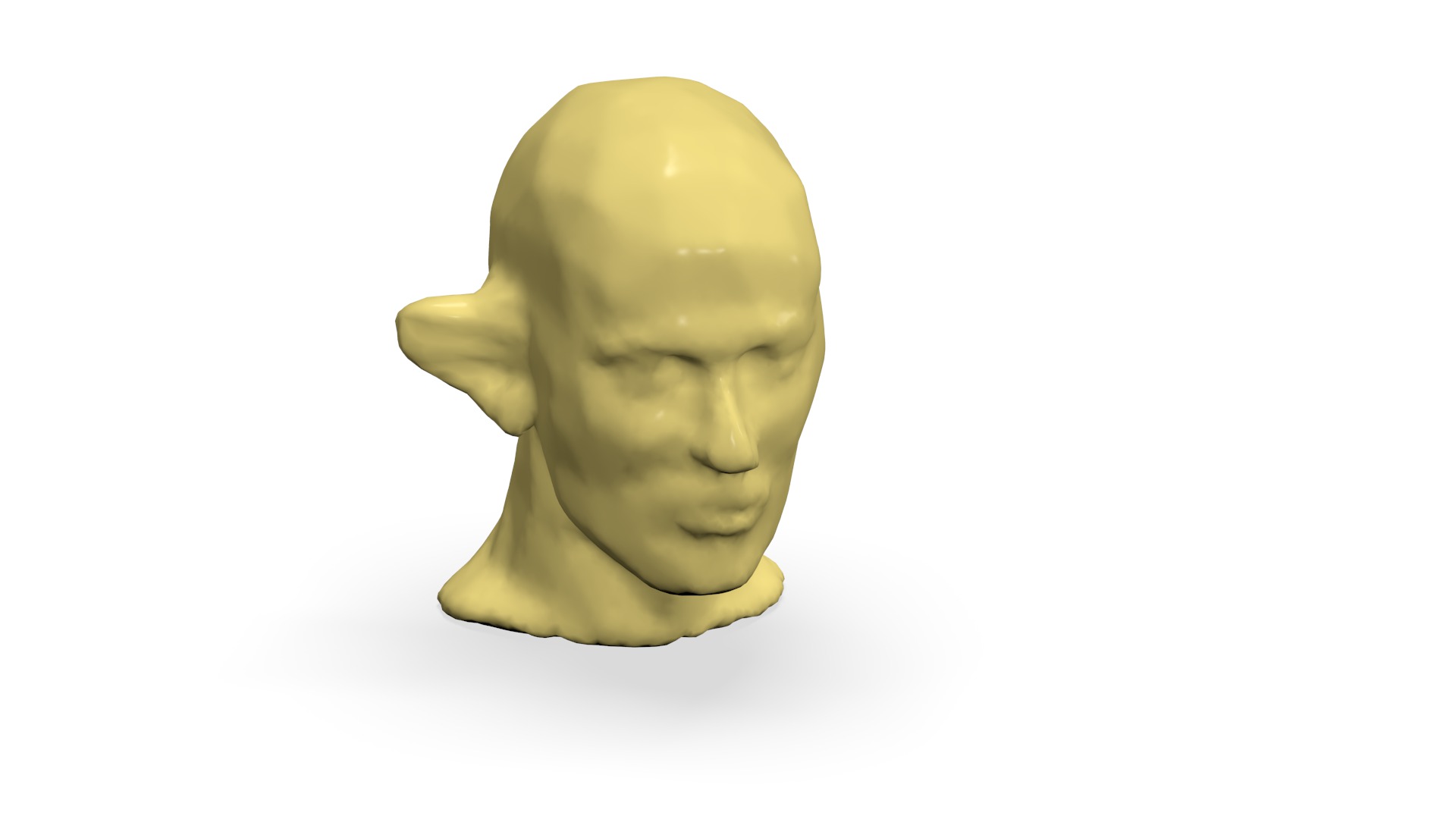}
	\end{overpic} 
	\begin{overpic}
	[trim=7cm 2cm 15cm 0cm,clip,width=0.12\columnwidth]{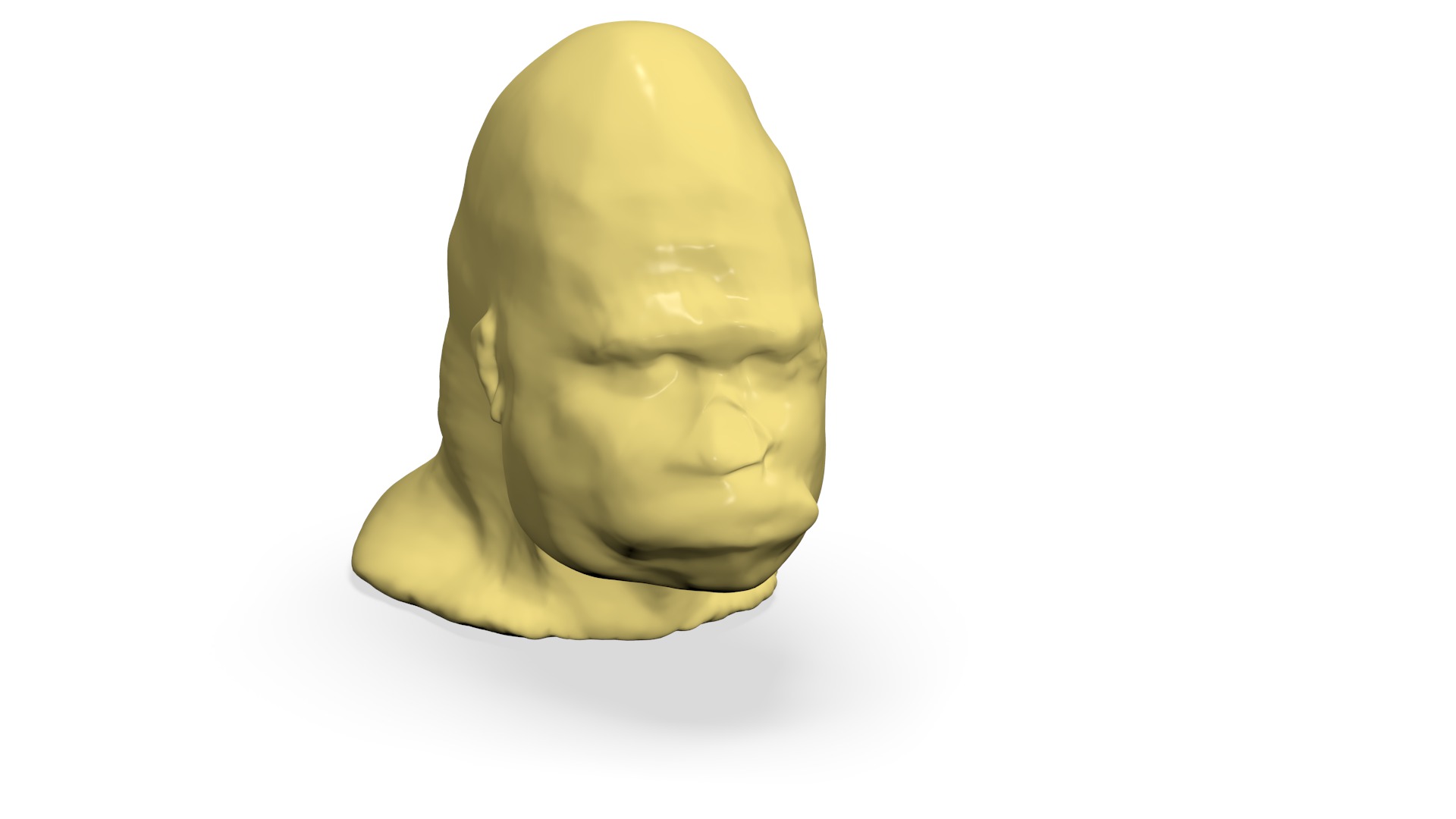}
	\end{overpic}
	\begin{overpic}
	[trim=7cm 2cm 15cm 0cm,clip,width=0.12\columnwidth]{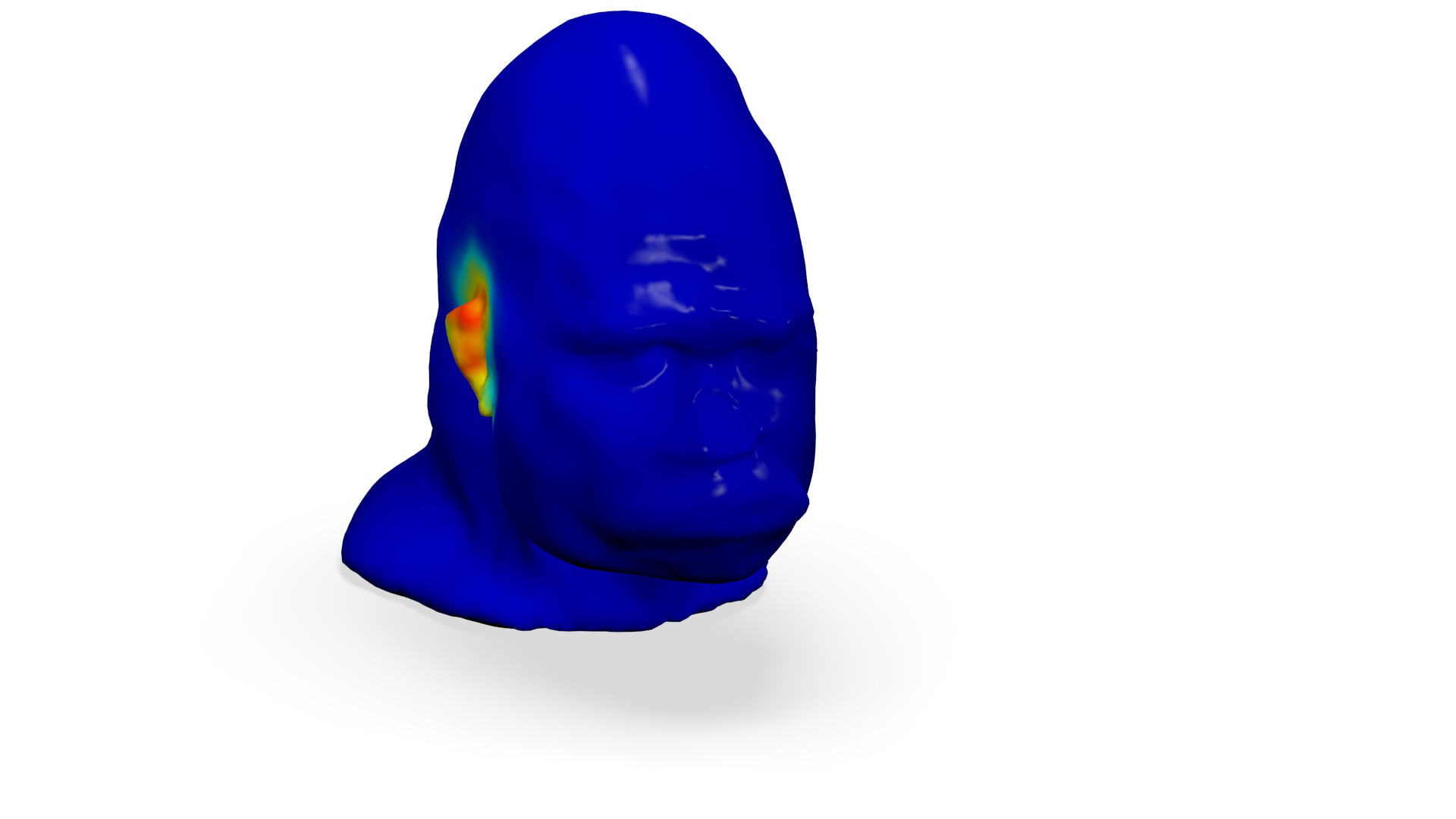}
	\put(42,-8){\fontsize{6}{8}\selectfont $1$}
	\end{overpic}
	\begin{overpic}
	[trim=7cm 2cm 15cm 0cm,clip,width=0.12\columnwidth]{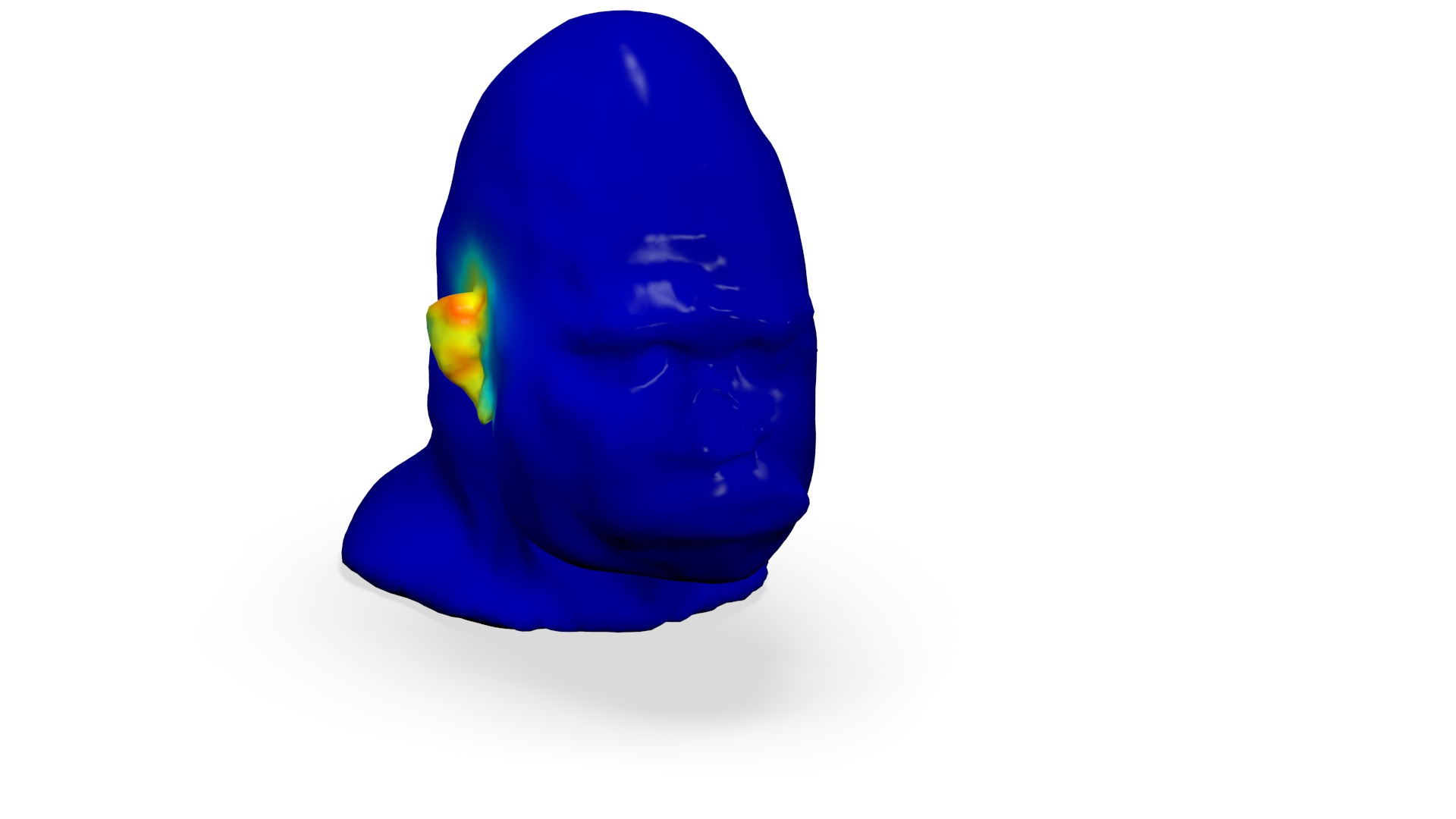}
	\put(35,-8){\fontsize{6}{8}\selectfont $10$}
	\end{overpic}
	\begin{overpic}
	[trim=7cm 2cm 15cm 0cm,clip,width=0.12\columnwidth]{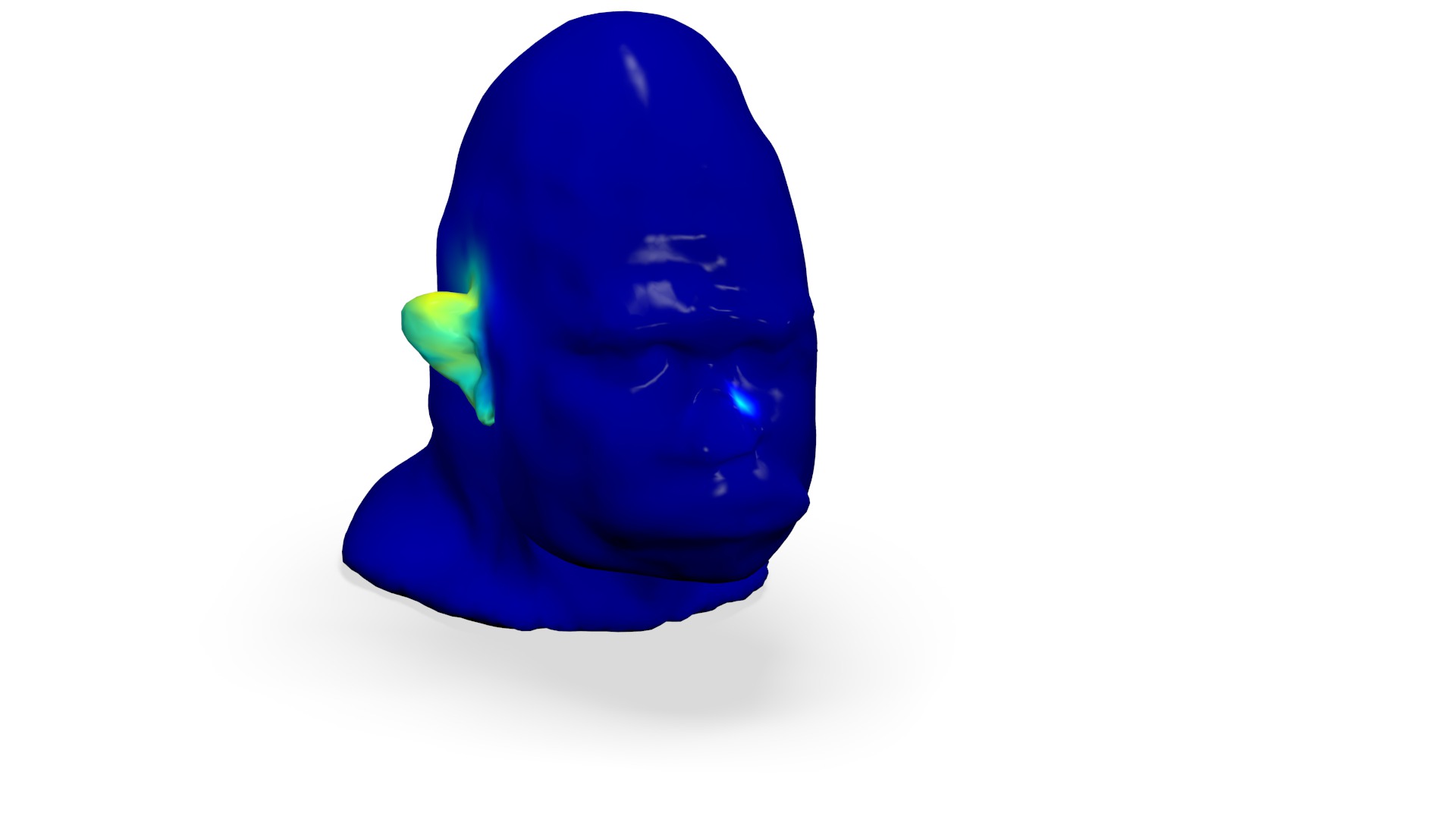}
	\put(35,-8){\fontsize{6}{8}\selectfont $50$}
	\end{overpic}
	\begin{overpic}
	[trim=7cm 2cm 15cm 0cm,clip,width=0.12\columnwidth]{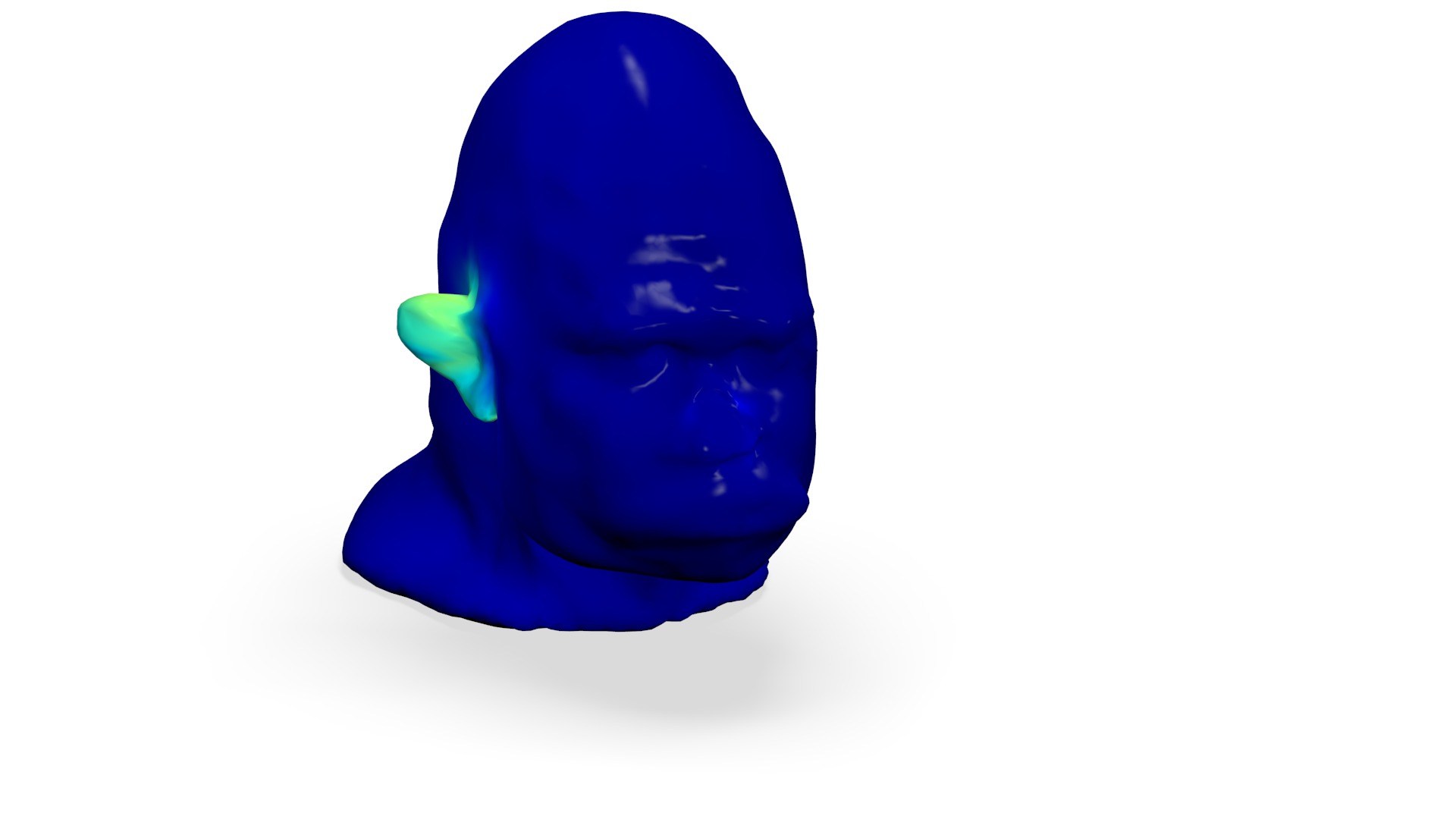}
	\put(35,-8){\fontsize{6}{8}\selectfont $200$}
	\end{overpic}
	\begin{overpic}
	[trim=7cm 2cm 15cm 0cm,clip,width=0.12\columnwidth]{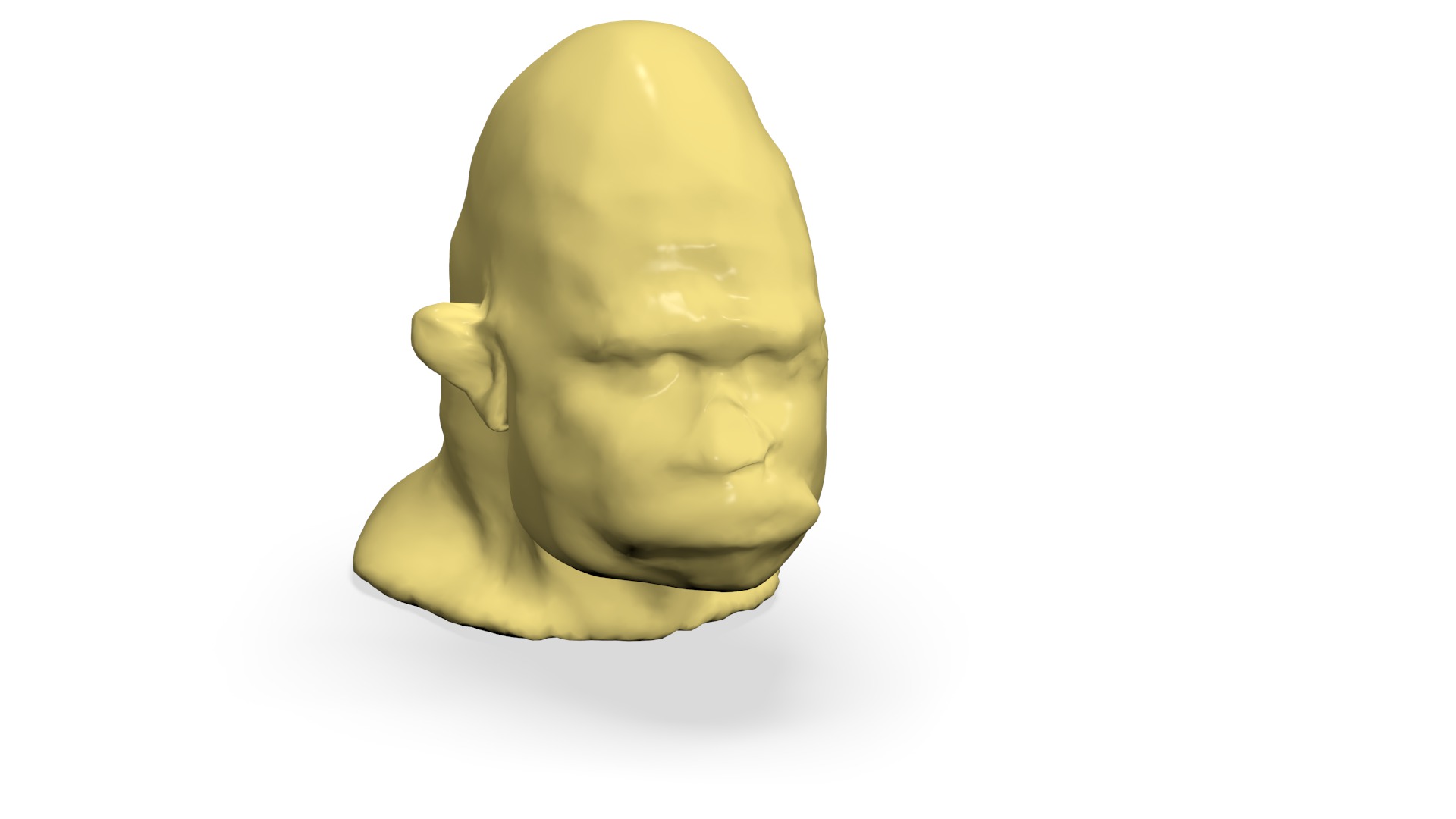}
	%
	%
	\put(-13,22){\begin{overpic}[height=12mm,width=22mm]{./figures/colorbar}\end{overpic}}
	\put(-5,10){\fontsize{6}{8}\selectfont $0$}
	\put(-5,77){\fontsize{6}{8}\selectfont $10^{1.7}$}
	\end{overpic}
\end{minipage} \\ \vspace{0.4cm}%
%
\begin{minipage}{\textwidth}
	\begin{overpic}
	[trim=17cm 0cm 18cm 0cm,clip,width=0.12\columnwidth]{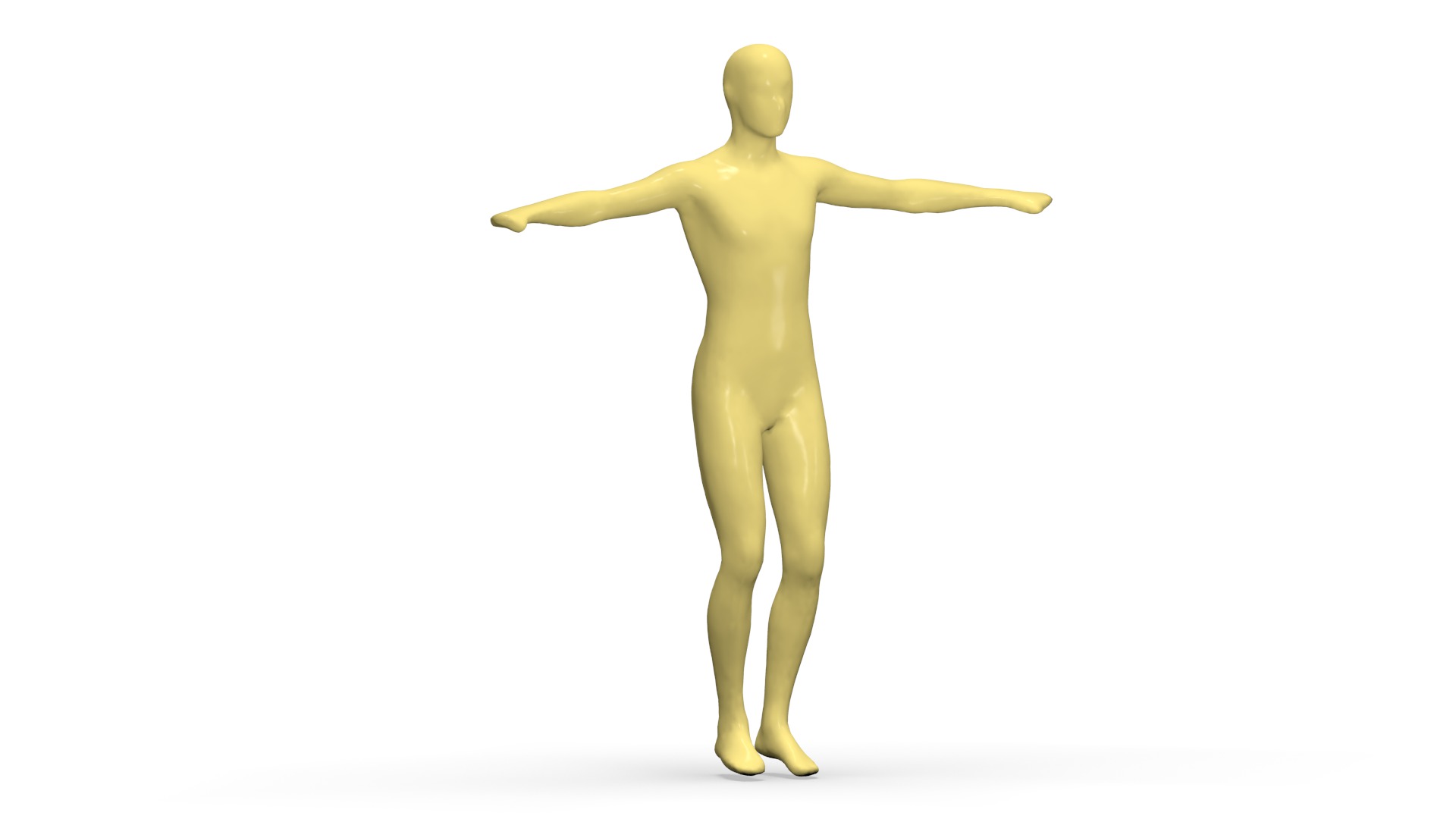}
	\end{overpic}
	\begin{overpic}
	[trim=17cm 0cm 18cm 0cm,clip,width=0.12\columnwidth]{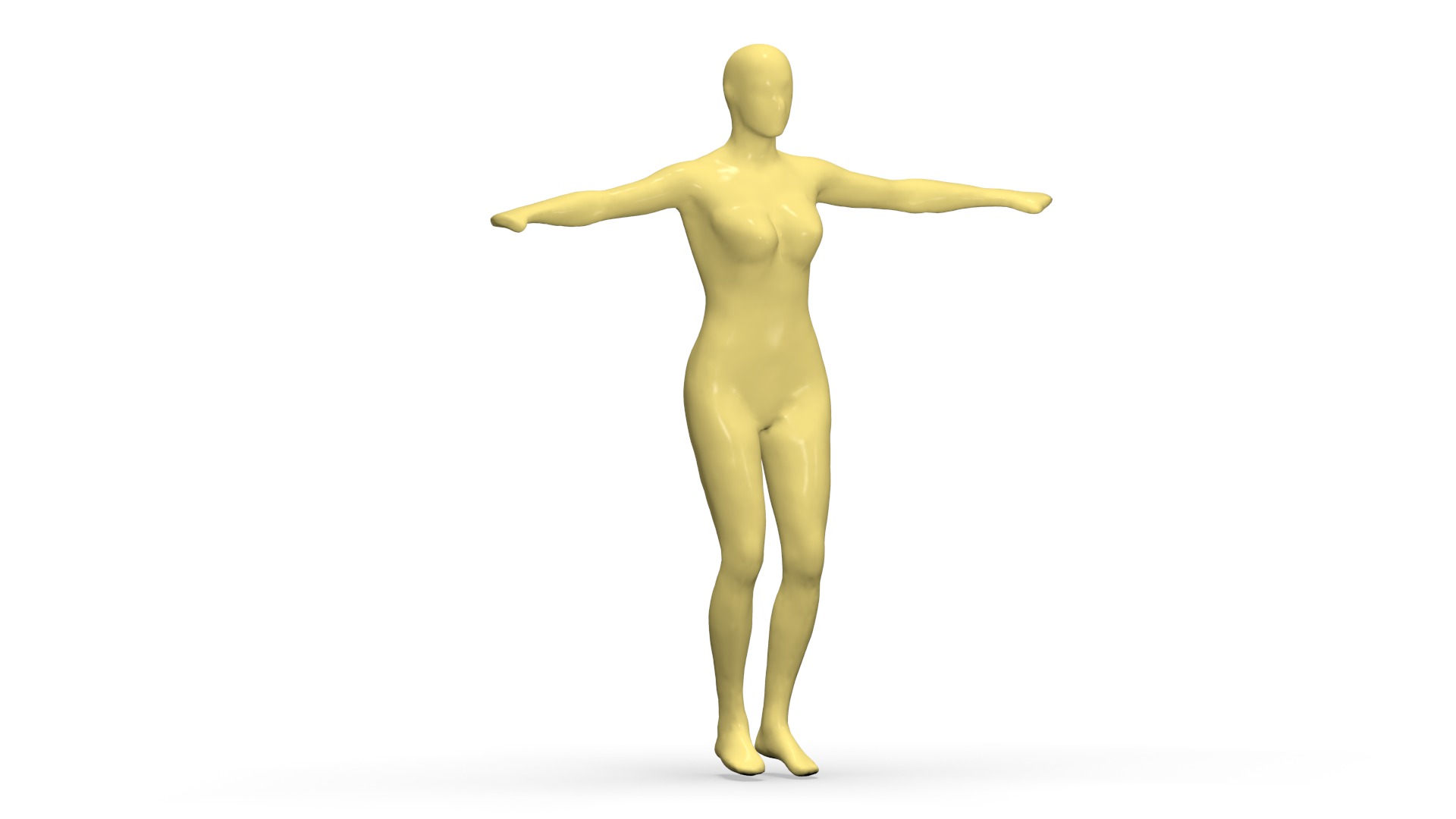}
	\end{overpic} 
	\begin{overpic}
	[trim=17cm 0cm 18cm 0cm,clip,width=0.12\textwidth]{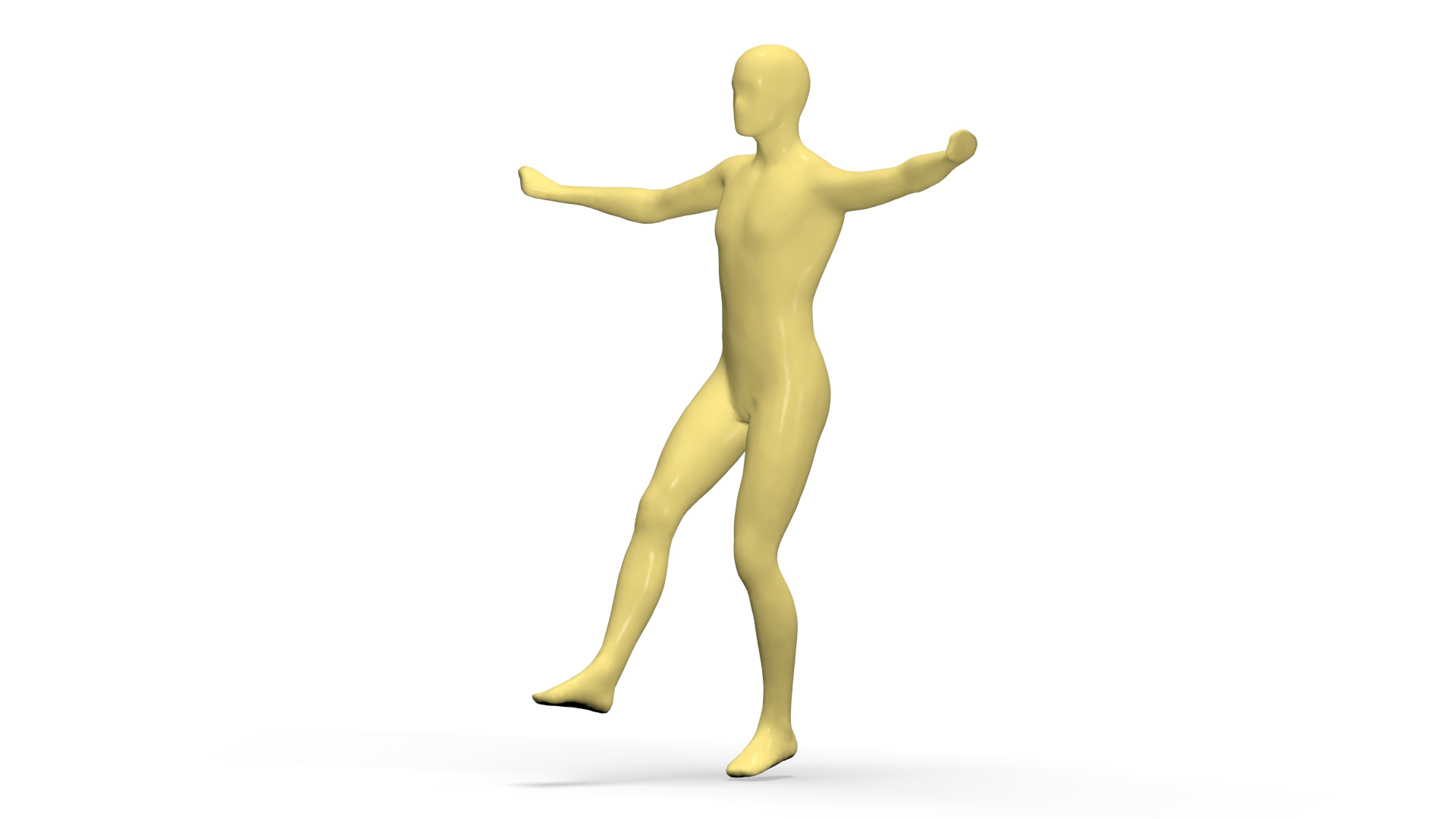}
	\end{overpic}
	\begin{overpic}
	[trim=21cm 0cm 14cm 0cm,clip,width=0.12\textwidth]{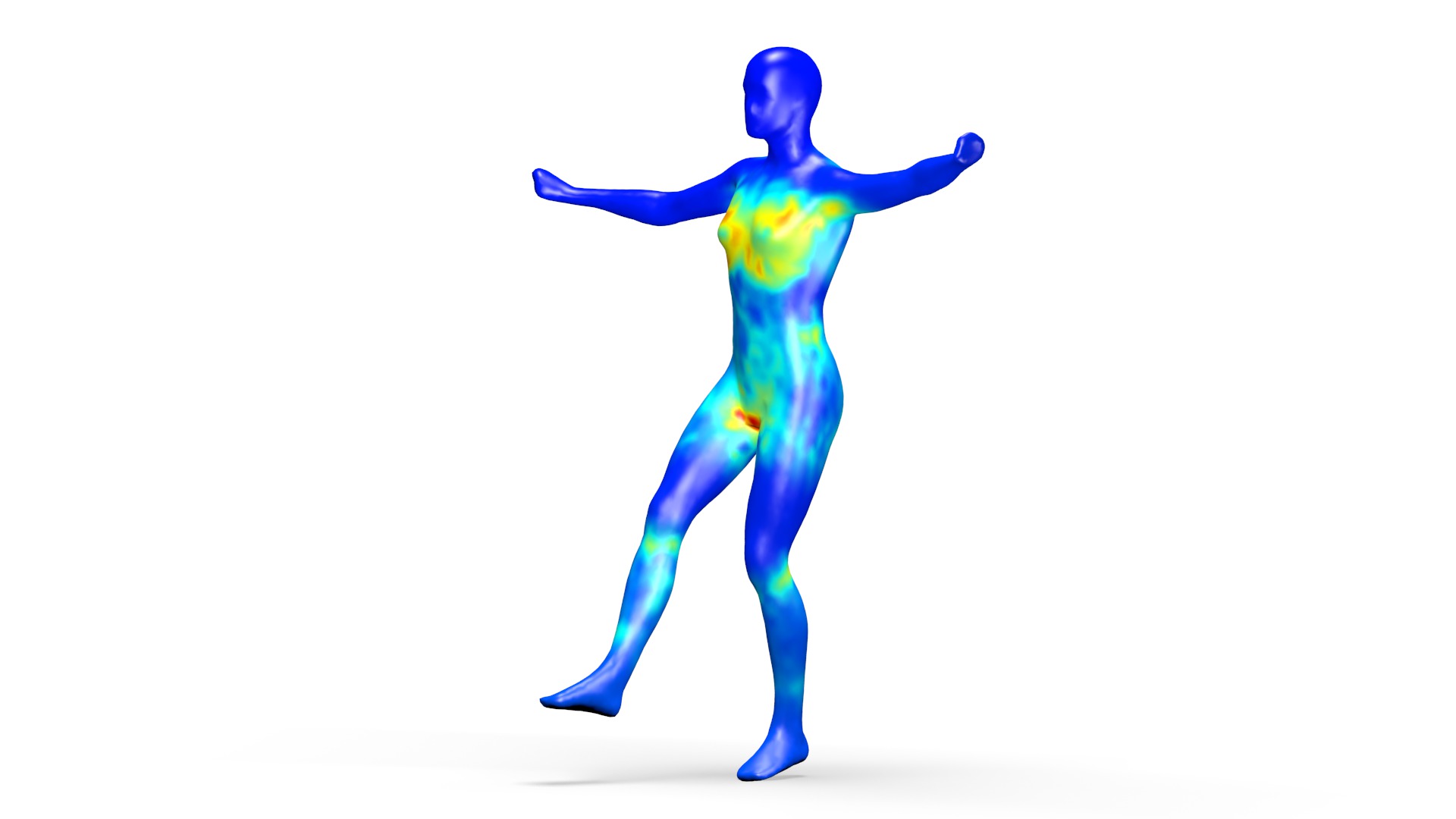}
	\put(36,-8){\fontsize{6}{8}\selectfont $1$}
	\end{overpic}
	\begin{overpic}
	[trim=21cm 0cm 14cm 0cm,clip,width=0.12\textwidth]{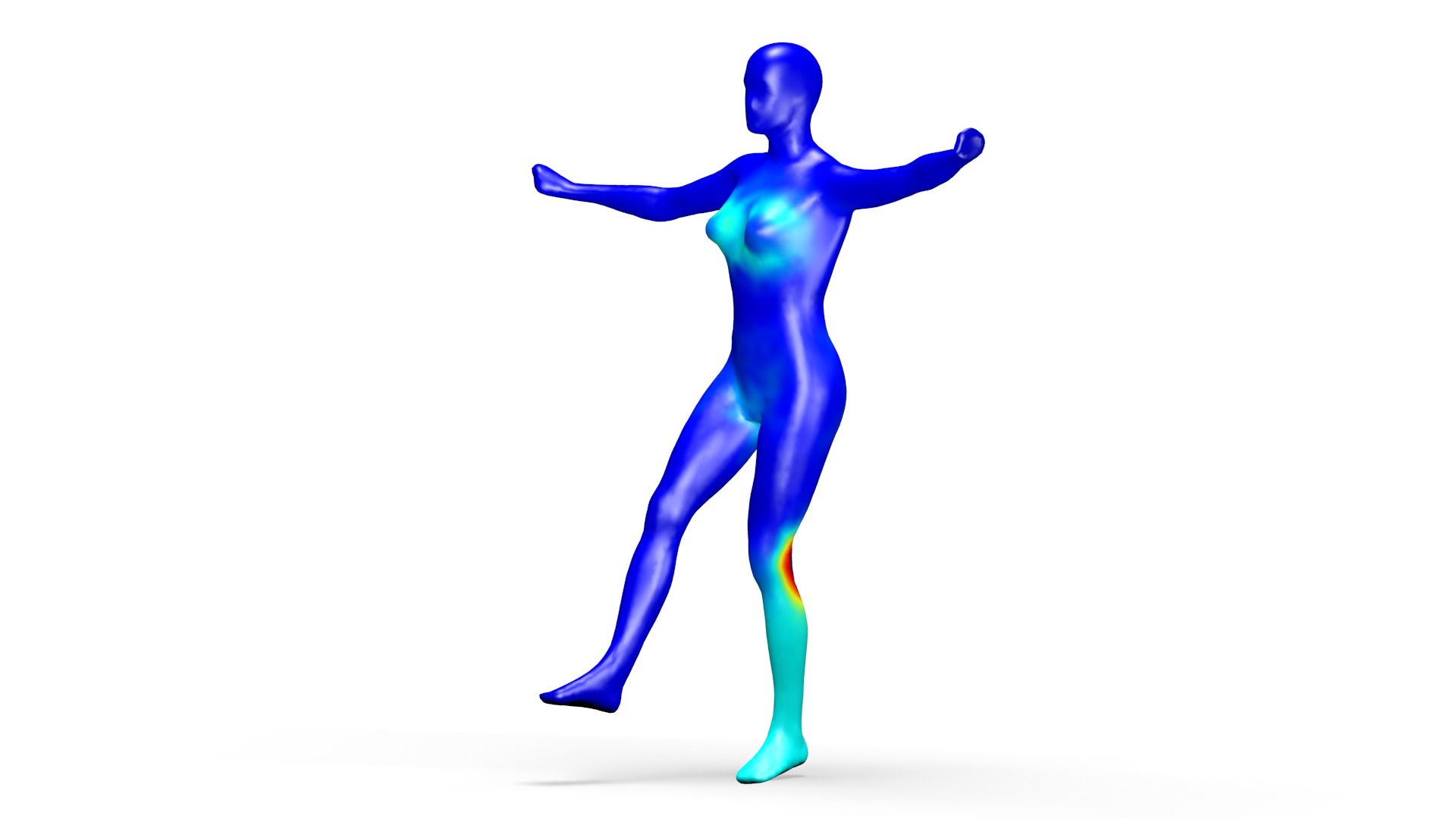}
	\put(33,-8){\fontsize{6}{8}\selectfont $20$}
	\end{overpic}
	\begin{overpic}
	[trim=21cm 0cm 14cm 0cm,clip,width=0.12\textwidth]{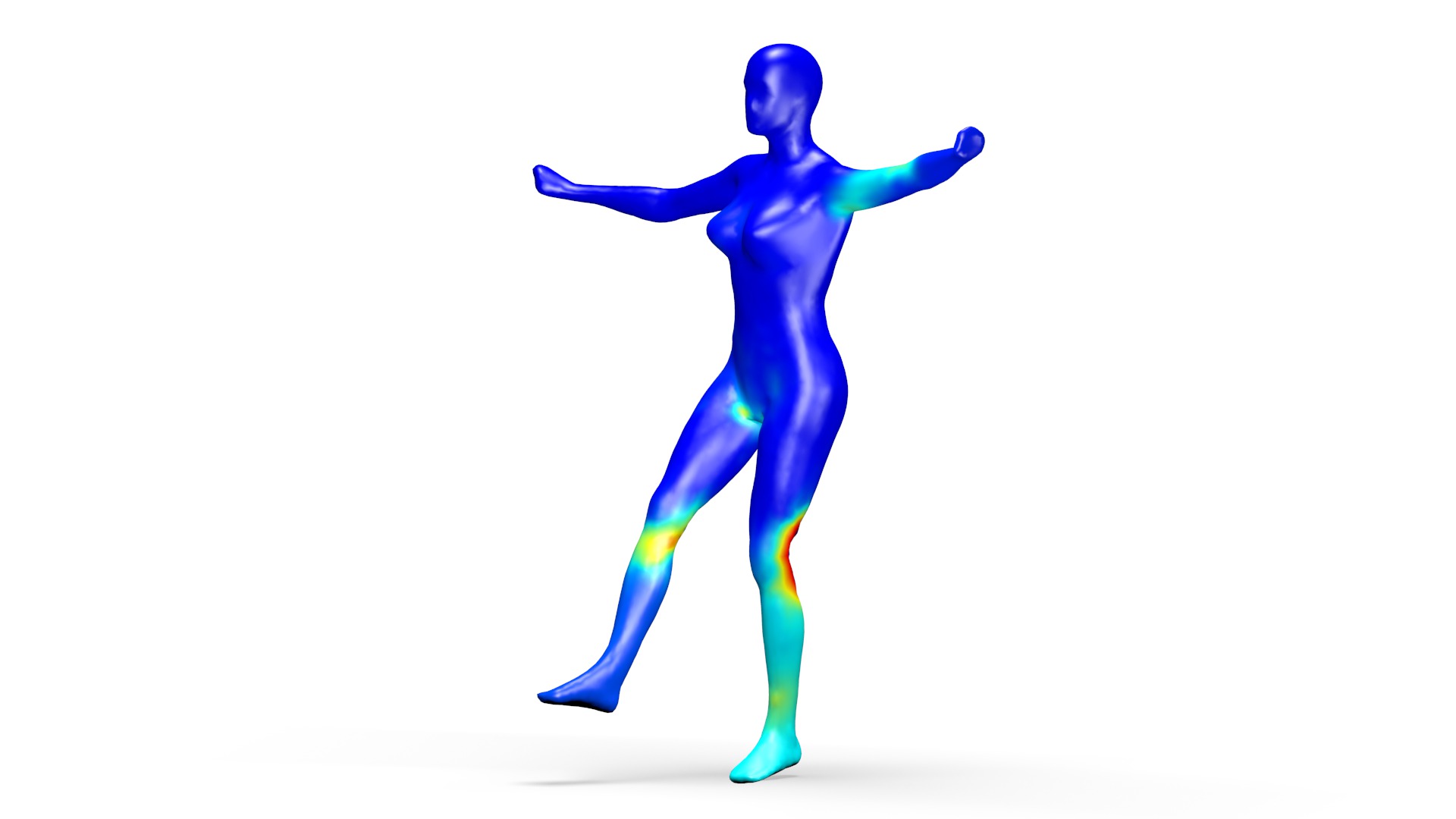}
	\put(30,-8){\fontsize{6}{8}\selectfont $100$}
	\end{overpic}
	\begin{overpic}
	[trim=21cm 0cm 14cm 0cm,clip,width=0.12\textwidth]{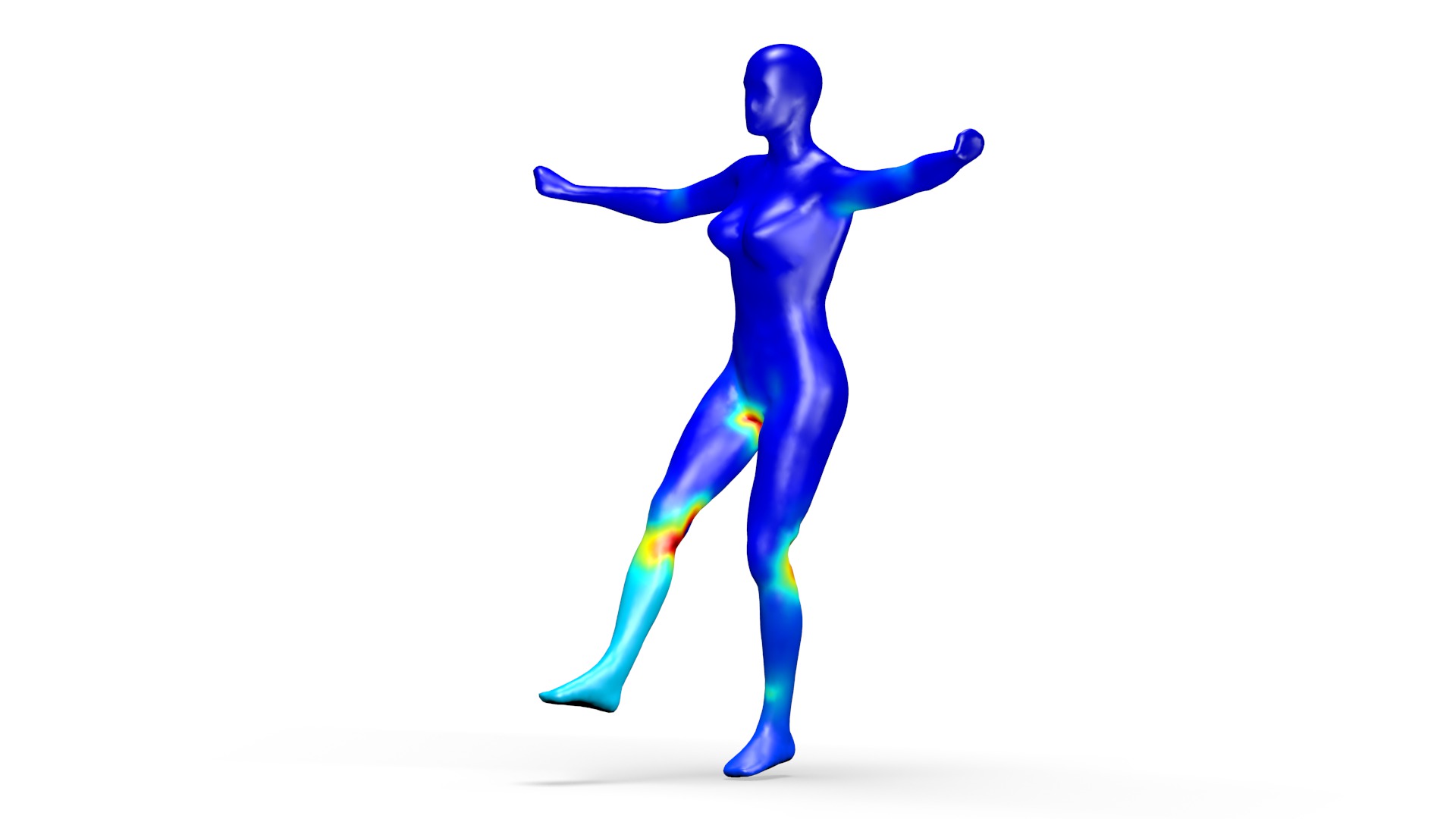}
	\put(30,-8){\fontsize{6}{8}\selectfont $200$}
	\end{overpic}
	\begin{overpic}
	[trim=15cm 0cm 20cm 0cm,clip,width=0.12\textwidth]{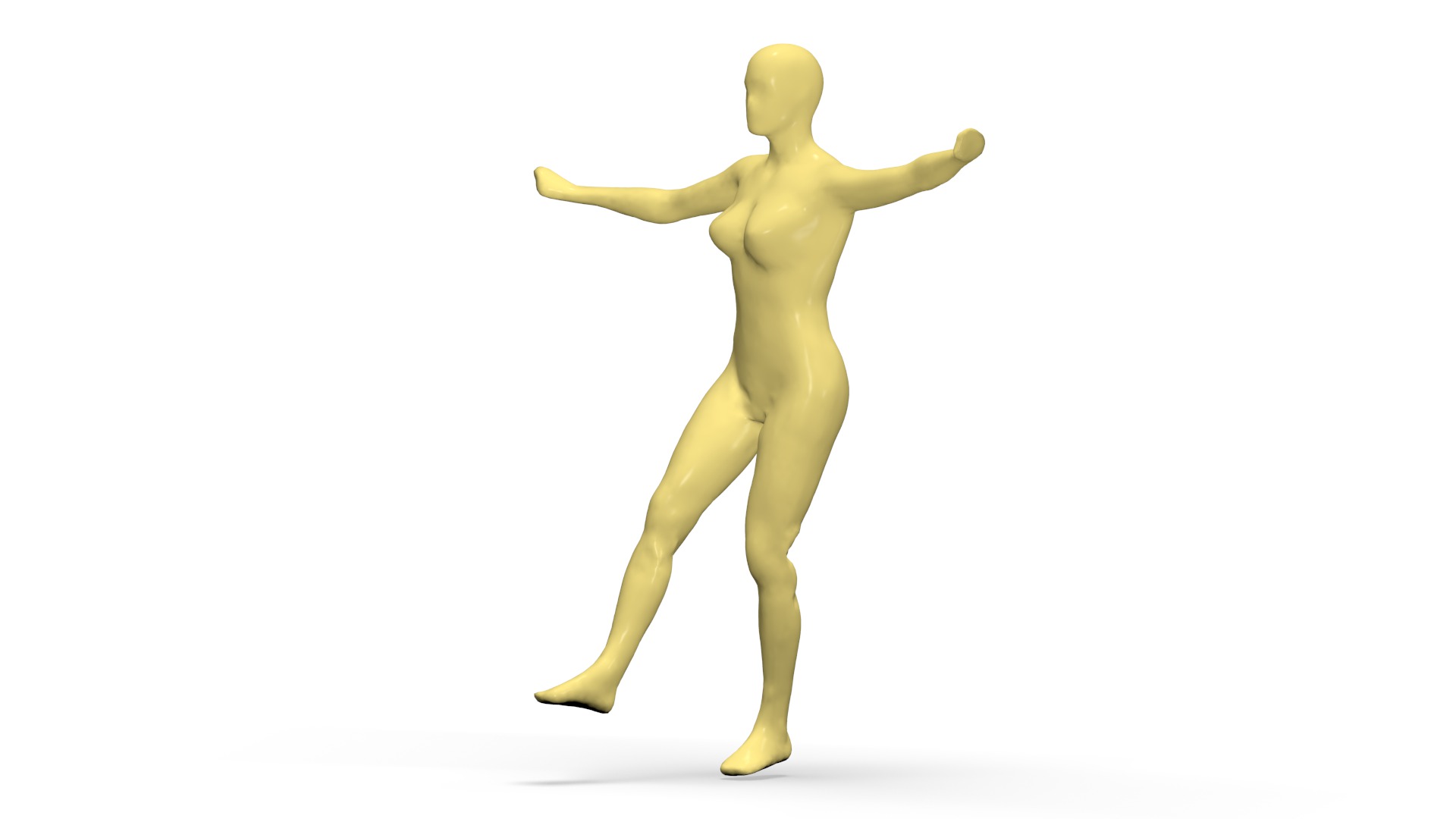}
	%
	%
	\put(-144,12){\begin{overpic}[width=0.2\textwidth,angle=180]{./figures/colorbar}\end{overpic}}
	\put(2,10){\fontsize{6}{8}\selectfont $0$}
	\put(2,77){\fontsize{6}{8}\selectfont $10^{1.6}$}
	%
	%
	\end{overpic}
\end{minipage} \\ \vspace{0.4cm}
%
%
\begin{minipage}{\textwidth}
	\begin{overpic}
	[trim=17cm 0cm 18cm 0cm,clip,width=0.12\columnwidth]{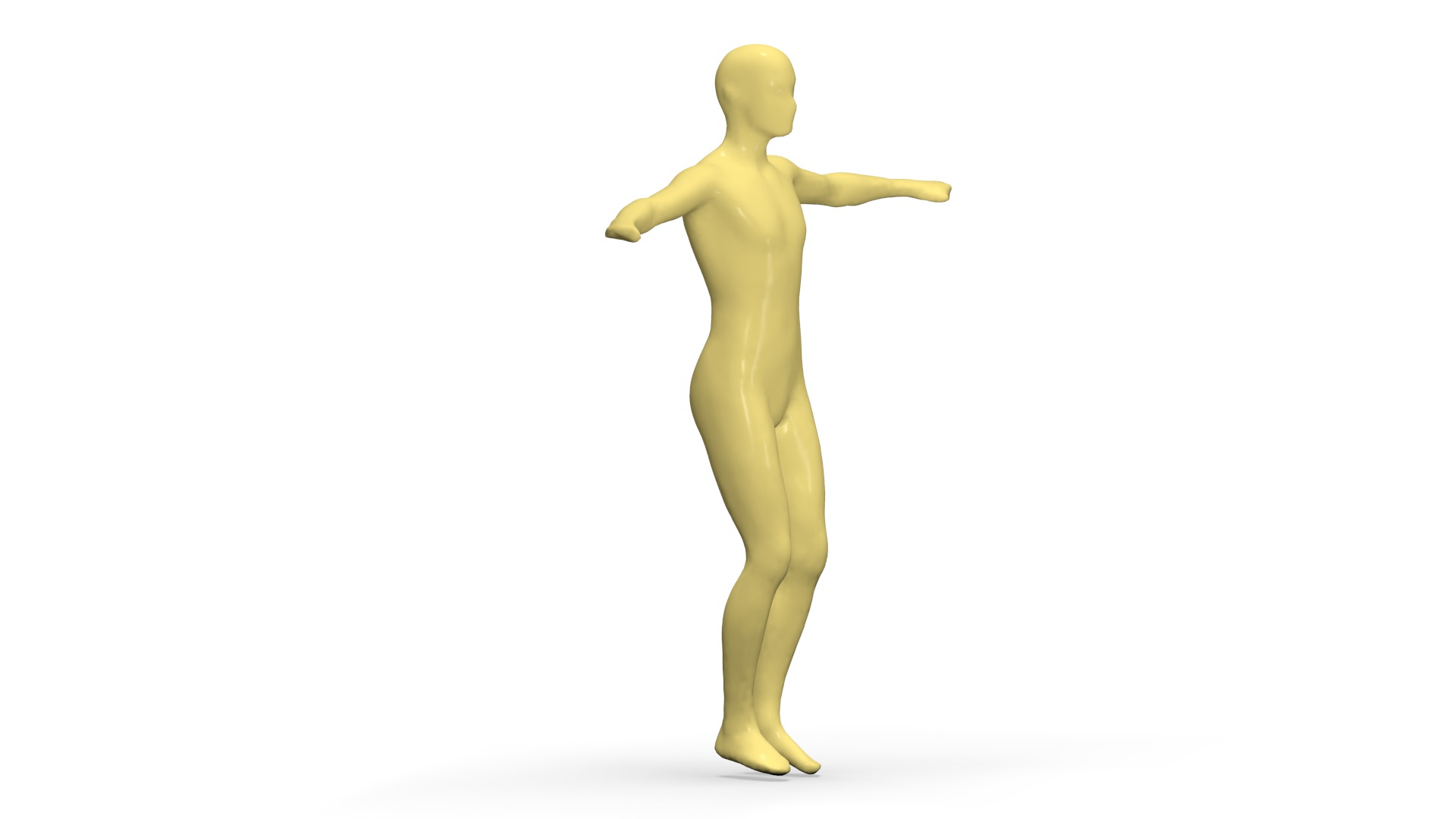}
	\end{overpic}
	\begin{overpic}
	[trim=17cm 0cm 18cm 0cm,clip,width=0.12\columnwidth]{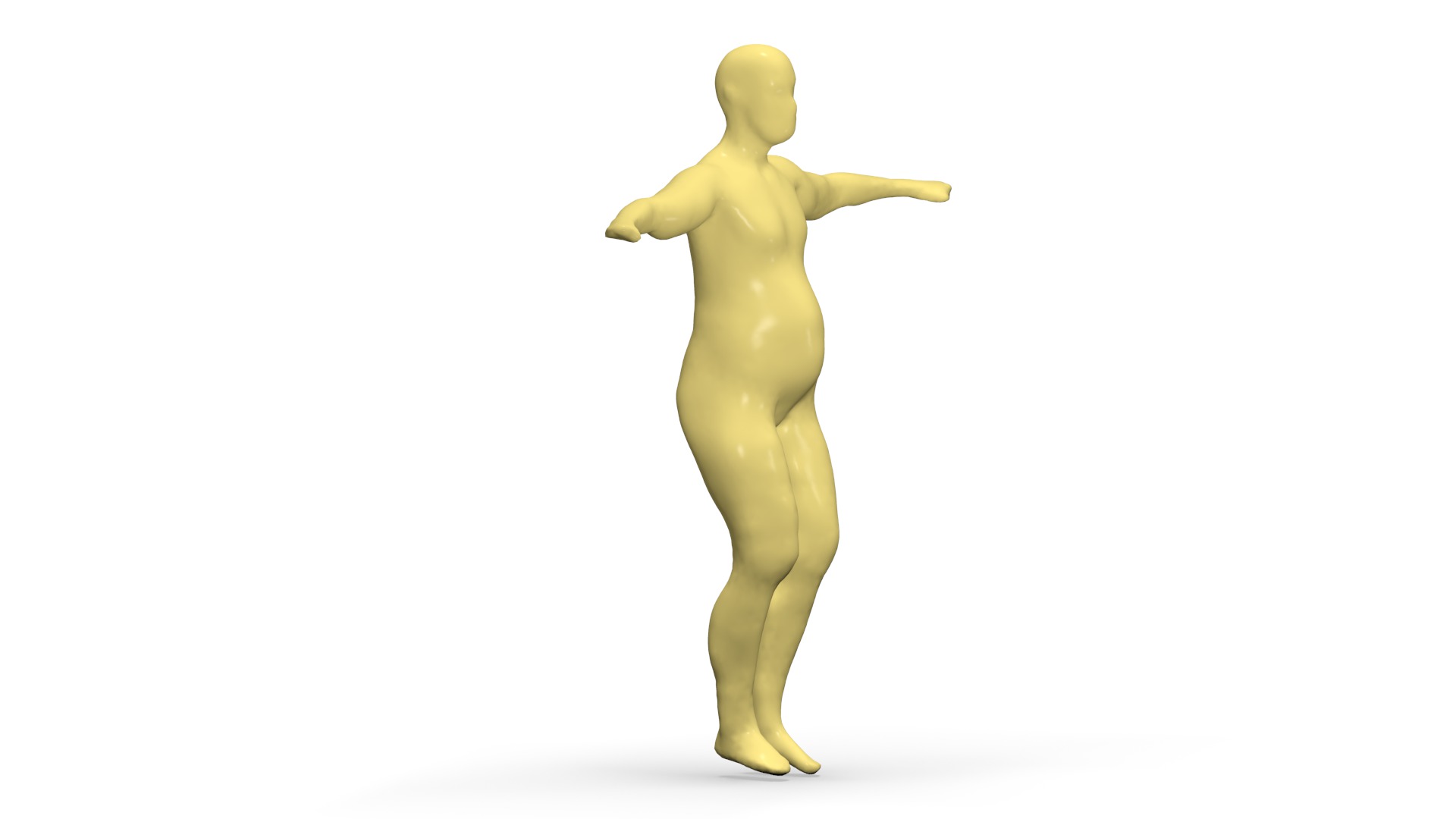}
	\end{overpic} 
	\begin{overpic}
	[trim=17cm 0cm 18cm 0cm,clip,width=0.12\textwidth]{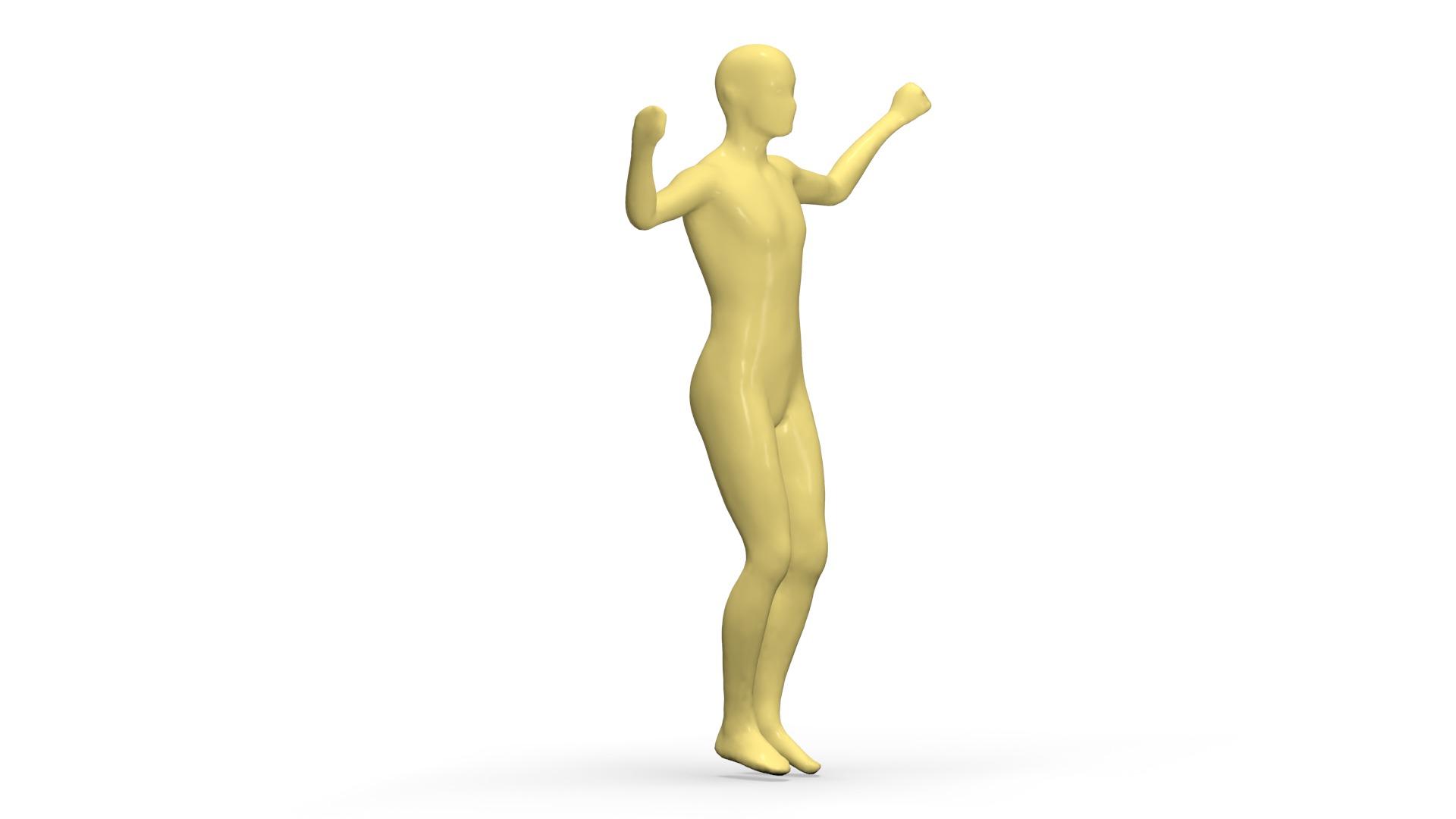}
	\end{overpic}
	\begin{overpic}
	[trim=21cm 0cm 14cm 0cm,clip,width=0.12\textwidth]{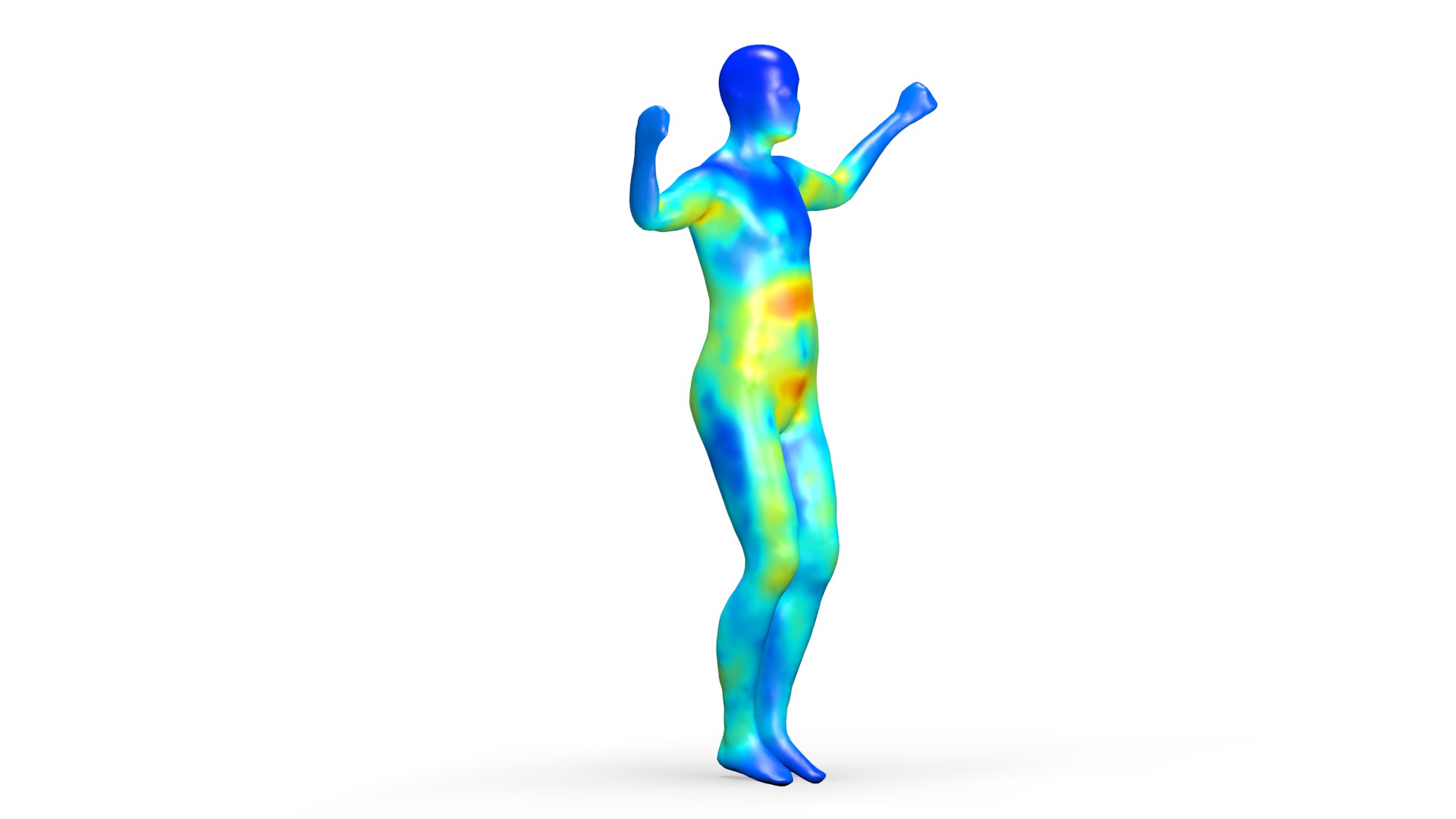}
	\put(33,-8){\fontsize{6}{8}\selectfont $10$}
	\end{overpic}
	\begin{overpic}
	[trim=21cm 0cm 14cm 0cm,clip,width=0.12\textwidth]{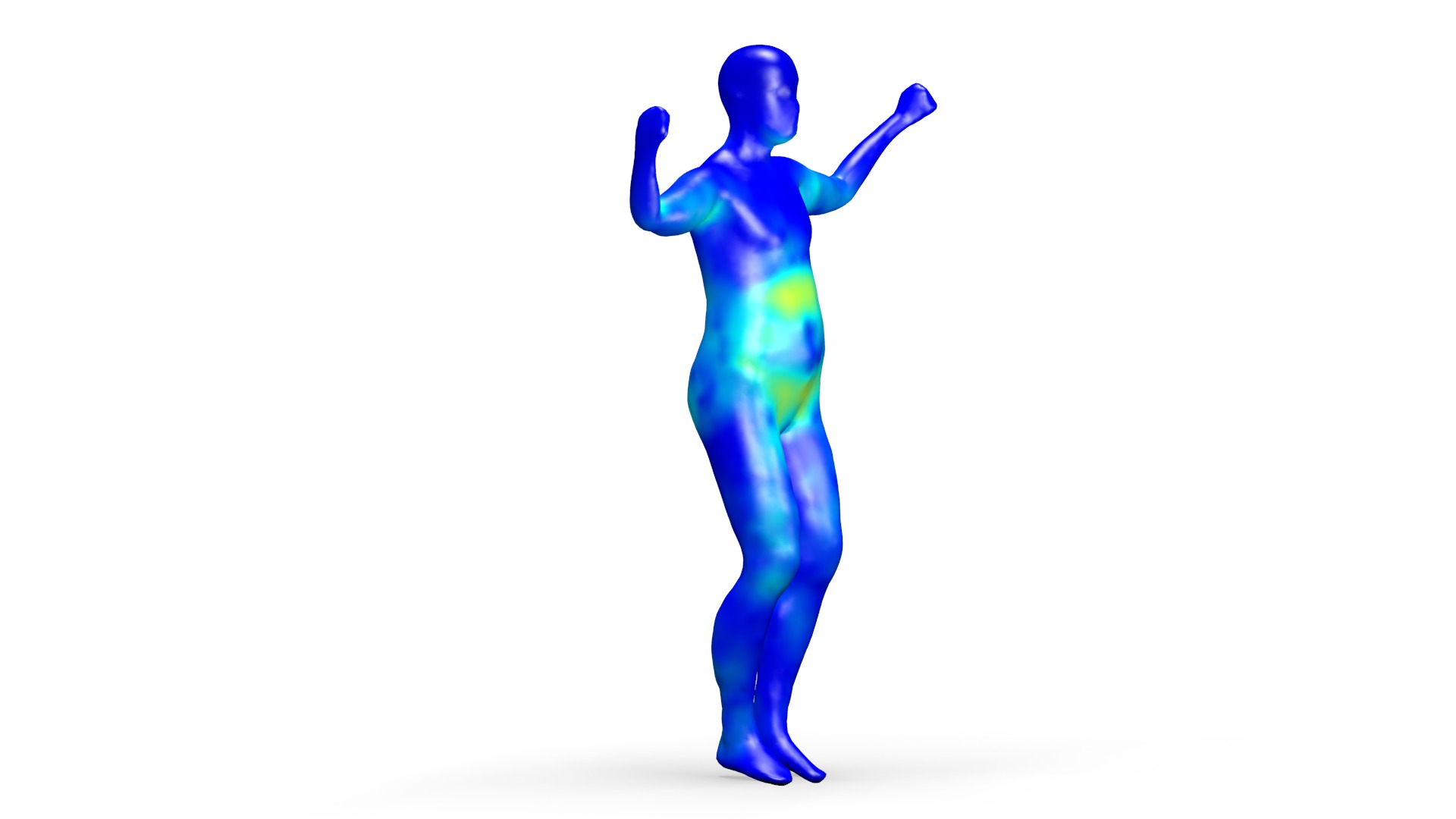}
	\put(33,-8){\fontsize{6}{8}\selectfont $50$}
	\end{overpic}
	\begin{overpic}
	[trim=21cm 0cm 14cm 0cm,clip,width=0.12\textwidth]{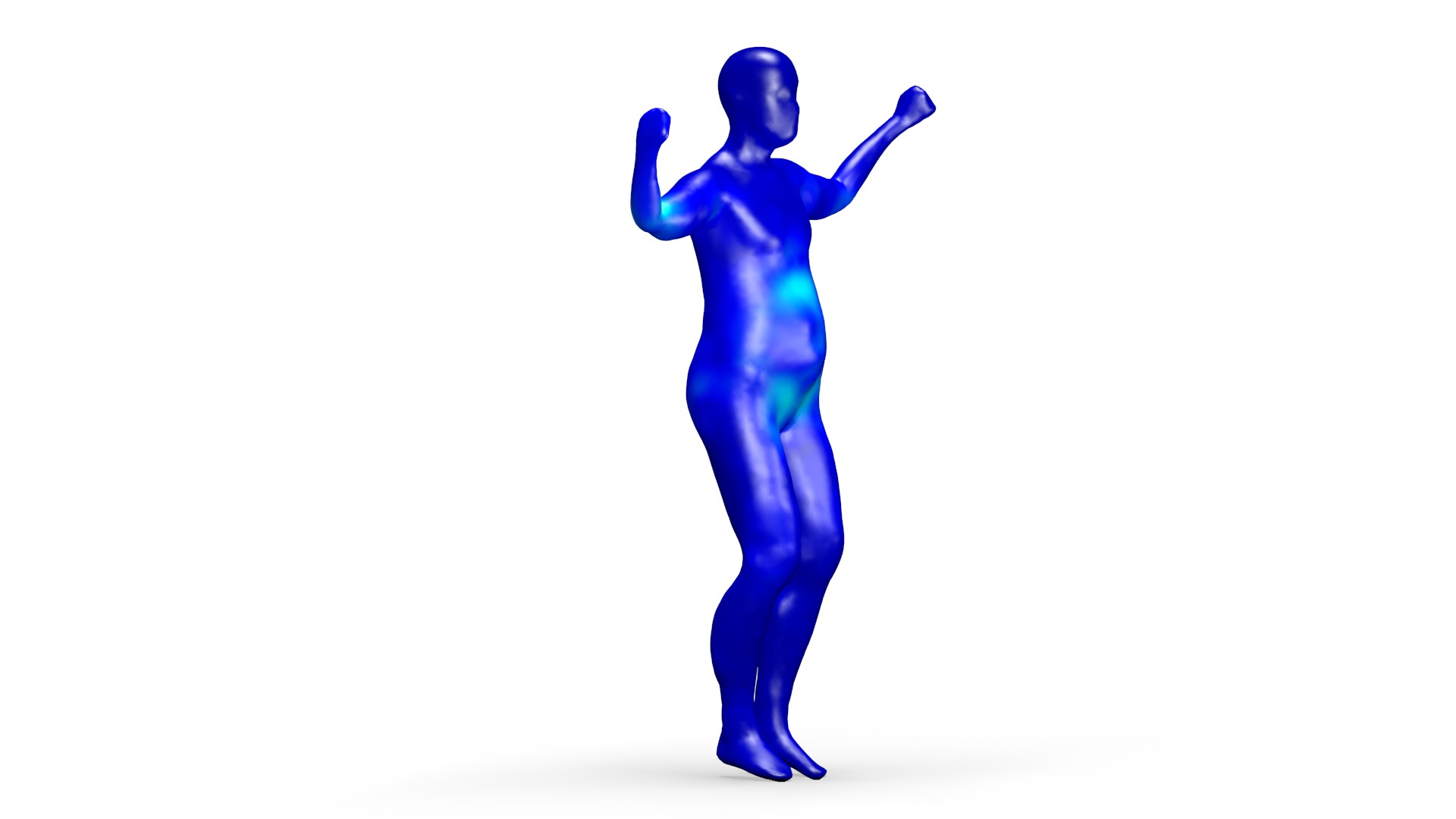}
	\put(30,-8){\fontsize{6}{8}\selectfont $100$}
	\end{overpic}
	\begin{overpic}
	[trim=21cm 0cm 14cm 0cm,clip,width=0.12\textwidth]{./figures/man_woman/man_manfat_manpose1_output_200iter}
	\put(30,-8){\fontsize{6}{8}\selectfont $200$}
	\end{overpic}
	\begin{overpic}
	[trim=15cm 0cm 20cm 0cm,clip,width=0.12\textwidth]{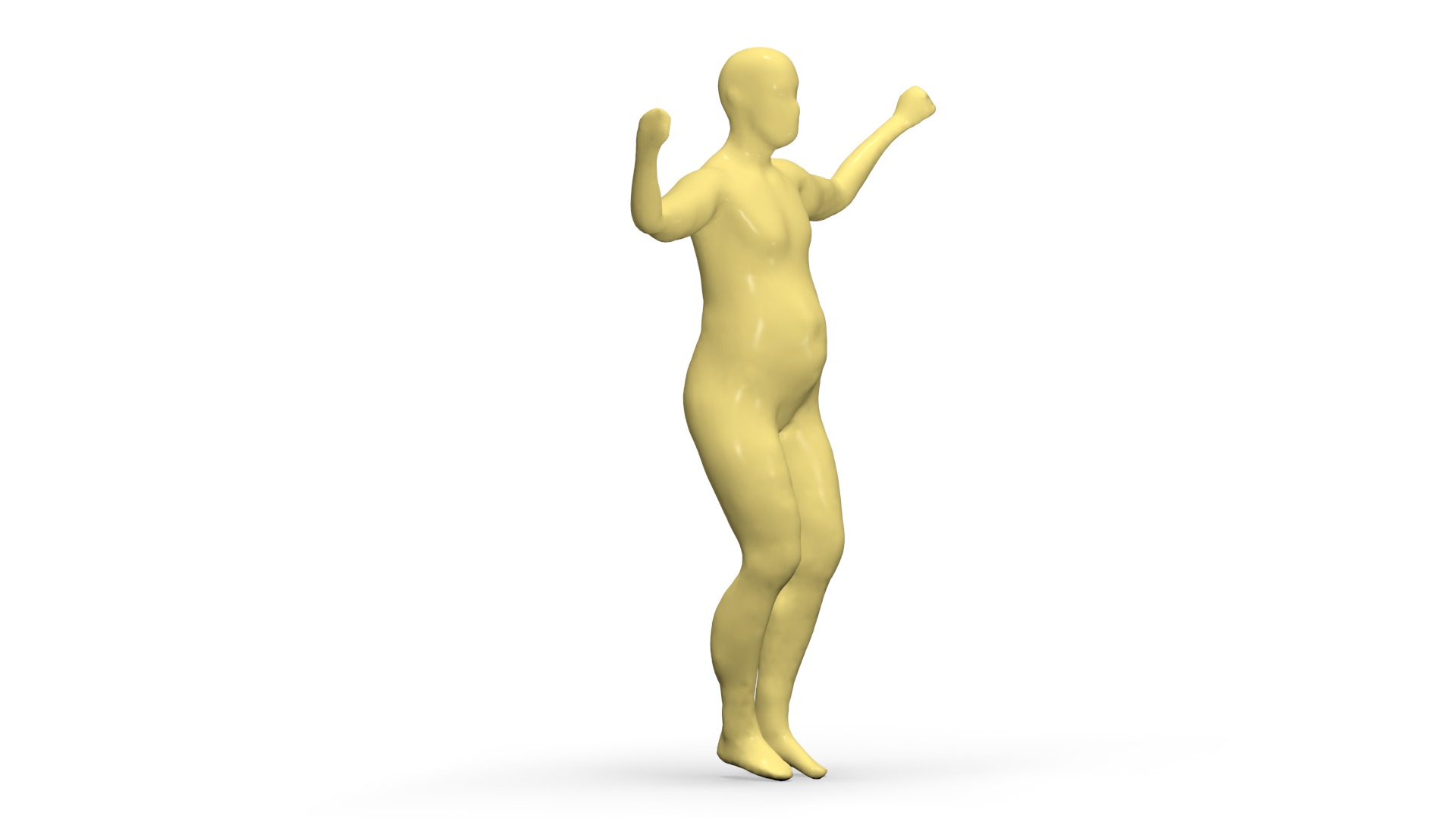}
	%
	%
	\put(-144,12){\begin{overpic}[width=0.2\textwidth,angle=180]{./figures/colorbar}\end{overpic}}
	\put(2,10){\fontsize{6}{8}\selectfont $0$}
	\put(2,77){\fontsize{6}{8}\selectfont $10^{1.2}$}
	%
	%
	\end{overpic}
\end{minipage} \\ \vspace{0.4cm}
%
%
\begin{minipage}{\textwidth}
	\begin{overpic}
	[trim=17cm 0cm 18cm 0cm,clip,width=0.12\columnwidth]{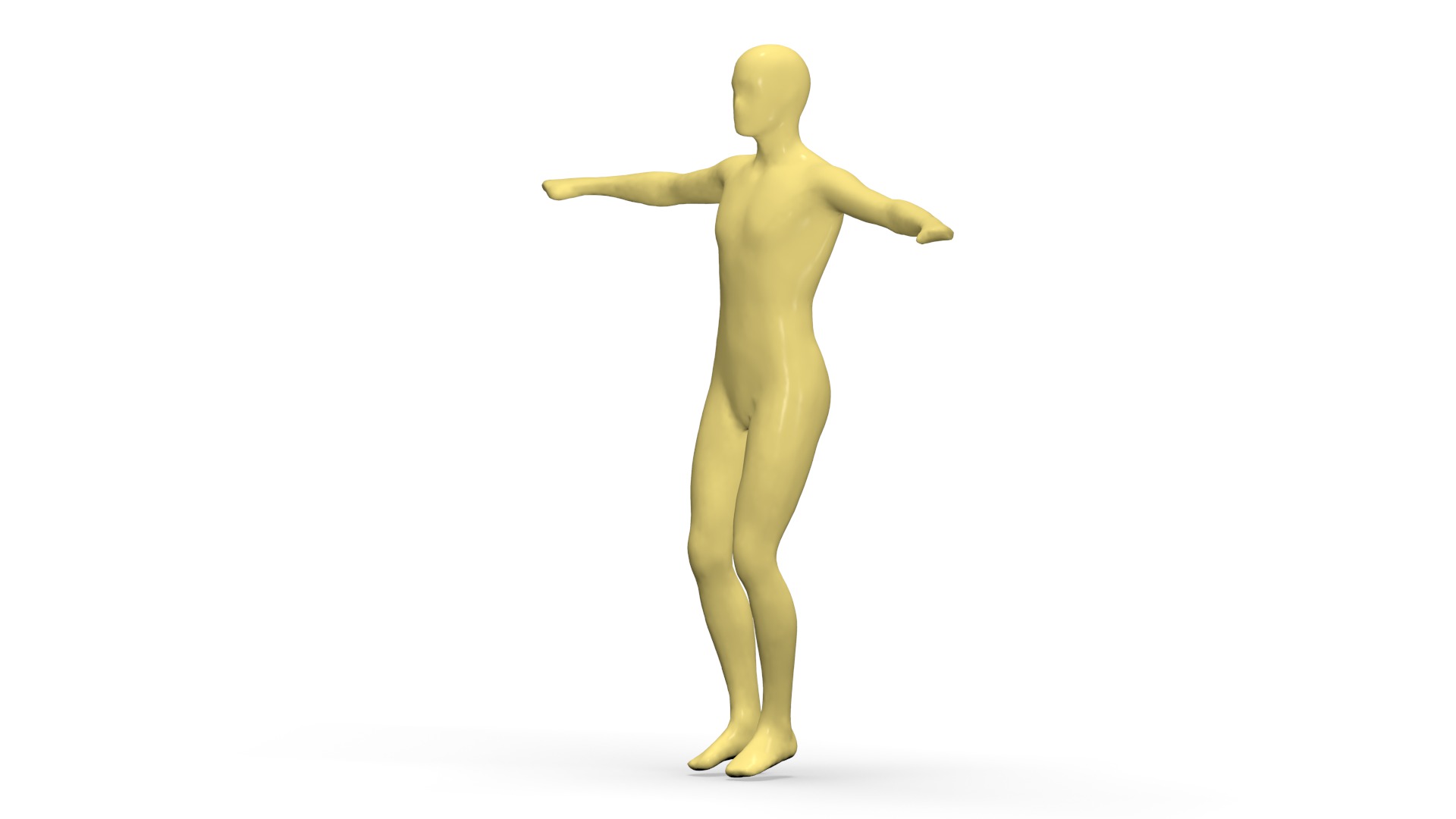}
	\put(45,-8){\fontsize{6}{8}\selectfont $A$}
	\end{overpic}
	\begin{overpic}
	[trim=17cm 0cm 18cm 0cm,clip,width=0.12\columnwidth]{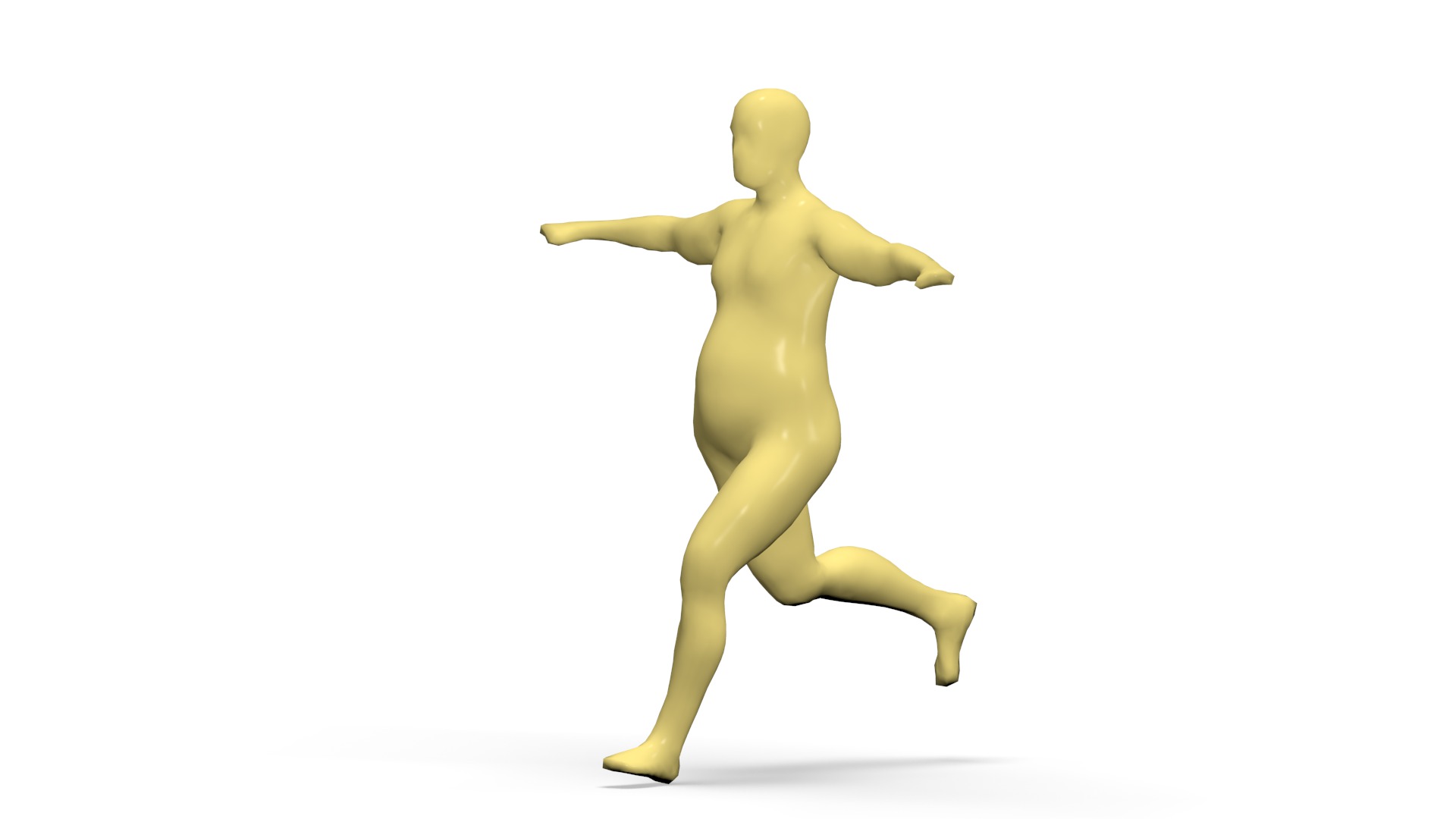}
	\put(45,-8){\fontsize{6}{8}\selectfont $B$}
	\end{overpic} 
	\begin{overpic}
	[trim=17cm 0cm 18cm 0cm,clip,width=0.12\textwidth]{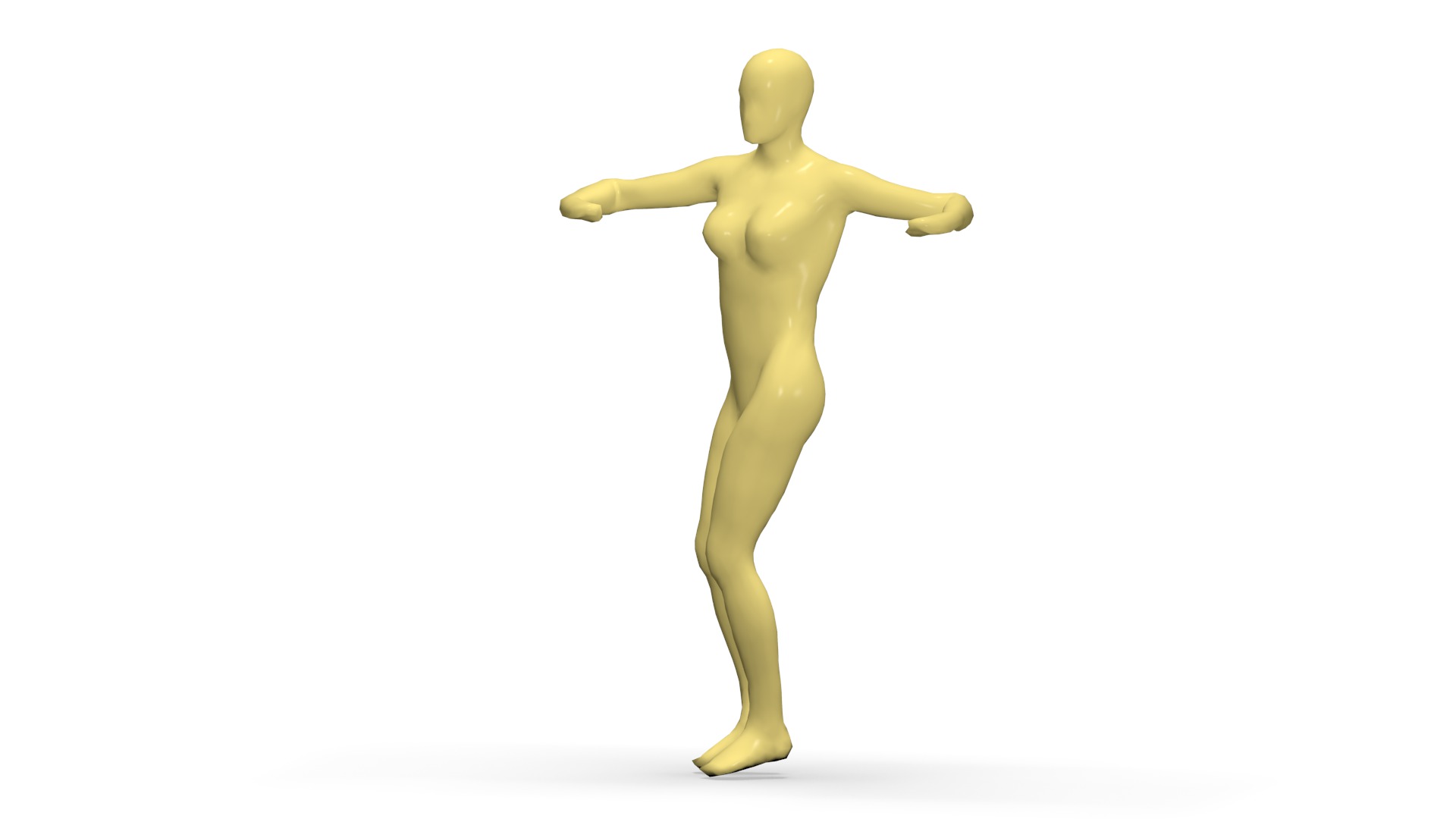}
	\put(45,-8){\fontsize{6}{8}\selectfont $C$}
	\end{overpic}
	\begin{overpic}
	[trim=21cm 0cm 14cm 0cm,clip,width=0.12\textwidth]{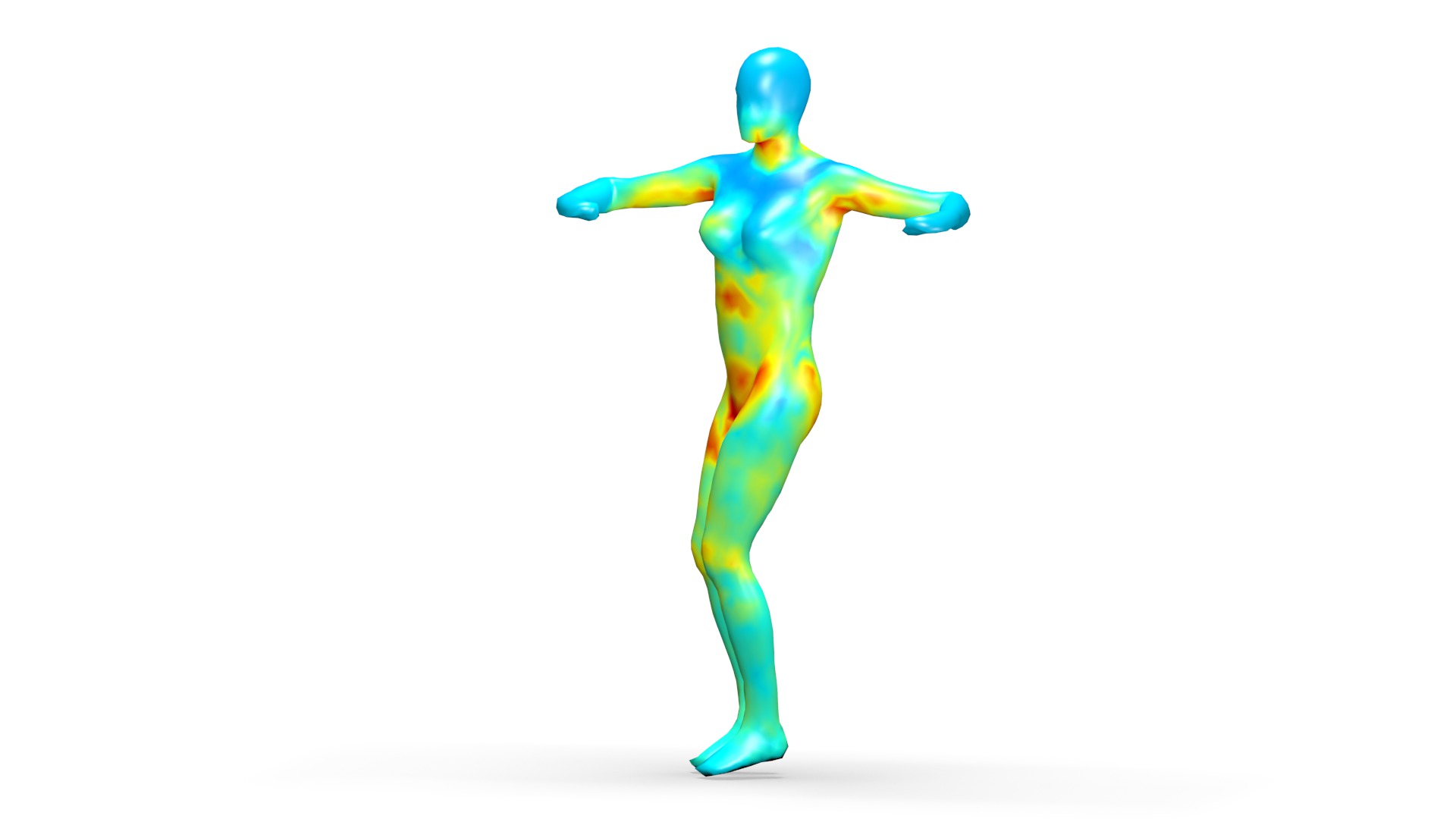}
	\put(35,-8){\fontsize{6}{8}\selectfont $1$}
	\end{overpic}
	\begin{overpic}
	[trim=21cm 0cm 14cm 0cm,clip,width=0.12\textwidth]{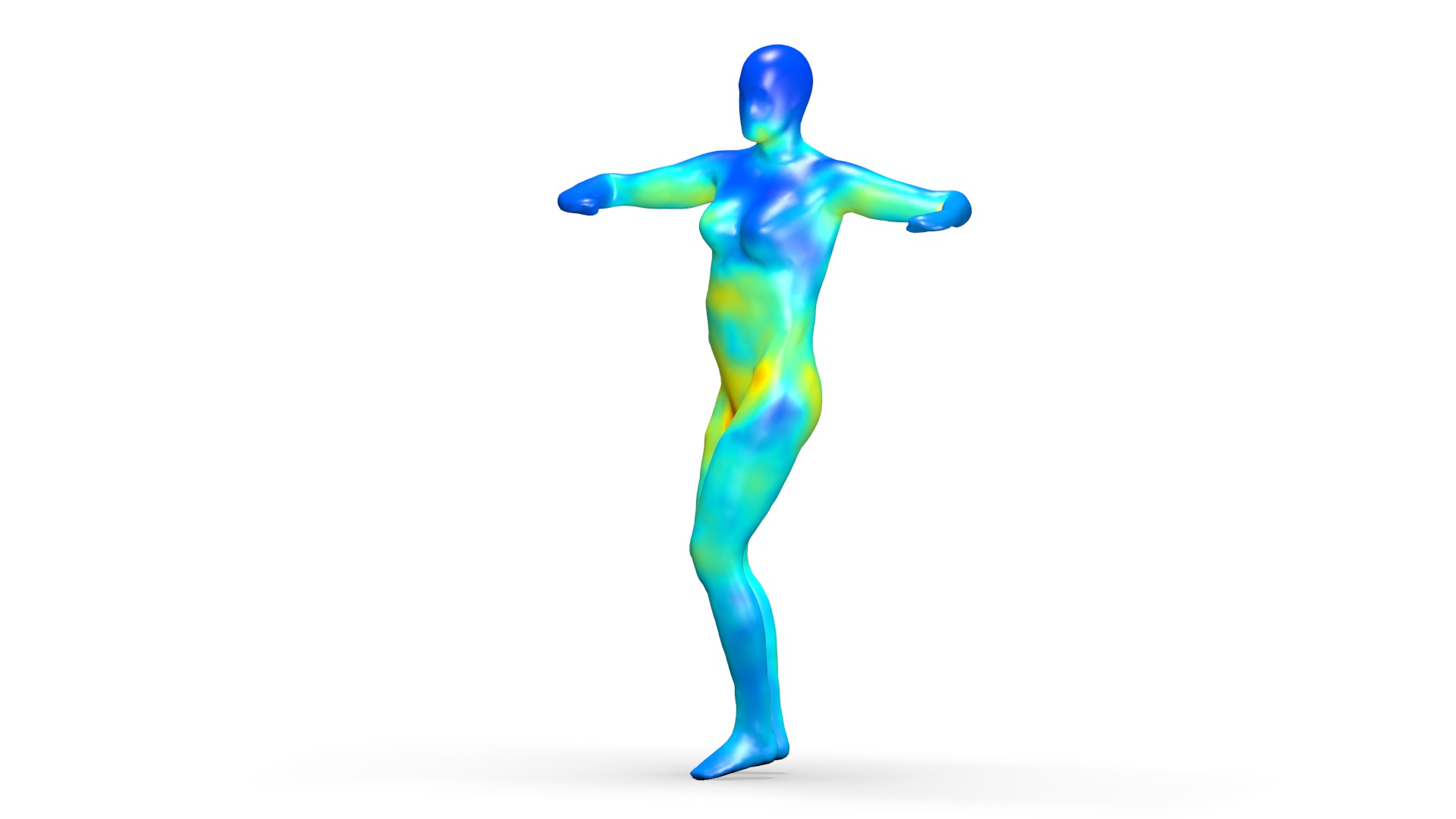}
	\put(33,-8){\fontsize{6}{8}\selectfont $10$}
	\end{overpic}
	\begin{overpic}
	[trim=21cm 0cm 14cm 0cm,clip,width=0.12\textwidth]{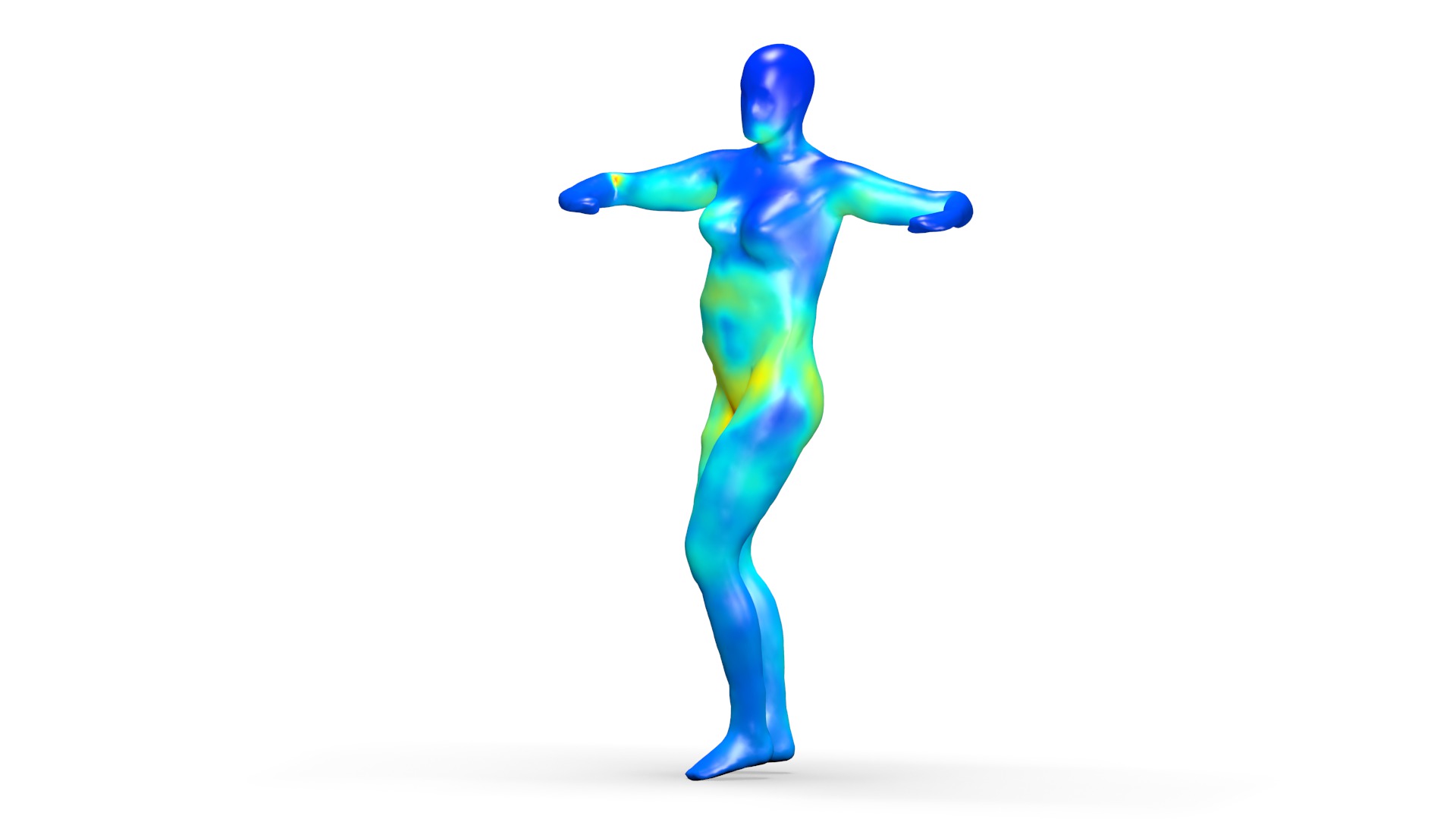}
	\put(33,-8){\fontsize{6}{8}\selectfont $20$}
	\end{overpic}
	\begin{overpic}
	[trim=21cm 0cm 14cm 0cm,clip,width=0.12\textwidth]{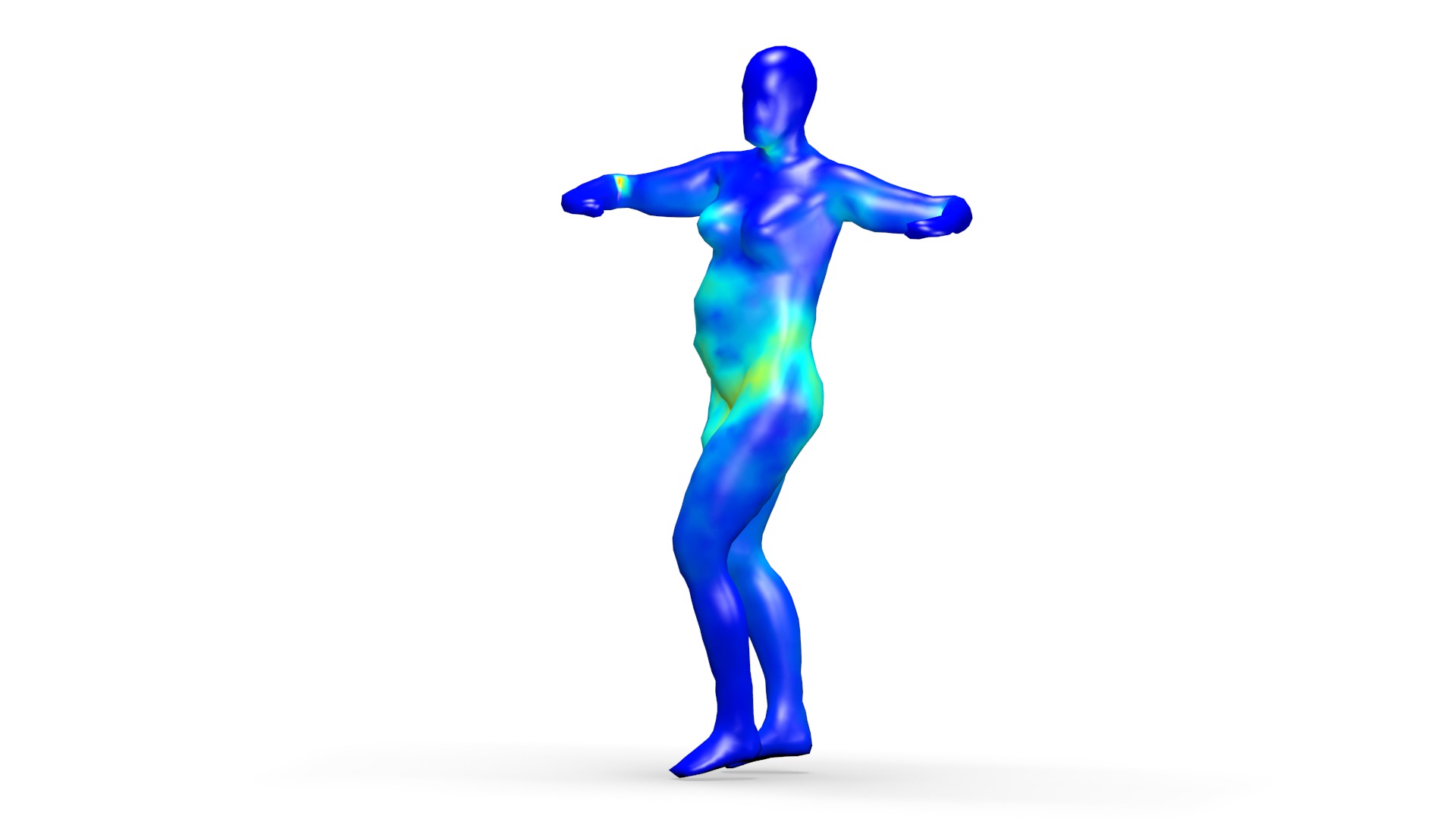}
	\put(33,-8){\fontsize{6}{8}\selectfont $50$}
	\end{overpic}
	\begin{overpic}
	[trim=15cm 0cm 20cm 0cm,clip,width=0.12\textwidth]{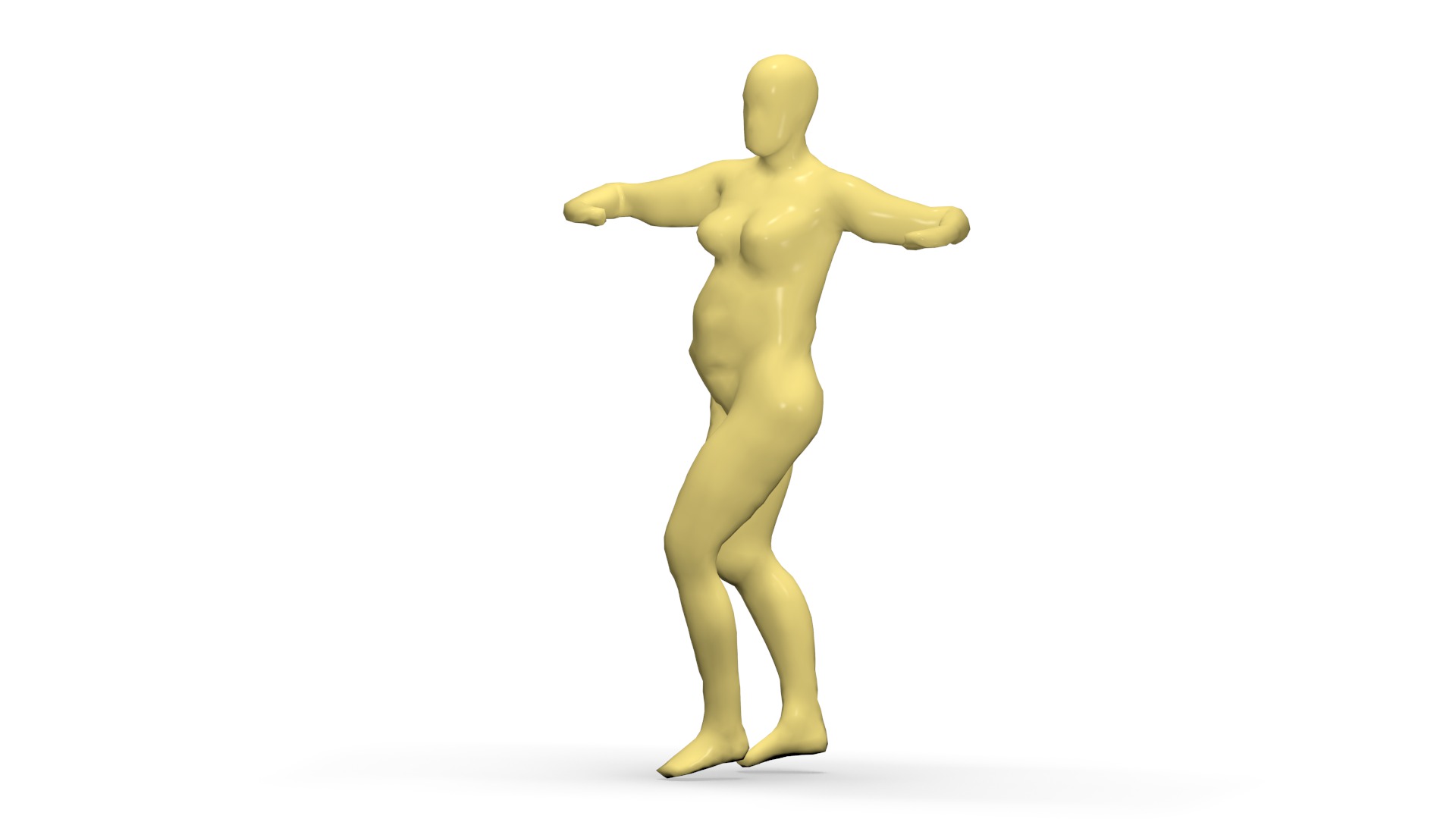}
	\put(47,-8){\fontsize{6}{8}\selectfont $X$}
	%
	\put(-144,12){\begin{overpic}[width=0.2\textwidth,angle=180]{./figures/colorbar}\end{overpic}}
	\put(2,10){\fontsize{6}{8}\selectfont $0$}
	\put(2,77){\fontsize{6}{8}\selectfont $10^{1.6}$}
	\put(-440,-10){\color{gray}\line(1,0){535}} 
	\put(-440,102){\color{gray}\line(1,0){535}} 
	\put(-440,-10){\color{gray}\line(0,1){112}} 
	\put(95,-10){\color{gray}\line(0,1){112}}   
	\put(-440,107){\color{gray}\line(1,0){535}} 
	\put(-440,219){\color{gray}\line(1,0){535}} 
	\put(-440,107){\color{gray}\line(0,1){112}} 
	\put(95,107){\color{gray}\line(0,1){112}}   
	\put(-440,224){\color{gray}\line(1,0){535}} 
	\put(-440,336){\color{gray}\line(1,0){535}} 
	\put(-440,224){\color{gray}\line(0,1){112}} 
	\put(95,224){\color{gray}\line(0,1){112}}   
	\put(-440,341){\color{gray}\line(1,0){535}} 
	\put(-440,435){\color{gray}\line(1,0){535}} 
	\put(-440,341){\color{gray}\line(0,1){94}} 
	\put(95,341){\color{gray}\line(0,1){94}}   
	\end{overpic}
\end{minipage} \\ \vspace{0.7cm}
\caption{Synthesis of shape analogies: given $A$, $B$, and $C$ (leftmost columns), find $X$ (rightmost column). 
Columns 4-7 show a few intermediate steps of the optimization method described in this paper. Colors represent the vertex-wise contribution to the energy; hotter colors correspond to higher values (log scale). }
\label{fig:analogies2}
\end{figure*}

\textbf{Shape-from-difference operator } amounts to computing an unknown shape $X$ from a shape $C$, given a difference operator between the analogous shapes $A$ and $B$, as described in Section~\ref{sec:prob}. 
Figure \ref{fig:analogies1} presents some shape analogies synthesized using our algorithm. 
Figures \ref{fig:analogies2} and \ref{fig:teaser} additionally shows the intermediate step of the optimization, where, 
%
%
%
for visualization purposes, we plot the vertex-wise contribution to the energy $\mathcal{E}_\mathrm{dif}(\bb{X})$,  
\begin{equation}
\epsilon_{i}(\bb{X}) = \lambda \sum_{j} \lvert e_{ij}  \rvert + \lvert e_{ji} \rvert + (1 - \lambda) \sum_{j} \lvert e'_{ij} \rvert + \lvert e'_{ji} \lvert, 
%
\end{equation}
where $\bb{E} = \bb{A}_{C}^{-1}\bb{A}(\bb{\ell}(\bb{X})) \bb{G} - \bb{G} \bb{V}_{A,B} $ and 
$\bb{E}' = \bb{W}_{C}^{\dagger}\bb{W}(\bb{\ell}(\bb{X})) \bb{G} - \bb{G} \bb{R}_{A,B} $.  
Note that examples in the first two rows of Figure~\ref{fig:analogies2} include only intrinsic differences (or `style') of shapes $A, B$ (man vs woman or thin man vs fat man) that are transferred to a different pose $C$, resulting in $X$ being a shape of style of $B$ in the pose of $C$.  
In the last row, however, the difference between shapes $A$ and $B$ includes both `style' (thin vs fat man) and pose differences (standing vs running). 
Shape $C$ is a woman in a standing pose, and the synthesized analogy $X$ is a fat woman in running pose (though the pose of $X$ is attenuated compared to $B$). 
Obviously, since the operators in our problem are intrinsic, if the pose transformation were a perfect isometry, the pose of $C$ would not change. However, the real pose transformations result in local non-isometric deformations around the knee joint, affecting the metric and the resulting difference operators. Consequently, our optimization tries to account for such a difference by bending the leg of the woman $X$. 
This result is consistent with the experiments of \cite{Rustamov2013}, who were able to capture extrinsic shape transformations (poses) by means of intrinsic shape difference operators.

%


\begin{figure}[b!]
\centering
%
%
\begin{minipage}{\columnwidth}
	\begin{overpic}
	[trim=16cm 0cm 19cm 0cm,clip,width=0.24\columnwidth]{./figures/man_woman/man}
	\end{overpic}
	\begin{overpic}
	[trim=16cm 0cm 19cm 0cm,clip,width=0.24\columnwidth]{./figures/man_woman/woman}
	\end{overpic}
	\begin{overpic}
	[trim=15cm 0cm 20cm 0cm,clip,width=0.24\columnwidth]{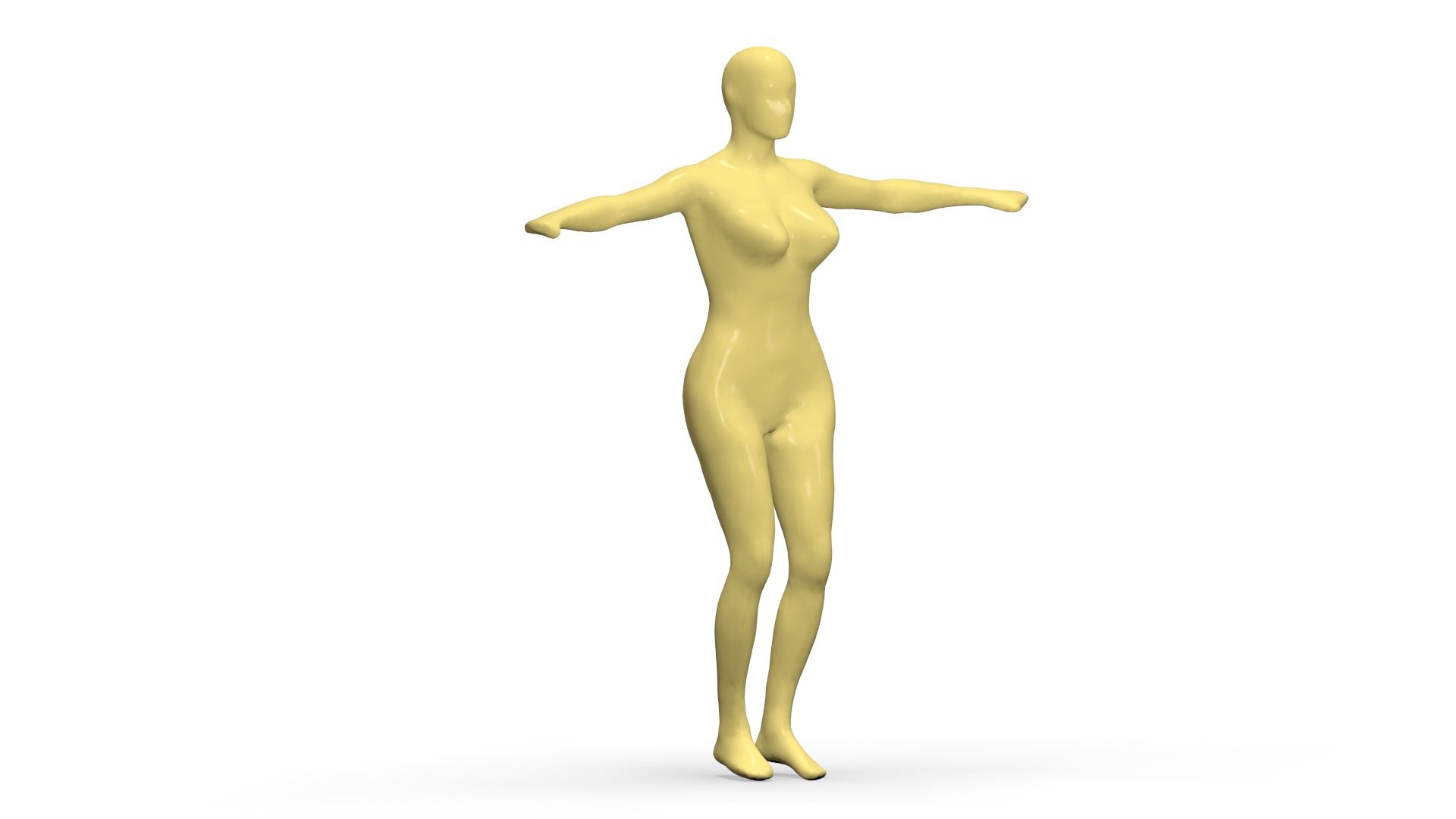}
	\end{overpic}
	\begin{overpic}
	[trim=15cm 0cm 20cm 0cm,clip,width=0.24\columnwidth]{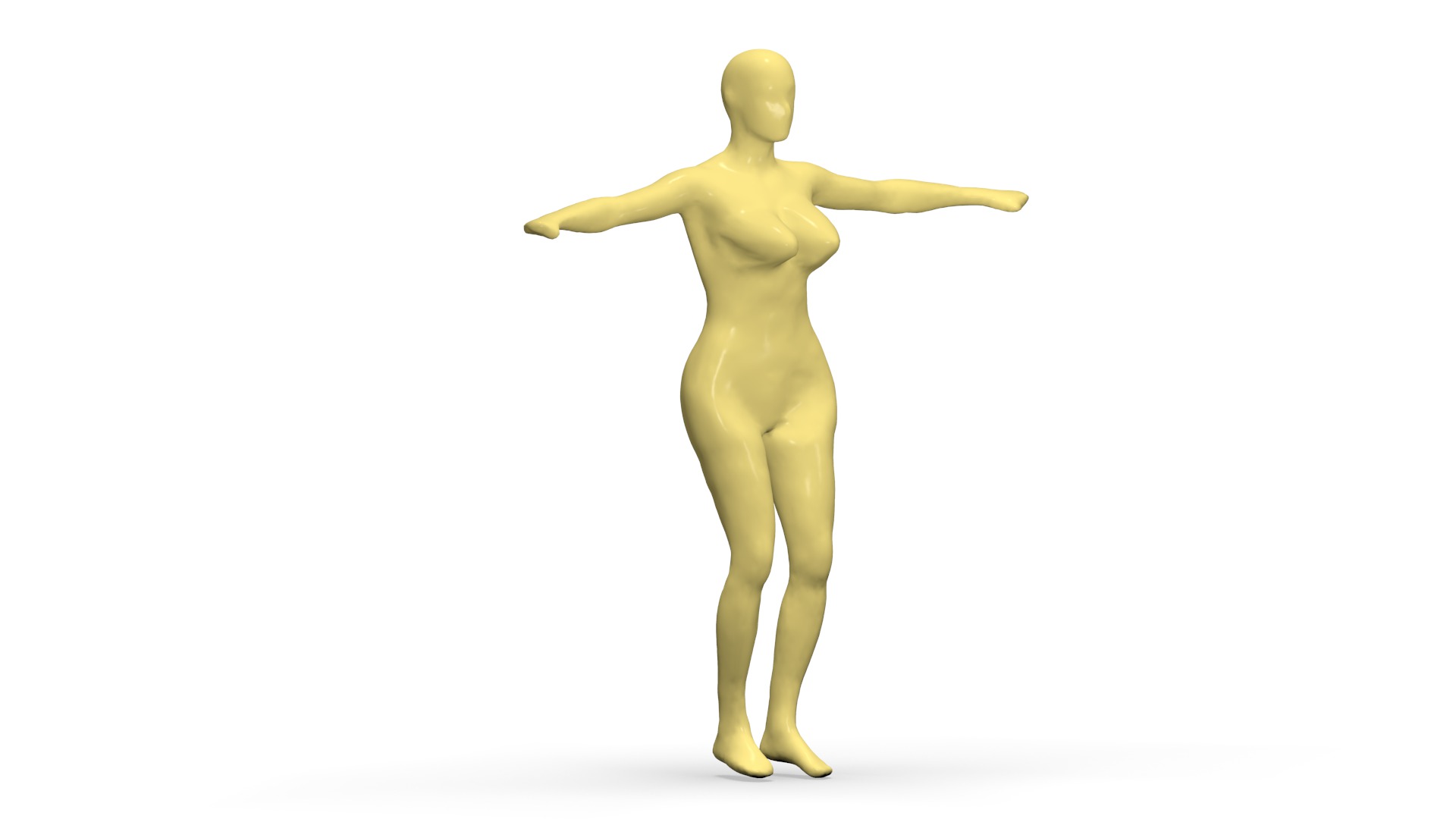}
	\end{overpic}
\end{minipage} \\ 
%
%
\begin{minipage}{\columnwidth}
	\begin{overpic}
	[trim=15cm 0cm 20cm 0cm,clip,width=0.24\columnwidth]{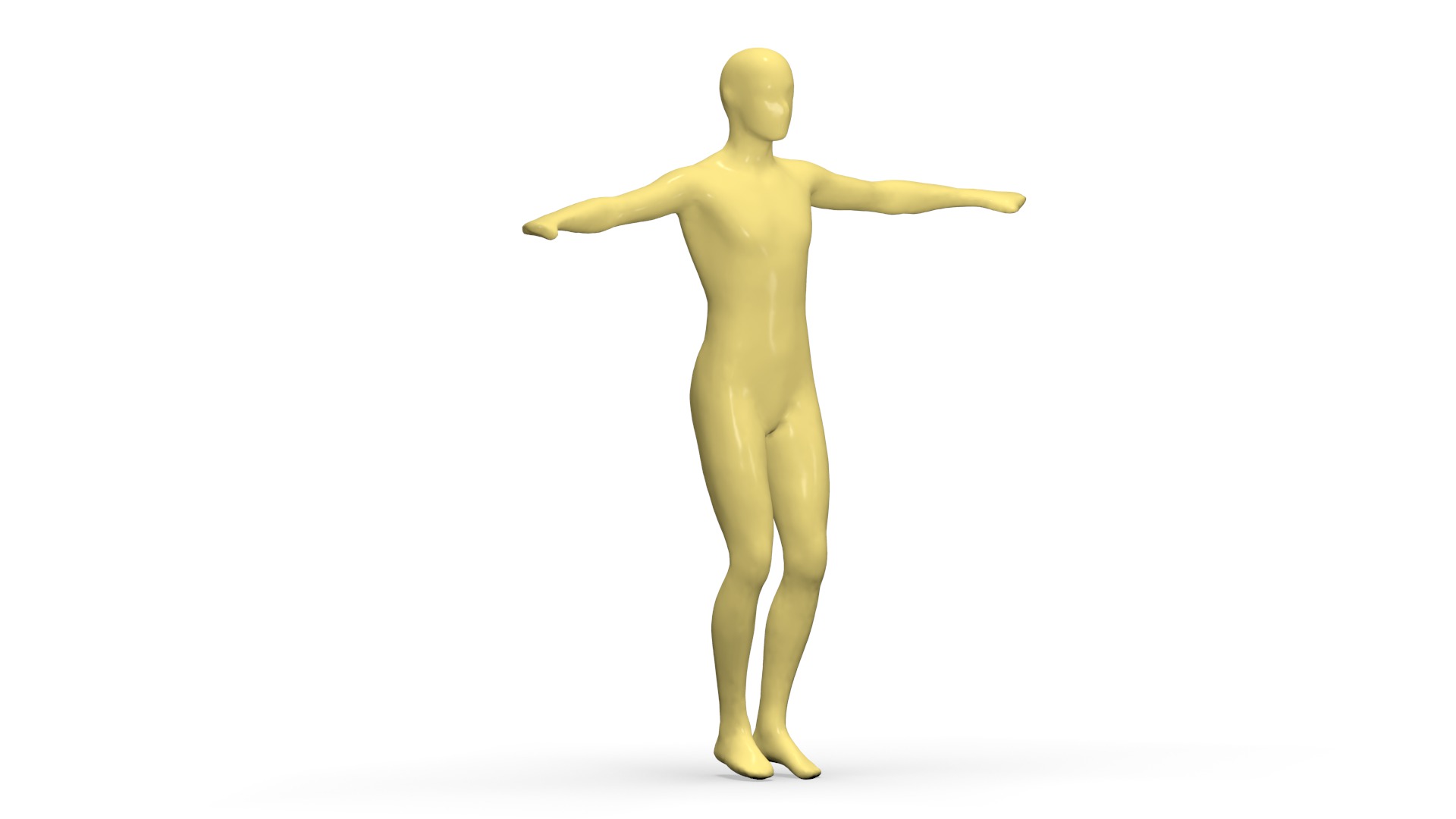}
	\put(50,-12){{\fontsize{6}{8}\selectfont $A$}}
	\end{overpic}
	\begin{overpic}
	[trim=15cm 0cm 20cm 0cm,clip,width=0.24\columnwidth]{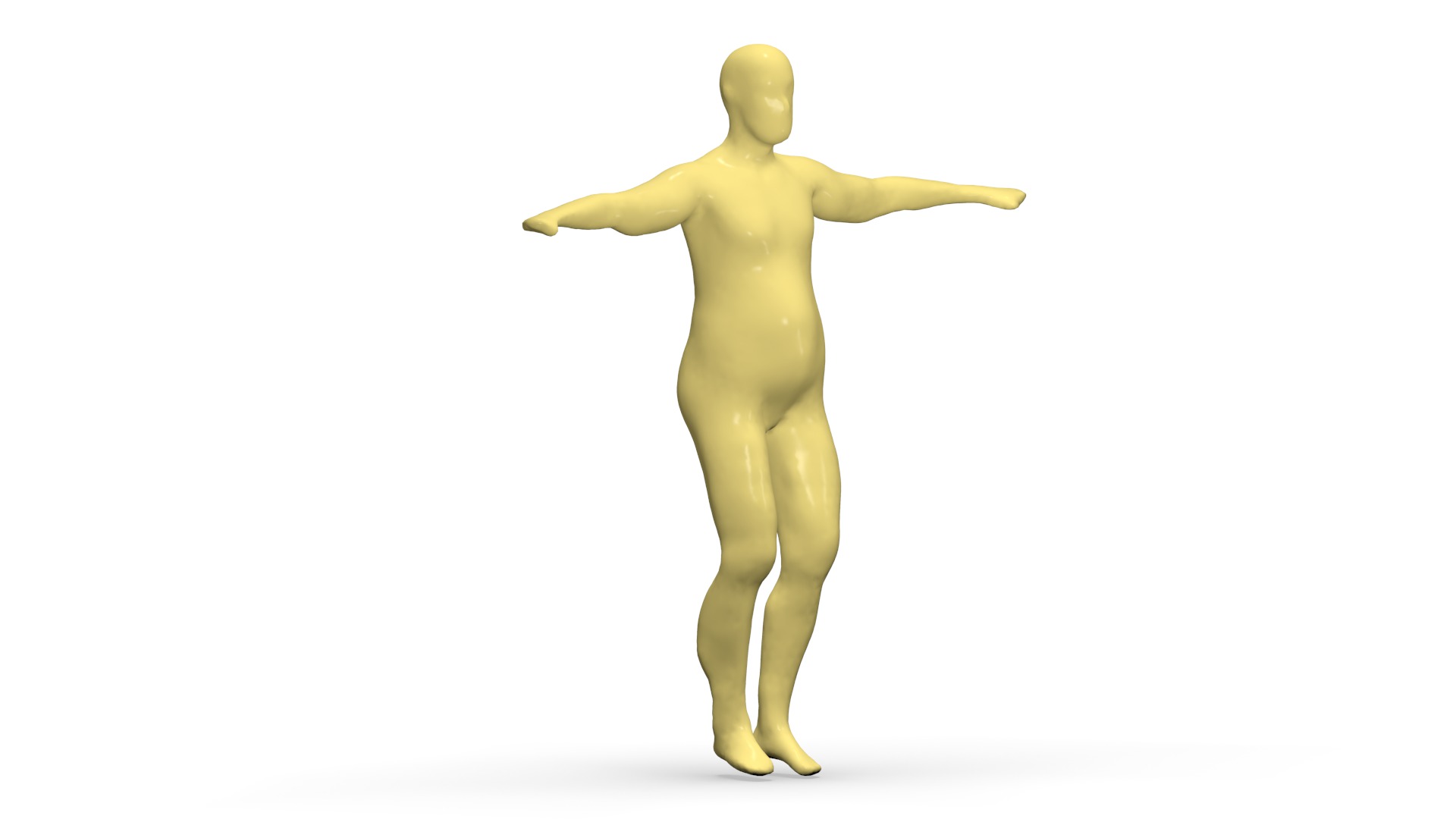}
	\put(50,-12){{\fontsize{6}{8}\selectfont $B$}}
	\end{overpic}
	\begin{overpic}
	[trim=15cm 0cm 20cm 0cm,clip,width=0.24\columnwidth]{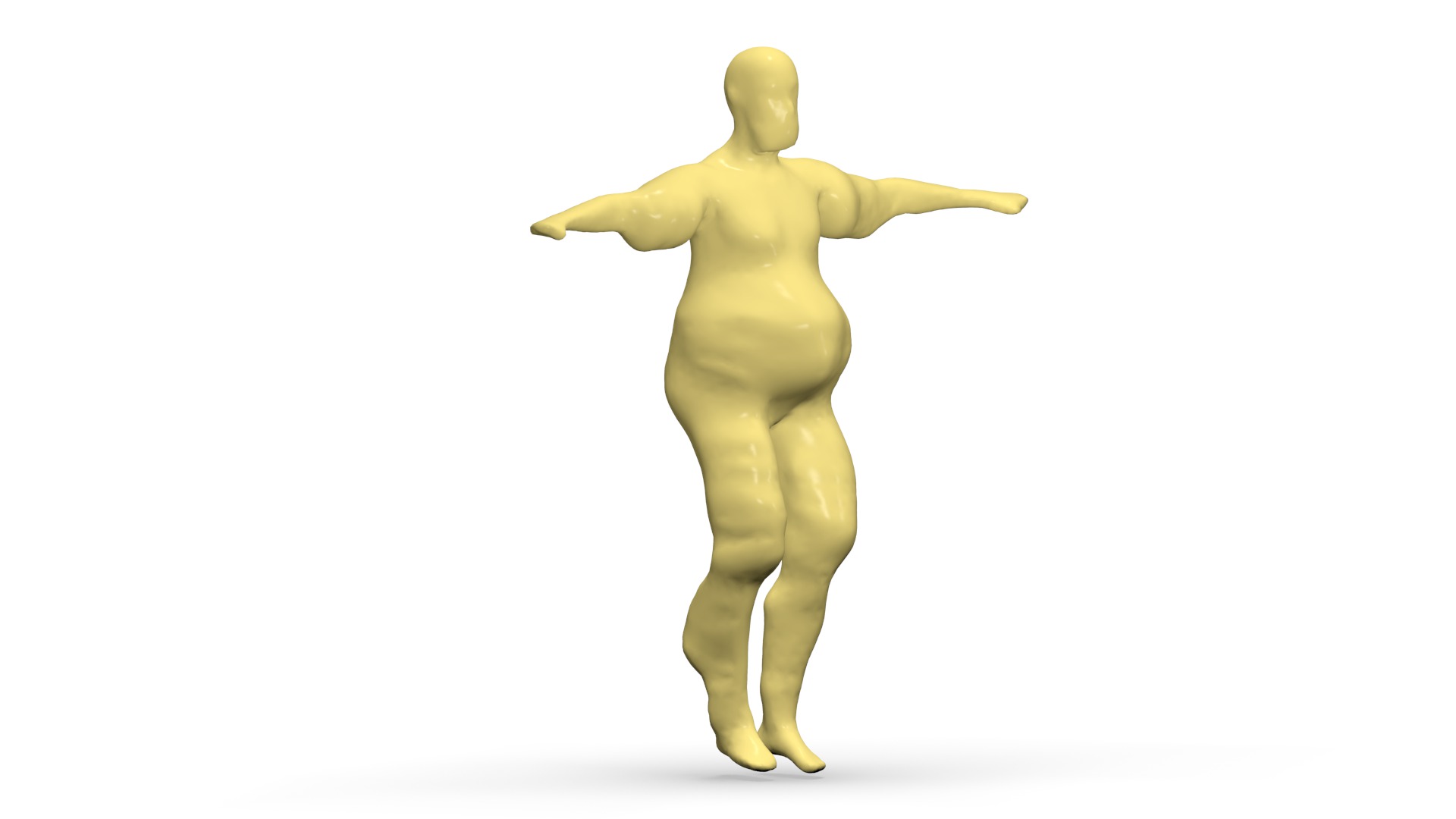}
	\put(50,-12){{\fontsize{6}{8}\selectfont $X$}}
	\end{overpic}
	\begin{overpic}
	[trim=15cm 0cm 20cm 0cm,clip,width=0.24\columnwidth]{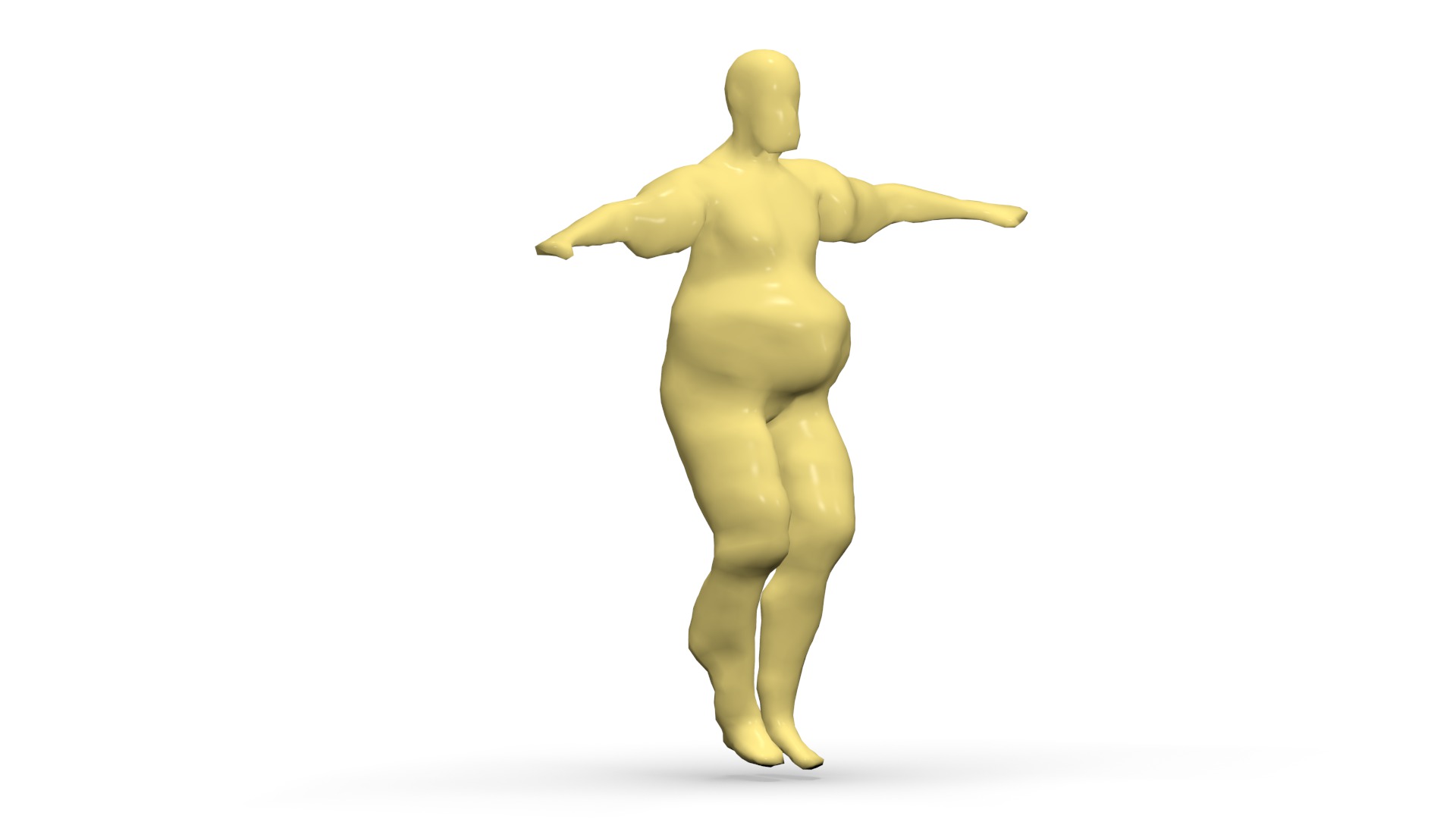}
	\put(47,-12){{\fontsize{6}{8}\selectfont ``$X^2$"}}
	\end{overpic}
\end{minipage} \\ \vspace{0.5cm}
\caption{Shape exaggeration is the setting of the shape-from-difference operator with $C=B$, resulting in an exaggeration of the difference between shapes $A$ and $B$. In this example, we caricaturize the difference between a man and a woman (top), and a man and its fat version (bottom).
Shown left-to-right: shape $A$ (man), shape $B$ (woman or fat man, respectively); result of the shape $X$ synthesis (exaggerated woman or fat man); result of the same process repeated once again. 
}
\label{fig:caricaturization}
\end{figure}

\textbf{Shape exaggeration} is an interesting setting of the shape-from-difference operator problem where $B=C$. In this case, the shape difference between $A$ and $B$ is applied to $B$ itself, `caricaturizing' this difference. Repeating the process several time, an even stronger effect is obtained. 
Figure \ref{fig:caricaturization} shows an example of such caricaturization of the difference between a man and a woman (top row), resulting in an exaggeratedly female shape with large breast and hips, and the difference between a thin and fat man (bottom row), resulting in a very fat man. 


\textbf{Shape-from-Laplacian } amounts to 
%
deforming shape $A$ into a new shape $X$ (with embedding $\bb{X}$) in such a way that the resulting $\bb{W}({\bb{\ell}(\bb{X})}) \approx \bb{W}_B$, as described in Section~\ref{sec:prob}. 
%
%
 %
%
%
%
%
Figure \ref{fig:shapes_from_Laplacian} shows the results of our experiments. The initial shapes $A$ are shown in the leftmost column, shapes $B$ used to compute the reference Laplacian are shown in the second column from left. The result of the optimization $X$ is shown in the rightmost column. Intermediate steps of the optimization with vertex-wise energy are shown in columns 3-6  (in this case, $\epsilon_{i}(\bb{X})$ simply boils down to the mismatch between the $i$th row and column of $\bb{W}(\bb{\ell}(\bb{X})) - \bb{W}_B$). 
Examples of rows 1-6 show that the Laplacian encodes the `style' of the shape, in the sense that if we e.g. start from a human head and deform it to make its Laplacian close to that of a gorilla, we obtain gorilla's head. 
The example of row 8  shows a pose transformation (open vs bent fingers). As we observed in our shape-from-difference operator experiments, since such deformations are not perfectly isometric, the reconstructed result $X$ also has a bent finger. 
Finally, row 7 shows a combination of `style' and pose transformation (thin man in standing pose used as initialization $A$ vs fat man in running pose whose Laplacian is used as a reference $B$). Here again, the reconstructed $X$ has both the style and the pose (albeit attenuated) of the reference shape $B$. 
%
%
%

\section{Discussion and conclusions}\label{sec:disc}

We presented a framework for reconstructing shapes from intrinsic operators, focusing on the problem of shape-from-difference operator, as this problem includes other important problems such as shape-from-Laplacian and shape exaggeration as its particular instances. 

In our experiments, we have encountered two important factors that affect the quality of the obtained results, and which can be considered as limitations of our approach. 

\textbf{Sensitivity to mesh quality.} 
The definition of cotangent weights~(\ref{eq:cotan}) produces $w_{ij} < 0$ if $\alpha_{ij} + \beta_{ij} > \pi$, an issue known to be problematic in many applications (e.g. in harmonic parametrization and texture mapping where it leads to triangle flips \cite{bobenko2007discrete}). 
In our case, negative weights have adverse effect on the convergence of the MoF optimization, since the step size $\mu$ in Algorithm~\ref{algo:mfo} becomes very small.

\begin{figure}[t!]
\centering
\begin{minipage}{\columnwidth}
\setlength\figureheight{4cm} 
\setlength\figurewidth{0.9\columnwidth}
\input{./sections/convergence_plot.tikz}
\end{minipage} \\ \vspace{0.35cm}
\begin{flushright}
\begin{minipage}{1.2\columnwidth}\hspace{-10mm}
	\begin{overpic}
	[trim=0cm 0cm 0cm 0cm,clip,width=0.32\textwidth]{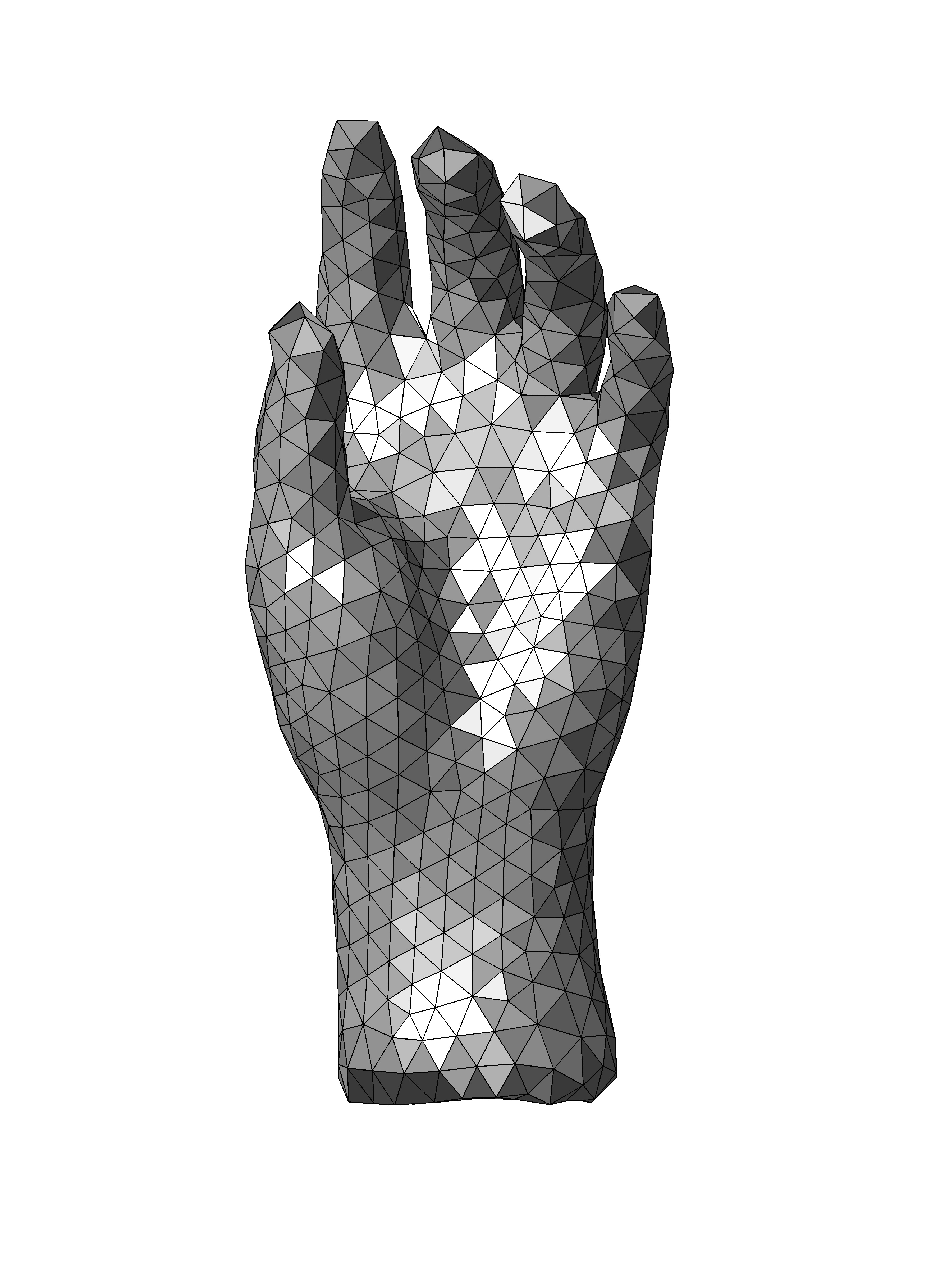}
	\end{overpic} 
	\begin{overpic}
	[trim=0cm 0cm 0cm 0cm,clip,width=0.32\textwidth]{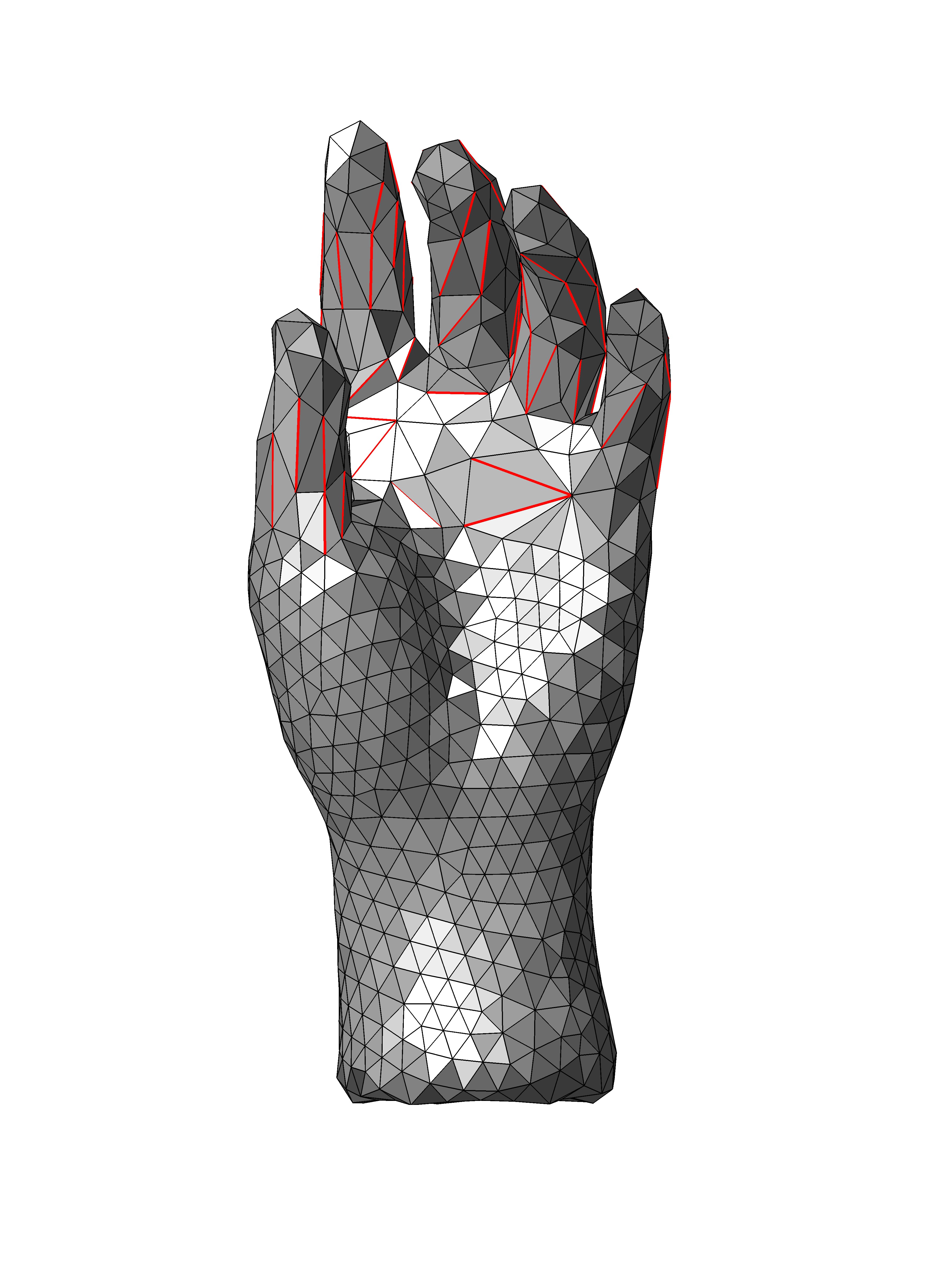}
	\end{overpic} 
	\begin{overpic}
	[trim=0cm 0cm 0cm 0cm,clip,width=0.32\textwidth]{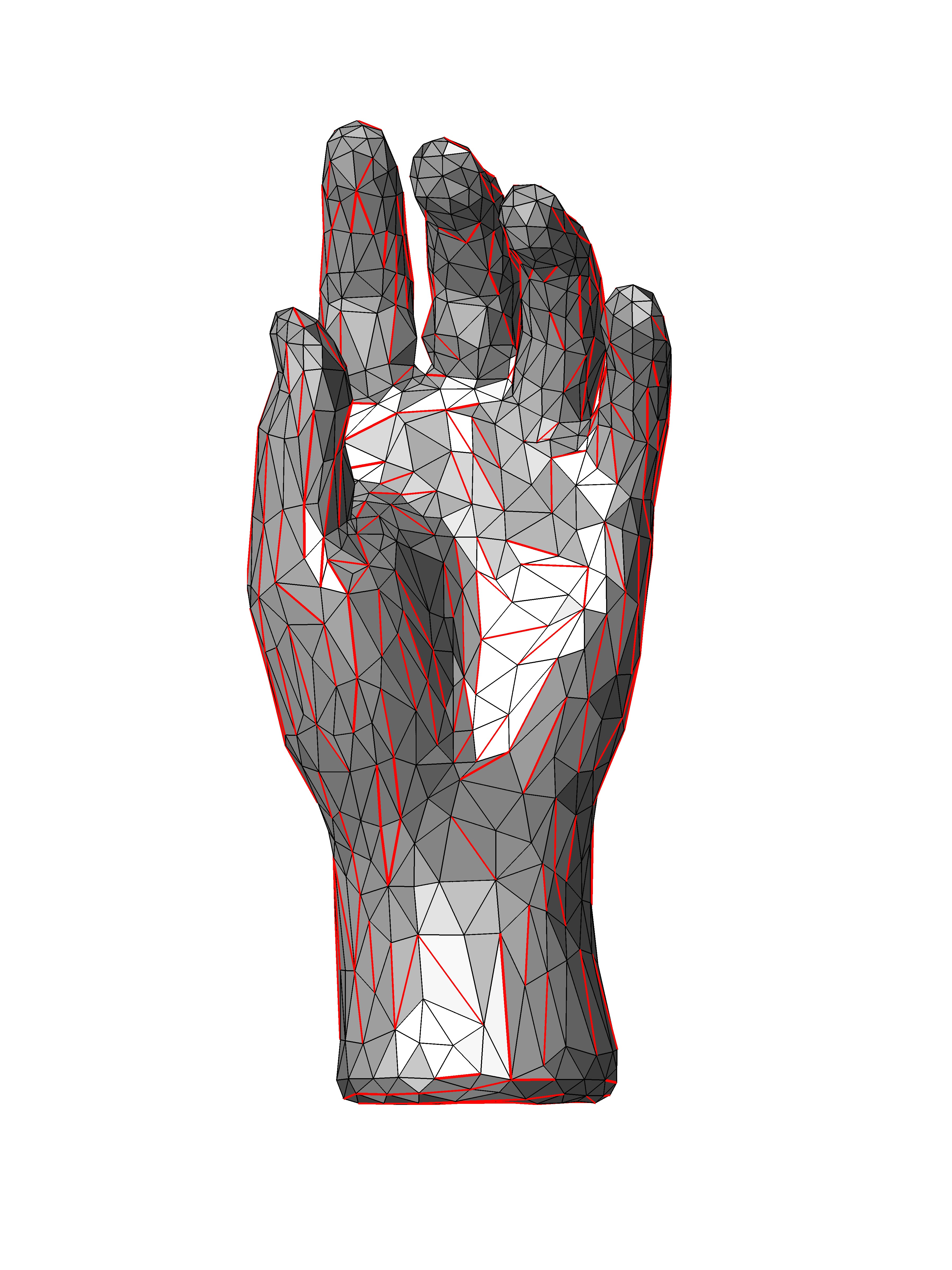}
	\put(-50,105){\fontsize{8}{8}\selectfont iterations}
	\put(-125,145){\rotatebox{90}{\fontsize{8}{8}\selectfont MfO energy}}
	\put(-125,0){\fontsize{9}{8}\selectfont \color{red}{\textbf{`good'}}}
	\put(-45,0){\fontsize{9}{8}\selectfont \color{green!50!black}{\textbf{`bad'}}}
	\put(30,0){\fontsize{9}{8}\selectfont \color{blue}{\textbf{`ugly'}}}
	\end{overpic}
\end{minipage}
\end{flushright}
\caption{
Convergence of our optimization problem depending on the quality of the mesh (top; shown is MfO energy). 
Red: a `good' mesh (bottom left) obtained with isotropic remeshing and containing no obtuse triangles results in all positive cotangent weights, which leads to fast convergence. 
Green: a `bad' mesh (bottom center) contains few obtuse triangles, resulting in some negative weights (respective edges are shown in red), leading to a slower convergence. 
Blue: an `ugly' mesh (bottom right) containing many obtuse triangles breaks the convergence. }
\label{fig:remeshing}
\end{figure}

In Figure \ref{fig:remeshing}, we exemplify this behavior by showing the plot of the energy $\mathcal{E}_\mathrm{dif}$ as function of internal iteration (for simplicity, only MoF iterations are shown) in the shape-from-Laplacian problem. 
When the mesh is completely isotropic (all triangles are acute, Figure \ref{fig:remeshing} left), convergence is very fast (red curve). 
Adding a few triangles with obtuse angles produces negative weights (marked in red in Figure \ref{fig:remeshing}, center) and slows down the convergence of the algorithm (green). 
Finally, when too many obtuse triangles are present(Figure \ref{fig:remeshing}, right), the algorithm fails to converge (blue).

\textbf{Sensitivity to functional map quality.}
Rustamov et al. \cite{Rustamov2013} remark that ``the quality of the information one gets from [shape difference operators] depends on the quality and density of the shape maps''. 
We also found that the result of shape synthesis with our approach largely depends on the accuracy of the functional maps between the shapes.

\begin{figure*}[t!]
\centering
%
%
\begin{minipage}{\textwidth}
	\hspace{-0.4cm}
	\begin{overpic}
	[trim=4cm 1cm 14cm 1cm,clip,width=0.14\textwidth]{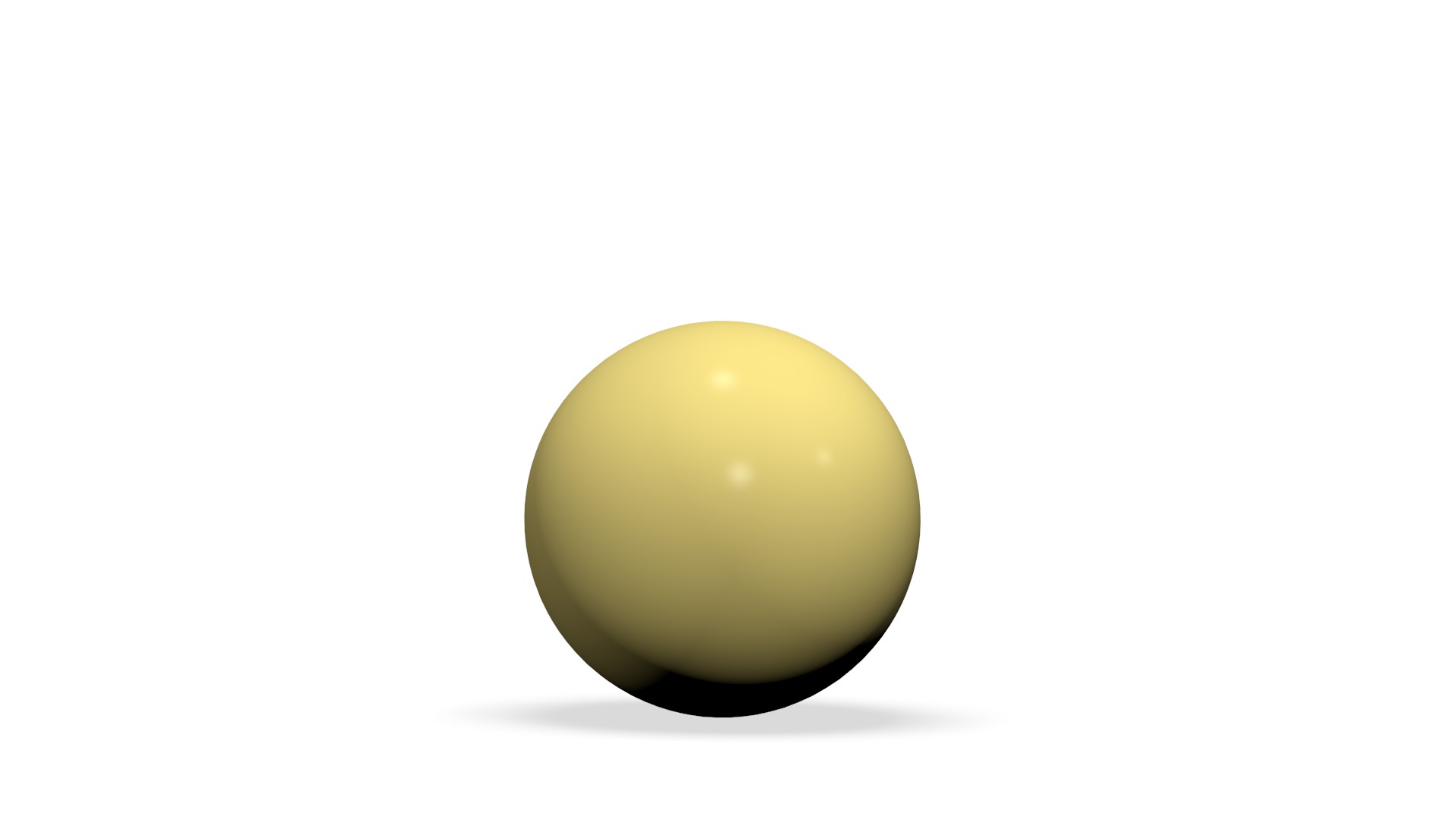}
	\end{overpic} 
	\hspace{-0.2cm}
	\begin{overpic}
	[trim=6cm 1cm 12cm 1cm,clip,width=0.14\textwidth]{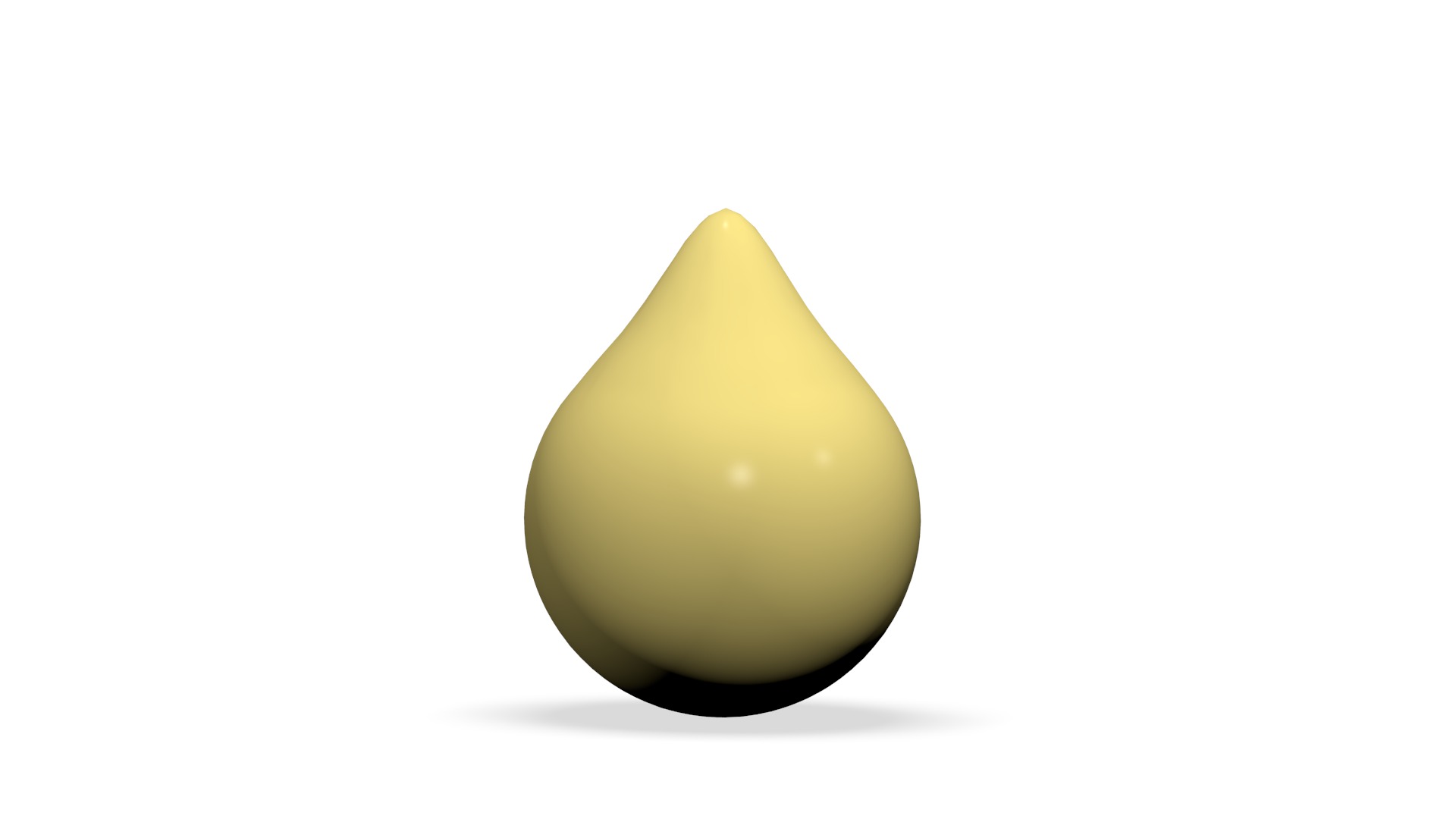}
	\end{overpic}
	\hspace{0.1cm}
	\begin{overpic}
	[trim=9cm 1cm 7cm 1cm,clip,width=0.14\textwidth]{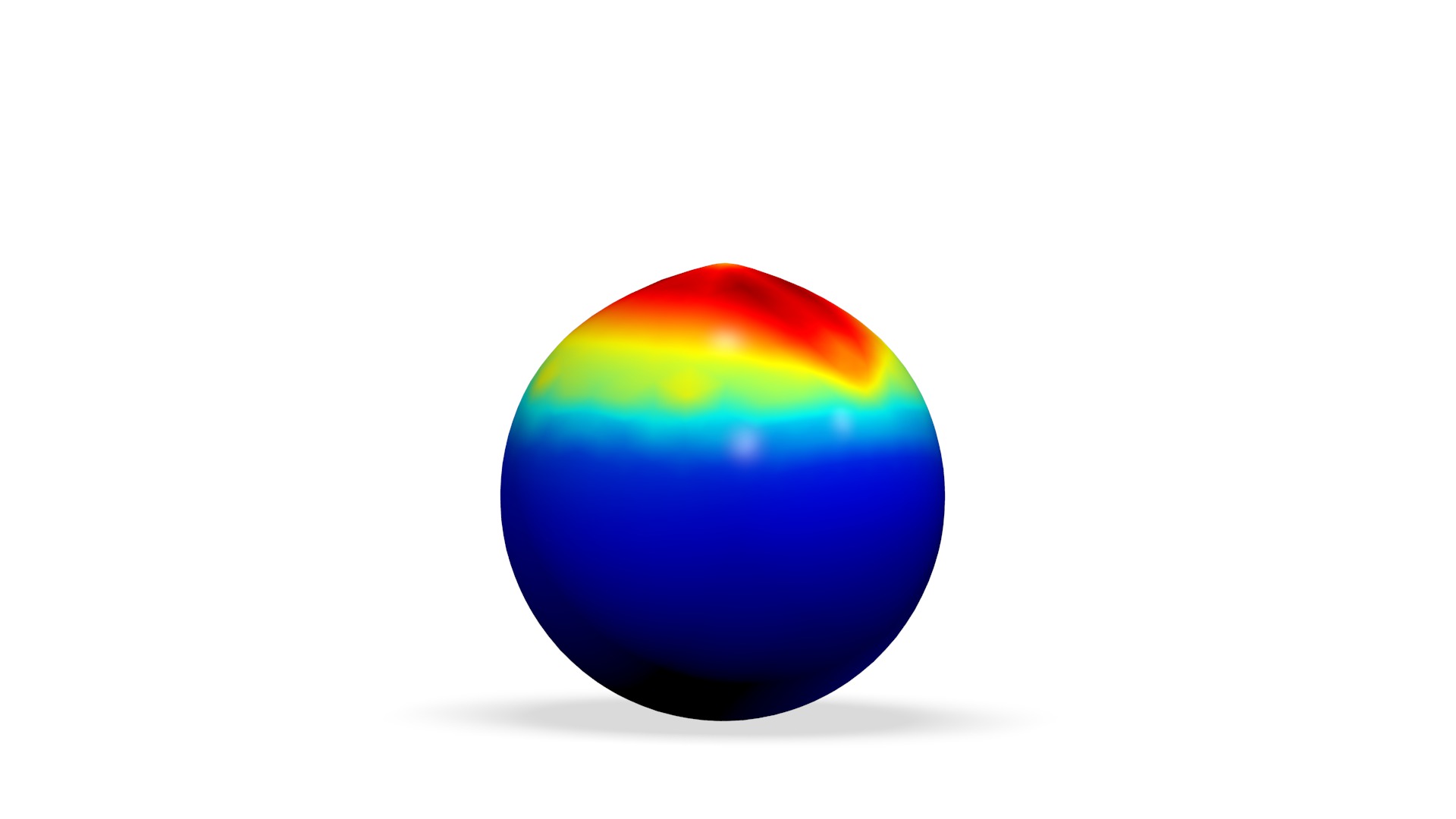}
	\end{overpic} 
	\hspace{-0.25cm}
	\begin{overpic}
	[trim=12cm 1cm 6cm 1cm,clip,clip,width=0.14\textwidth]{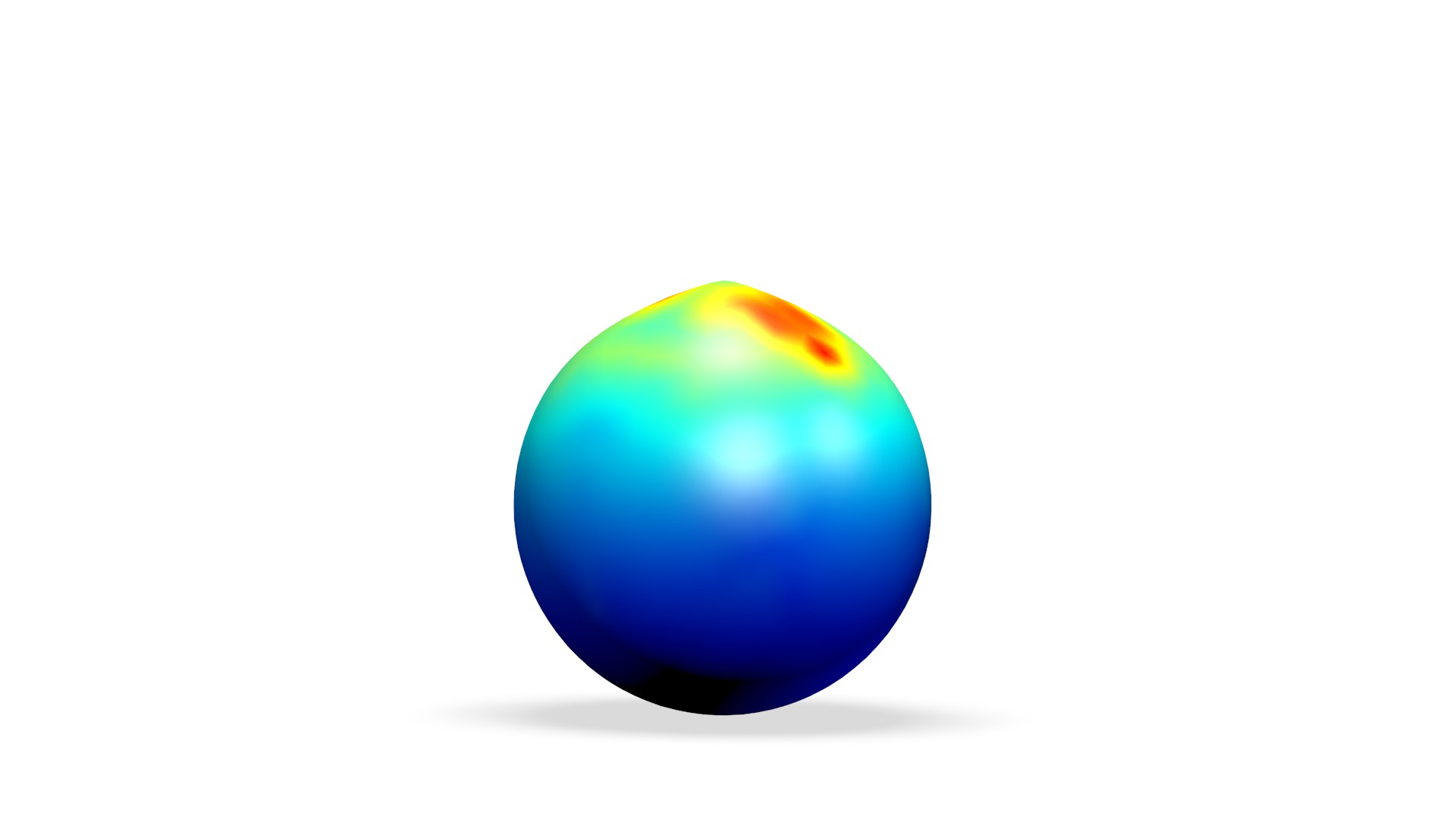}
	\end{overpic}
	\hspace{-0.25cm}
	\begin{overpic}
	[trim=12cm 1cm 6cm 1cm,clip,width=0.14\textwidth]{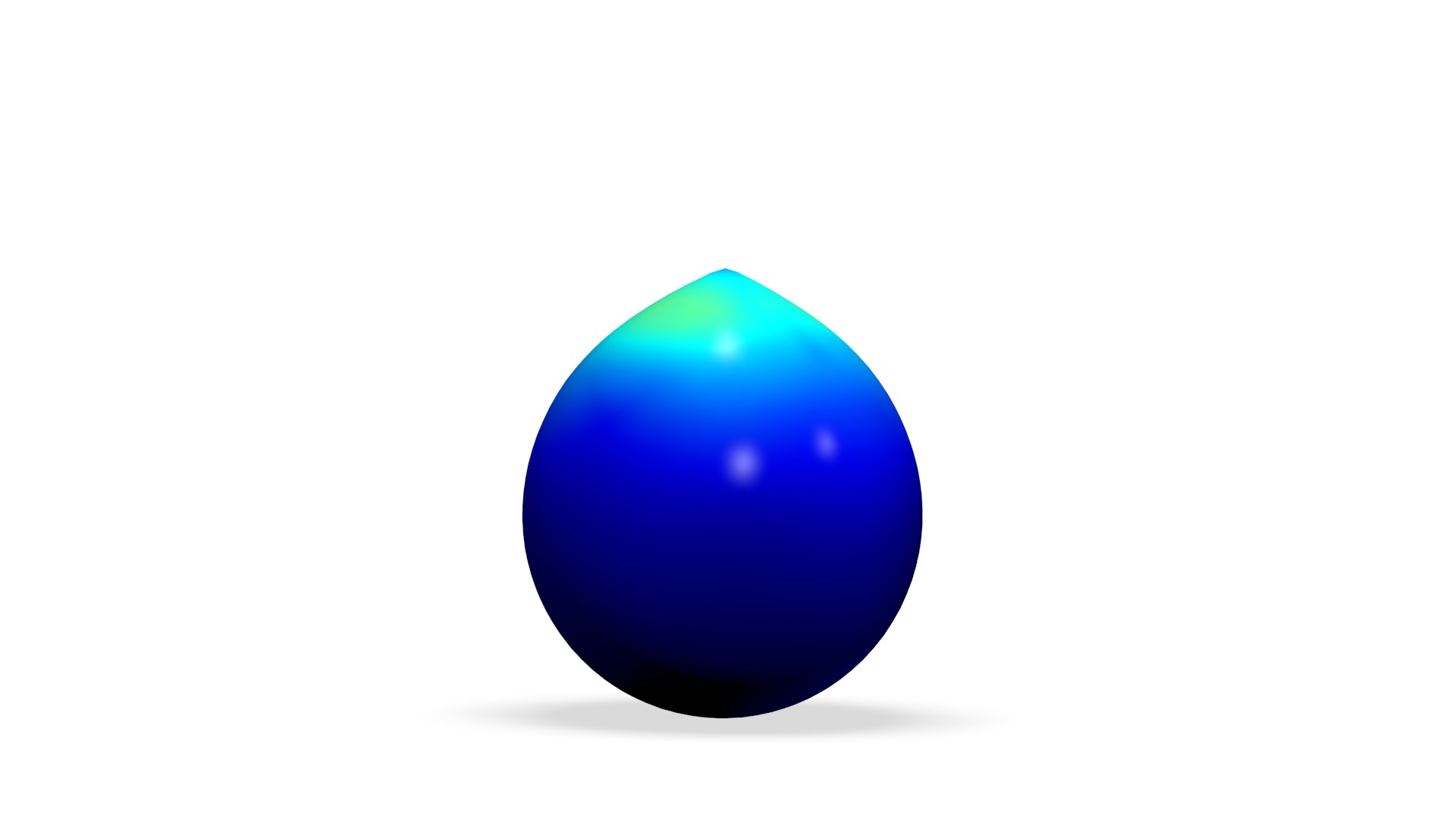}
	\end{overpic}
	\hspace{-0.25cm}
	\begin{overpic}
	[trim=12cm 1cm 6cm 1cm,clip,clip,width=0.14\textwidth]{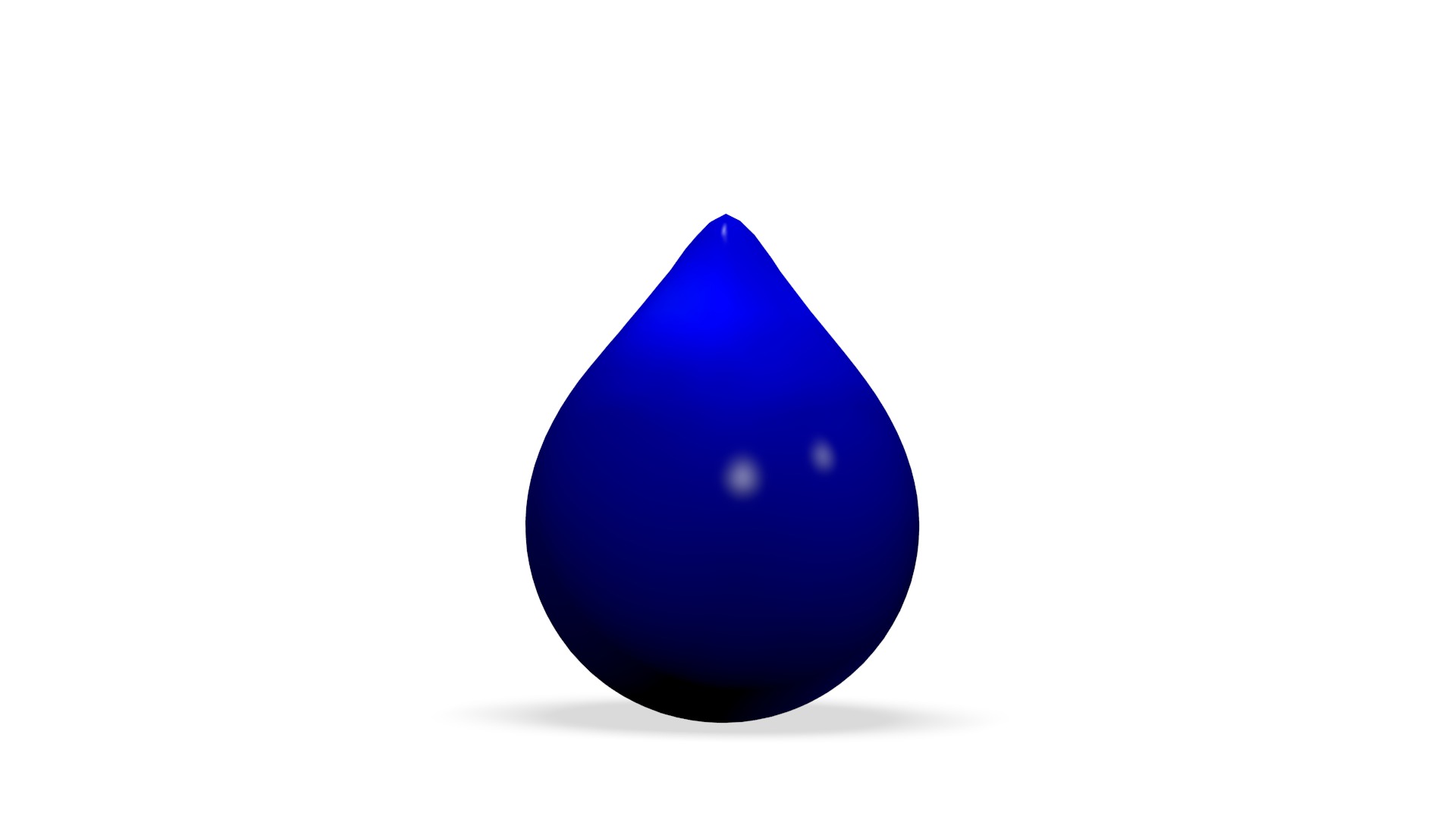}
	\end{overpic}
	\hspace{0.2cm}
	\begin{overpic}
	[trim=6cm 1cm 12cm 1cm,clip,clip,width=0.14\textwidth]{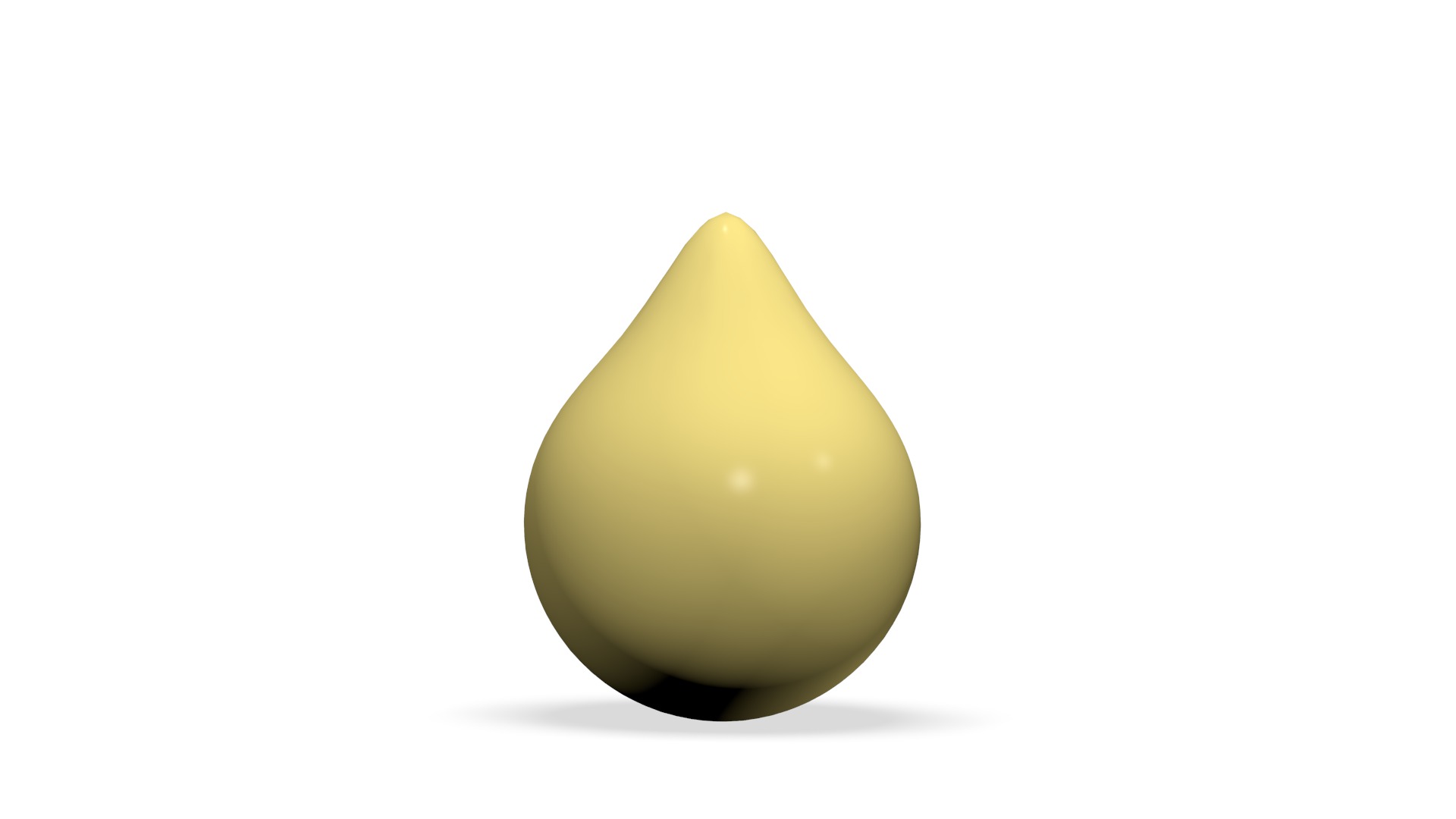}
	\end{overpic}
\end{minipage} \\ \vspace{0.2cm}
%
%
%
%
\begin{minipage}{\textwidth}
	\hspace{-0.5cm}
	\begin{overpic}
	[trim=2cm 1cm 16cm 1cm,clip,width=0.14\textwidth]{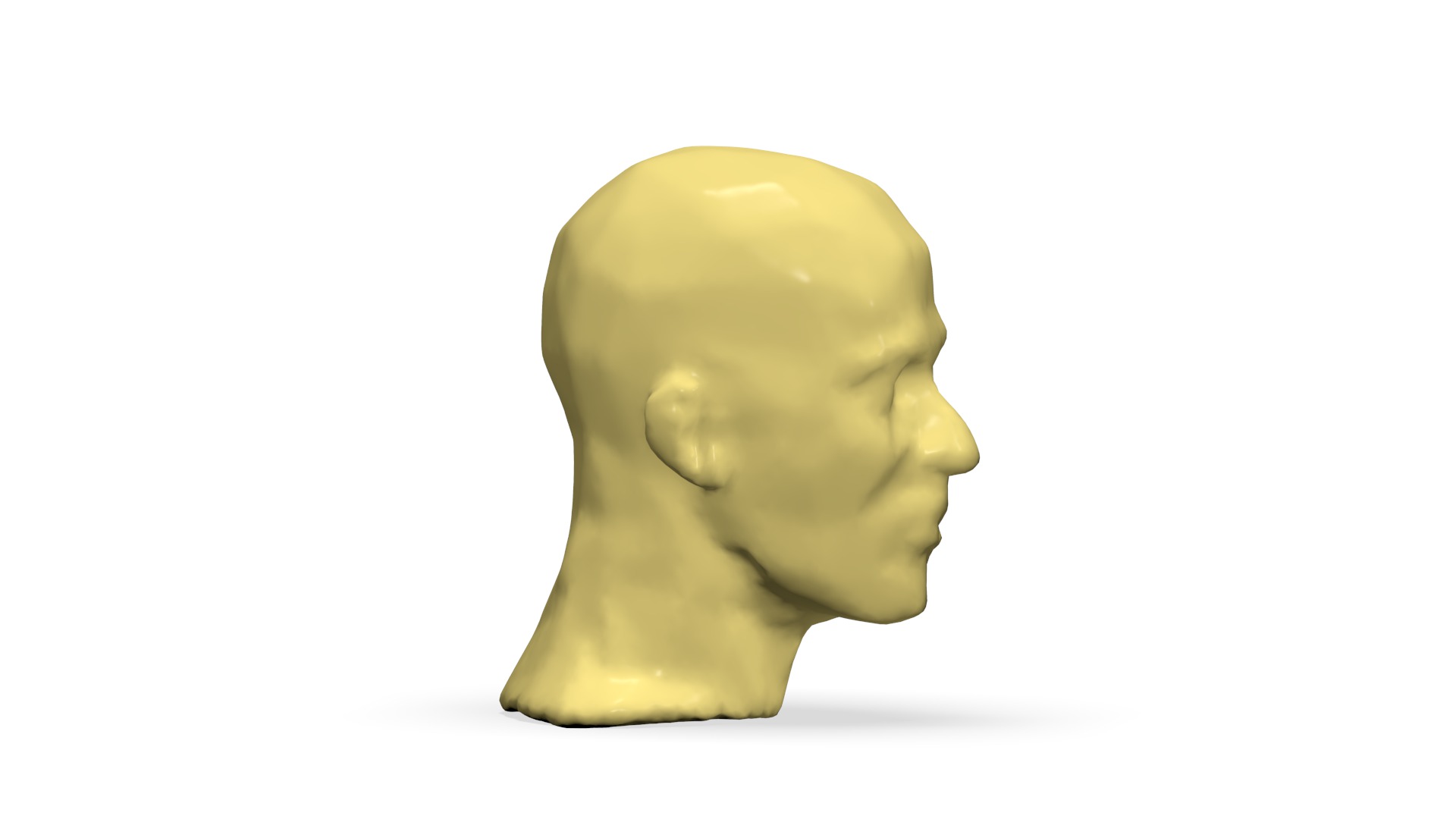}
	\end{overpic} 
	\hspace{-0.2cm}
	\begin{overpic}
	[trim=6cm 1cm 12cm 1cm,clip,width=0.14\textwidth]{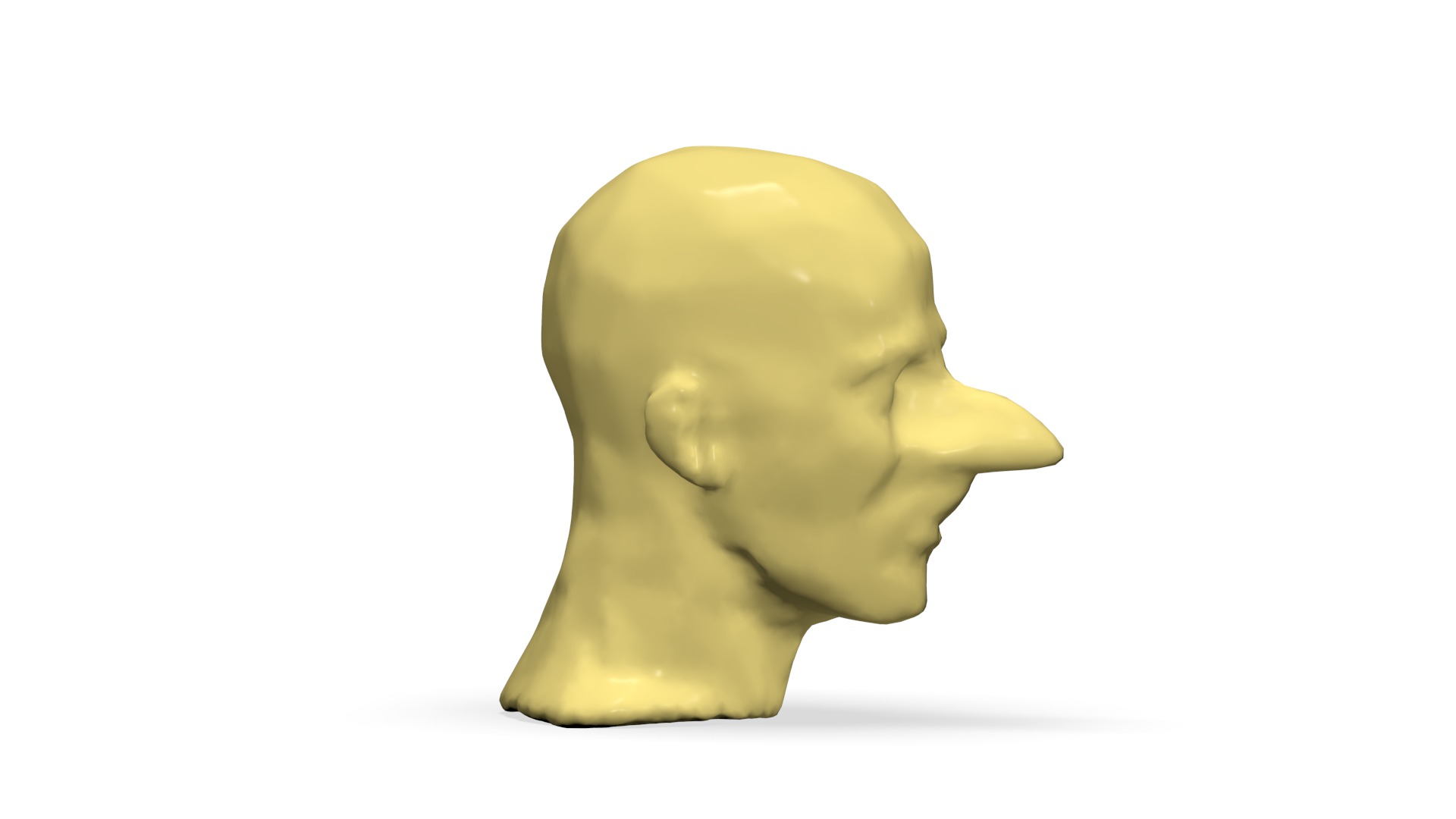}
	\end{overpic}
	\hspace{0.1cm}
	\begin{overpic}
	[trim=12cm 1cm 6cm 1cm,clip,width=0.14\textwidth]{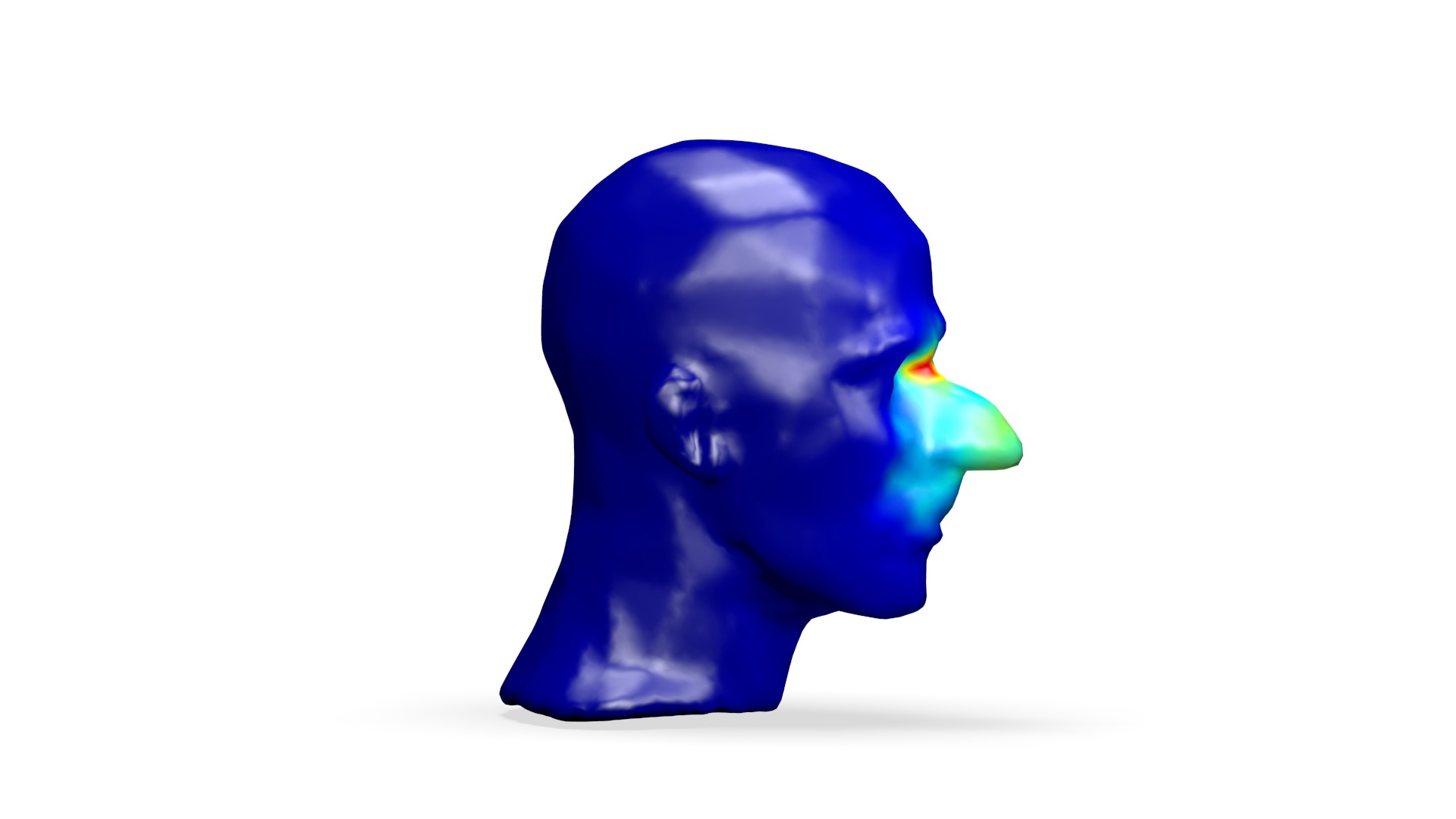}
	\end{overpic} 
	\hspace{-0.3cm}
	\begin{overpic}
	[trim=12cm 1cm 6cm 1cm,clip,clip,width=0.14\textwidth]{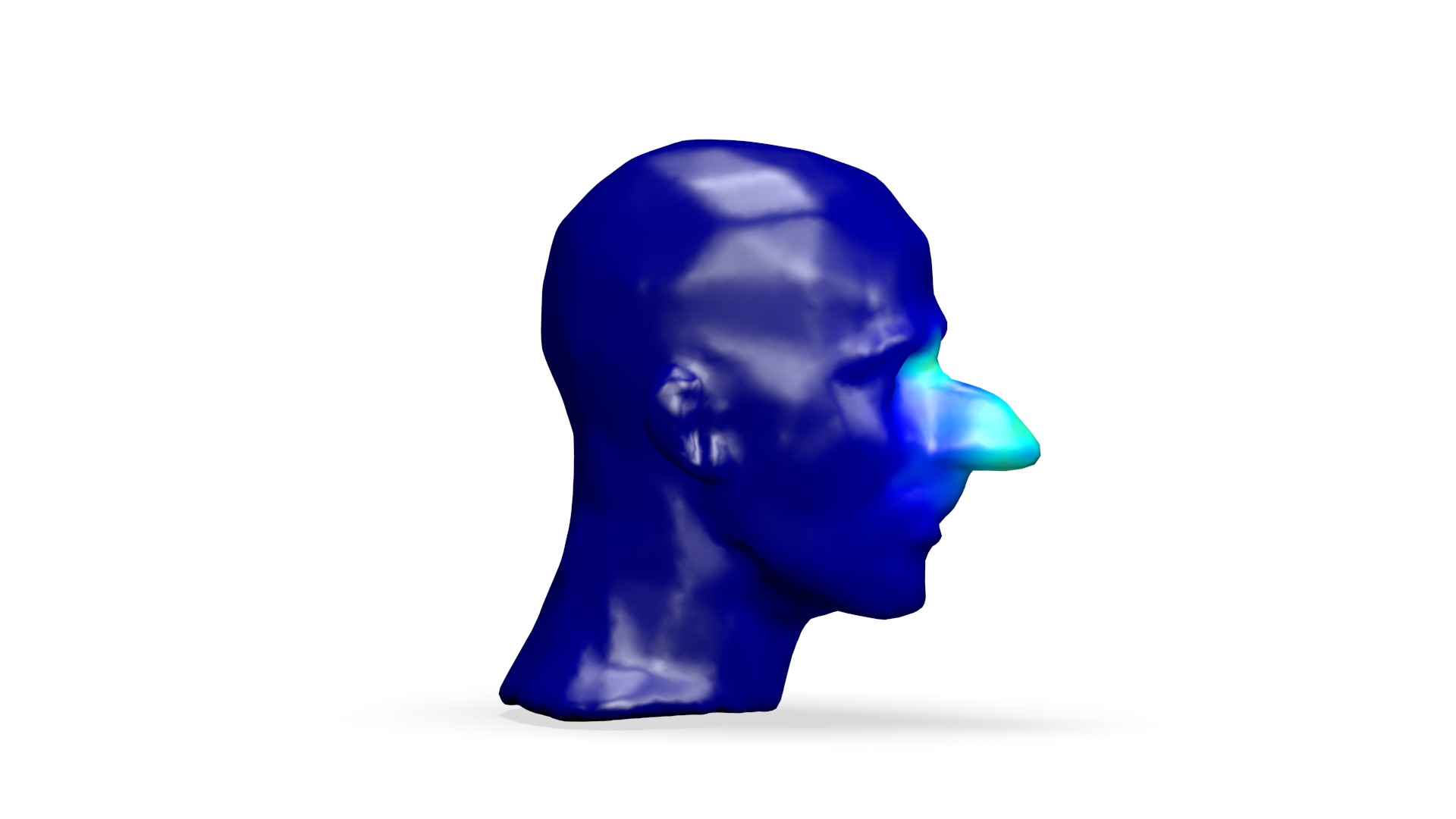}
	\end{overpic}
	\hspace{-0.3cm}
	\begin{overpic}
	[trim=12cm 1cm 6cm 1cm,clip,width=0.14\textwidth]{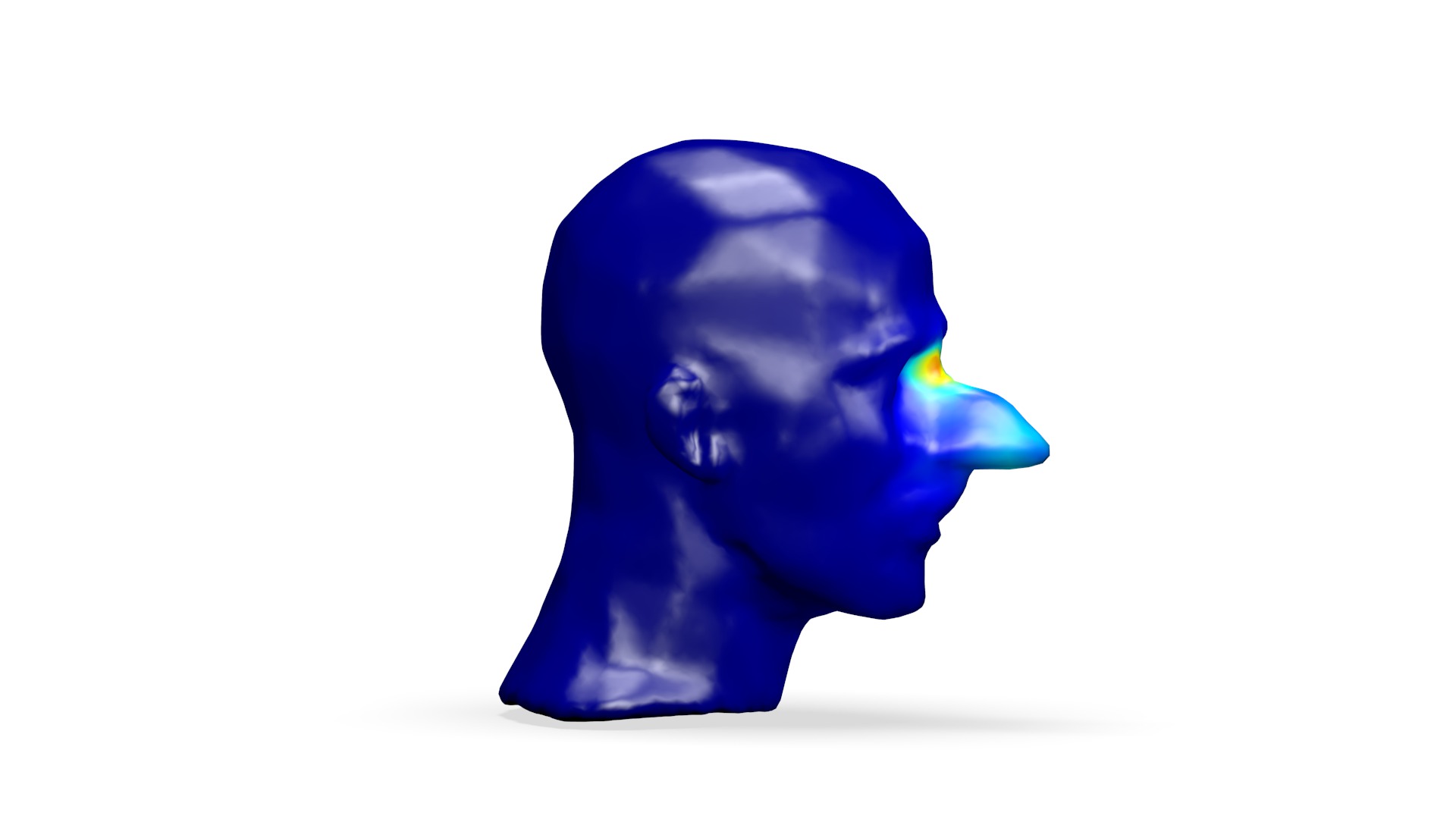}
	\end{overpic}
	\hspace{-0.3cm}
	\begin{overpic}
	[trim=12cm 1cm 6cm 1cm,clip,clip,width=0.14\textwidth]{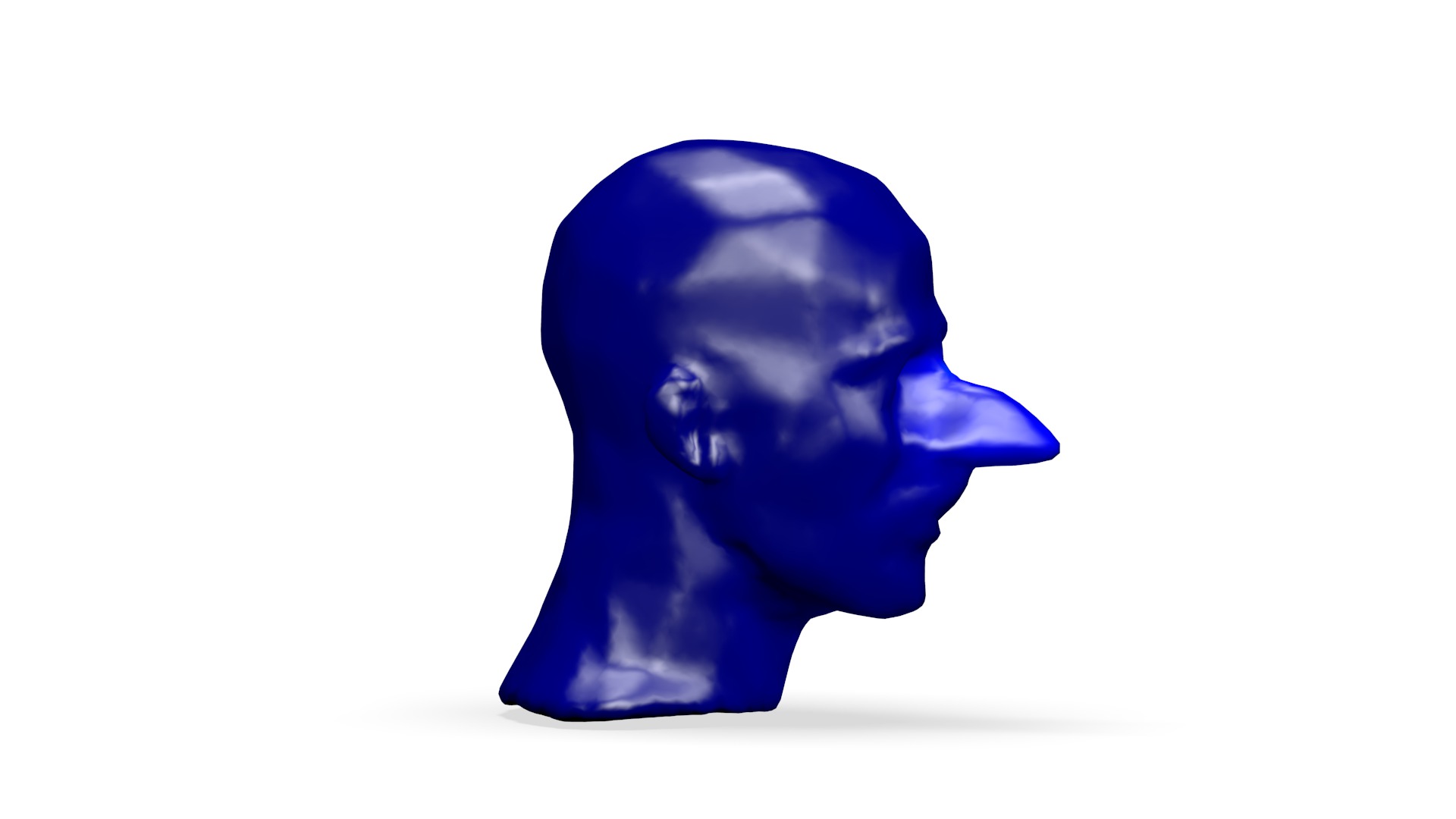}
	\end{overpic}
	\hspace{0.2cm}
	\begin{overpic}
	[trim=2cm 1cm 16cm 1cm,clip,clip,width=0.14\textwidth]{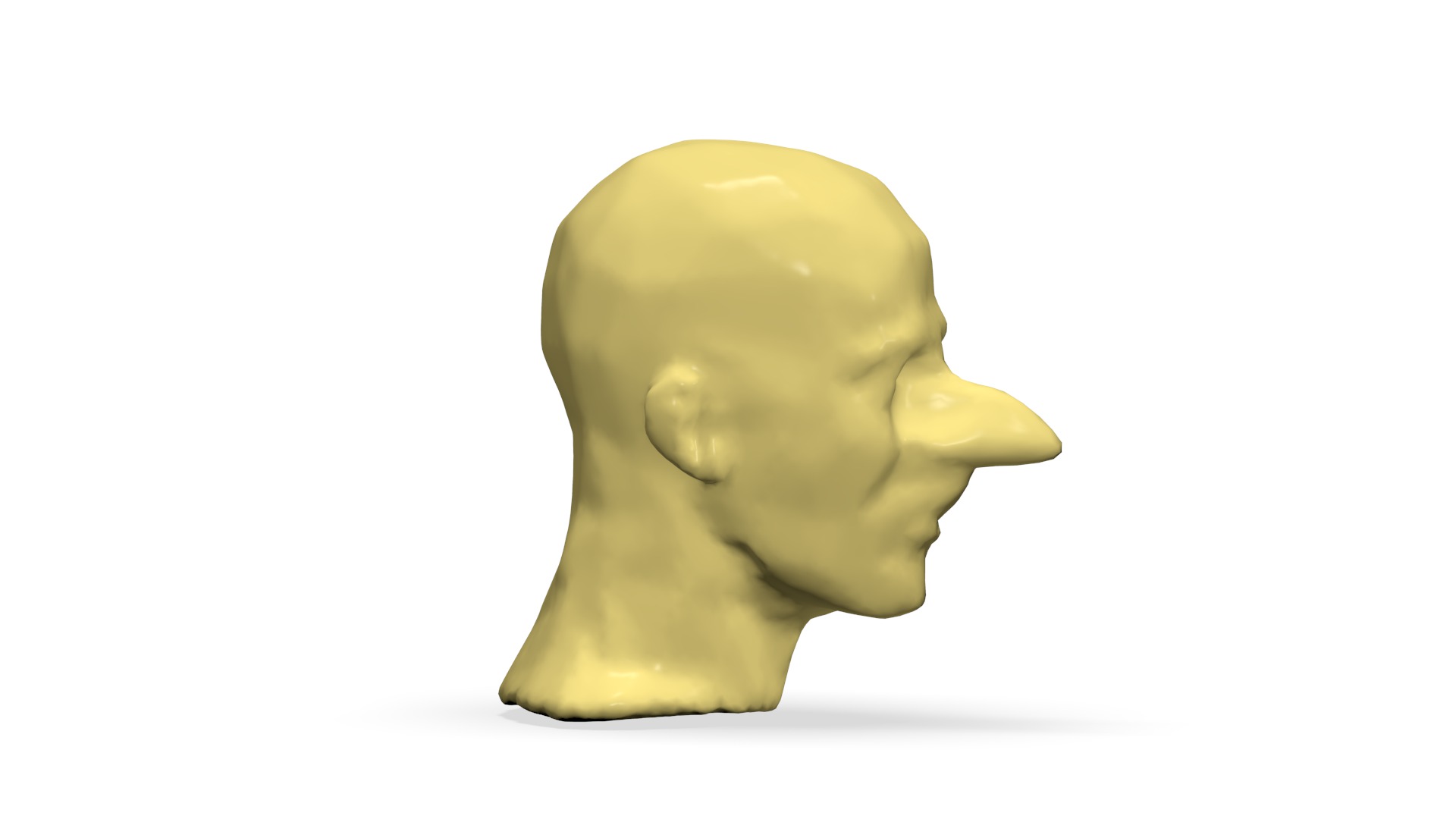}
	\end{overpic}
\end{minipage} \\ \vspace{0.4cm}
%
%
%
%
\begin{minipage}{\textwidth}
	\hspace{-0.5cm}
	\begin{overpic}
	[trim=2cm 2cm 19cm 2cm,clip,width=0.14\textwidth]{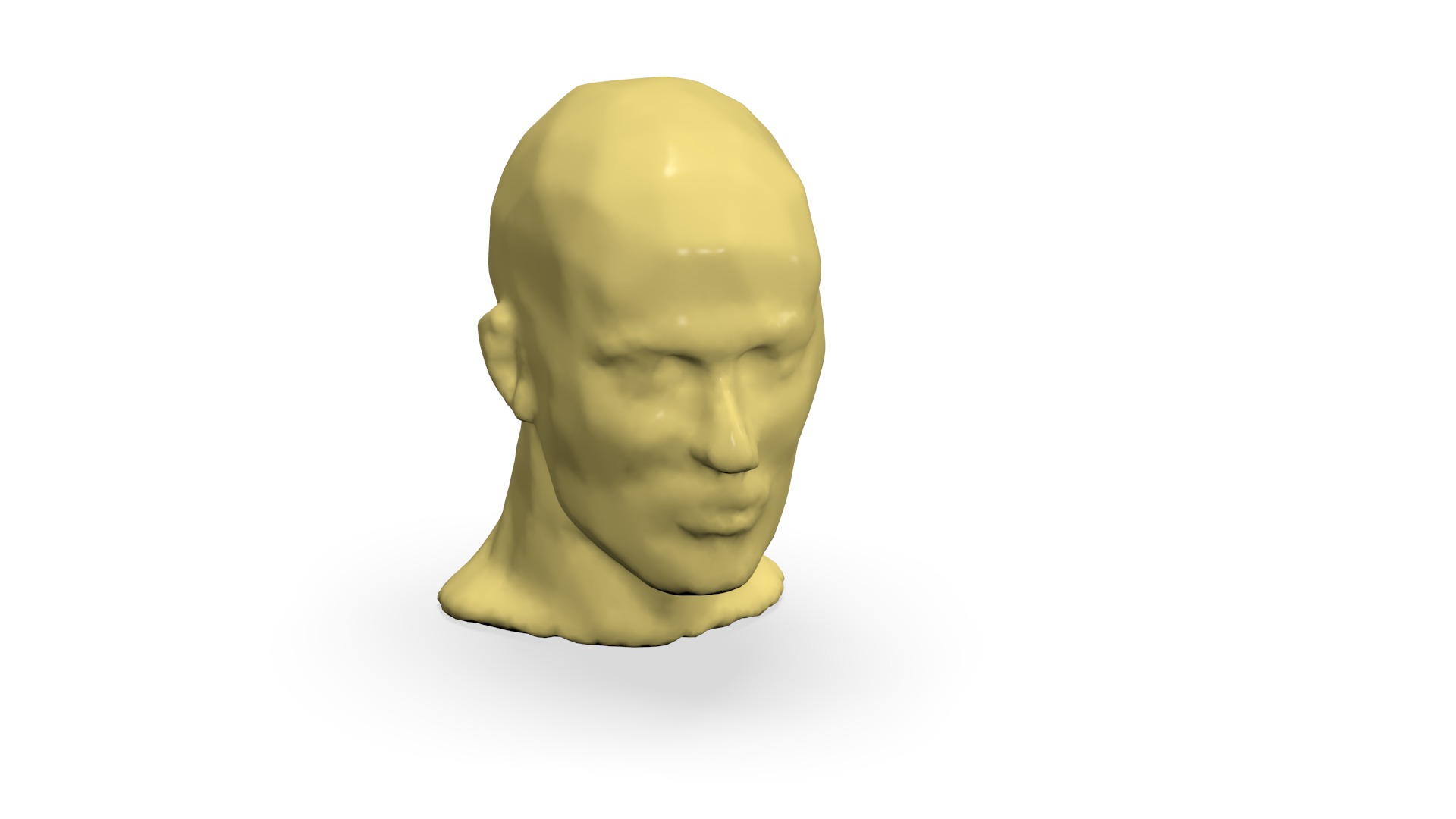}
	\end{overpic} 
	\begin{overpic}
	[trim=6cm 2cm 15cm 2cm,clip,width=0.14\textwidth]{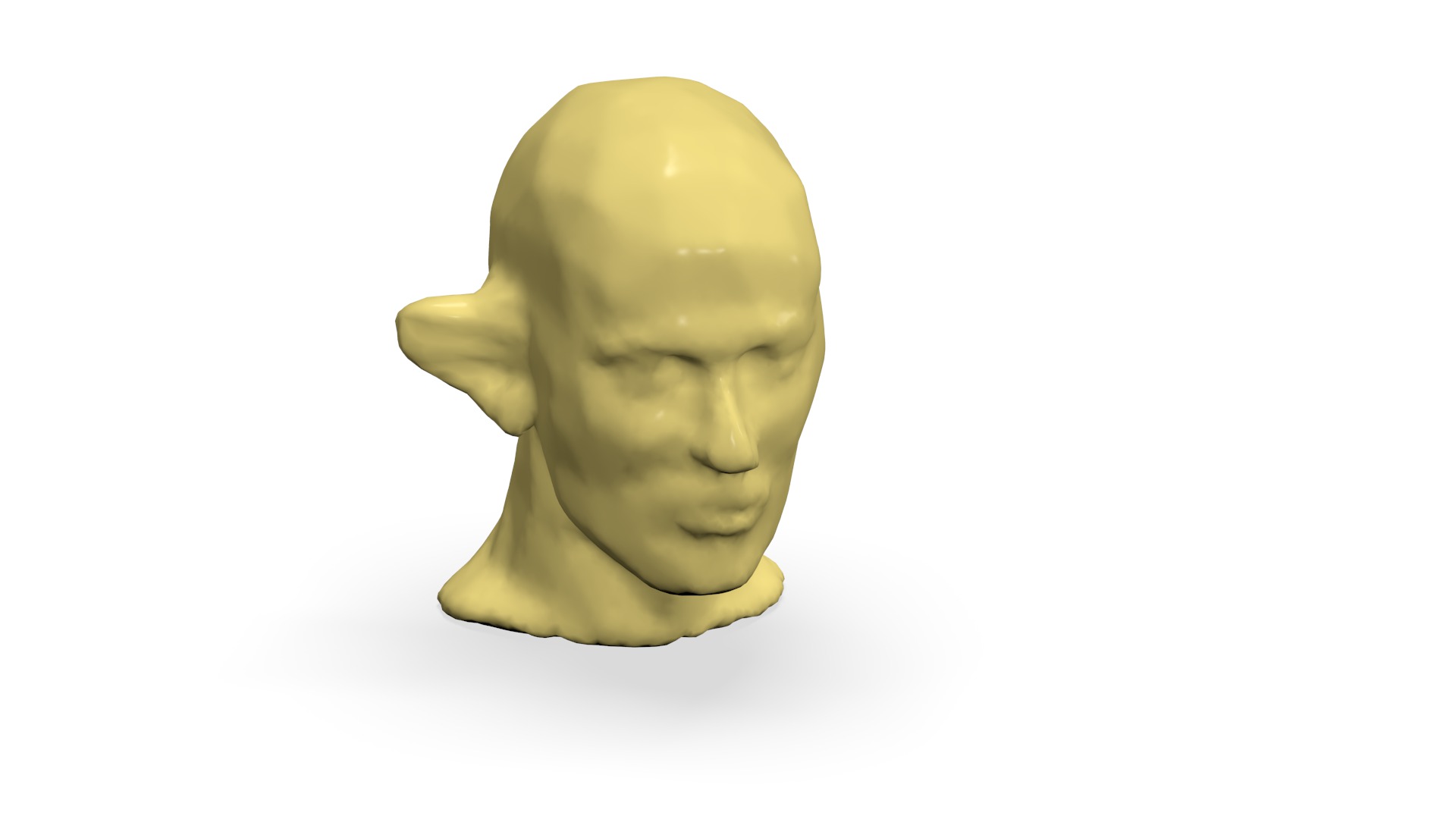}
	\end{overpic}
	\hspace{0.1cm}
	\begin{overpic}
	[trim=7cm 2cm 14cm 2cm,clip,width=0.14\textwidth]{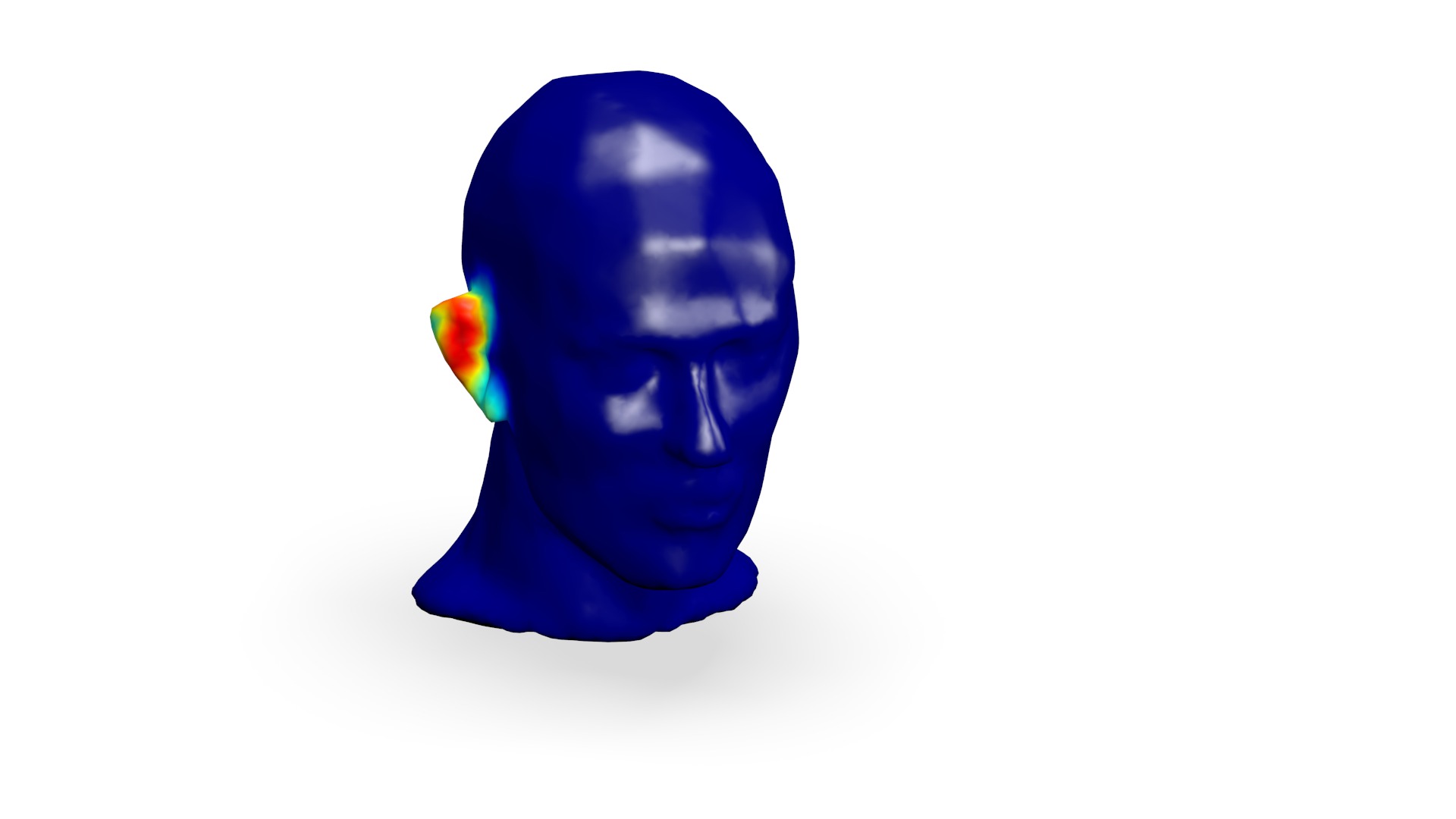}
	\end{overpic} 
	\hspace{-0.3cm}
	\begin{overpic}
	[trim=7cm 2cm 14cm 2cm,clip,clip,width=0.14\textwidth]{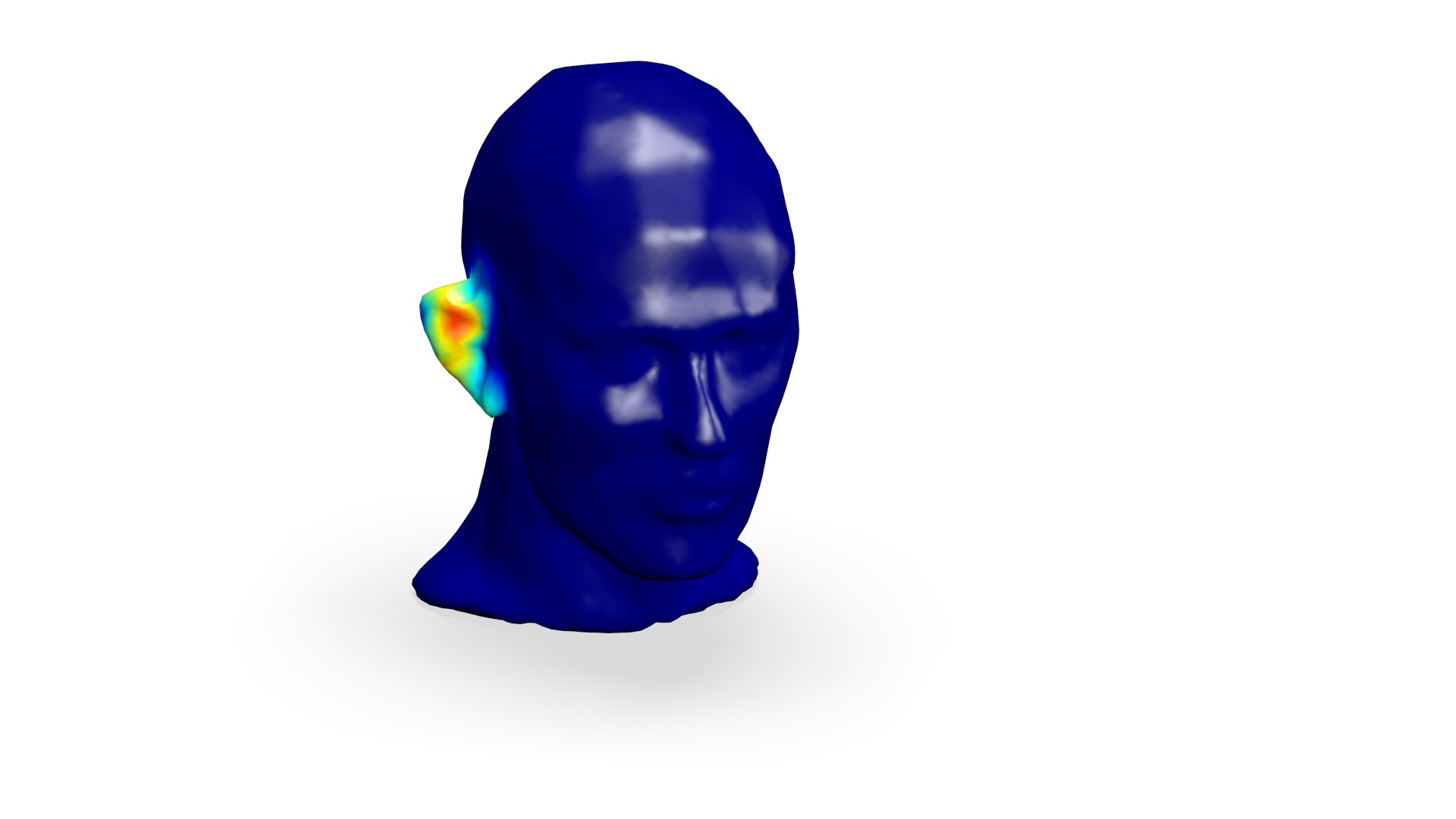}
	\end{overpic}
	\hspace{-0.3cm}
	\begin{overpic}
	[trim=7cm 2cm 14cm 2cm,clip,width=0.14\textwidth]{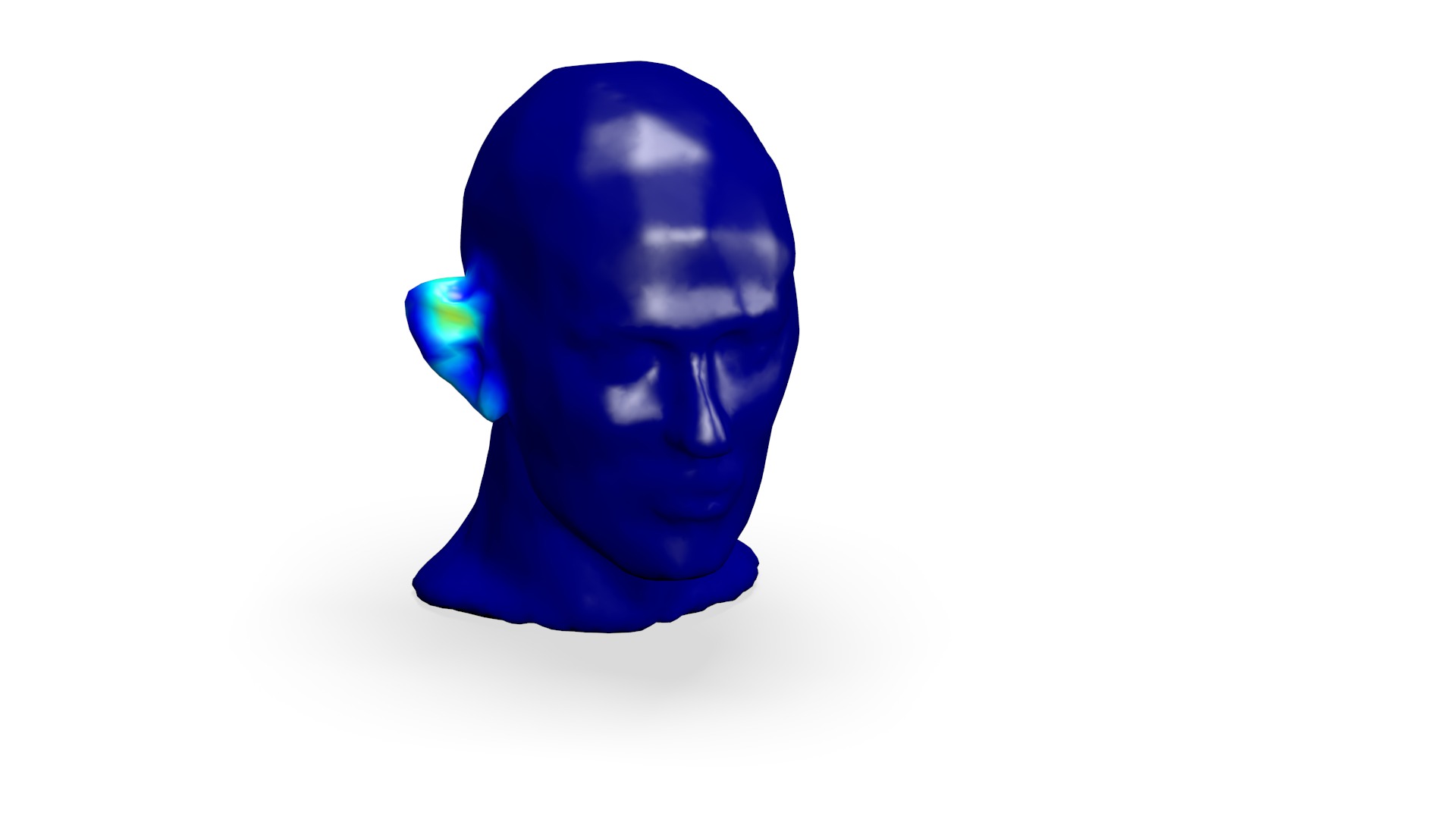}
	\end{overpic}
	\hspace{-0.3cm}
	\begin{overpic}
	[trim=7cm 2cm 14cm 2cm,clip,clip,width=0.14\textwidth]{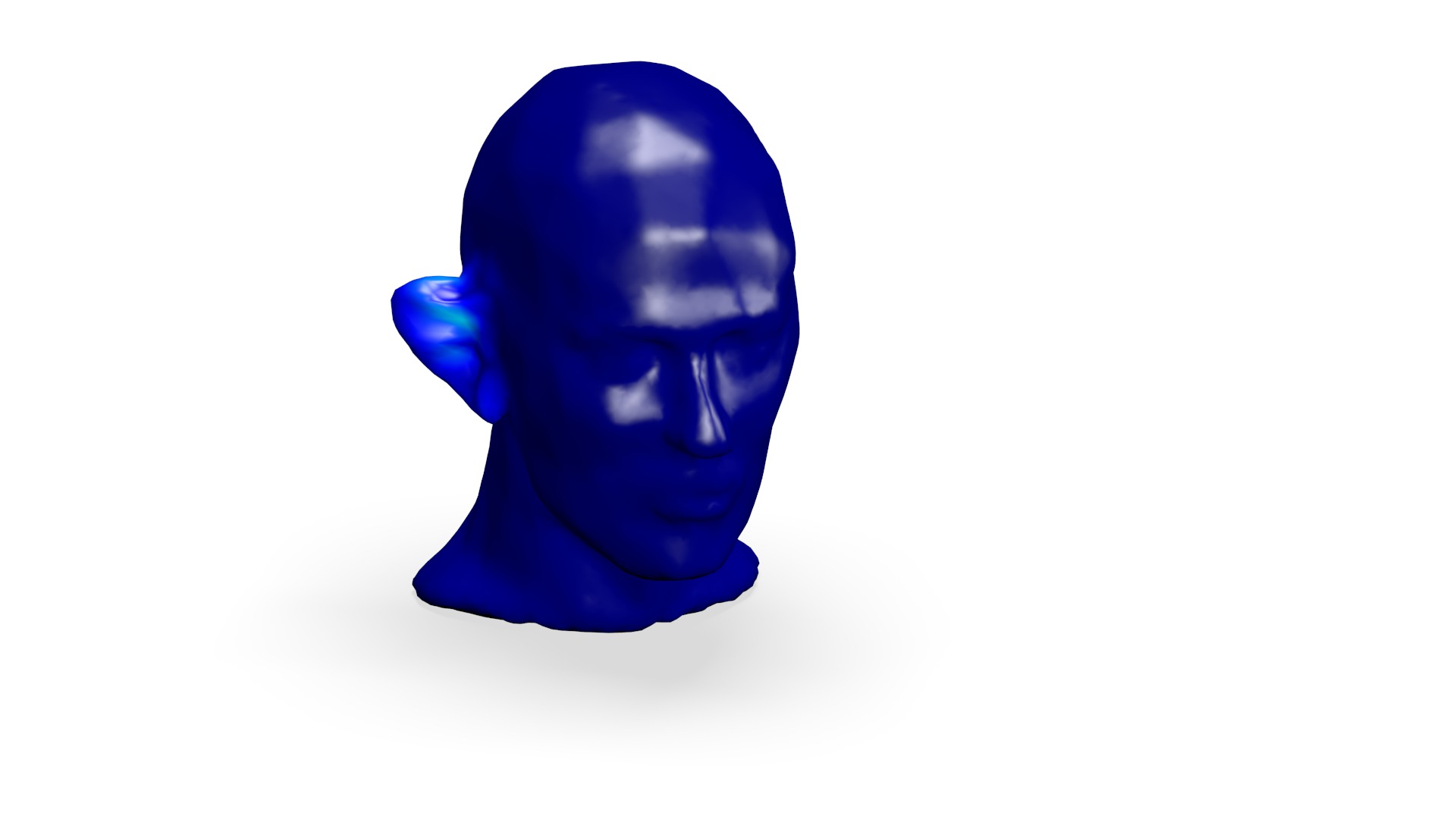}
	\end{overpic}
	\hspace{0.6cm}
	\begin{overpic}
	[trim=6cm 2cm 15cm 2cm,clip,clip,width=0.14\textwidth]{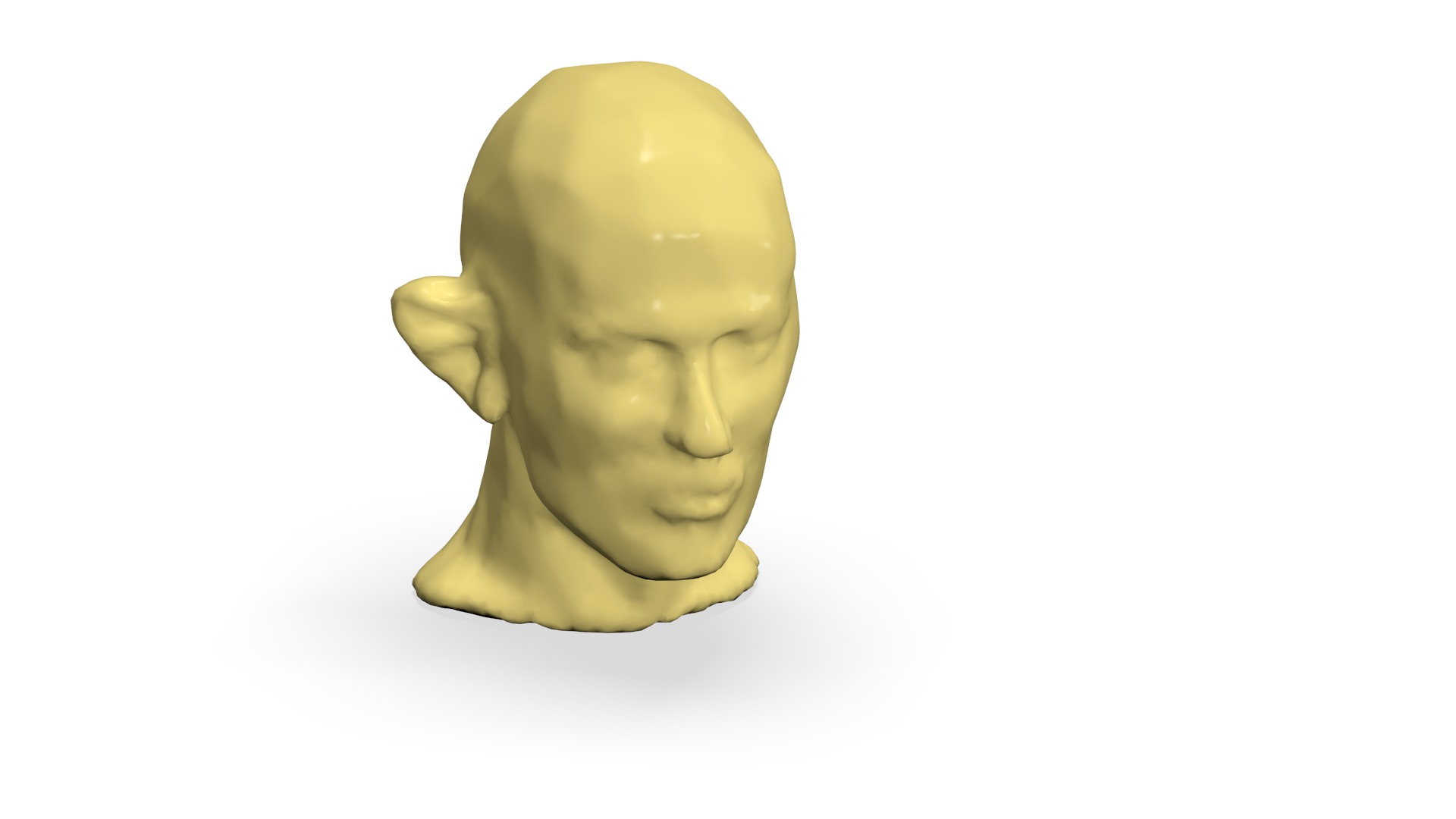}
	\end{overpic}
\end{minipage} \\ \vspace{0.1cm}
%
%
%
%
\begin{minipage}{\textwidth}
	\hspace{-0.3cm}
	\begin{overpic}
	[trim=8cm 2cm 13cm 2cm,clip,width=0.14\textwidth]{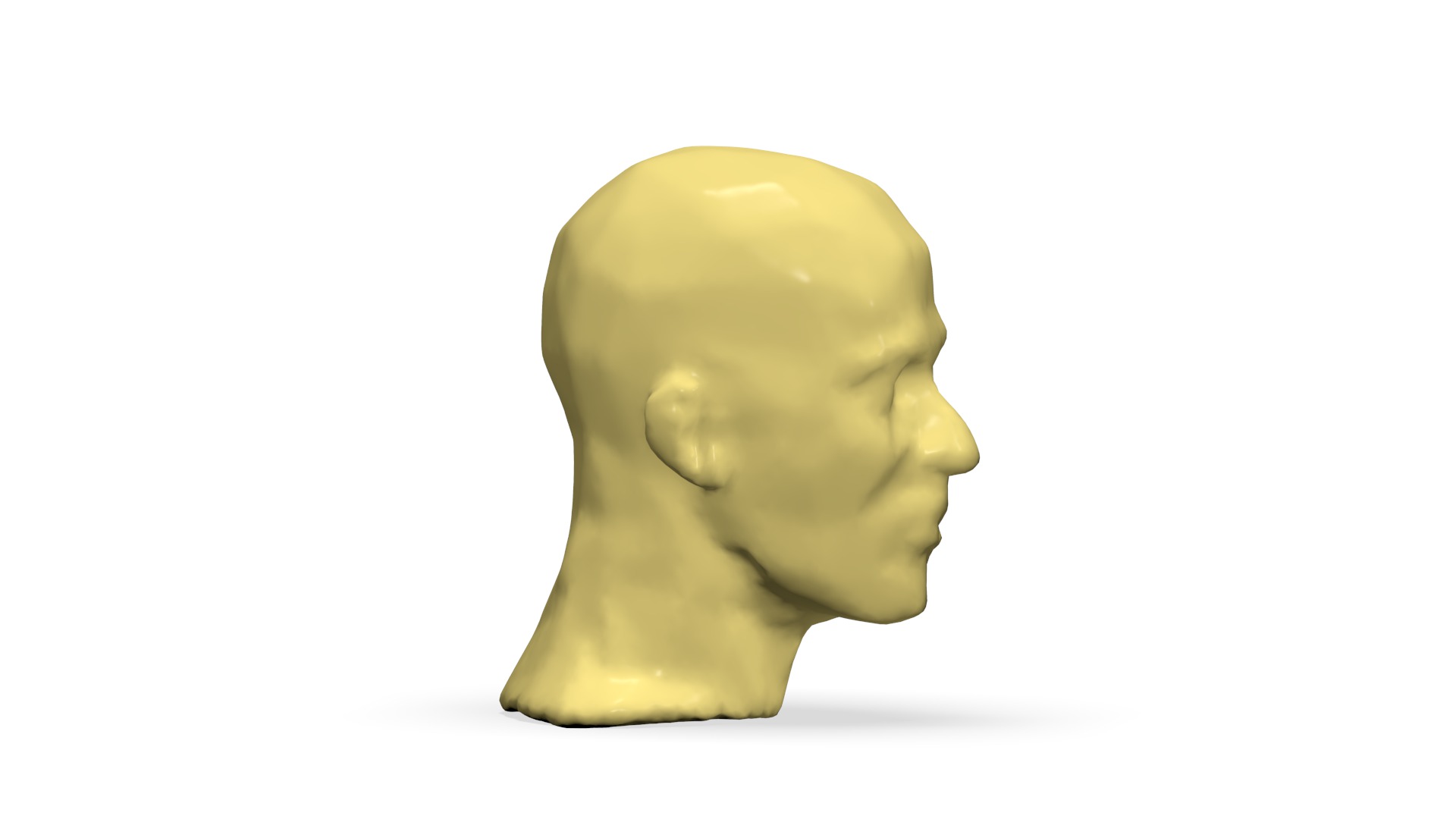}
	\end{overpic} 
	\begin{overpic}
	[trim=10cm 2cm 11cm 2cm,clip,width=0.14\textwidth]{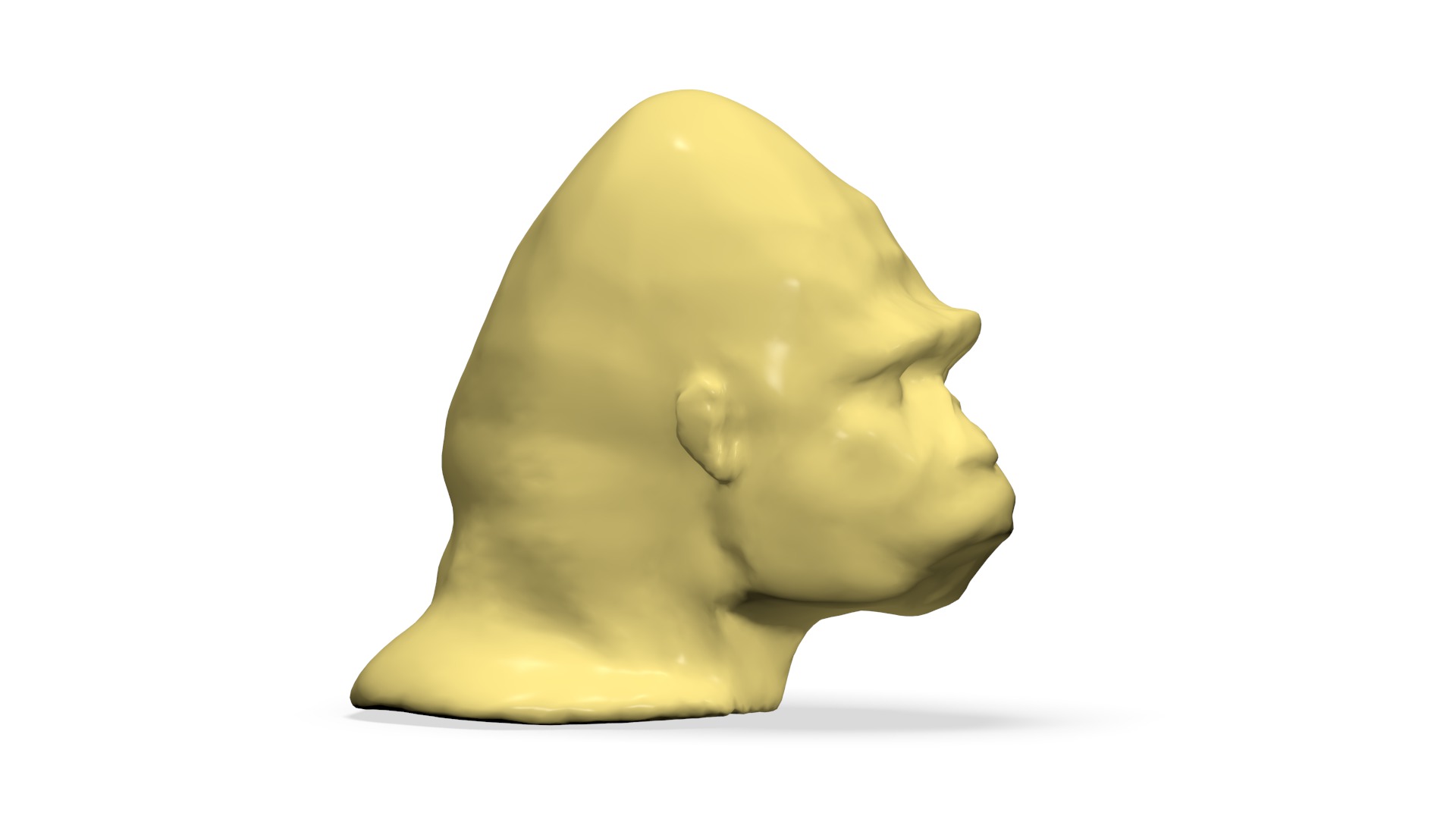}
	\end{overpic}
%
	\begin{overpic}
	[trim=14cm 2cm 7cm 2cm,clip,width=0.14\textwidth]{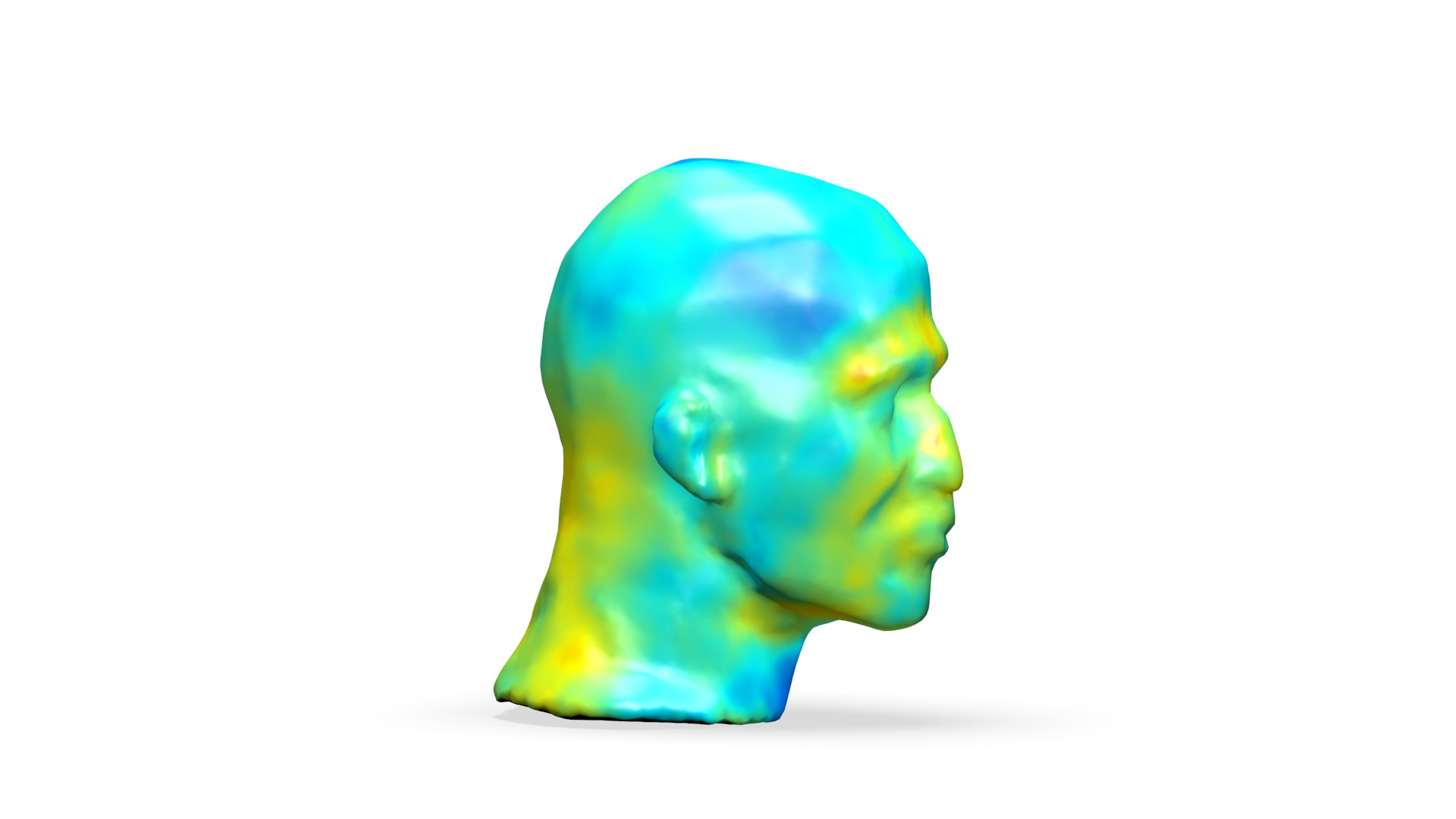}
	\end{overpic} 
	\hspace{-0.3cm}
	\begin{overpic}
	[trim=14cm 2cm 7cm 2cm,clip,clip,width=0.14\textwidth]{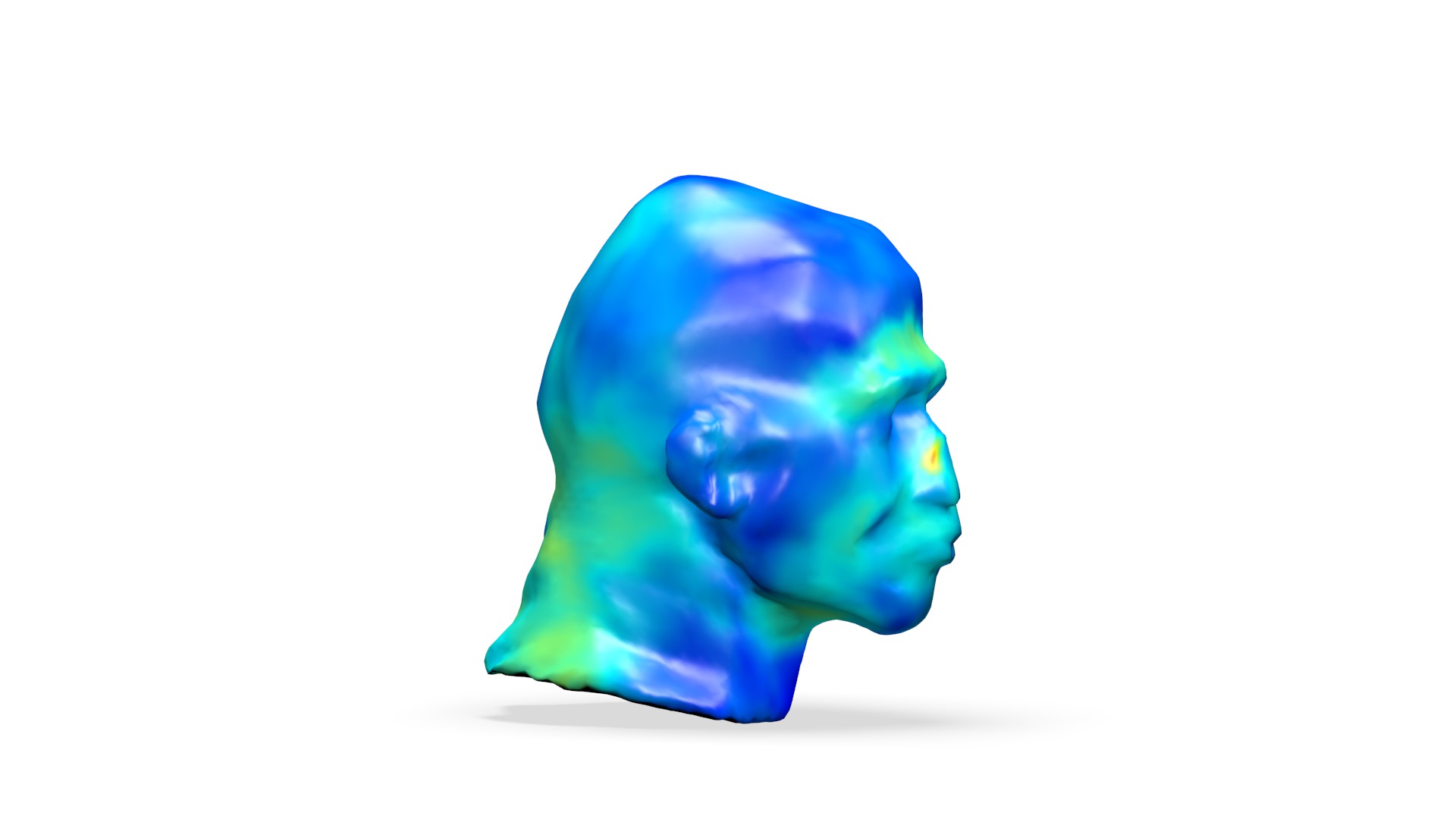}
	\end{overpic}
	\hspace{-0.3cm}
	\begin{overpic}
	[trim=14cm 2cm 7cm 2cm,clip,width=0.14\textwidth]{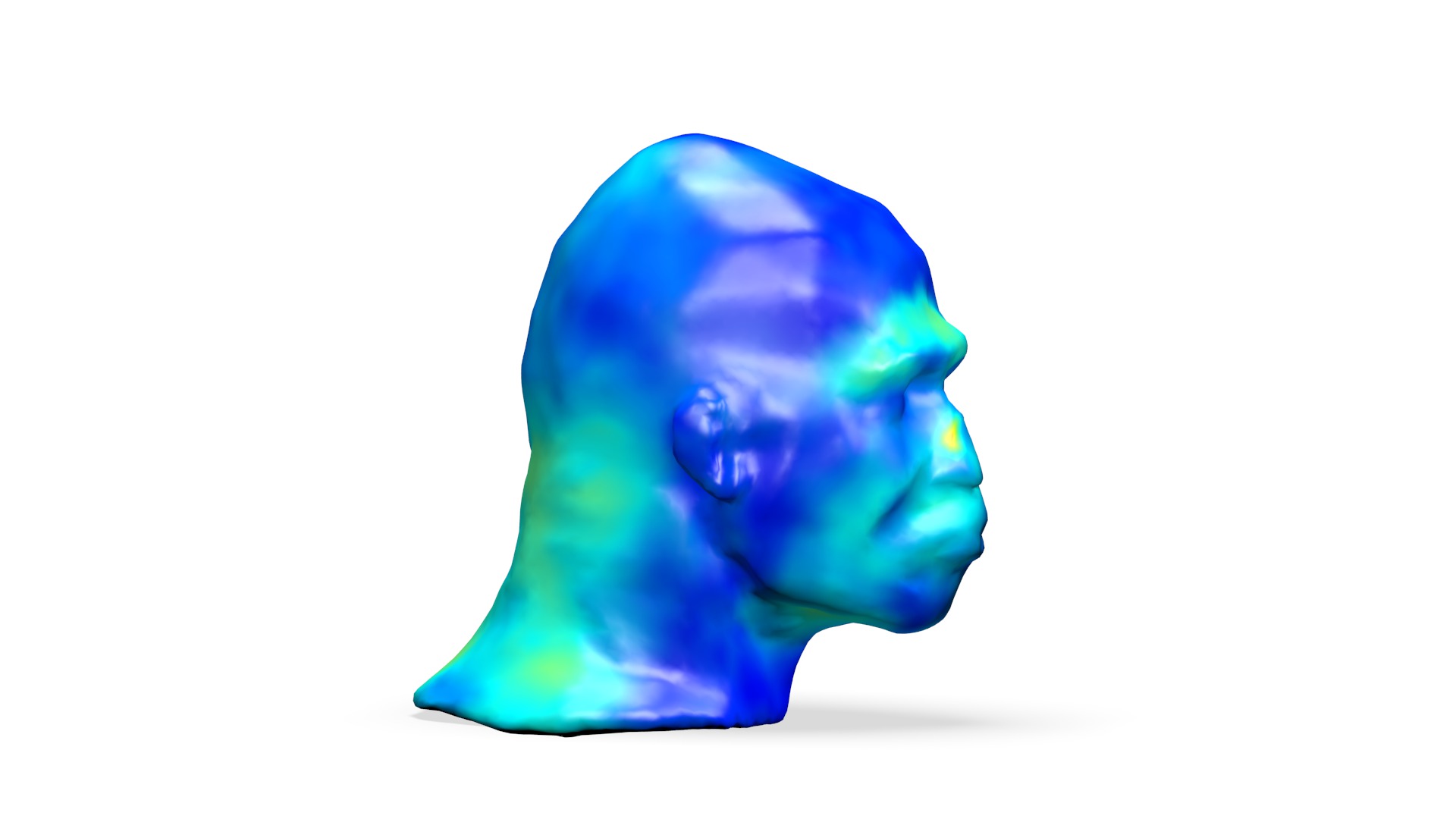}
	\end{overpic}
	\hspace{-0.3cm}
	\begin{overpic}
	[trim=12cm 2cm 7cm 2cm,clip,clip,width=0.14\textwidth]{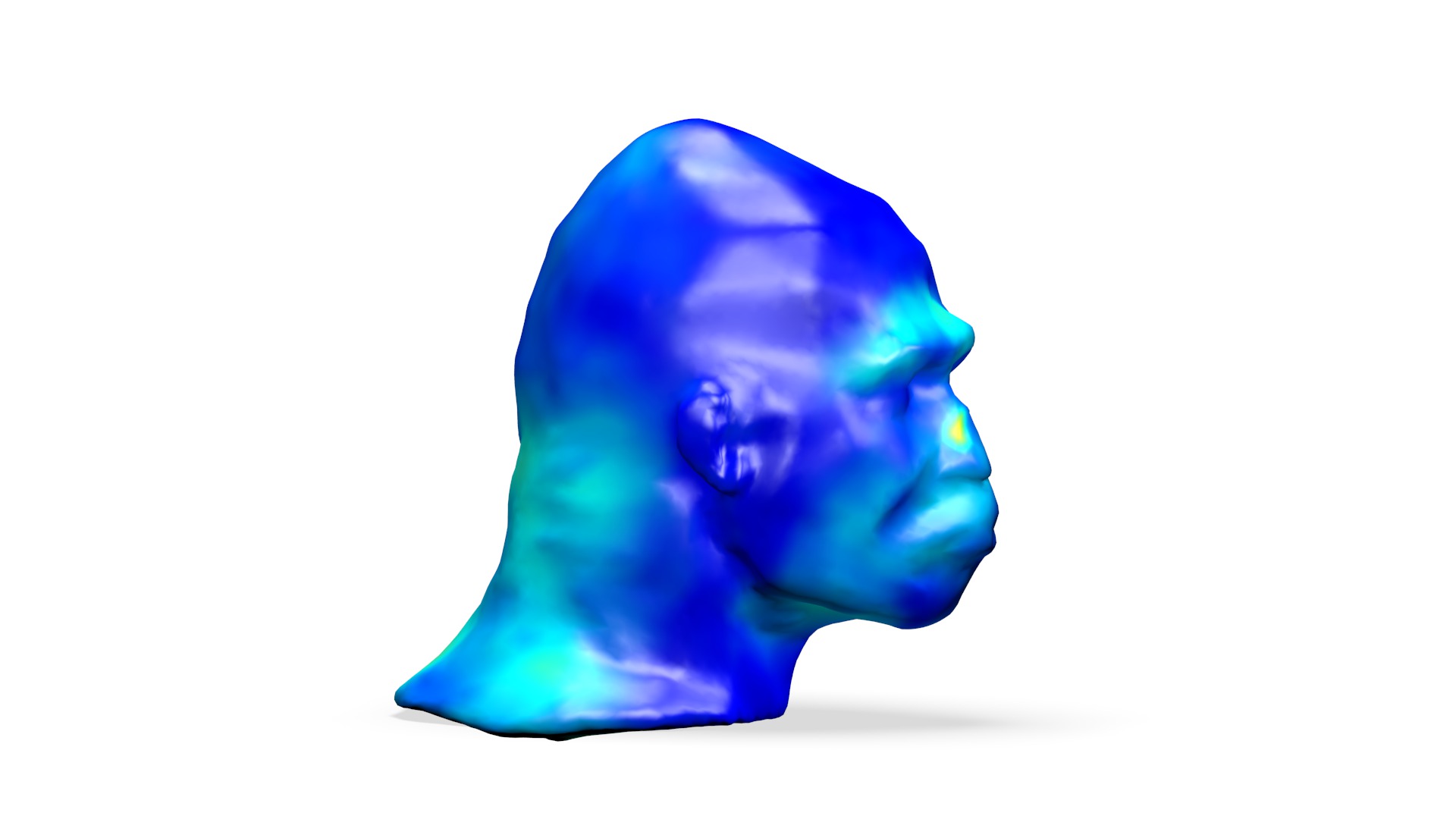}
	\end{overpic}
	\hspace{0.3cm}
	\begin{overpic}
	[trim=2cm 2cm 18cm 2cm,clip,clip,width=0.14\textwidth]{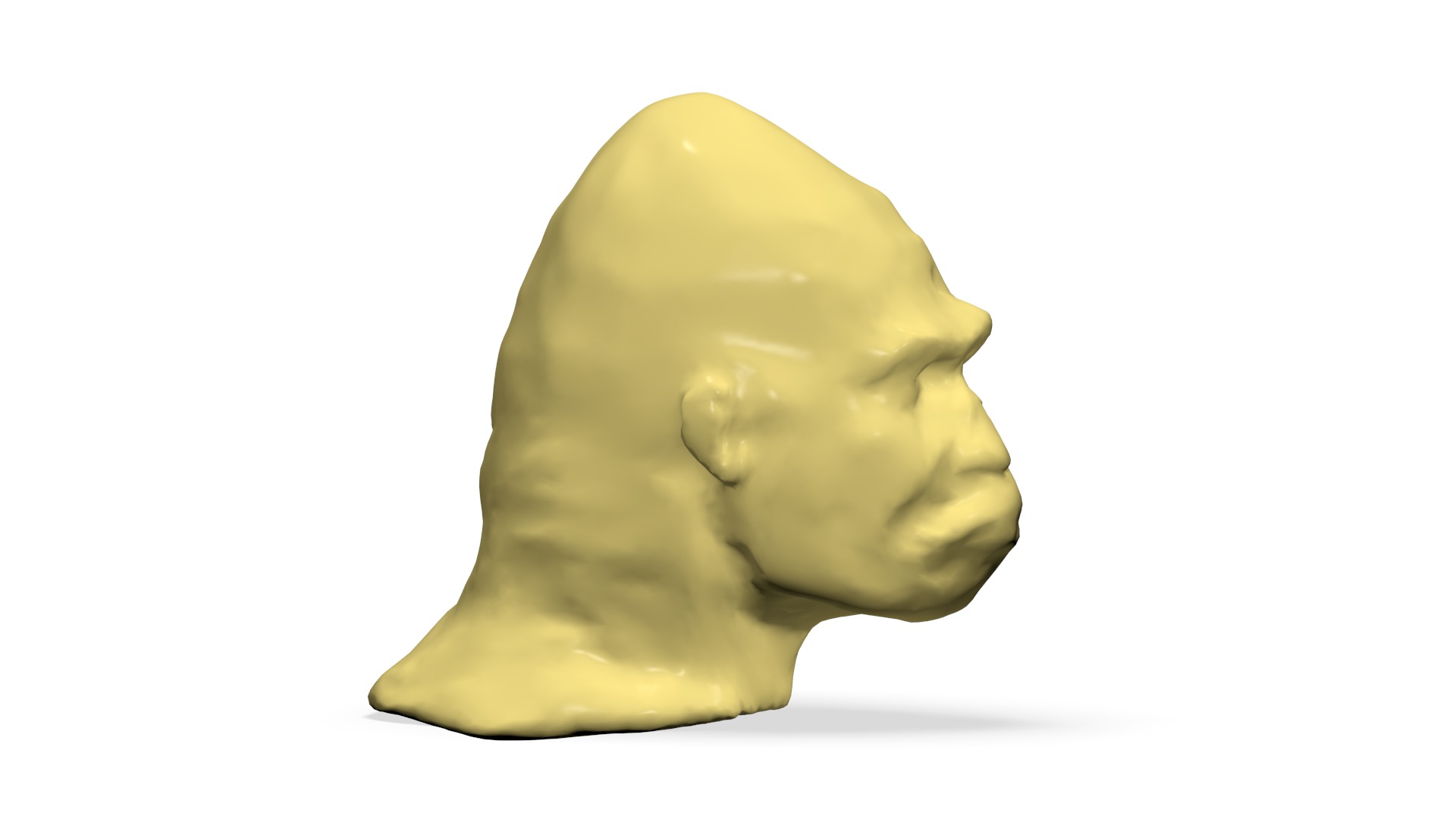}
	\put(32.5,-30){output}
	\end{overpic}
\end{minipage} \\ \vspace{0.5cm}
%
%
%
%
\begin{minipage}{\textwidth}
	\hspace{-0.5cm}
	\begin{overpic}
	[trim=12cm 1cm 18cm 1cm,clip,width=0.14\textwidth]{./figures/man_woman/man}
	\end{overpic} 
	\hspace{-0.2cm}
	\begin{overpic}
	[trim=12cm 1cm 18cm 1cm,clip,width=0.14\textwidth]{./figures/man_woman/woman}
	\end{overpic}
	\hspace{0.1cm}
	\begin{overpic}
	[trim=18cm 1cm 12cm 1cm,clip,width=0.14\textwidth]{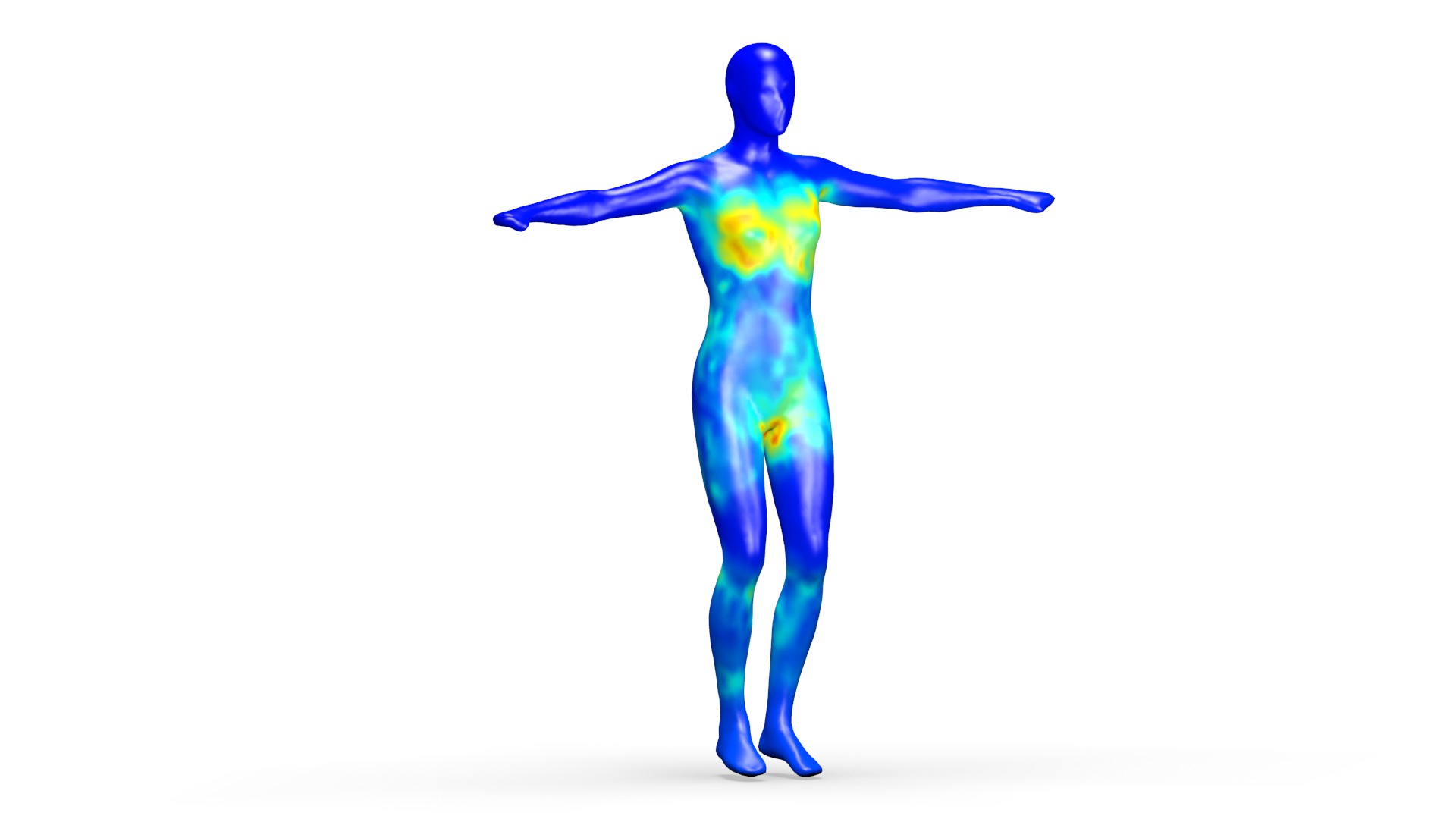}
	\end{overpic} 
	\hspace{-0.2cm}
	\begin{overpic}
	[trim=18cm 1cm 12cm 1cm,clip,clip,width=0.14\textwidth]{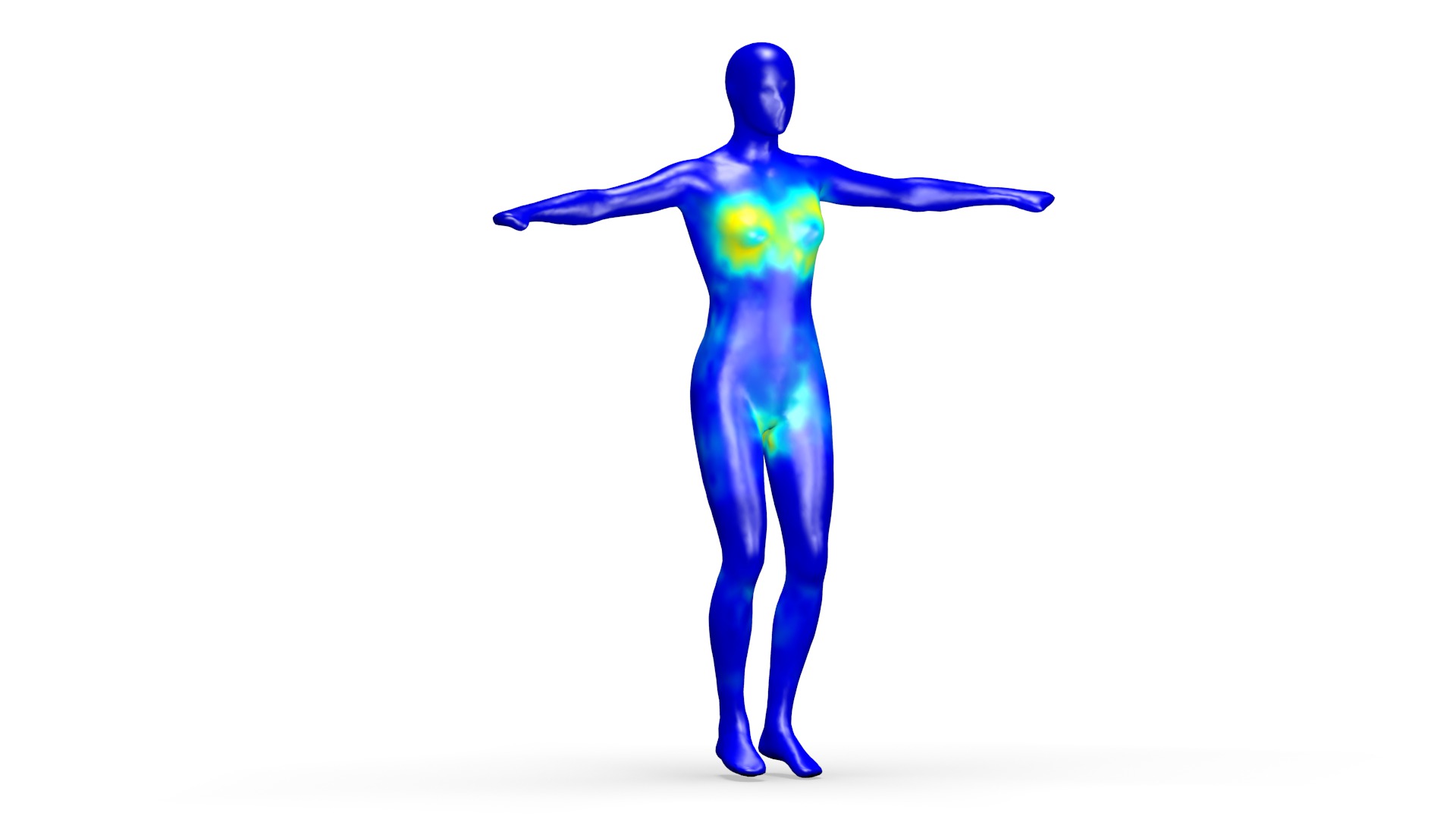}
	\end{overpic}
	\hspace{-0.2cm}
	\begin{overpic}
	[trim=18cm 1cm 12cm 1cm,clip,width=0.14\textwidth]{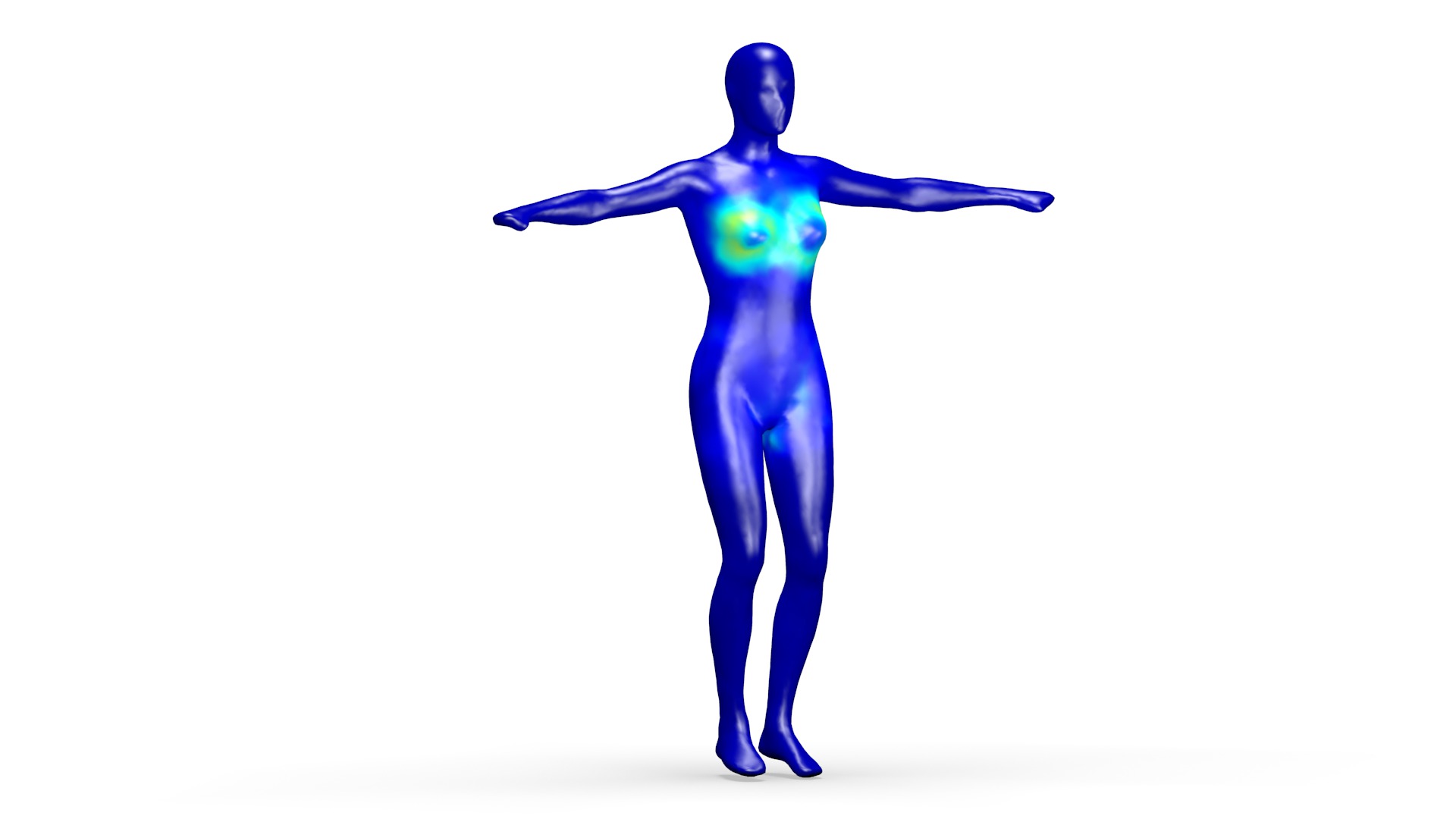}
	\end{overpic}
	\hspace{-0.2cm}
	\begin{overpic}
	[trim=18cm 1cm 12cm 1cm,clip,clip,width=0.14\textwidth]{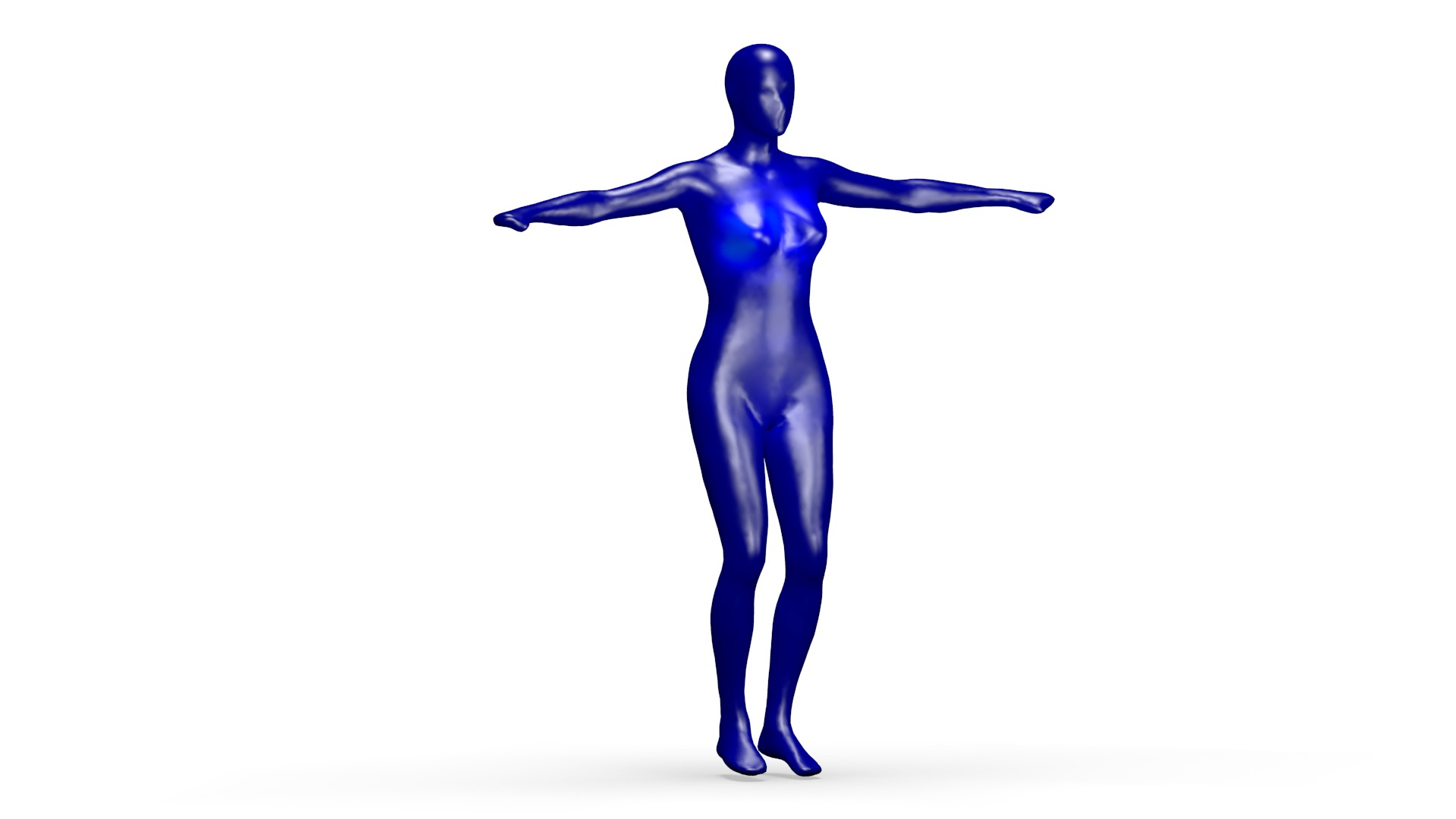}
	\end{overpic}
	\hspace{0.2cm}
	\begin{overpic}
	[trim=12cm 1cm 18cm 1cm,clip,clip,width=0.14\textwidth]{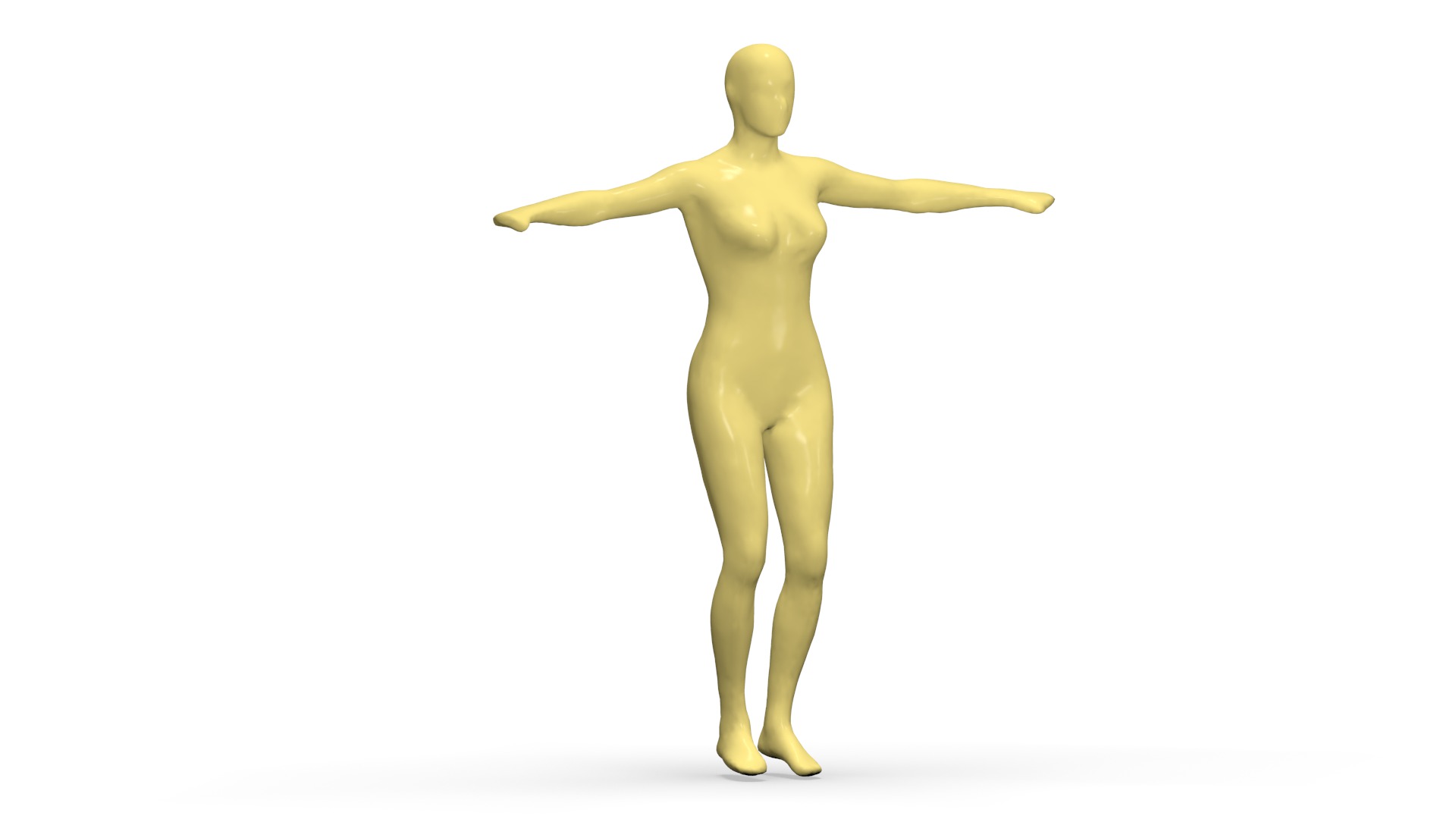}
	\end{overpic}
\end{minipage} \\ \vspace{0.4cm}
%
%
%
%
\begin{minipage}{\textwidth}
	\hspace{-0.5cm}
	\begin{overpic}
	[trim=12cm 1cm 18cm 1cm,clip,width=0.14\textwidth]{./figures/man_woman/man_160}
	\end{overpic} 
	\hspace{-0.2cm}
	\begin{overpic}
	[trim=12cm 1cm 18cm 1cm,clip,width=0.14\textwidth]{./figures/man_woman/man_fat}
	\end{overpic}
	\hspace{0.1cm}
	\begin{overpic}
	[trim=18cm 1cm 12cm 1cm,clip,width=0.14\textwidth]{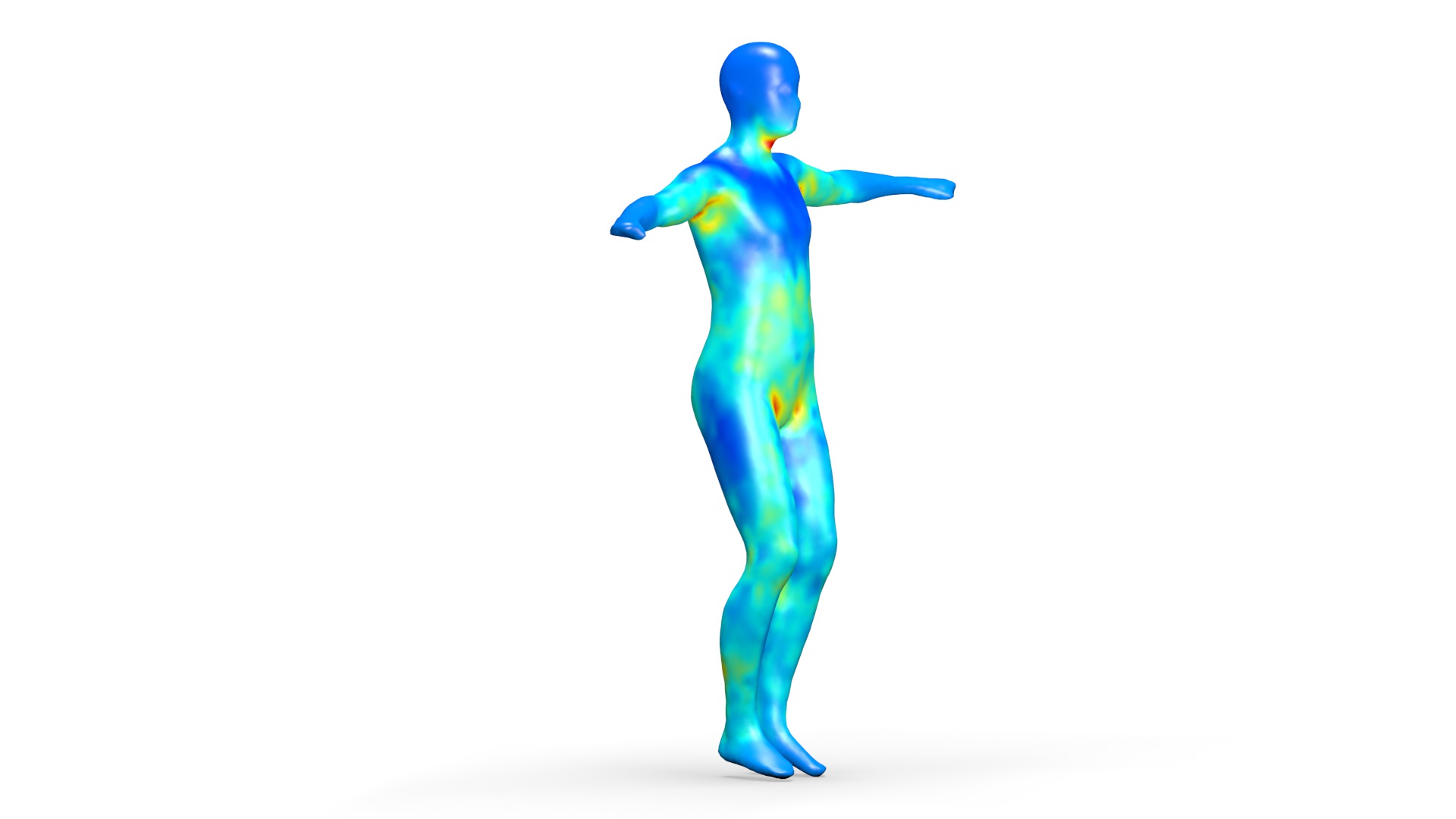}
	\end{overpic} 
	\hspace{-0.2cm}
	\begin{overpic}
	[trim=18cm 1cm 12cm 1cm,clip,clip,width=0.14\textwidth]{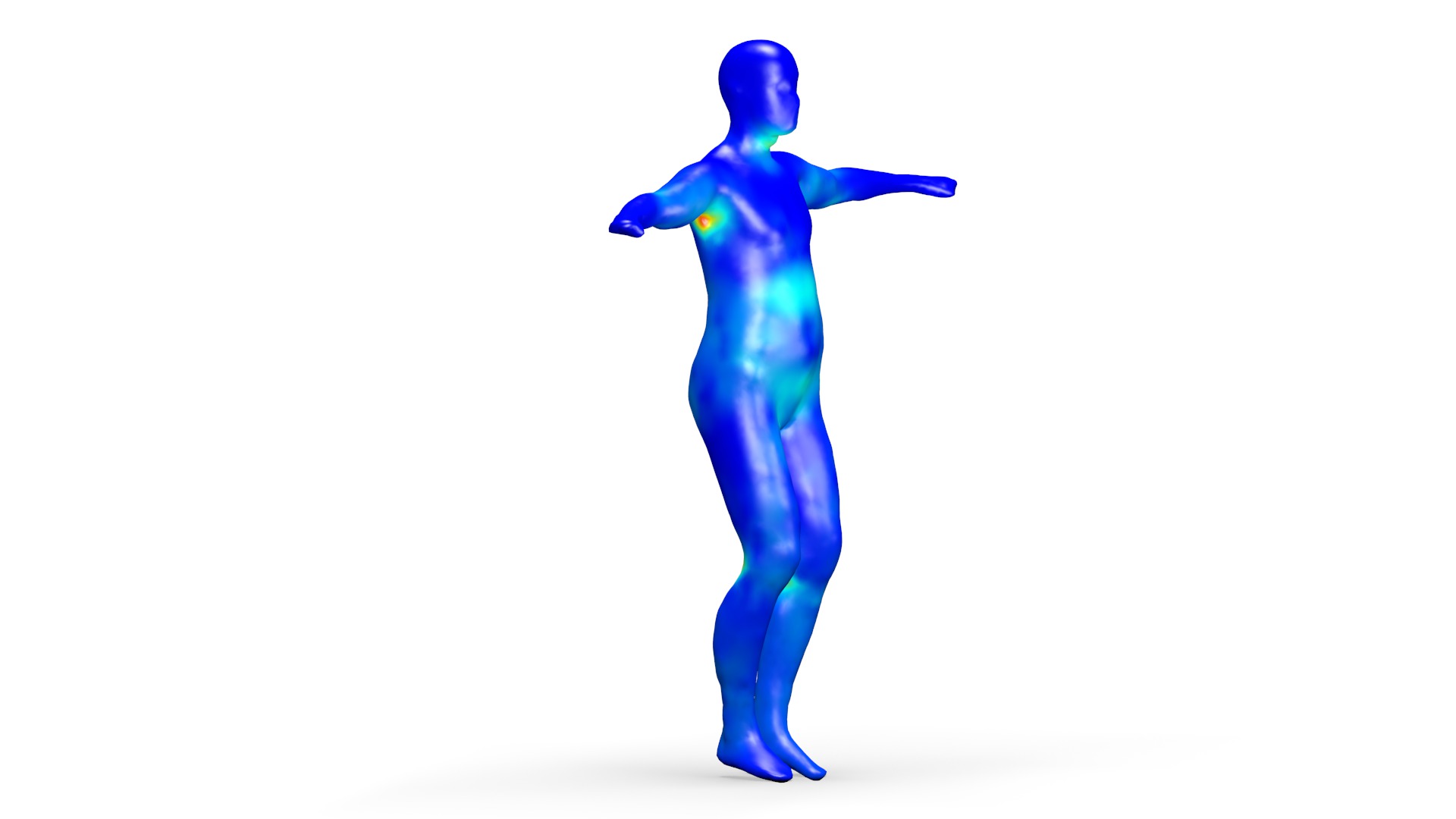}
	\end{overpic}
	\hspace{-0.2cm}
	\begin{overpic}
	[trim=18cm 1cm 12cm 1cm,clip,width=0.14\textwidth]{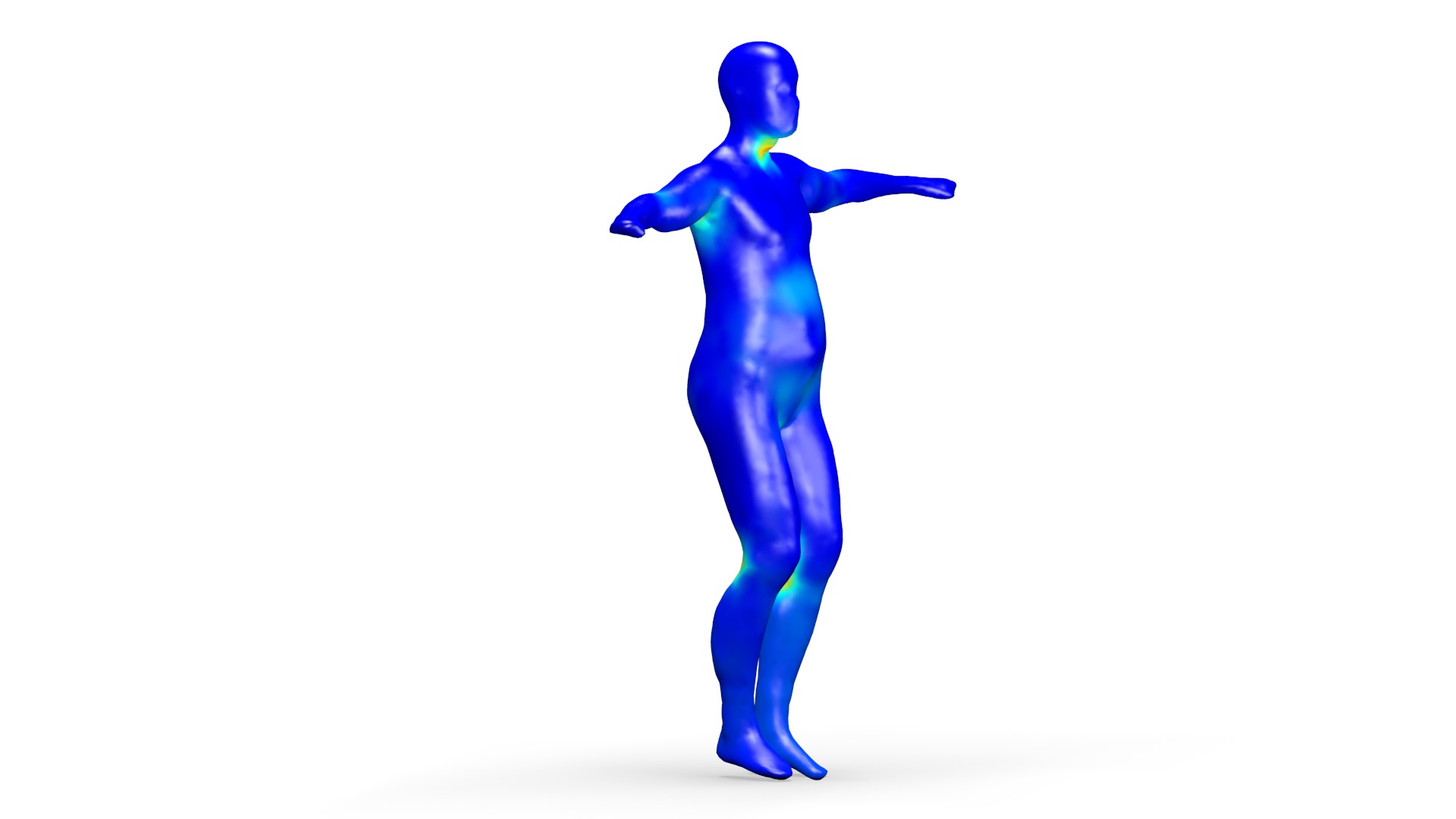}
	\end{overpic}
	\hspace{-0.2cm}
	\begin{overpic}
	[trim=18cm 1cm 12cm 1cm,clip,clip,width=0.14\textwidth]{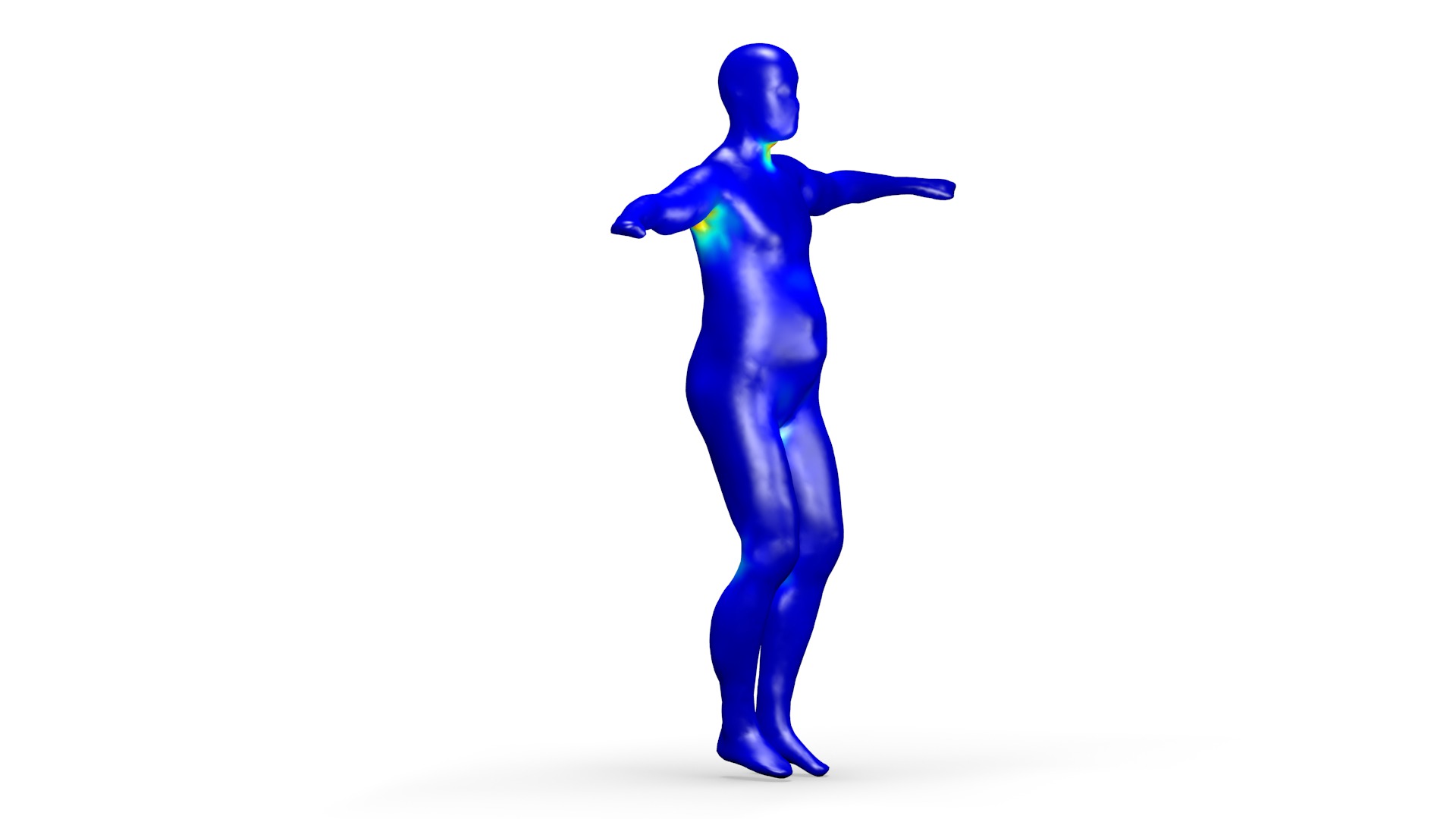}
	\end{overpic}
	\hspace{0.2cm}
	\begin{overpic}
	[trim=12cm 1cm 18cm 1cm,clip,clip,width=0.14\textwidth]{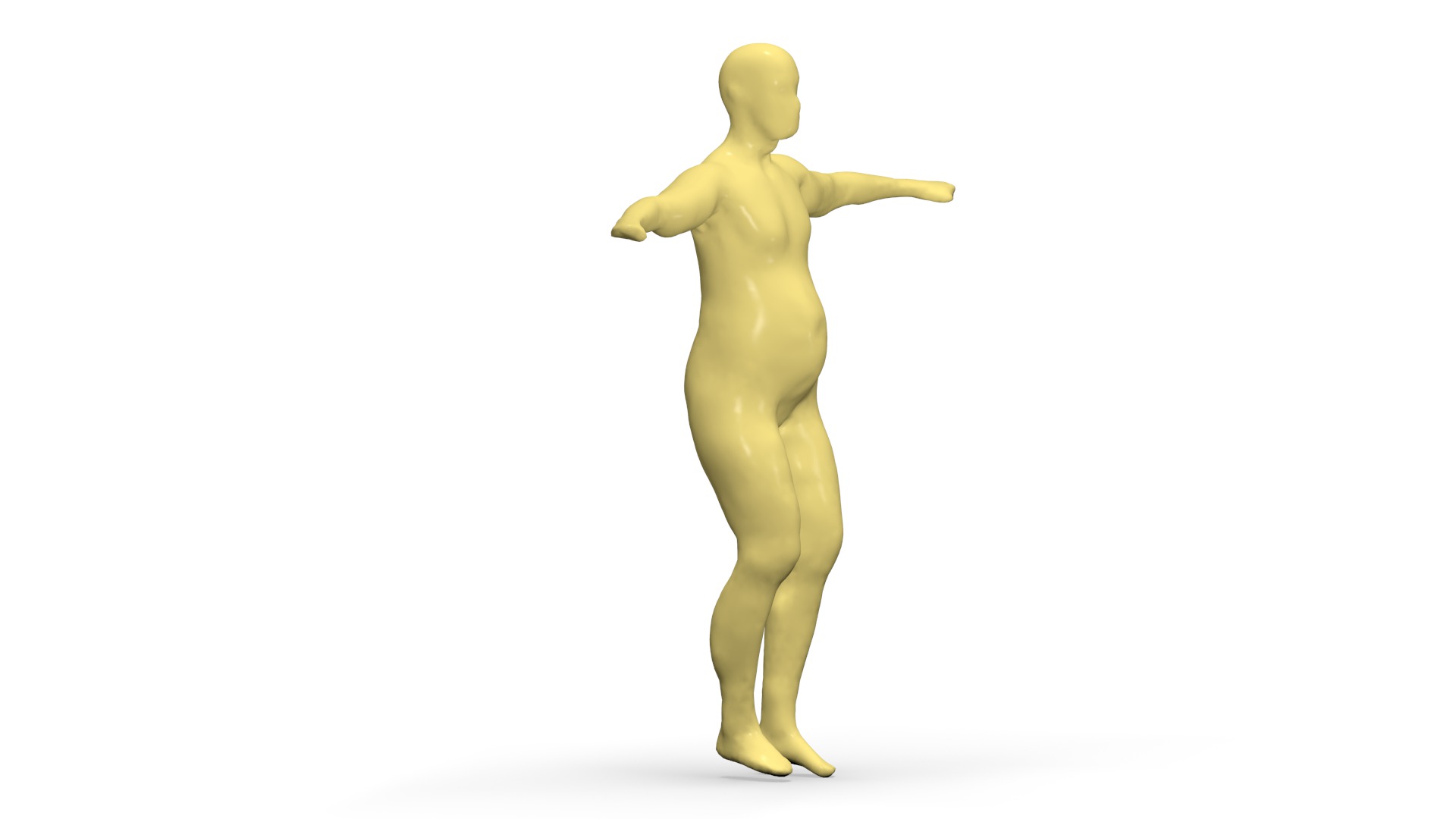}
	\end{overpic}
\end{minipage}  \\ \vspace{0.4cm}
%
%
%
%
\begin{minipage}{\textwidth}
	\hspace{-0.5cm}
	\begin{overpic}
	[trim=12cm 1cm 18cm 1cm,clip,width=0.14\textwidth]{./figures/man_woman/man_50}
	\end{overpic} 
	\hspace{-0.2cm}
	\begin{overpic}
	[trim=12cm 1cm 18cm 1cm,clip,width=0.14\textwidth]{./figures/man_woman/man_ciccione_pose1}
	\end{overpic}
	\hspace{0.1cm}
	\begin{overpic}
	[trim=18cm 1cm 12cm 1cm,clip,width=0.14\textwidth]{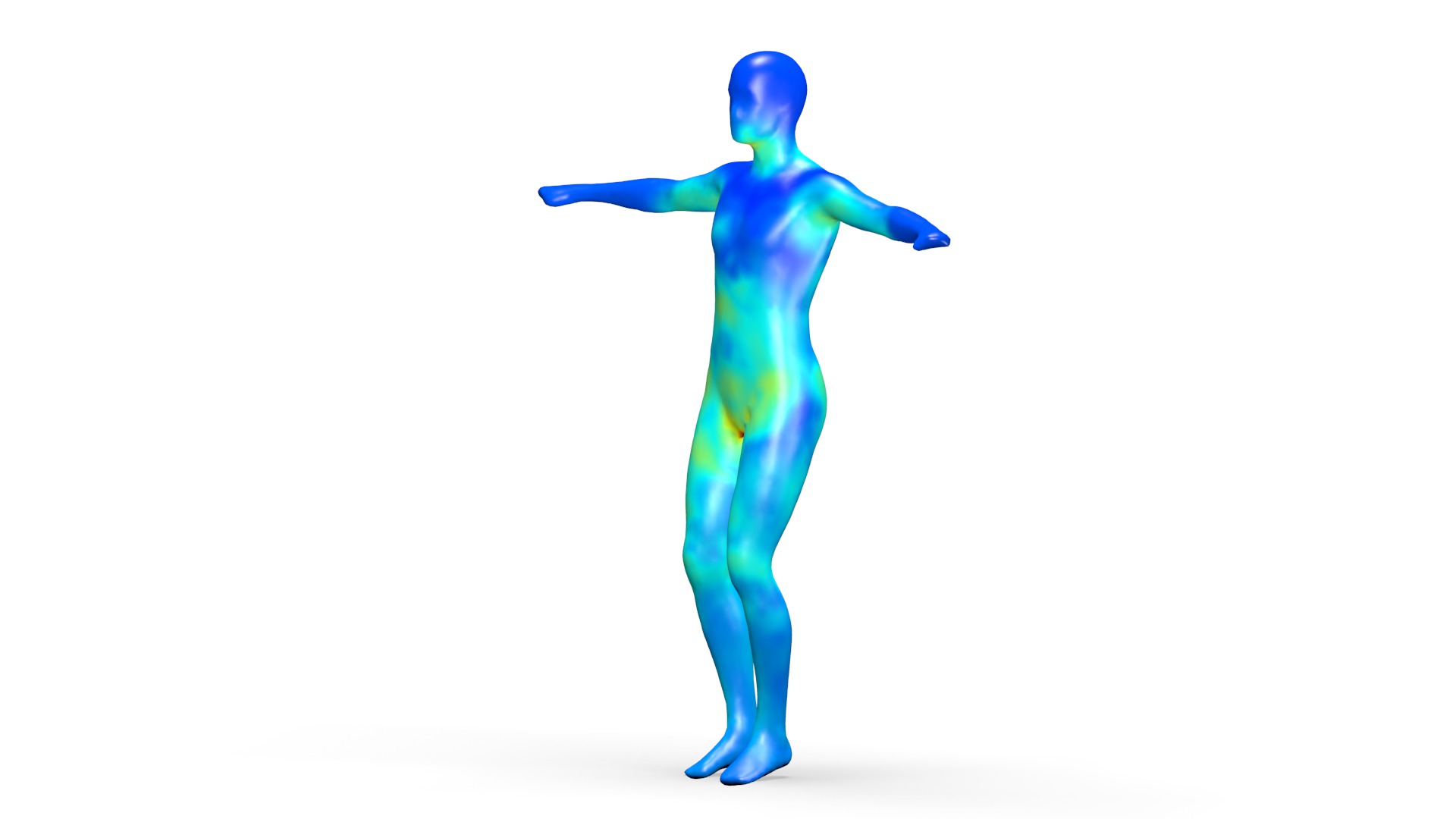}
	\end{overpic} 
	\hspace{-0.2cm}
	\begin{overpic}
	[trim=18cm 1cm 12cm 1cm,clip,clip,width=0.14\textwidth]{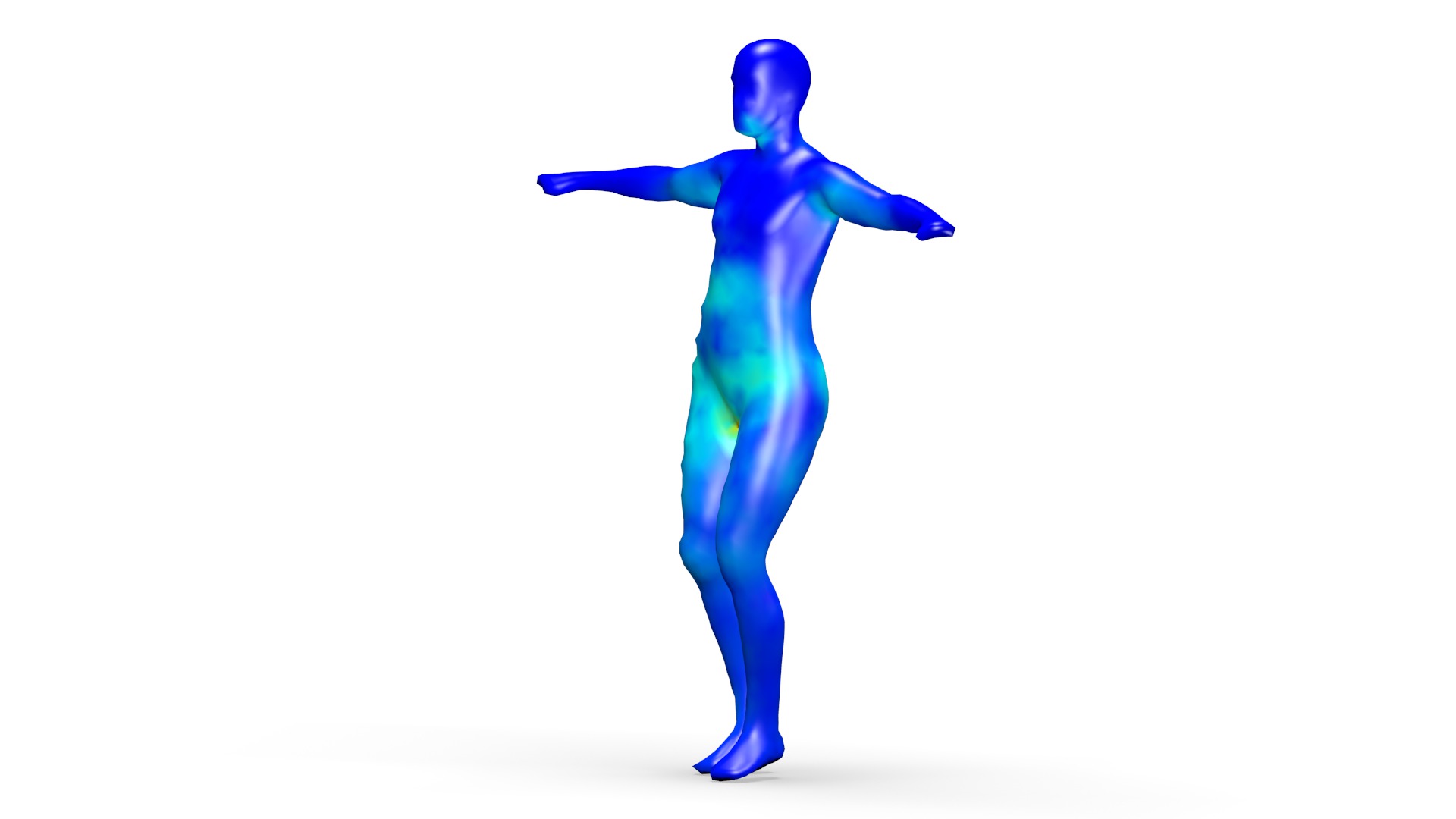}
	\end{overpic}
	\hspace{-0.2cm}
	\begin{overpic}
	[trim=18cm 1cm 12cm 1cm,clip,width=0.14\textwidth]{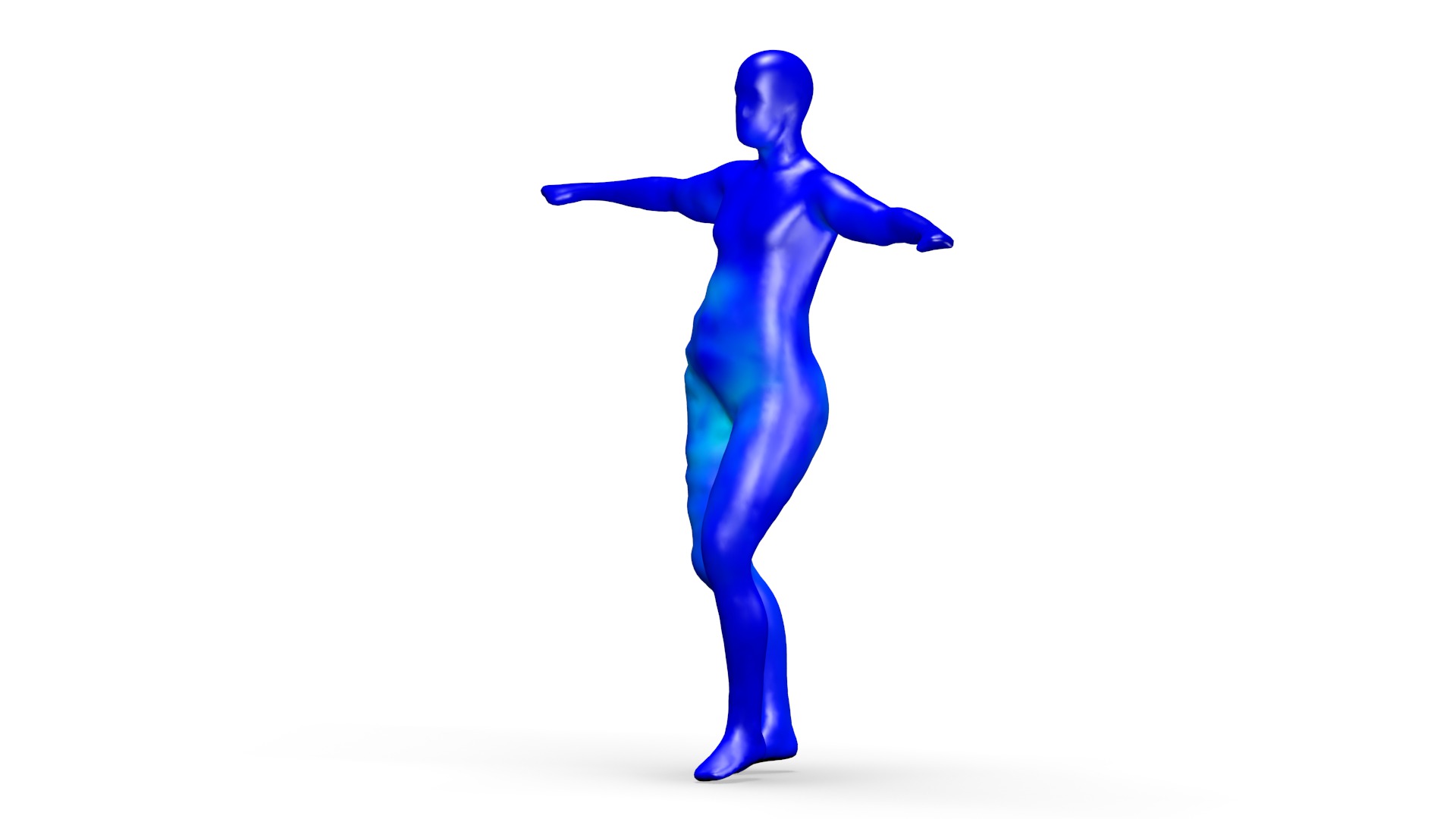}
	\end{overpic}
	\hspace{-0.2cm}
	\begin{overpic}
	[trim=18cm 1cm 12cm 1cm,clip,clip,width=0.14\textwidth]{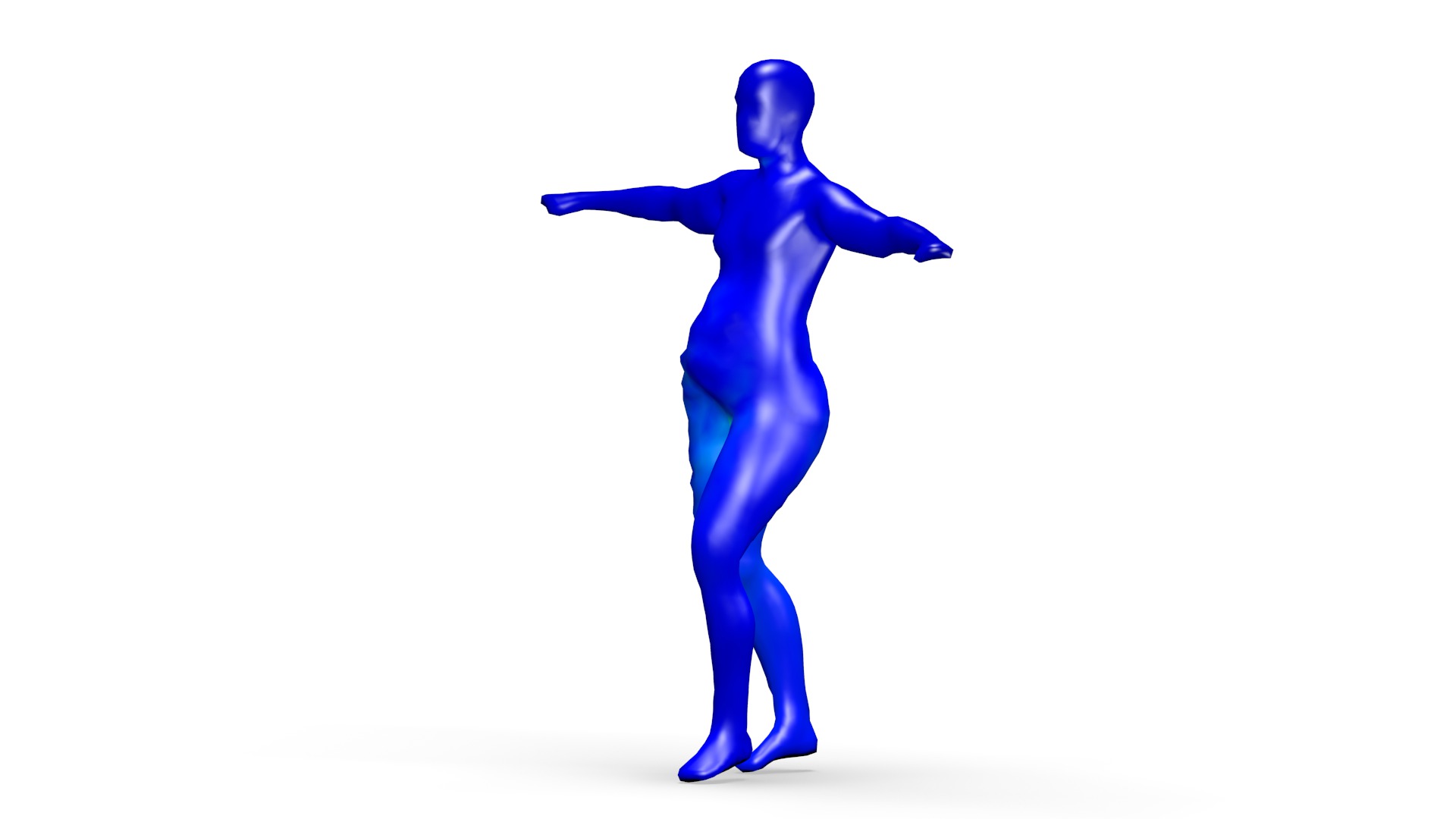}
	\end{overpic}
	\hspace{0.2cm}
	\begin{overpic}
	[trim=12cm 1cm 18cm 1cm,clip,clip,width=0.14\textwidth]{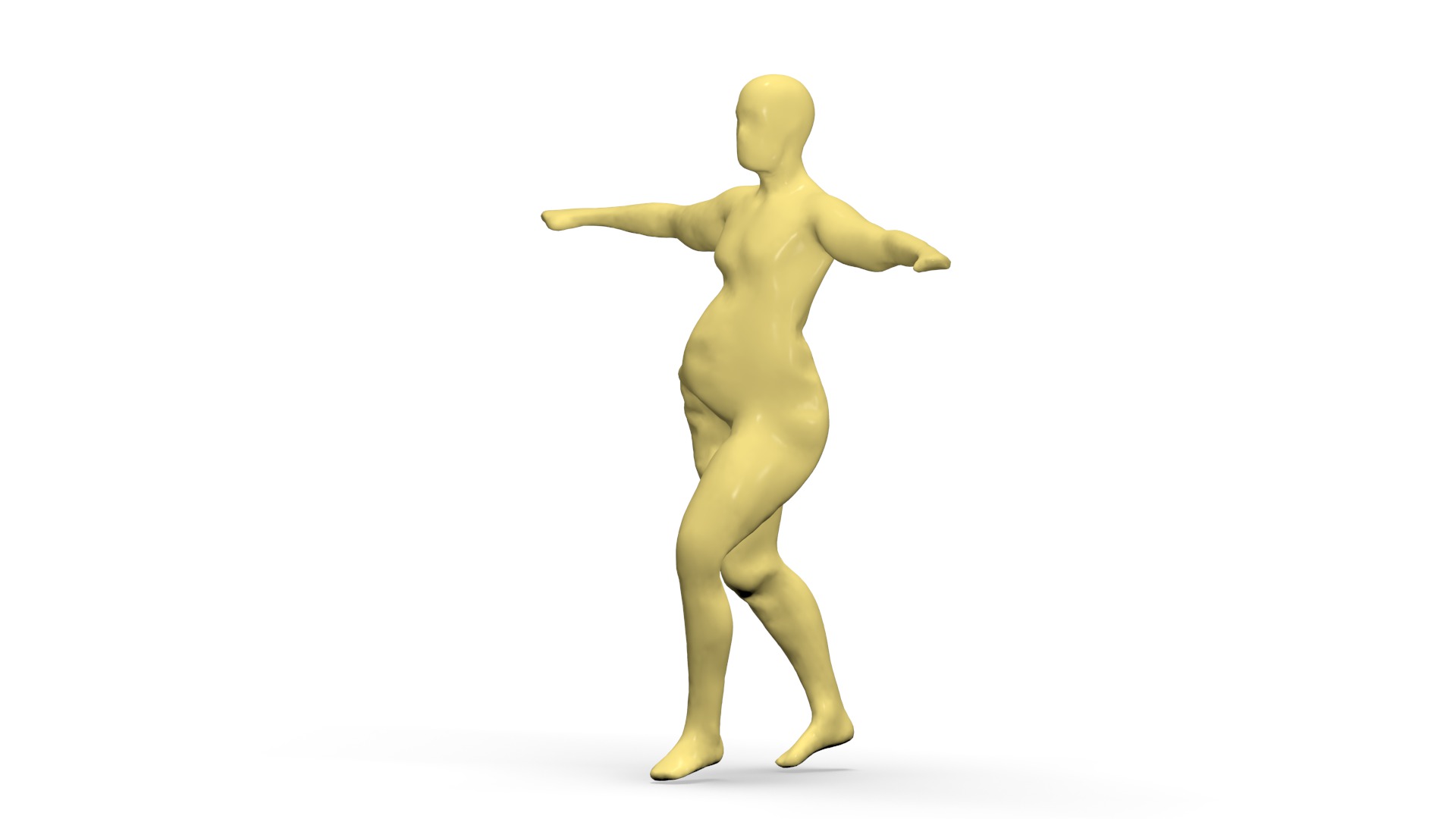}
	\end{overpic}
\end{minipage} \\ \vspace{0.2cm}
%
%
%
%
\begin{minipage}{\textwidth}
	\hspace{-0.3cm}
	\begin{overpic}
	[trim=12cm 1cm 18cm 6cm,clip,width=0.14\textwidth]{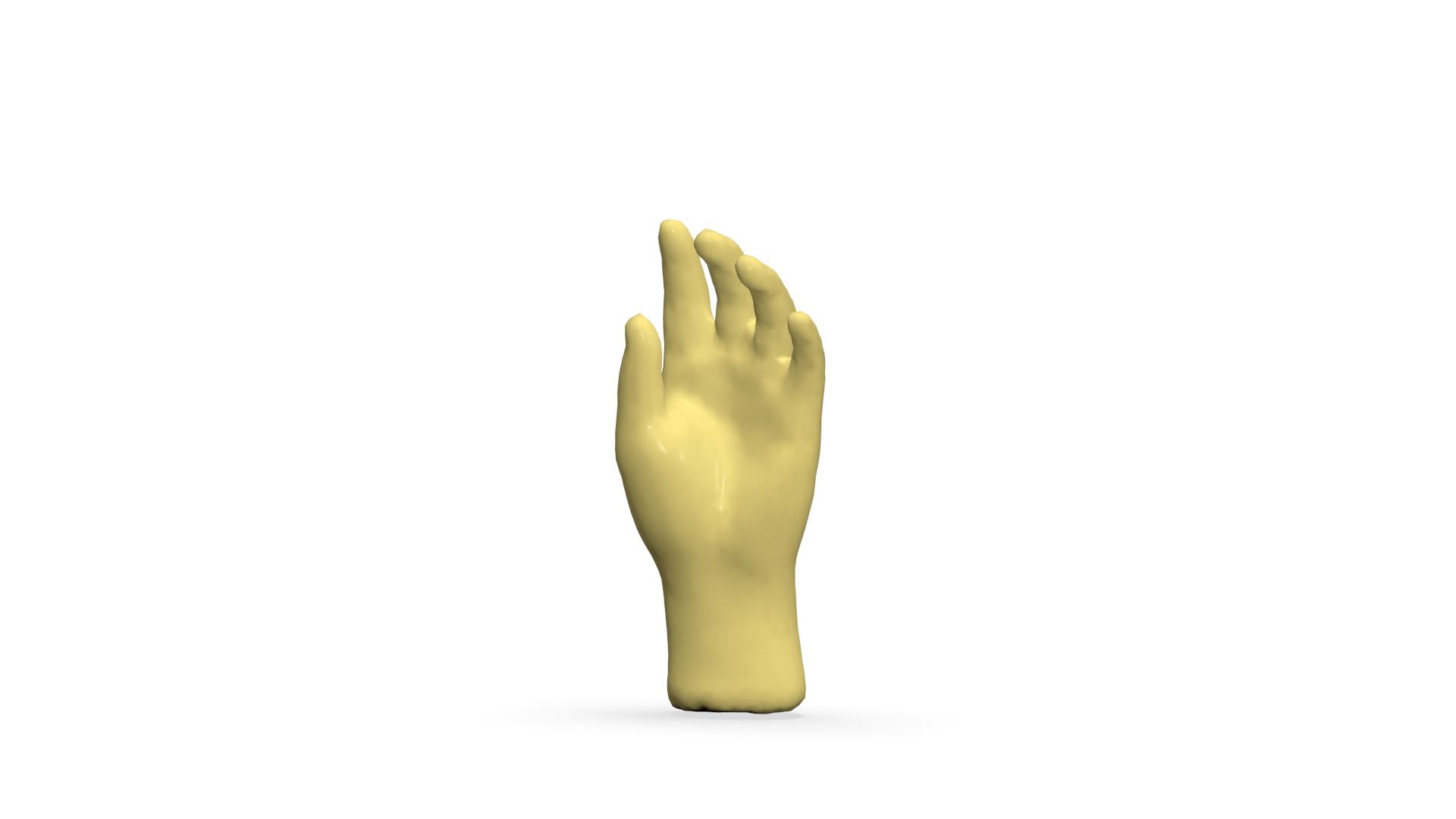}
	\put(55,-12){\fontsize{6}{8}\selectfont $A$}
	\end{overpic} 
	\hspace{-0.2cm}
	\begin{overpic}
	[trim=12cm 1cm 18cm 6cm,clip,width=0.14\textwidth]{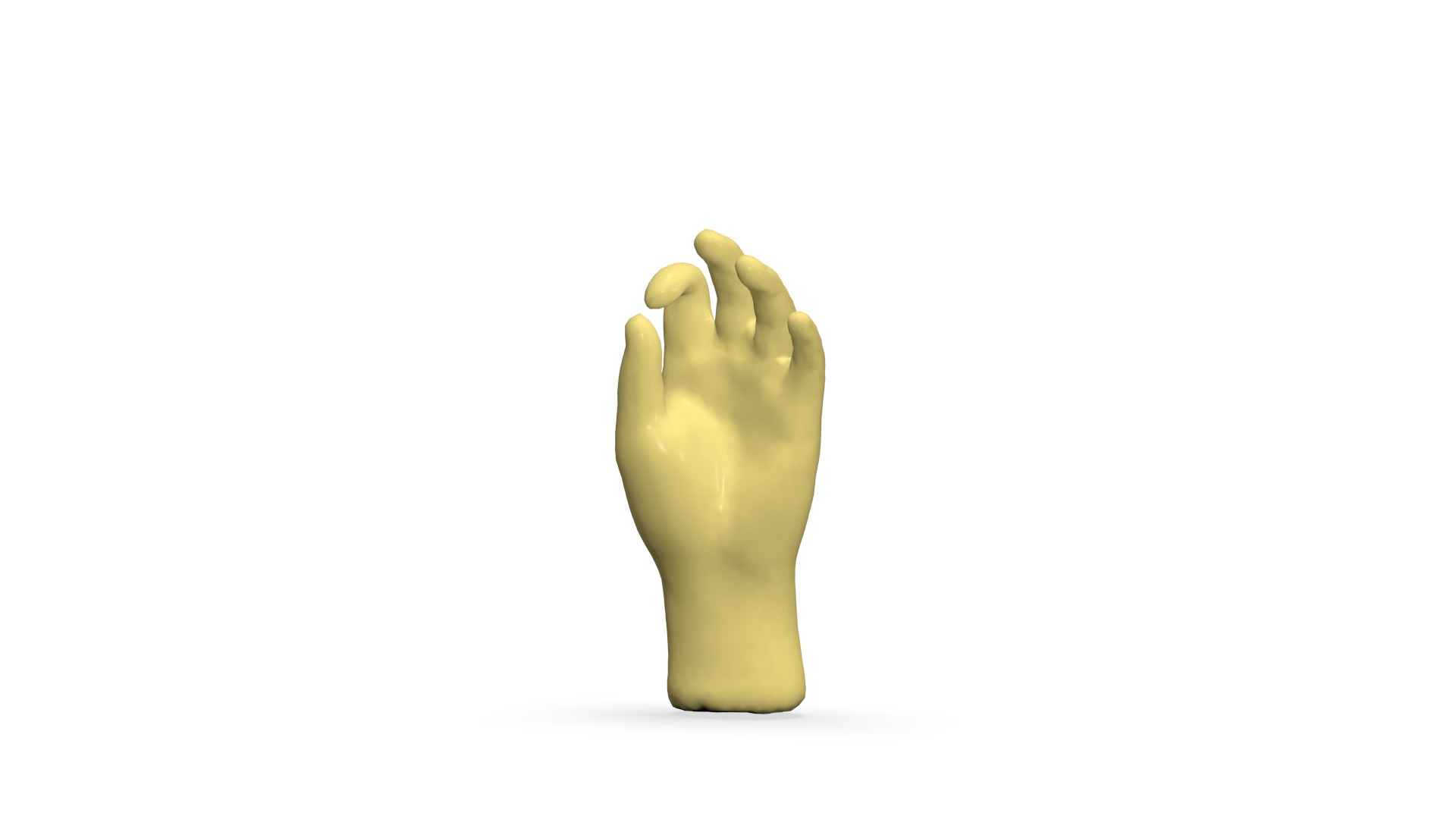}
	\put(55,-12){\fontsize{6}{8}\selectfont $B$}
	\end{overpic}
	\hspace{0.1cm}
	\begin{overpic}
	[trim=20cm 1cm 10cm 6cm,clip,width=0.14\textwidth]{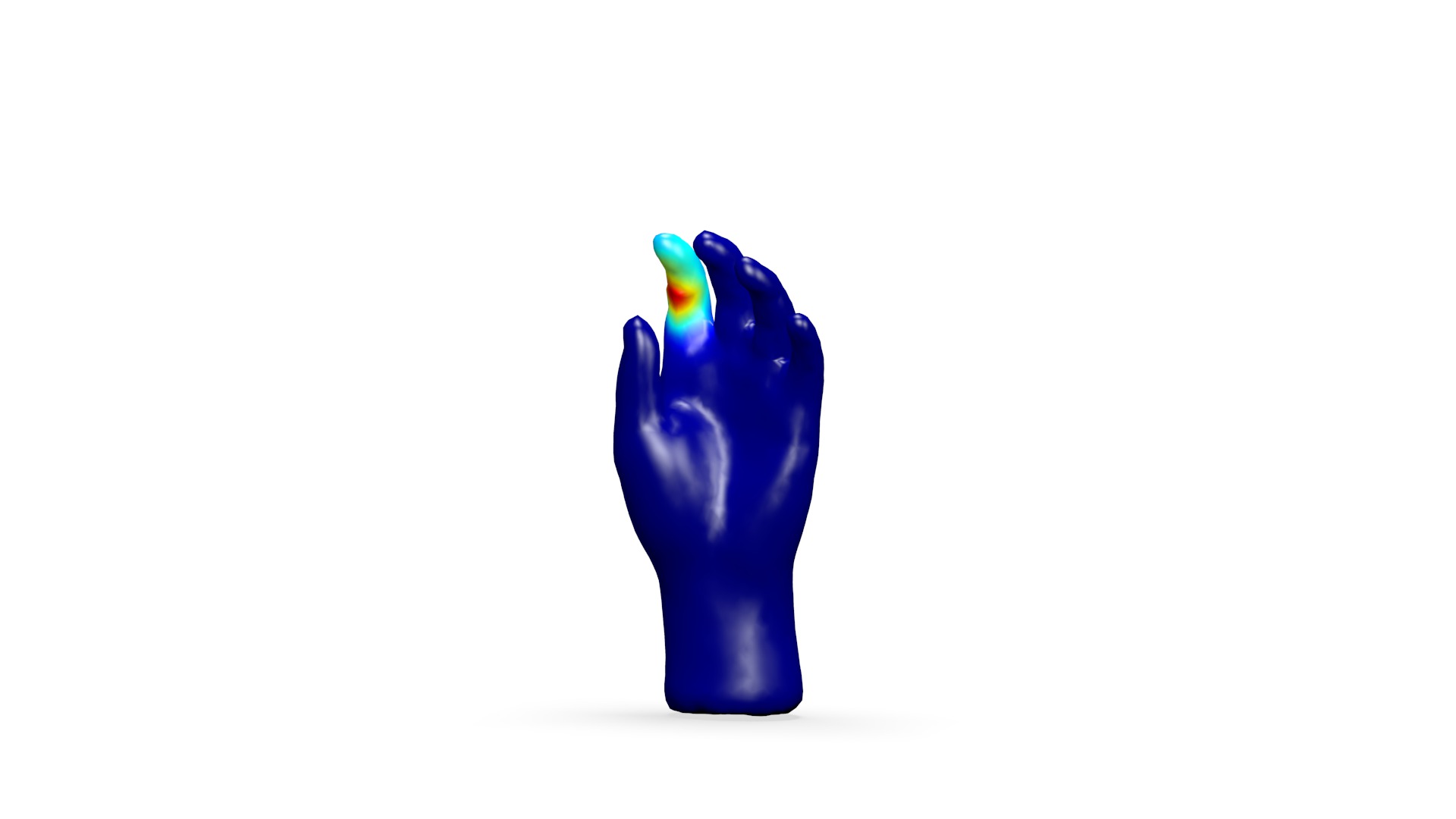}
	\put(38,0){\fontsize{6}{8}\selectfont $1$}
	\put(38,82){\fontsize{6}{8}\selectfont $1$}
	\put(38,194){\fontsize{6}{8}\selectfont $1$}
	\put(38,307){\fontsize{6}{8}\selectfont $1$}
	\put(30,425){\fontsize{6}{8}\selectfont $100$}
	\put(38,512){\fontsize{6}{8}\selectfont $1$}
	\put(38,595){\fontsize{6}{8}\selectfont $1$}
	\put(38,677){\fontsize{6}{8}\selectfont $1$}
	\put(-10,-12){\color{black}\line(0,1){770}}
	%
	%
	\end{overpic} 
	\hspace{-0.2cm}
	\begin{overpic}
	[trim=20cm 1cm 10cm 6cm,clip,clip,width=0.14\textwidth]{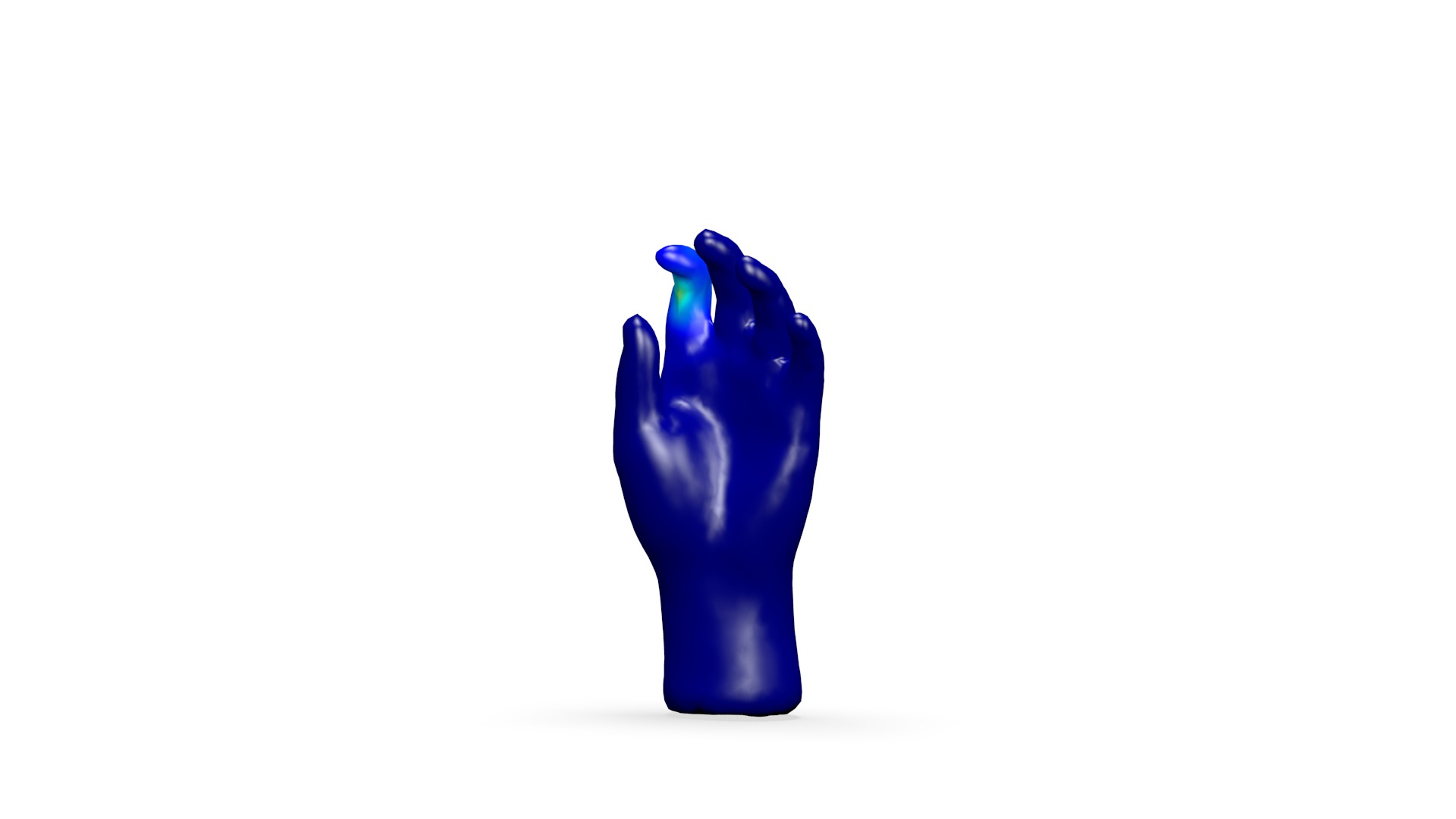}
	\put(32,0){\fontsize{6}{8}\selectfont $10$}
	\put(32,82){\fontsize{6}{8}\selectfont $10$}
	\put(32,194){\fontsize{6}{8}\selectfont $10$}
	\put(32,307){\fontsize{6}{8}\selectfont $10$}
	\put(27,425){\fontsize{6}{8}\selectfont $500$}
	\put(34,512){\fontsize{6}{8}\selectfont $5$}
	\put(34,595){\fontsize{6}{8}\selectfont $5$}
	\put(34,677){\fontsize{6}{8}\selectfont $2$}
	\end{overpic}
	\hspace{-0.2cm}
	\begin{overpic}
	[trim=20cm 1cm 10cm 6cm,clip,width=0.14\textwidth]{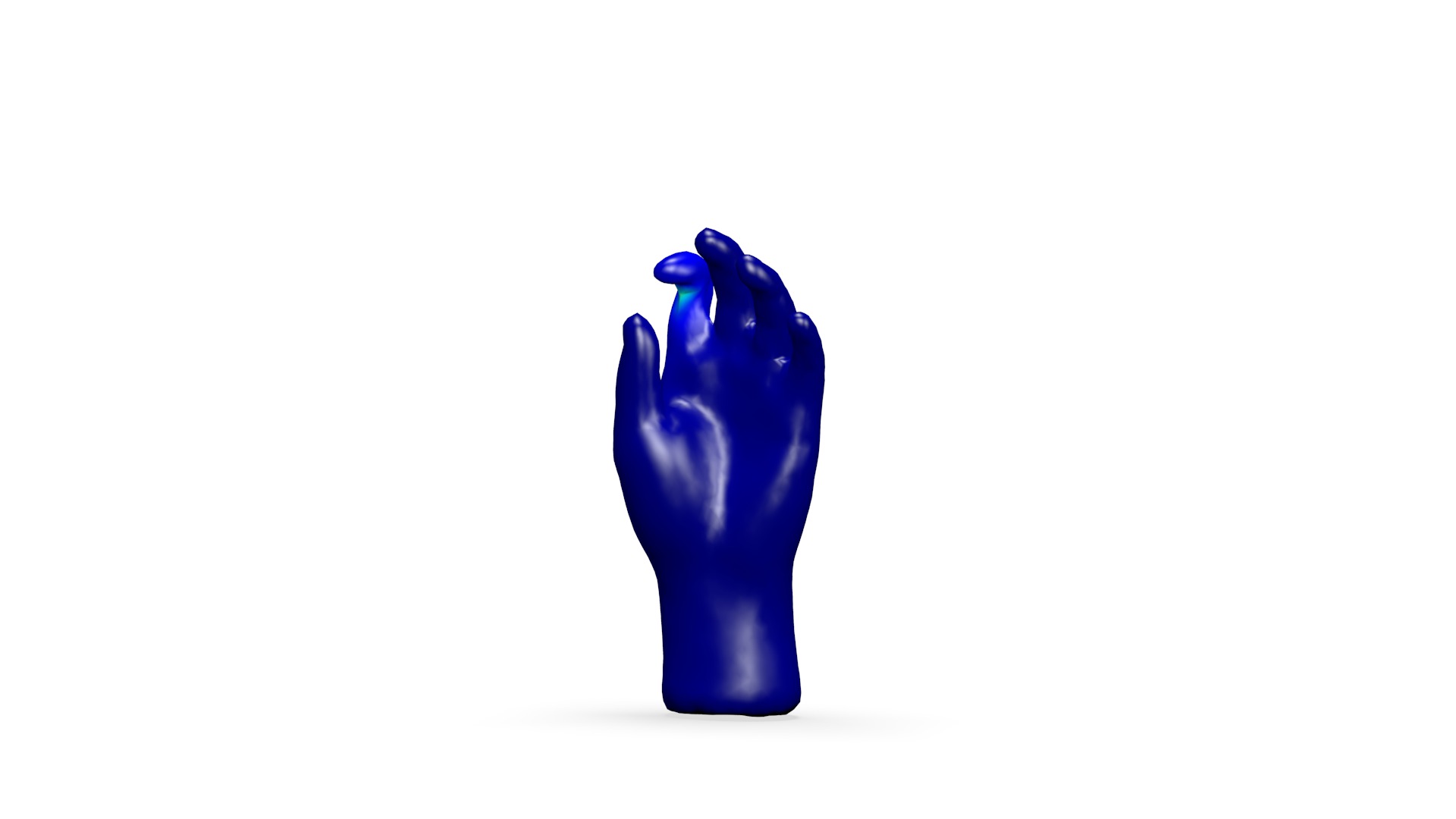}
	\put(32,0){\fontsize{6}{8}\selectfont $30$}
	\put(32,82){\fontsize{6}{8}\selectfont $50$}
	\put(32,194){\fontsize{6}{8}\selectfont $20$}
	\put(28,307){\fontsize{6}{8}\selectfont $50$}
	\put(28,425){\fontsize{6}{8}\selectfont $600$}
	\put(32,512){\fontsize{6}{8}\selectfont $20$}
	\put(32,595){\fontsize{6}{8}\selectfont $10$}
	\put(32,677){\fontsize{6}{8}\selectfont $10$}
	\end{overpic}
	\hspace{-0.2cm}
	\begin{overpic}
	[trim=20cm 1cm 10cm 6cm,clip,clip,width=0.14\textwidth]{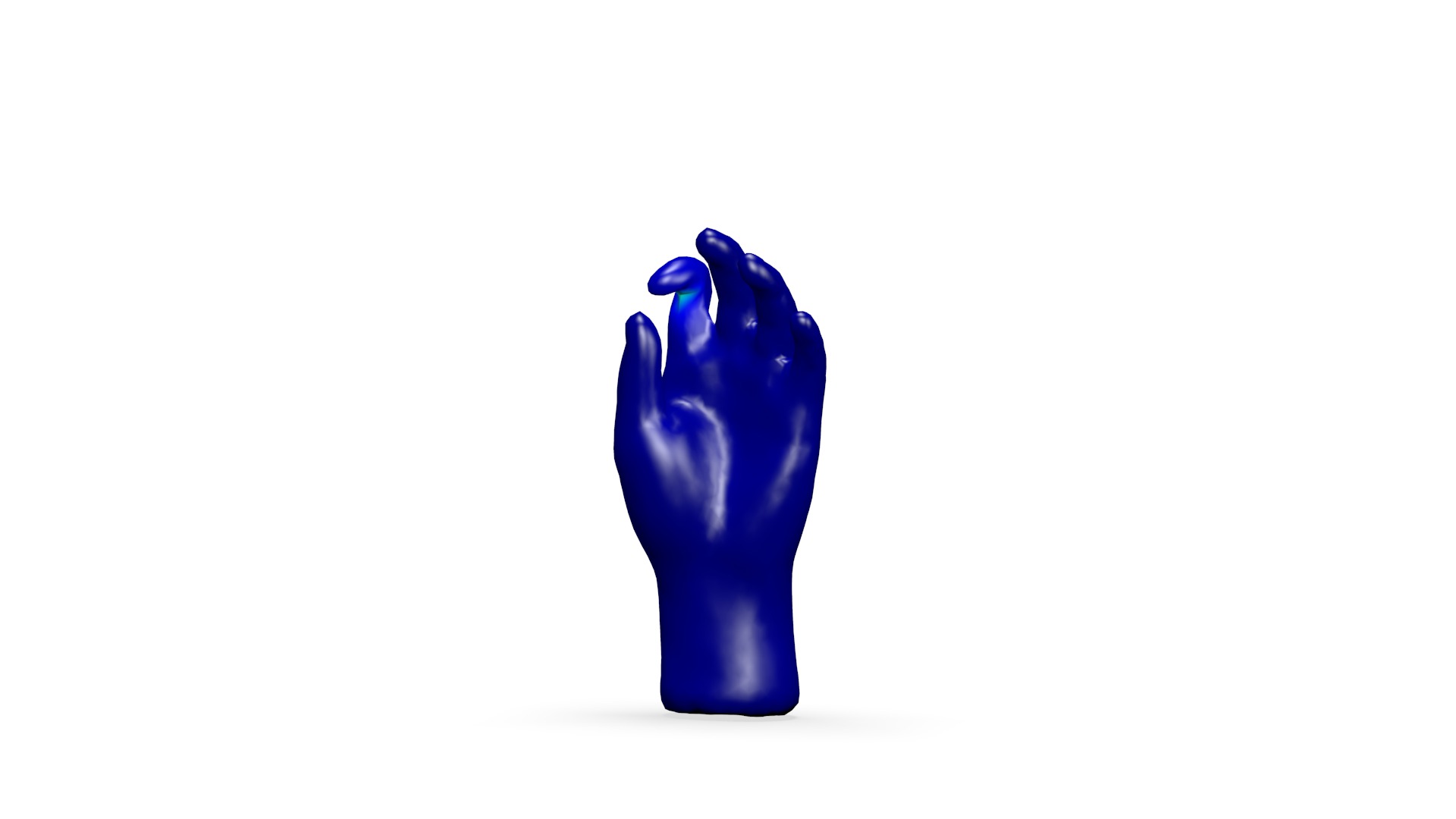}
	\put(28,0){\fontsize{6}{8}\selectfont $200$}
	\put(28,82){\fontsize{6}{8}\selectfont $100$}
	\put(32,194){\fontsize{6}{8}\selectfont $50$}
	\put(28,307){\fontsize{6}{8}\selectfont $300$}
	\put(28,425){\fontsize{6}{8}\selectfont $800$}
	\put(32,512){\fontsize{6}{8}\selectfont $50$}
	\put(32,595){\fontsize{6}{8}\selectfont $50$}
	\put(32,677){\fontsize{6}{8}\selectfont $50$}
	\end{overpic}
	\hspace{0.2cm}
	\begin{overpic}
	[trim=12cm 1cm 18cm 6cm,clip,clip,width=0.14\textwidth]{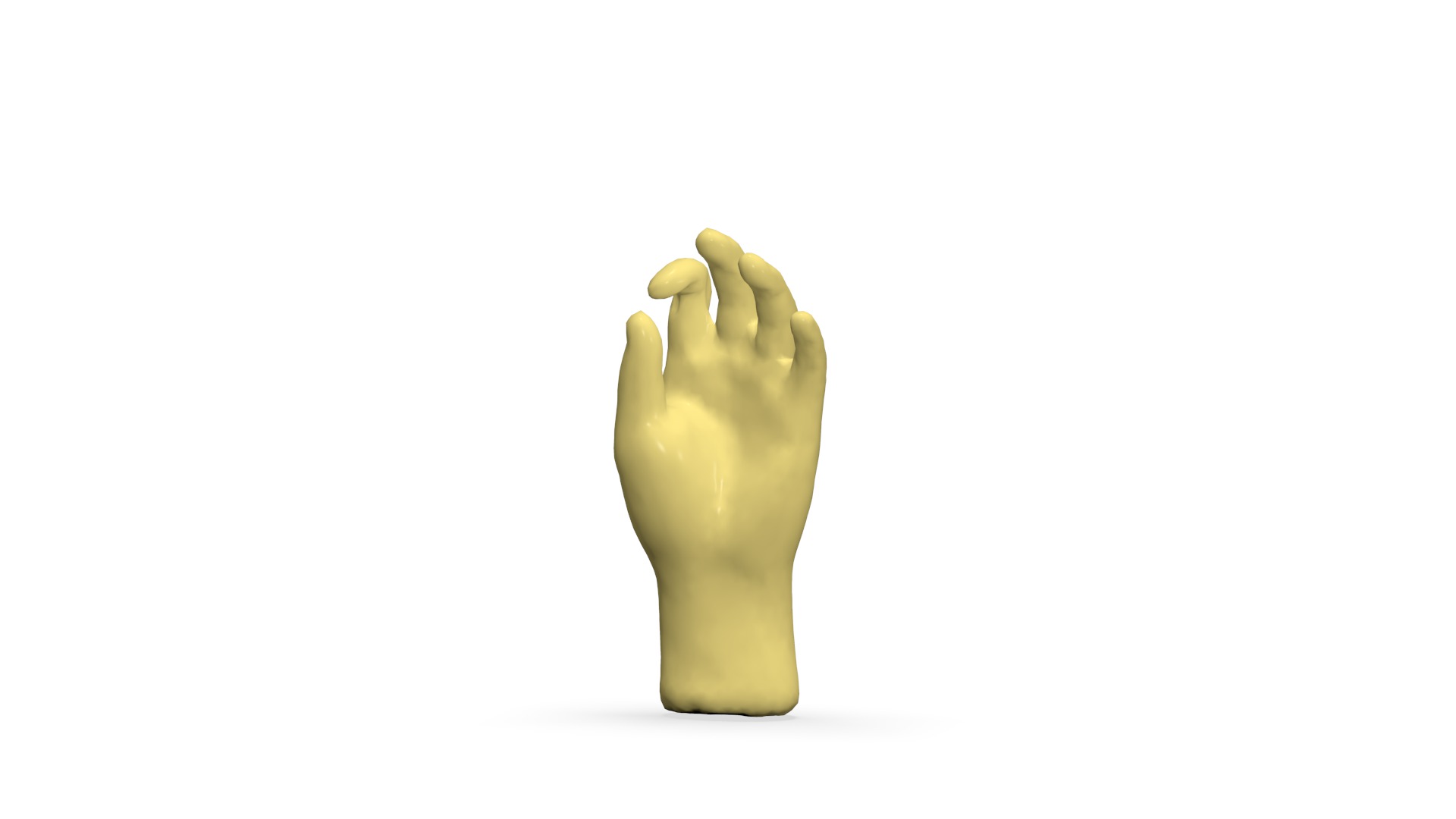}
	\put(55,-12){\fontsize{6}{8}\selectfont $X$}
	\put(-143,15){\begin{overpic}[width=0.15\textwidth,angle=180]{./figures/colorbar}\end{overpic}}
	\put(-32,10){\fontsize{6}{8}\selectfont $0$}
	\put(-32,67){\fontsize{6}{8}\selectfont $10^{1.4}$}
	\put(-143,100){\begin{overpic}[width=0.15\textwidth,angle=180]{./figures/colorbar}\end{overpic}}
	\put(-32,95){\fontsize{6}{8}\selectfont $0$}
	\put(-32,152){\fontsize{6}{8}\selectfont $10^{2.3}$}
	\put(-143,215){\begin{overpic}[width=0.15\textwidth,angle=180]{./figures/colorbar}\end{overpic}}
	\put(-32,210){\fontsize{6}{8}\selectfont $0$}
	\put(-32,267){\fontsize{6}{8}\selectfont $10^{1.9}$}
	\put(-143,325){\begin{overpic}[width=0.15\textwidth,angle=180]{./figures/colorbar}\end{overpic}}
	\put(-32,320){\fontsize{6}{8}\selectfont $0$}
	\put(-32,377){\fontsize{6}{8}\selectfont $10^{1.7}$}
	\put(-143,438){\begin{overpic}[width=0.15\textwidth,angle=180]{./figures/colorbar}\end{overpic}}
	\put(-32,435){\fontsize{6}{8}\selectfont $0$}
	\put(-32,492){\fontsize{6}{8}\selectfont $10^{1.9}$}
	\put(-143,525){\begin{overpic}[width=0.15\textwidth,angle=180]{./figures/colorbar}\end{overpic}}
	\put(-32,520){\fontsize{6}{8}\selectfont $0$}
	\put(-32,577){\fontsize{6}{8}\selectfont $10^{1.8}$}
	\put(-143,610){\begin{overpic}[width=0.15\textwidth,angle=180]{./figures/colorbar}\end{overpic}}
	\put(-32,605){\fontsize{6}{8}\selectfont $0$}
	\put(-32,662){\fontsize{6}{8}\selectfont $10^{2.1}$}
	\put(-143,690){\begin{overpic}[width=0.15\textwidth,angle=180]{./figures/colorbar}\end{overpic}}
	\put(-32,685){\fontsize{6}{8}\selectfont $0$}
	\put(-32,742){\fontsize{6}{8}\selectfont $10^{1.3}$}
	\put(0,-12){\color{black}\line(0,1){770}}
	\end{overpic}
\end{minipage} \\ \vspace{0.4cm}
%
%
\caption{Shape-from-Laplacian: given an initial shape $A$ (leftmost column) and the Laplacian of shape $B$ (second column from left), produce a new shape $X$ (rightmost column) whose Laplacian is as close as possible to the given one, by the proposed optimization algorithm initialized with $A$.  
Columns 3-6 show the intermediate steps of the optimization. Colors represent the vertex-wise energy $\epsilon_i$; hotter colors correspond to larger values (log scale). 
}
\label{fig:shapes_from_Laplacian}
\end{figure*}

\begin{figure}[b!]
\centering
%
%
\begin{minipage}{\columnwidth}
	\begin{overpic}
	[trim=24cm 1cm 8cm 5cm,clip,width=0.24\textwidth]{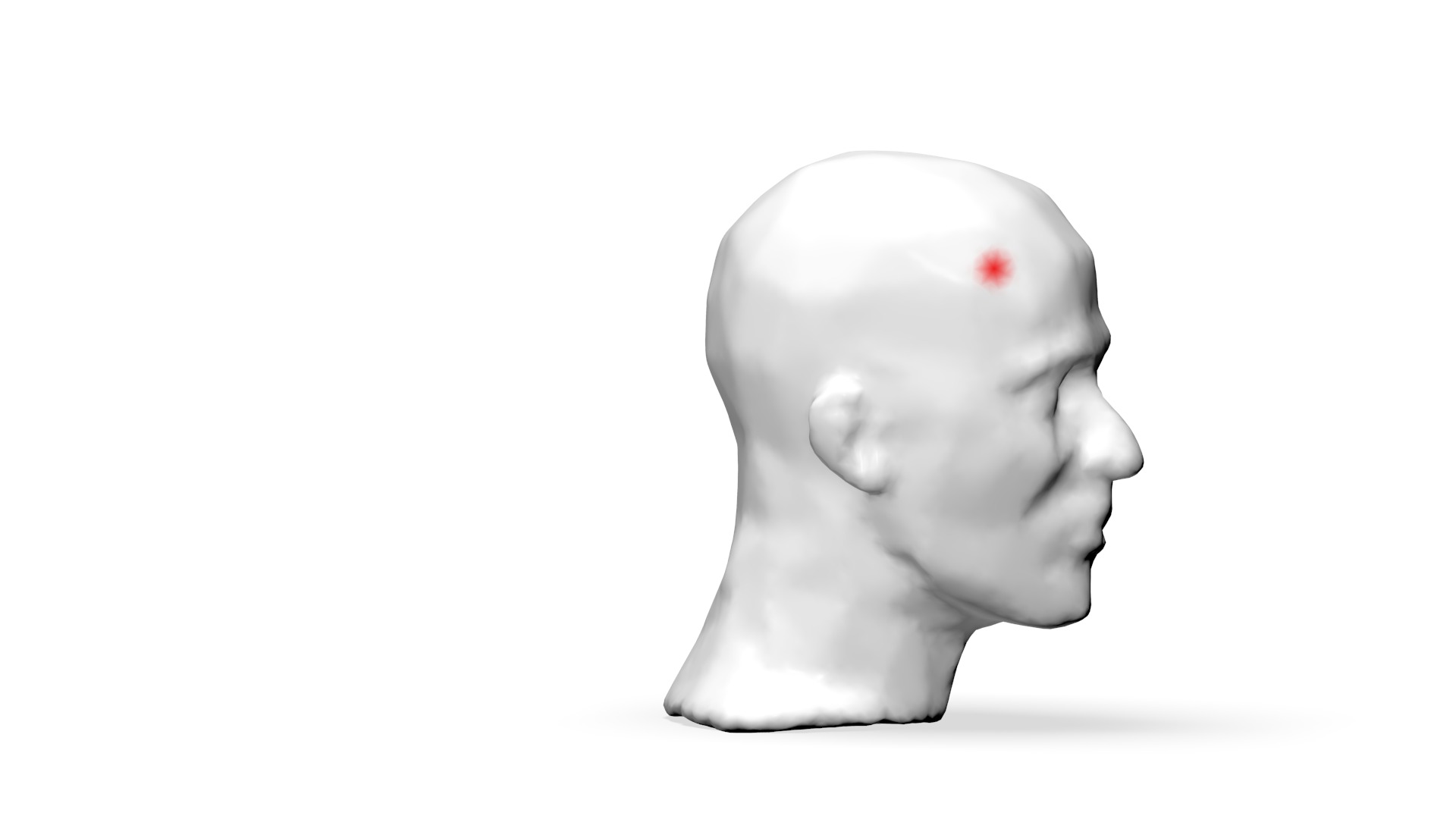}
	\end{overpic} 
	\begin{overpic}
	[trim=26cm 1cm 6cm 5cm,clip,width=0.24\textwidth]{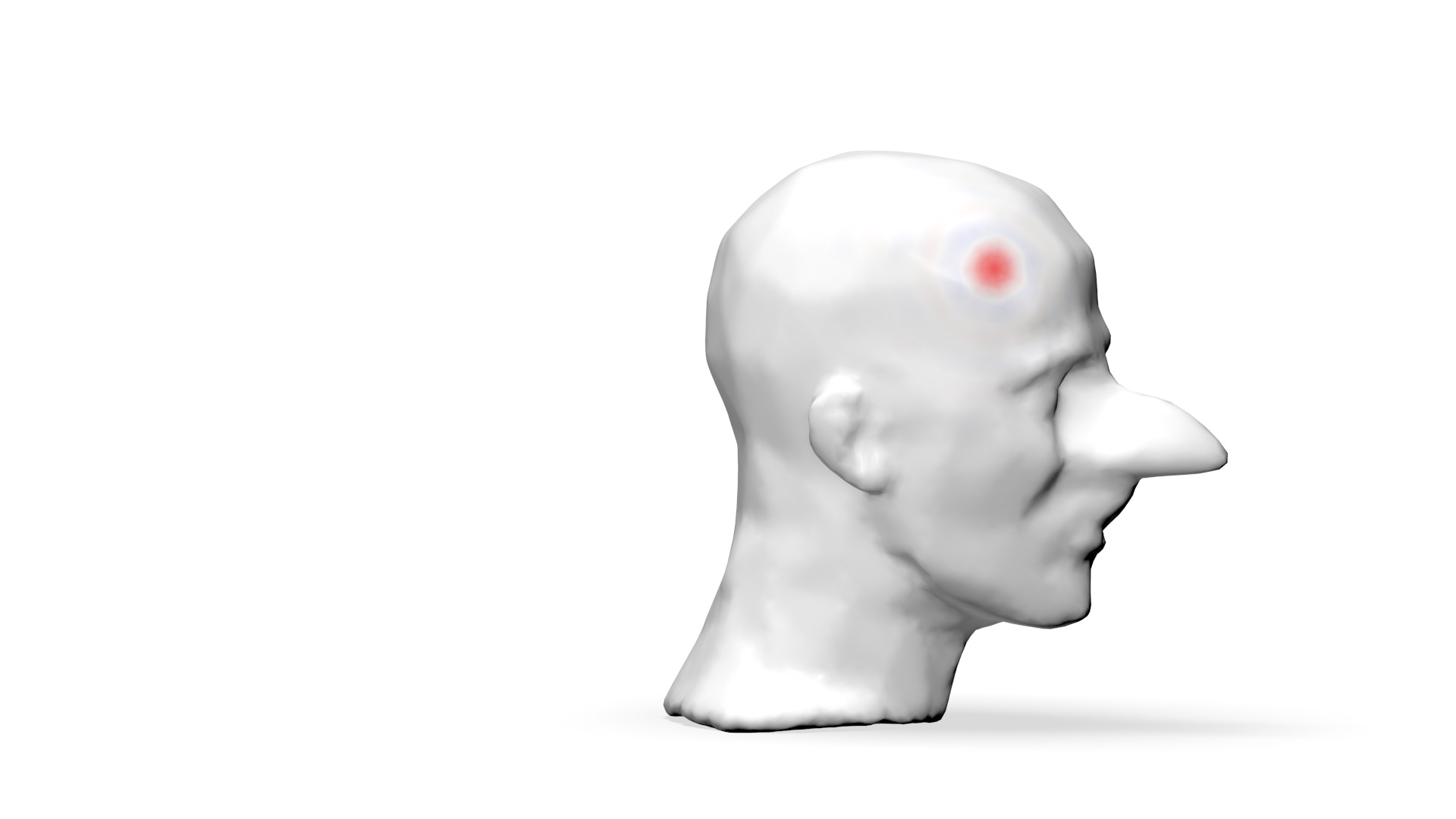}
	\end{overpic}
	\begin{overpic}
	[trim=28cm 1cm 4cm 5cm,clip,width=0.24\textwidth]{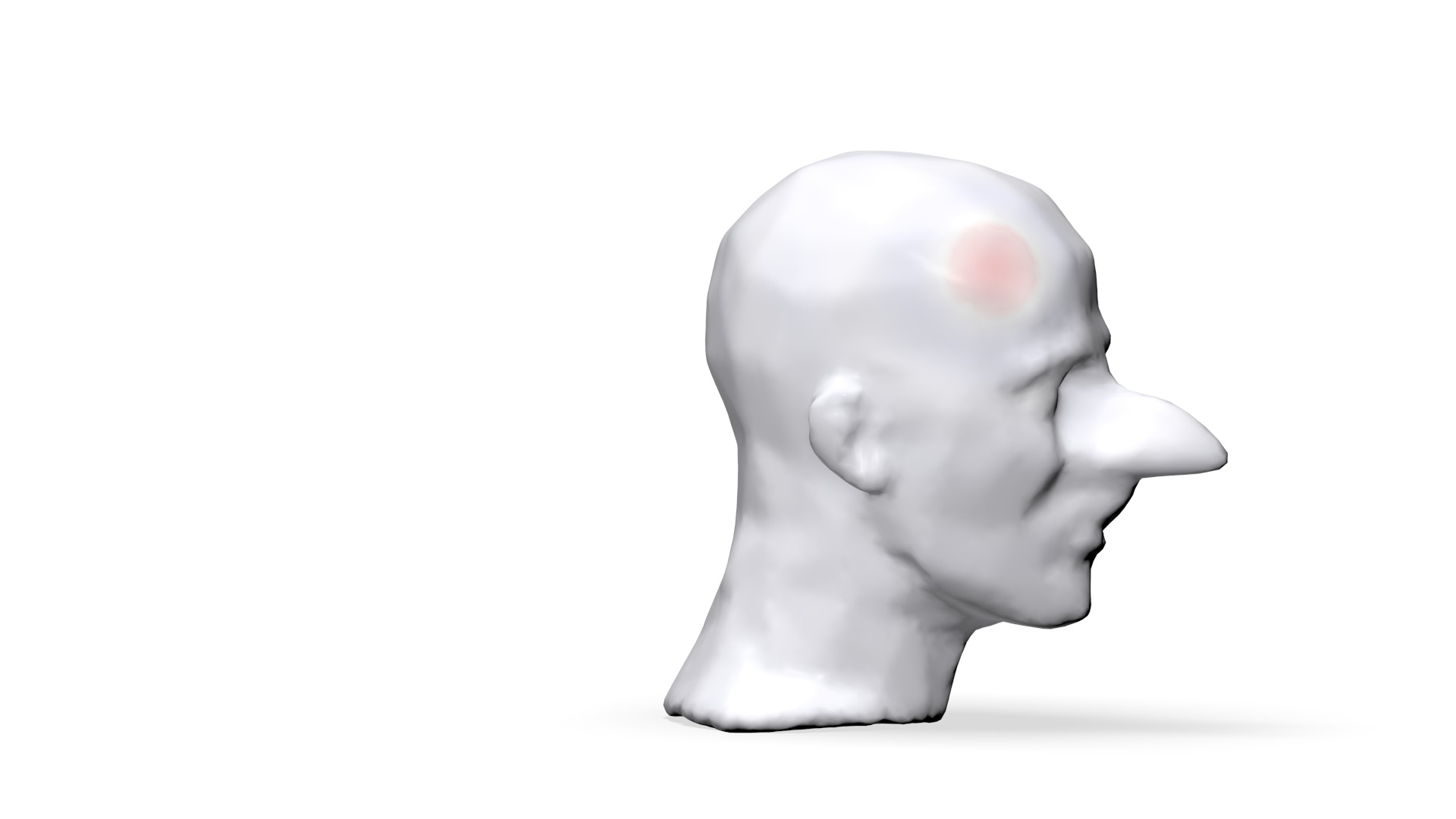}
	\end{overpic}
	\begin{overpic}
	[trim=31cm 1cm 0cm 6cm,clip,width=0.24\textwidth]{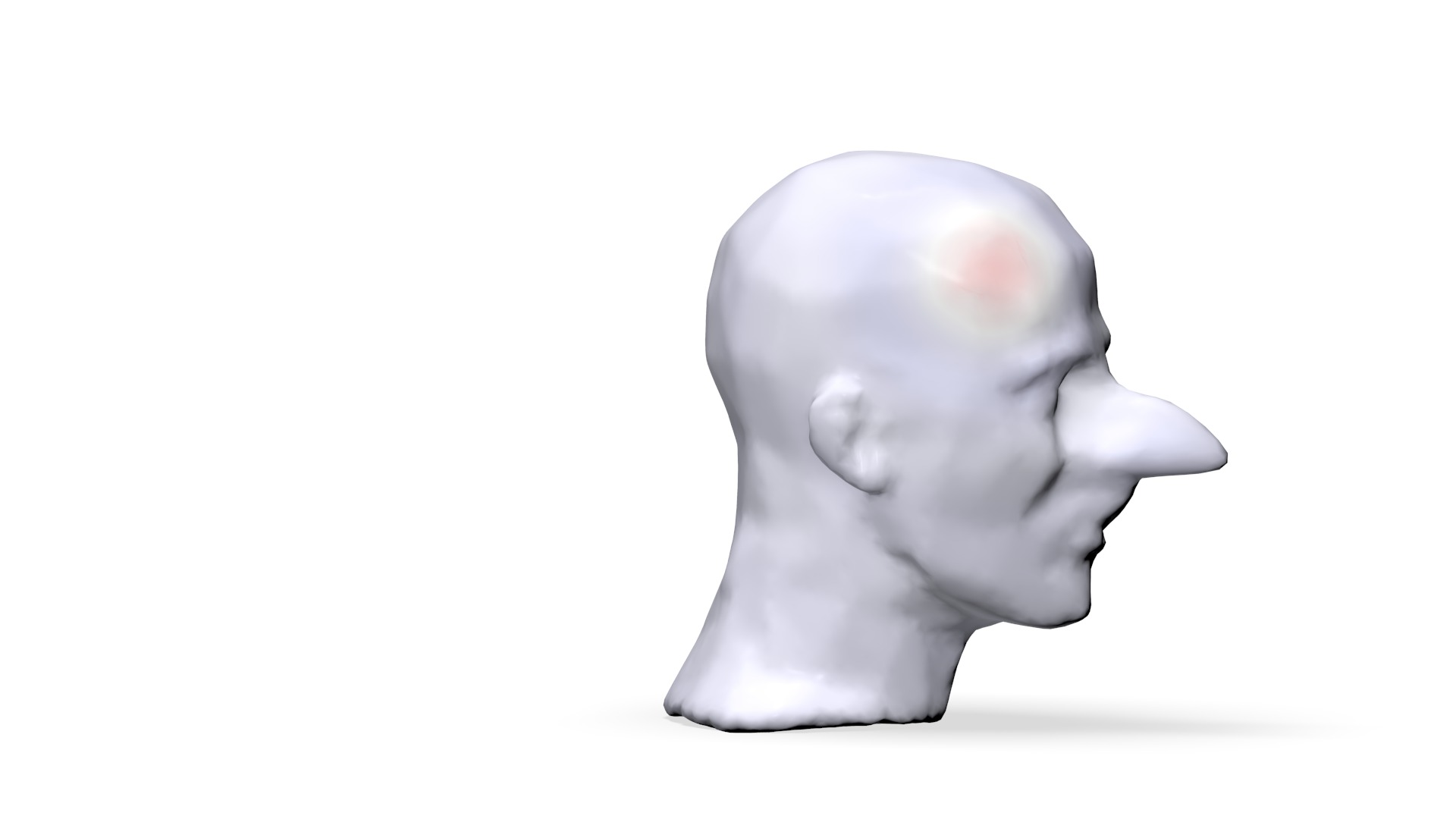}
	\put(-35,15){\begin{overpic}[width=0.3\textwidth,angle=180]{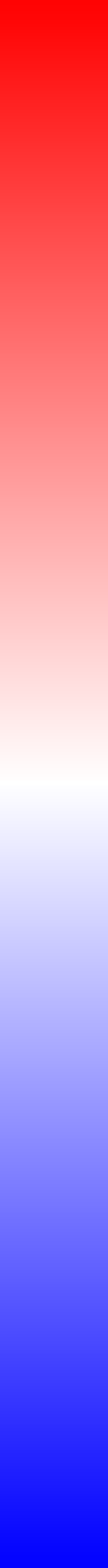}\end{overpic}}
	\put(95,10){\fontsize{6}{8}\selectfont $-1$}
	\put(95,40){\fontsize{6}{8}\selectfont $0$}
	\put(95,75){\fontsize{6}{8}\selectfont $1$}
	\end{overpic}
\end{minipage} \\ \vspace{2mm}
%
%
\begin{minipage}{\columnwidth}
	\begin{overpic}
	[trim=18cm 1cm 14cm 5cm,clip,width=0.24\textwidth]{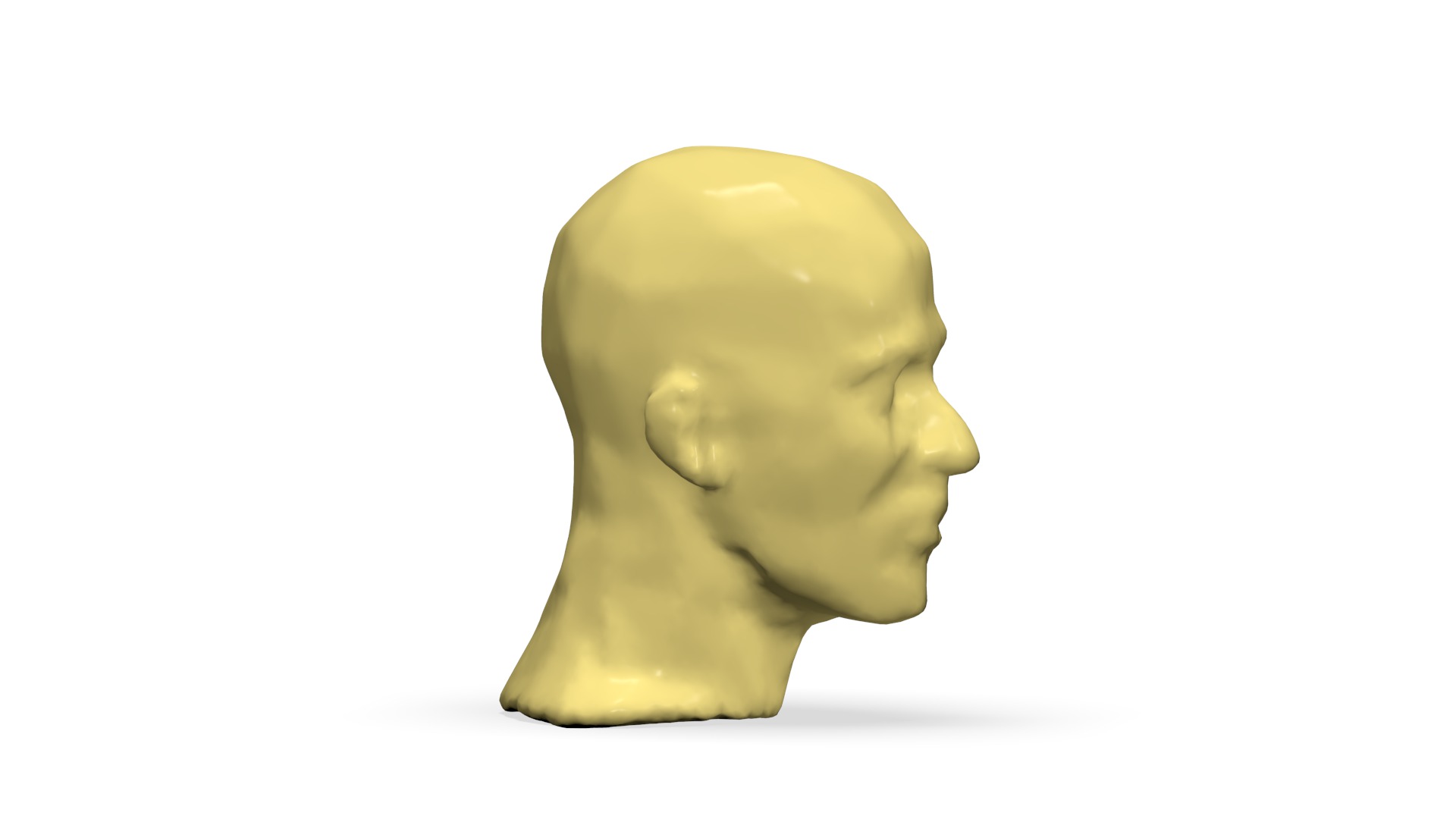}
	\end{overpic} 
	\begin{overpic}
	[trim=20cm 1cm 12cm 5cm,clip,width=0.24\textwidth]{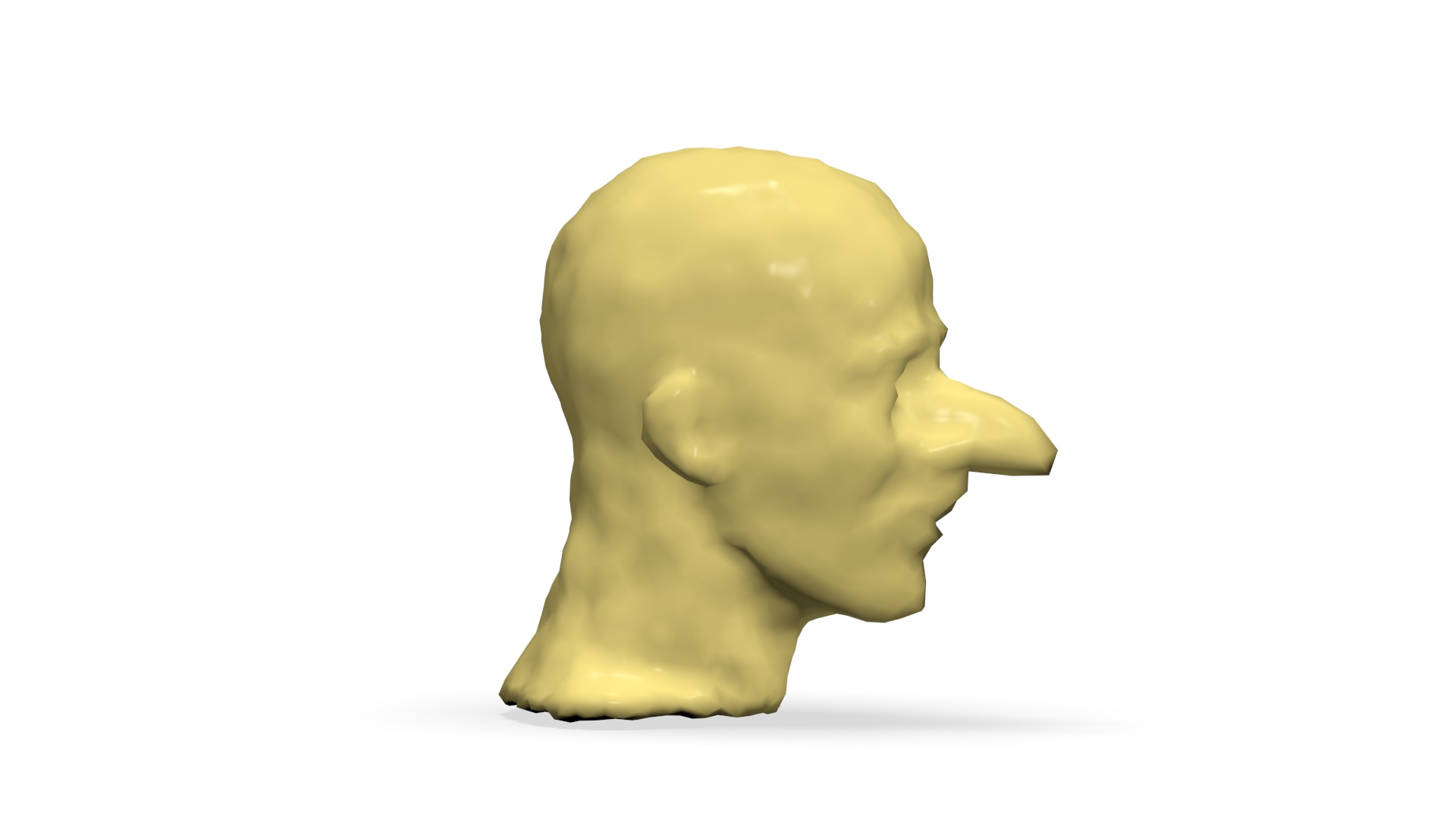}
	\put(11,-5){\fontsize{6}{8}\selectfont $K=1000$}
	\end{overpic}
	\begin{overpic}
	[trim=21cm 1cm 11cm 5cm,clip,width=0.24\textwidth]{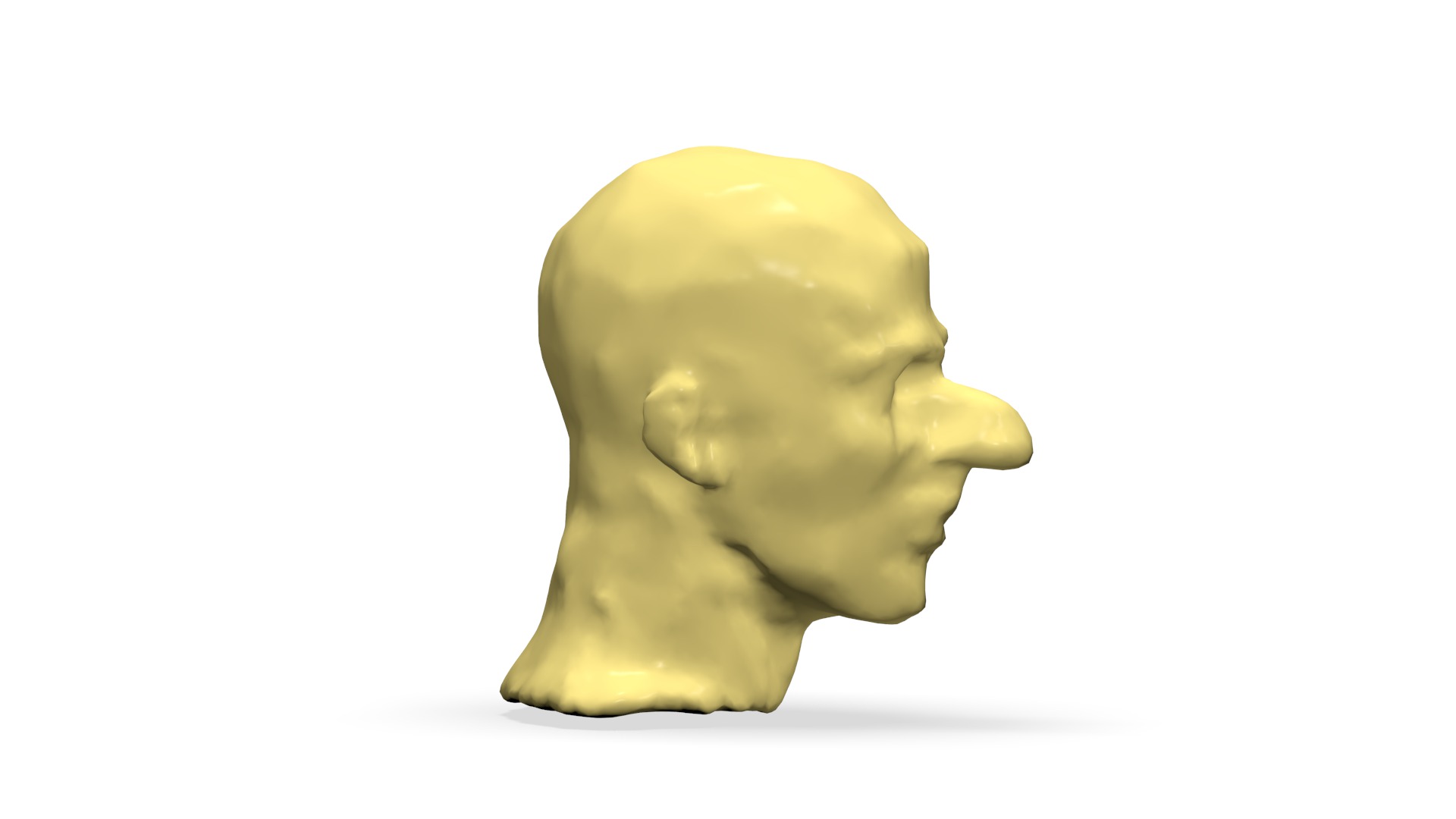}
	\put(25,-5){\fontsize{6}{8}\selectfont $200$}
	\end{overpic}
	\begin{overpic}
	[trim=23cm 1cm 8cm 5cm,clip,width=0.24\textwidth]{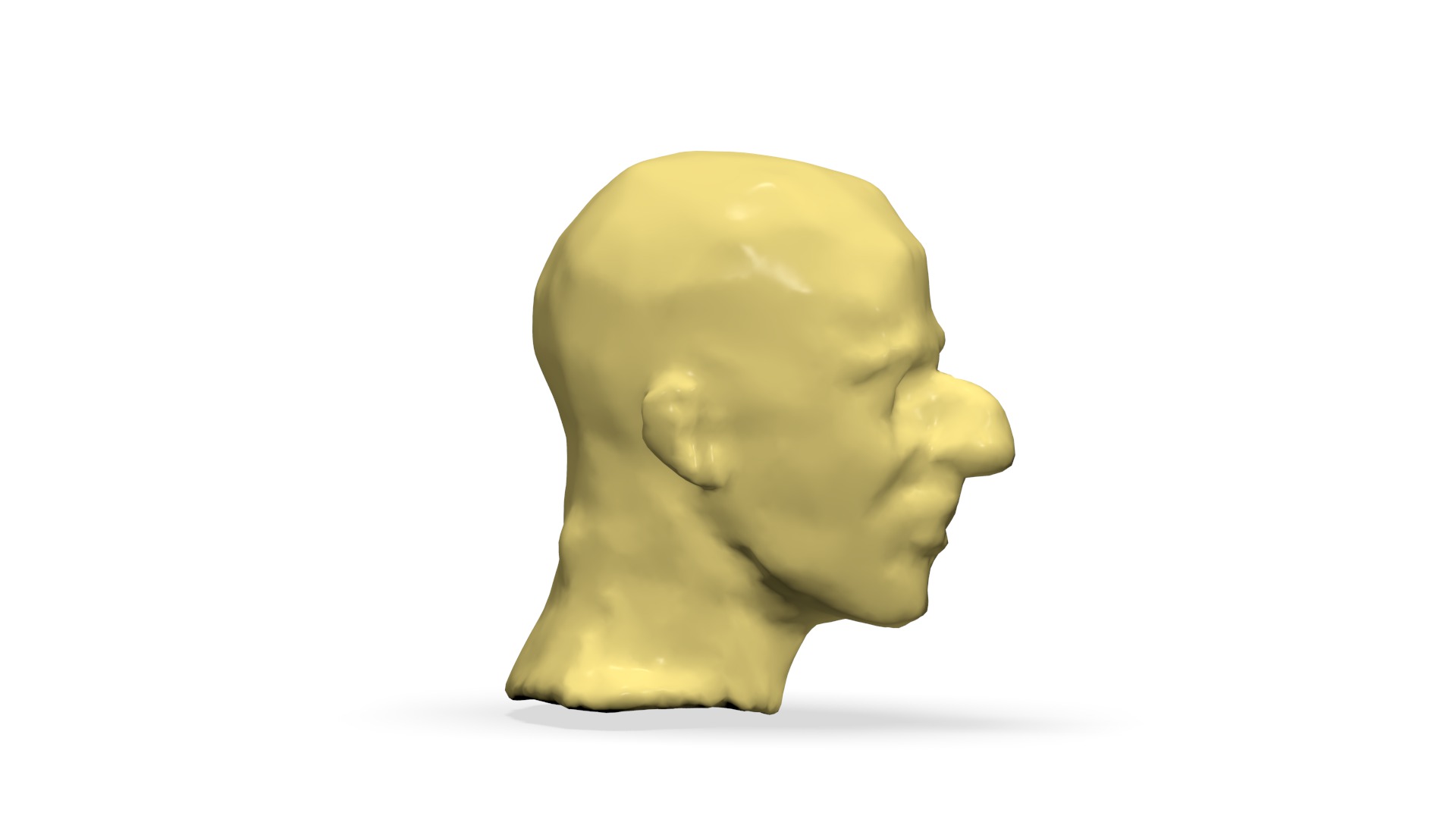}
	\put(18,-5){\fontsize{6}{8}\selectfont $100$}
	\end{overpic}
\end{minipage}
\caption{Sensitivity to functional map quality. Top: functional maps of different quality depending on the number of basis functions used in the spectral expansion, visualized by showing the image of a delta-function (leftmost) under the map (columns 2-4; the worse the map, the `blobbier' the result becomes). 
Bottom: result of shape-from-Laplacian reconstruction using the same shapes $A, B$ from Figure \ref{fig:shapes_from_Laplacian} (second row), related by the above functional maps. Result quality starts suffering when the map becomes inaccurate. 
}
\label{fig:funcmaps}
\end{figure}

To illustrate this sensitivity, we show in Figure \ref{fig:funcmaps} the result of shape-from-Laplacian reconstruction using as shapes $A$ and $B$ the faces from Figure \ref{fig:shapes_from_Laplacian} (second row) and varying the quality of the functional map between them. 
Functional maps were approximated in the bases of the first $K$ Laplace-Beltrami eigenfunctions according to \cite{ovsjanikov2012functional}. Larger values of $K$ result in a better expansion and consequently in better maps (Figure \ref{fig:funcmaps}, top). 
Figure \ref{fig:funcmaps} (bottom) shows the shape reconstruction result. The output quality is good as long as the functional maps are accurate, and deteriorates only when the map becomes very rough.



\vspace{1cm}

\section*{Acknowledgements}

The authors are grateful to Randolf Sch{\"a}rfig for help with generating the experimental datasets and rendering the results, 
to Daniele Panozzo for help with isotropic remeshing,
to Fernando De Goes, David Gu, and Maks Ovsjanikov for helpful discussion on their prior works, 
and to Klaus Glashov and Artiom Kovnatsky for their useful comments on an early version of the manuscript. 
This research was supported by ERC Starting Grant no. 307047 (COMET).

\bibliographystyle{acmsiggraph}
\bibliography{./sections/biblio,./sections/intro}

\begin{thebibliography}{\protect\citename{Ovsjanikov et~al\mbox{.} }2012}

\bibitem[\protect\citename{Aflalo and Kimmel }2013]{aflalo2013spectral}
{\sc Aflalo, Y., and Kimmel, R.}
\newblock 2013.
\newblock Spectral multidimensional scaling.
\newblock {\em PNAS 110}, 45, 18052--18057.

\bibitem[\protect\citename{{Aim}@{S}hape }]{aimatshape}
{\sc {Aim}@{S}hape}.
\newblock {A}im@{S}hape shape repository.
\newblock \url{http://shapes.aimatshape.net/}.

\bibitem[\protect\citename{Alhashim et~al\mbox{.} }2014]{alhashim_sig14}
{\sc Alhashim, I., Li, H., Xu, K., Cao, J., Ma, R., and Zhang, H.}
\newblock 2014.
\newblock Topology-varying 3{D} shape creation via structural blending.
\newblock {\em TOG 33}, 4.

\bibitem[\protect\citename{Azencot et~al\mbox{.} }2013]{azencot2013operator}
{\sc Azencot, O., Ben-Chen, M., Chazal, F., and Ovsjanikov, M.}
\newblock 2013.
\newblock An operator approach to tangent vector field processing.
\newblock {\em CGF 32}, 5, 73--82.

\bibitem[\protect\citename{Bobenko and Springborn }2007]{bobenko2007discrete}
{\sc Bobenko, A.~I., and Springborn, B.~A.}
\newblock 2007.
\newblock A discrete {L}aplace--{B}eltrami operator for simplicial surfaces.
\newblock {\em Discrete \& Computational Geometry 38}, 4, 740--756.

\bibitem[\protect\citename{Borg and Groenen }2005]{borg2005modern}
{\sc Borg, I., and Groenen, P.~J.}
\newblock 2005.
\newblock {\em Modern multidimensional scaling: Theory and applications}.
\newblock Springer.

\bibitem[\protect\citename{Brady and Yuille }1984]{brady1984extremum}
{\sc Brady, M., and Yuille, A.}
\newblock 1984.
\newblock An extremum principle for shape from contour.
\newblock {\em PAMI}, 3, 288--301.

\bibitem[\protect\citename{Bronstein et~al\mbox{.}
  }2006]{bronstein2006multigrid}
{\sc Bronstein, M.~M., Bronstein, A.~M., Kimmel, R., and Yavneh, I.}
\newblock 2006.
\newblock Multigrid multidimensional scaling.
\newblock {\em Numerical linear algebra with applications 13}, 2-3, 149--171.

\bibitem[\protect\citename{Bronstein et~al\mbox{.} }2008]{tosca}
{\sc Bronstein, A.~M., Bronstein, M.~M., and Kimmel, R.}
\newblock 2008.
\newblock {\em {N}umerical {G}eometry of {N}on-{R}igid {S}hapes}.
\newblock Springer.

\bibitem[\protect\citename{Clarke et~al\mbox{.} }2011]{Clarke2011}
{\sc Clarke, L., Chen, M., and Mora, B.}
\newblock 2011.
\newblock Automatic generation of 3{D} caricatures based on artistic
  deformation styles.
\newblock {\em TVCG 17}, 6 (June), 808--821.

\bibitem[\protect\citename{de~Goes et~al\mbox{.} }2014]{deGoes2014}
{\sc de~Goes, F., Memari, P., Mullen, P., and Desbrun, M.}
\newblock 2014.
\newblock Weighted triangulation for geometry processing.
\newblock {\em TOG\/}.

\bibitem[\protect\citename{Forsyth }2001]{Forsyth01}
{\sc Forsyth, D.~A.}
\newblock 2001.
\newblock Shape from texture and integrability.
\newblock In {\em Proc. ICCV}.

\bibitem[\protect\citename{Ikeuchi and Horn }1981]{IkeuchiHorn1981}
{\sc Ikeuchi, K., and Horn, B. K.~P.}
\newblock 1981.
\newblock Numerical shape from shading and occluding boundaries.
\newblock {\em Artificial Intelligence 17}, 1-3, 141--184.

\bibitem[\protect\citename{Ikeuchi }1984]{ikeuchi1984}
{\sc Ikeuchi, K.}
\newblock 1984.
\newblock Shape from regular patterns.
\newblock {\em Artificial Intelligence 22}, 1, 49 -- 75.

\bibitem[\protect\citename{Jacobson and Sorkine }2012]{Jacobson2012}
{\sc Jacobson, A., and Sorkine, O.}
\newblock 2012.
\newblock A cotangent {L}aplacian for images as surfaces.
\newblock Tech. Rep. 757, ETH Zurich.

\bibitem[\protect\citename{Kanatani }1985]{Kanatani1985}
{\sc Kanatani, K.-I.}
\newblock 1985.
\newblock Structure from motion without correspondence: General principle.
\newblock In {\em Proc. IJCAI}.

\bibitem[\protect\citename{Karpenko and Hughes }2006]{Karpenko2006}
{\sc Karpenko, O.~A., and Hughes, J.~F.}
\newblock 2006.
\newblock Smoothsketch: 3{D} free-form shapes from complex sketches.
\newblock {\em TOG 25}, 3 (July), 589--598.

\bibitem[\protect\citename{Kruskal }1964]{kruskal1964multidimensional}
{\sc Kruskal, J.~B.}
\newblock 1964.
\newblock Multidimensional scaling by optimizing goodness of fit to a nonmetric
  hypothesis.
\newblock {\em Psychometrika 29}, 1, 1--27.

\bibitem[\protect\citename{Leeuw et~al\mbox{.} }1977]{de1977applications}
{\sc Leeuw, J.~D., Barra, I. J.~R., Brodeau, F., Romier, G., and (eds), B.
  V.~C.}
\newblock 1977.
\newblock Applications of convex analysis to multidimensional scaling.
\newblock In {\em Recent Developments in Statistics}, North Holland Publishing
  Company, 133--146.

\bibitem[\protect\citename{Lewiner et~al\mbox{.} }2011]{Lewiner2011}
{\sc Lewiner, T., Vieira, T., Mart\'{\i}nez, D., Peixoto, A., Mello, V., and
  Velho, L.}
\newblock 2011.
\newblock Interactive 3{D} caricature from harmonic exaggeration.
\newblock {\em Computers \& Graphics 35}, 3, 586--595.

\bibitem[\protect\citename{Ma et~al\mbox{.} }2014]{Ma2014}
{\sc Ma, C., Huang, H., Sheffer, A., Kalogerakis, E., and Wang, R.}
\newblock 2014.
\newblock {Analogy-Driven 3{D} Style Transfer}.
\newblock {\em CGF 33}, 2, 175--184.

\bibitem[\protect\citename{Meyer et~al\mbox{.} }2003]{meyer2003:ddg}
{\sc Meyer, M., Desbrun, M., Schr{\"o}der, P., and Barr, A.~H.}
\newblock 2003.
\newblock Discrete differential-geometry operators for triangulated
  2-manifolds.
\newblock {\em Visualization$\&$Mathematics\/}, 35--57.

\bibitem[\protect\citename{Ovsjanikov et~al\mbox{.}
  }2012]{ovsjanikov2012functional}
{\sc Ovsjanikov, M., Ben-Chen, M., Solomon, J., Butscher, A., and Guibas, L.}
\newblock 2012.
\newblock Functional maps: A flexible representation of maps between shapes.
\newblock {\em TOG 31}, 4.

\bibitem[\protect\citename{Peyre }2007]{Peyre_toolbox}
{\sc Peyre, G.}, 2007.
\newblock Graph toolbox.
\newblock
  \url{http://www.mathworks.com/matlabcentral/fileexchange/5355-toolbox-graph}.

\bibitem[\protect\citename{Pinkall and Polthier }1993]{Pinkall1993}
{\sc Pinkall, U., and Polthier, K.}
\newblock 1993.
\newblock Computing discrete minimal surfaces and their conjugates.
\newblock {\em Experimental Mathematics 2}, 1, 15--36.

\bibitem[\protect\citename{Poelman and Kanade }1997]{Poelman1997}
{\sc Poelman, C.~J., and Kanade, T.}
\newblock 1997.
\newblock A paraperspective factorization method for shape and motion recovery.
\newblock {\em PAMI 19}, 3, 206--218.

\bibitem[\protect\citename{Rong et~al\mbox{.} }2008]{Rong2008}
{\sc Rong, G., Cao, Y., and Guo, X.}
\newblock 2008.
\newblock Spectral mesh deformation.
\newblock {\em Visual Computer 24}, 7, 787--796.

\bibitem[\protect\citename{Rosenberg }1997]{rosenberg1997laplacian}
{\sc Rosenberg, S.}
\newblock 1997.
\newblock {\em The Laplacian on a Riemannian manifold: an introduction to
  analysis on manifolds}.
\newblock No.~31. Cambridge University Press.

\bibitem[\protect\citename{Rosenholtz and Malik }1997]{Rosenholtz1997}
{\sc Rosenholtz, R., and Malik, J.}
\newblock 1997.
\newblock {Surface orientation from texture: Isotropy or homogeneity (or
  both)?}
\newblock {\em Vision Research 37}, 16, 2283--2293.

\bibitem[\protect\citename{Rustamov et~al\mbox{.} }2013]{Rustamov2013}
{\sc Rustamov, R.~M., Ovsjanikov, M., Azencot, O., Ben-Chen, M., Chazal, F.,
  and Guibas, L.}
\newblock 2013.
\newblock Map-based exploration of intrinsic shape differences and variability.
\newblock {\em TOG 32}, 4, 72:1--72:12.

\bibitem[\protect\citename{Snavely et~al\mbox{.} }2006]{Snavely2006}
{\sc Snavely, N., Seitz, S.~M., and Szeliski, R.}
\newblock 2006.
\newblock Photo tourism: Exploring photo collections in 3{D}.
\newblock {\em TOG 25}, 3, 835--846.

\bibitem[\protect\citename{Sorkine et~al\mbox{.} }2004]{sorkine2004laplacian}
{\sc Sorkine, O., Cohen-Or, D., Lipman, Y., Alexa, M., R{\"o}ssl, C., and
  Seidel, H.-P.}
\newblock 2004.
\newblock Laplacian surface editing.
\newblock In {\em Proc. SGP}.

\bibitem[\protect\citename{Sumner and Popovi\'{c} }2004]{Sumner2004}
{\sc Sumner, R.~W., and Popovi\'{c}, J.}
\newblock 2004.
\newblock Deformation transfer for triangle meshes.
\newblock {\em TOG 23}, 3, 399--405.

\bibitem[\protect\citename{Valgaerts et~al\mbox{.} }2012]{Valgaerts2012}
{\sc Valgaerts, L., Wu, C., Bruhn, A., Seidel, H.-P., and Theobalt, C.}
\newblock 2012.
\newblock Lightweight binocular facial performance capture under uncontrolled
  lighting.
\newblock {\em TOG 31}, 6, 187:1--187:11.

\bibitem[\protect\citename{Welnicka et~al\mbox{.} }2011]{Welnicka2011}
{\sc Welnicka, K., B{\ae}rentzen, J., Aan{\ae}s, H., and Larsen, R.}
\newblock 2011.
\newblock {\em Descriptor Based Classification of Shapes in Terms of Style and
  Function}.
\newblock IMM-Technical Report-2011. Technical University of Denmark.

\bibitem[\protect\citename{Witkin }1980]{Witkin1980}
{\sc Witkin, A.~P.}
\newblock 1980.
\newblock {\em Shape from contour}.
\newblock PhD thesis, MIT.

\bibitem[\protect\citename{Woodham }1980]{Woodham1989}
{\sc Woodham, R.~J.}
\newblock 1980.
\newblock Photometric method for determining surface orientation from multiple
  images.
\newblock {\em Optical Engineering 19}, 1, 139--144.

\bibitem[\protect\citename{Yu et~al\mbox{.} }2013]{Yu2013}
{\sc Yu, L.-F., Yeung, S.-K., Tai, Y.-W., and Lin, S.}
\newblock 2013.
\newblock Shading-based shape refinement of rgb-d images.
\newblock In {\em Proc. CVPR}.

\bibitem[\protect\citename{Zeng et~al\mbox{.} }2012]{Zeng2012}
{\sc Zeng, W., Guo, R., Luo, F., and Gu, X.}
\newblock 2012.
\newblock Discrete heat kernel determines discrete {R}iemannian metric.
\newblock {\em Graphical Models 74}, 4, 121--129.

\end{thebibliography}

\section*{Appendix: Derivatives of the energy} 

\fontsize{8}{9}

In this section, we derive the gradient of the energy used in Section~\ref{sec:num}, considering  a generic energy of the form %
$\mathcal{E}(\bb{\ell}) = \lVert \bb{H}\bb{Q}(\bb{\ell}) \bb{K}-\bb{J} \rVert_\mathrm{F}^2$, %
with $\bb{Q}(\bb{\ell})$ being either $\bb{A}(\bb{\ell})$ or $\bb{W}(\bb{\ell})$.
%
%
By the chain rule, 
\[
\frac{\partial \mathcal{E}(\bb{\ell})}{\partial \bb{\ell}} = \frac{\partial \mathcal{E}(\bb{\ell})}{\partial \bb{Q}(\bb{\ell})} \frac{\partial \bb{Q}(\bb{\ell})}{\partial \bb{\ell}}, 
\]
where 
\[
\frac{\partial \mathcal{E}(\bb{\ell})}{\partial \bb{Q}(\bb{\ell})} = 2 \bb{H}^\top (\bb{H}\bb{Q}(\ell)\bb{K}-\bb{J})\bb{K}^\top 
\]
is an $n\times n$ matrix, which is row-stacked into an $1 \times n^2$. 
%
The gradient $\frac{\partial \bb{Q}(\bb{\ell})}{\partial \bb{\ell}}$ can be represented as an $n^2 \times |E|$ matrix, which, when multiplied by $\frac{\partial \mathcal{E}(\bb{\ell})}{\partial \bb{Q}(\bb{\ell})}$, produces a vector of size $\lvert E \rvert$.

For $\bb{Q} = \bb{A}$, a diagonal matrix of area elements~(\ref{eq:lap_area}), the computation of the gradient is based on the derivative of the area element~(\ref{eq:heron})
\[
\frac{\partial A_{ijk}}{\partial \ell_{i'j'}}  = 
\begin{cases}
\gamma(\ell_{ij},\ell_{jk},\ell_{ki}) & i'j' = ij \\
\gamma(\ell_{jk},\ell_{ij},\ell_{ki}) & i'j' = jk \\
\gamma(\ell_{ki},\ell_{ij},\ell_{jk})  & i'j' = ki  \\
0 & \text{else} 
\end{cases}
\]
%
%
where $\gamma$ is defined as 
\[
\begin{aligned}
\gamma(x,y,z) = \frac{1}{4 A_{ijk}} \Big{[} & (s-x)(s-y)(s-z) + s(s-x)(s-y) \\
& + s(s-x)(s-z) - s(s-y)(s-z) \Big{]}.
\end{aligned}
\]
Then, 
\[
\begin{aligned}
\frac{\partial a_{ij} }{\partial \ell_{i'j'} } = \left\{
\begin{array}{cc}
0 & i\neq j \\
\frac{1}{3}\sum_{kl : ikl\in F} \frac{\partial A_{ikl}}{\partial \ell_{i'j'}} & i=j 
\end{array}
\right.
\end{aligned}
\]
for $i'j' \in E$.

For $\bb{Q} = \bb{W}$, we have 
\[
\frac{\partial w_{ij} }{\partial \ell_{i'j'} } = 
\begin{cases}
\frac{\ell_{ij}}{2} \left[ \frac{ \frac{\partial A_{ijk}}{\partial \ell_{i'j'}} \ell_{ij} - 2A_{ijk}  }{A_{ijk}^2 } + 
\frac{ \frac{\partial A_{ijh}}{\partial \ell_{i'j'}} \ell_{ij} - 2 A_{ijh} }{A_{ijh}^2 }
\right]
 & i'j' = ij \\
\ell_{ij} \frac{\left( \frac{\partial A_{ijk}}{\partial \ell_{i'j'}} \ell_{ij} - 2 A_{ijk} \right) }{2A_{ijk}^2 } & \small ij \in \{ ik, jk, ih, jh \} \\ 
0 & \text{else} 
\end{cases}
\]
for $i\neq j$ and 
\[
\frac{\partial w_{ii} }{\partial \ell_{i'j'} } = - \sum_{j \neq i} \frac{\partial w_{ij} }{\partial \ell_{i'j'} }. 
\]
for the diagonal elements, where $i'j' \in E$.

%
%
%
%

\end{document}